\documentclass{article}

\PassOptionsToPackage{numbers, compress}{natbib}

\usepackage[final]{neurips_2022}

\usepackage[utf8]{inputenc} 
\usepackage[T1]{fontenc}    
\usepackage{hyperref}       
\usepackage{url}            
\usepackage{booktabs}       
\usepackage{amsfonts}       
\usepackage{nicefrac}       
\usepackage{microtype}      
\usepackage{xcolor}         
\usepackage{amsmath}
\usepackage{amssymb}
\usepackage{amsthm}
\usepackage{bm}
\usepackage{graphicx}
\usepackage{color}
\usepackage{multirow}
\usepackage{wrapfig}
\usepackage{here}

\usepackage[hang,small,bf]{caption}
\usepackage[subrefformat=parens]{subcaption}
\usepackage[toc,page,header]{appendix}
\usepackage{minitoc}

\definecolor{my-theme}{rgb}{0.372,0.537,0.537}

\usepackage{listings}
\usepackage{xcolor}
\usepackage[export]{adjustbox}

\definecolor{codegreen}{rgb}{0,0.6,0}
\definecolor{codegray}{rgb}{0.5,0.5,0.5}
\definecolor{codepurple}{rgb}{0.58,0,0.82}
\definecolor{backcolour}{rgb}{0.95,0.95,0.92}

\lstdefinestyle{mystyle}{
    backgroundcolor=\color{backcolour},   
    commentstyle=\color{codegreen},
    keywordstyle=\color{magenta},
    numberstyle=\tiny\color{codegray},
    stringstyle=\color{codepurple},
    basicstyle=\ttfamily\footnotesize,
    breakatwhitespace=false,         
    breaklines=true,                 
    captionpos=b,                    
    keepspaces=true,                 
    numbers=left,                    
    numbersep=5pt,                  
    showspaces=false,                
    showstringspaces=false,
    showtabs=false,                  
    tabsize=2
}

\title{On the Effect of Pre-training for Transformer in Different Modality on Offline Reinforcement Learning}

\author{
  Shiro Takagi
\thanks{
  \href{https://t46.github.io/}{https://t46.github.io/}
  } \hspace*{0.01em}
  \thanks{
  Code is available at 
  \href{https://github.com/t46/pre-training-different-modality-offline-rl}{https://github.com/t46/pre-training-different-modality-offline-rl}
  } \hspace*{0.01em}
  \thanks{
  LaTeX source is available at 
  \href{https://github.com/t46/paper-pre-training-different-modality-offline-rl}{https://github.com/t46/paper-pre-training-different-modality-offline-rl}
  } \\
  Independent Researcher \\
  \texttt{takagi4646@gmail.com}
}

\begin{document}
\doparttoc 
\faketableofcontents 


\maketitle

\begin{abstract}
  We empirically investigate how pre-training on data of different modalities, such as language and vision, affects fine-tuning of Transformer-based models to Mujoco offline reinforcement learning tasks. Analysis of the internal representation reveals that the pre-trained Transformers acquire largely different representations before and after pre-training, but acquire less information of data in fine-tuning than the randomly initialized one. A closer look at the parameter changes of the pre-trained Transformers reveals that their parameters do not change that much and that the bad performance of the model pre-trained with image data could partially come from large gradients and gradient clipping. To study what information the Transformer pre-trained with language data utilizes, we fine-tune this model with no context provided, finding that the model learns efficiently even without context information. Subsequent follow-up analysis supports the hypothesis that pre-training with language data is likely to make the Transformer get context-like information and utilize it to solve the downstream task.
\end{abstract}

\section{Introduction}
\label{section:introduction}

The past few years have witnessed the tremendous success of pre-trained Transformer-based models \cite{vaswani2017attention} on versatile natural language processing (NLP) tasks. This is because pre-training with a large corpus enables Transformer-based models to gain rich universal language representations transferable to vast downstream tasks \cite{devlin2018bert,radford2019language,brown2020language}. Remarkably, recent studies demonstrate that Transformers pre-trained with language data can efficiently solve even non-NLP tasks \cite{lu2021pretrained,noorbakhsh2021pretrained,li2022pre,huang2022language,reid2022can,tam2022semantic}. Because of its versatility, it has been suggested that Transformers pre-trained with a large corpus could be called a kind of \textit{universal computation engines} \cite{lu2021pretrained}. 

One area worthwhile attempting to study the benefits of pre-trained Transformer is reinforcement learning because the applicability of pre-training to this area still has room to be explored \cite{singh2020parrot,yang2021representation,stooke2021decoupling}. In this vein of research, Reid et al. studied the effect of pre-training of Transformers on offline reinforcement learning (offline RL) tasks \cite{reid2022can}. Their study demonstrates a surprising contrast that pre-training with image significantly deteriorates performance, while that with language data does not negatively affect downstream tasks, rather it even improves performance for some datasets. 

However, it is not yet known how the presence or absence of pre-training and its content leads to differences in the performance of downstream tasks. What information does/doesn't the model pre-trained on language data (\textit{language-pre-trained model}) leverage to solve downstream tasks, and why does the model pre-trained on image data (\textit{image-pre-trained model}) result in catastrophic performance? In this paper, we approach these questions by analyzing the inside of the Transformer that is fine-tuning to Mujoco \cite{todorov2012mujoco} offline RL tasks. Analyzing the internals of Transformer, we study how randomly initialized, language-pre-trained, and image-pre-trained models acquire different representations, and what the differences could be attributed to. Our contributions are as follows:

\begin{itemize}
    \item We study the internal representation of pre-trained and randomly initialized Transformers and find that the pre-trained models largely change their representation, while they encode less information of data during fine-tuning.
    \item The analysis of parameter change and gradient shows that the pre-trained models do not change parameters much perhaps due to gradient confusion and that the performance of the image-pre-trained model might be partly affected by gradient clipping on the large gradient.
    \item We examine context-dependence by fine-tuning models with giving no context information, finding that the language-pre-trained Transformer efficiently learns even without context and this could be the case by replacing just a single Transformer block with a pre-trained one.
    \item Subsequent analysis of the change in how far away the model process the tokens of input supports the possibility that some contextual equivalent information has been positively transferred from language pre-training.
\end{itemize}
\section{Related Works}
\label{section:related-works}

The study of transferability of neural representation has been one of the most important research topics in machine learning research. A bunch of studies have investigated how pre-training influences the downstream tasks, such as computer vision tasks \cite{yosinski2014transferable,djolonga2021robustness,ding2021analyzing,orhand2021quantification} and NLP tasks \cite{mou2016transferable,liu-etal-2019-linguistic,rogers2020primer}. Recently, several studies have investigated the transferability of Transformers to various NLP tasks since understanding its high transferability is of great interest. Previous research reveals that Transformer representation encodes some syntax \cite{tenney2018what}, semantics \cite{ettinger2020bert}, cross-lingual information \cite{artetxe-etal-2020-cross}, and world knowledge \cite{petroni-etal-2019-language}.
Also, it has been pointed out that Transformers obtains not only language-specific information but also more abstract inter-token dependencies \cite{ri2022pretraining,chiang2021transferability,chi-etal-2020-finding}. This finding suggests that pre-trained Transformers may apply to a wider variety of tasks than just NLP tasks.

In recent years, studies have emerged that apply the Transformer to tasks other than NLP tasks, reporting the versatility of the Transformer even in such multi-modal cases. Such prior works have shown that language-pre-trained Transformers can efficiently learn image classification \cite{lu2021pretrained}, symbolic mathematics \cite{noorbakhsh2021pretrained}, simple arithmetic \cite{brown2020language}, reinforcement learning tasks \cite{li2022pre,huang2022language,tam2022semantic,reid2022can}, and more generally, sequence modeling problems \cite{lu2021pretrained}. Our study is in line with the research that examines the applicability of pre-trained Transformers on reinforcement learning.

Most prior works applying pre-trained Transformer on reinforcement learning explicitly express goals, observations, or actions in a language \cite{yao2020keep,li2022pre,huang2022language,tam2022semantic}. On the other hand, the work of \cite{reid2022can} applies pre-trained Transformers on the offline RL tasks that language is unrelated to (e.g. Mujoco \cite{todorov2012mujoco}) and still reports that pre-training is helpful. This study is interesting because it suggests that language-pre-trained Transformers seem to exploit some similarity between the structure of natural language data and trajectory data, which is more abstract and fundamental than linguistic information. 
Also, exploring the applicability of pre-trained Transformers, which have already been successful, is of great engineering significance since pre-training in offline RL has room for improvement compared to other modalities. Examining the internals of Transformers, we further delve into the findings from this study by Reid et al. \cite{reid2022can} and explores the applicability of Transformers to offline RL.

\section{Background}
\label{section:background}

\subsection{Transformer}
\label{subsection:transformer}
Transformers are sequence-to-sequence models composed of stacked identical blocks, called \textit{Transformer blocks}. Each block consists of self-attention \cite{vaswani2017attention}, multi-layer perceptron, skip-connection \cite{he2016deep}, and layer normalization (layer norm) \cite{ba2016layer}. Denoting the linear projection of input sequence of length $n$ to query, key, values of length $m$ by $Q \in \mathbb{R}^{n \times d_Q}$, $K \in \mathbb{R}^{m \times d_K}$, and $V \in \mathbb{R}^{m \times d_V}$, respectively, the self-attention is $\text{Attention}(Q, K, V) = \text{softmax}\left( \frac{QK^{\top}}{\sqrt{d_K}} \right) V$. We use GPTs \cite{radford2018improving}, and hence the autoregressive language model, where queries head to only keys before their position.

\subsection{Offline Reinforcement Learning and Decision Transformer}
\label{subsection:decision-transformer}
Reinforcement learning is the problem to learn to make optimal decisions in a dynamic environment. For ease of mathematical handling, the environment is usually described as a Markov decision process (MDP). The MDP consists of states $\bm{s} \in \mathcal{S}$, actions $\bm{a} \in \mathcal{A}$, transition probability $P(\bm{s}^{\prime}|\bm{s}, \bm{a})$, and reward function $r(\bm{s}, \bm{a}): \mathcal{S} \times \mathcal{A} \to \mathbb{R}$. The interaction of the agent in the environment is represented as $N$ length trajectory, or sequence of state, action, and reward $(\bm{s}_0, \bm{a}_0, r_0, \bm{s}_1, \bm{a}_1, r_1, ..., \bm{s}_N, \bm{a}_N, r_N)$, where $\bm{s}_t$, $\bm{a}_t$, and $r_t$ are state, action, and reward at time step $t$. The goal is to find an optimal policy $\pi(\bm{a}|\bm{s})$ to maximize the expected cumulative rewards $\mathbb{E}[\sum_{t=0}^Nr_t]$ 
for the trajectory.

Offline RL aims to achieve objectives of reinforcement learning from the trajectory data collected by some policy \cite{levine2020offline}. Because offline RL is purely characterized by trajectory data, several studies have proposed formulating offline RL as a sequence modeling problem \cite{janner2021offline,chen2021decision}. Decision Transformer is a seminal work that attempts to use a causal transformer (in other words, GPT architecture or decoder of BERT \cite{devlin2018bert}) to solve such a sequence modeling problem \cite{chen2021decision}. In Decision Transformer, the input trajectory representation is $\tau = (\hat{R}_0,  \bm{s}_0, \bm{a}_0, \hat{R}_1, \bm{s}_1, \bm{a}_1, ..., \hat{R}_N,  \bm{s}_N, \bm{a}_N)$, where return-to-go $\hat{R}_t = \sum_{i = t}^{N} r_i$ is the cumulative reward from time step $t$. The previous studies \cite{reid2022can,chen2021decision} employ the above problem set up to discuss the applicability of Transformer for offline RL. We also follow these studies and use the setup for our analysis.
\section{Experimental Setup}
\label{subsection:experimental-setup}

Unless otherwise noted, all of our experimental setups, including model configuration, dataset, and hyperparameters follow the previous work \cite{reid2022can} \footnote{We use the following code for experiments: \href{https://github.com/machelreid/can-wikipedia-help-offline-rl}{https://github.com/machelreid/can-wikipedia-help-offline-rl}}. The sanity check that our trained models achieve comparable performance with the previous work is shown in Table \ref{table:sanity-check}. Details are in Appendix \ref{appendix:fine-tuning-result}. 

We compare the model with GPT2 architecture \cite{radford2019language} that is randomly initialized (\textit{randomly initialized model}), pre-trained with language data (\textit{GPT2}), and with image data (\textit{iGPT}) to study the effect of pre-training. Pre-trained models are from the \lstinline{Transformers} library from HuggingFace \cite{wolf-etal-2020-transformers}. Like the previous study, the model code for GPT2 and iGPT are \lstinline{gpt2} and \lstinline{openai/imagegpt-small}.

For offline RL tasks, we use some of Mujoco tasks \cite{todorov2012mujoco} in OpenAI Gym \cite{brockman2016openai} provided by D4RL \cite{fu2020d4rl} \footnote{\href{https://github.com/rail-berkeley/d4rl}{https://github.com/rail-berkeley/d4rl}}, an offline RL dataset library. Specifically, we employ \textit{medium} datasets of \textit{HalfCheetah}, \textit{Walker2d}, and \textit{Hopper} environments. Unless otherwise specified, we show the result for \textit{Hopper-medium}. We put the results for other environments in Appendix \ref{appendix:results-for-other-conditions}.
 
Following the previous study \cite{reid2022can}, we train the models for 40 epochs, each of which consists of 2500 steps. In the following sections, we refer to the 40th epoch checkpoint as the \textit{post-fine-tuning model} and the 0th epoch model (the initial state) as the \textit{pre- fine-tuning model}. The training details, including hyperparameters and the optimizer, are detailed in Appendix \ref{appendix:training-detail}.

\begin{table}[H]
  \caption{Normalized mean return.}
  \label{table:sanity-check}
  \centering
  \scalebox{0.8}{
  \begin{tabular}{llllllll}
    \toprule
    \cmidrule(r){1-8}
    Dataset     & Environment & GPT2 (\cite{reid2022can}) & iGPT (\cite{reid2022can}) & DT (\cite{reid2022can})  & GPT2 (ours) & iGPT (ours) & Random Init (ours) \\
    \midrule
    \multirow{3}{*}{Medium} & Hopper & {$79.1 \pm  1.1$} & {$5.7 \pm  1.5$} & {$67.6$}  & {$79.5 \pm  1.3$} & {$2.6 \pm 0.3$} & {$67.7 \pm 3.0$} \\
    & HalfCheetah     & {$42.8 \pm  0.1$} & {$1.5 \pm  0.1$} & {$42.6$} & {$48.4 \pm 0.0$} & {$1.3 \pm 0.1$} & {$48.6 \pm 0.2$}  \\
    & Walker 2D     & {$78.3 \pm  1.5$} & {$0.4 \pm  0.4$} & {$74.0$} & {$71.2 \pm 1.3$} & {$4.3 \pm 1.9$} & {$71.0 \pm 0.5$}  \\
    \bottomrule
  \end{tabular}
}
\end{table}

\section{Results and Analysis}
\label{section:results-and-analysis}

\subsection{Activation Similarity}
\label{section:activation-similarity}
We first study how the pre-trained and randomly initialized models shape the representation in fine-tuning. To that end, we compare the centered kernel alignment (CKA) \cite{kornblith2019similarity} of each layer's activation between pre and post-fine-tuning, investigating how each layer's representation changes by learning offline RL data \footnote{We use the following code to compute CKA: \href{https://github.com/google-research/google-research/tree/master/representation_similarity}{https://github.com/google-research/google-research/tree/master/representation\_similarity} \cite{kornblith2019similarity}}. The CKA has been used in various studies and has provided numerous insights into the understanding of neural networks \cite{Raghu2020Rapid,wu-etal-2020-similarity,Neyshabur20,raghu2021vision,ramasesh2021anatomy}. In particular, we compute the linear CKA, following previous studies \cite{kornblith2019similarity,nguyen2021do}, for a subset of the dataset. We use the activation of each layer from the last time step of the context for return-to-go, state, and action, respectively. We show the result of state, putting the results for the others in Appendix \ref{appendix:results-for-other-conditions-activation-similarity}. Following the previous study \cite{raghu2021vision}, we compute CKA of activation not only for Transformer block output but for all internal activation of the 
Transformer blocks. The definition of the CKA and the details are in Appendices \ref{appendix:cka} and \ref{appendix:detail-of-experiments-activation-similarity}. 

\begin{figure}[h]
    \centering
    \begin{minipage}[b]{0.32\linewidth}
        \includegraphics[width=\linewidth]{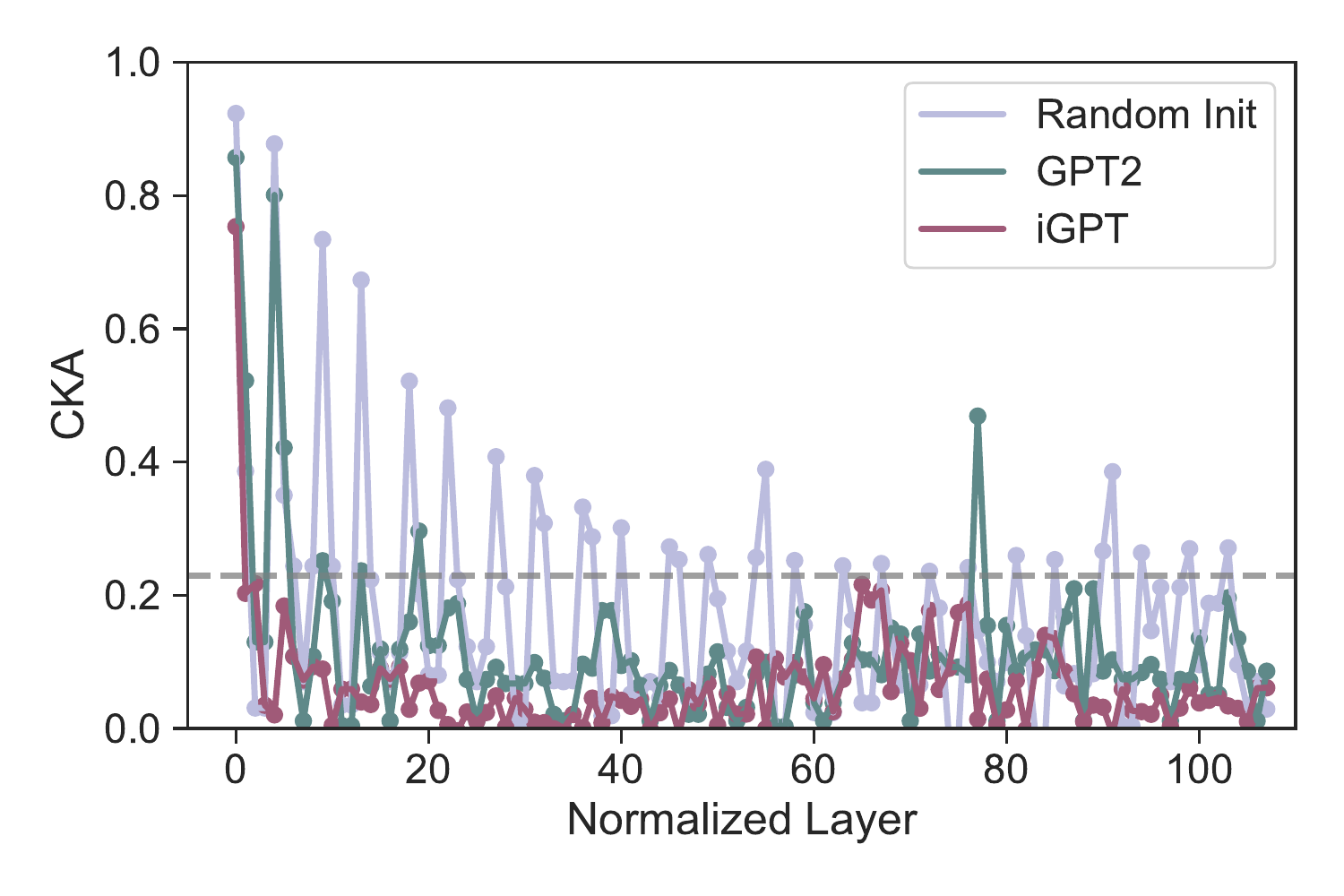}
        \subcaption{Hopper}
    \end{minipage}
    \begin{minipage}[b]{0.32\linewidth}
        \includegraphics[width=\linewidth]{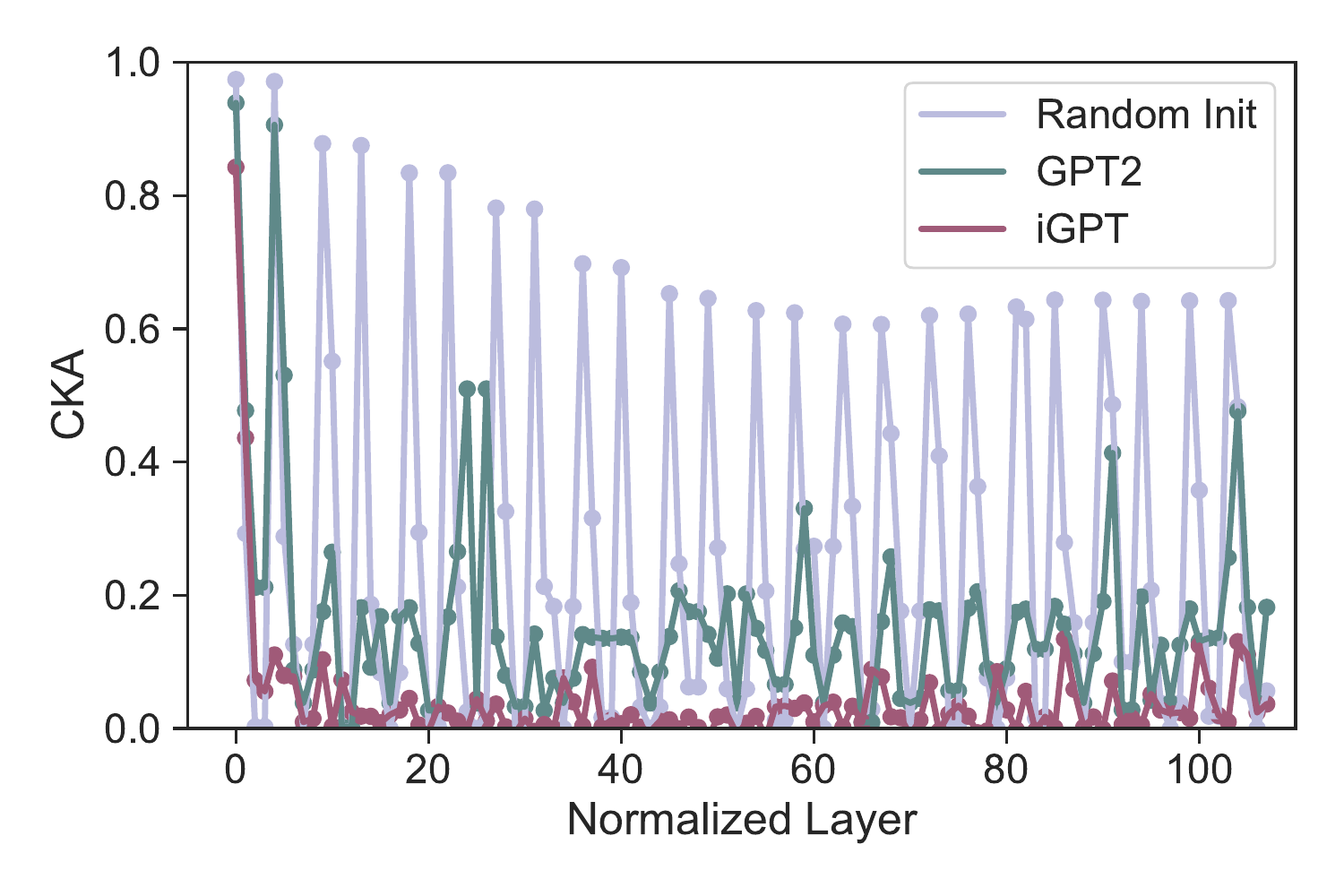}
        \subcaption{HalfCheetah}
    \end{minipage}
    \begin{minipage}[b]{0.32\linewidth}
        \includegraphics[width=\linewidth]{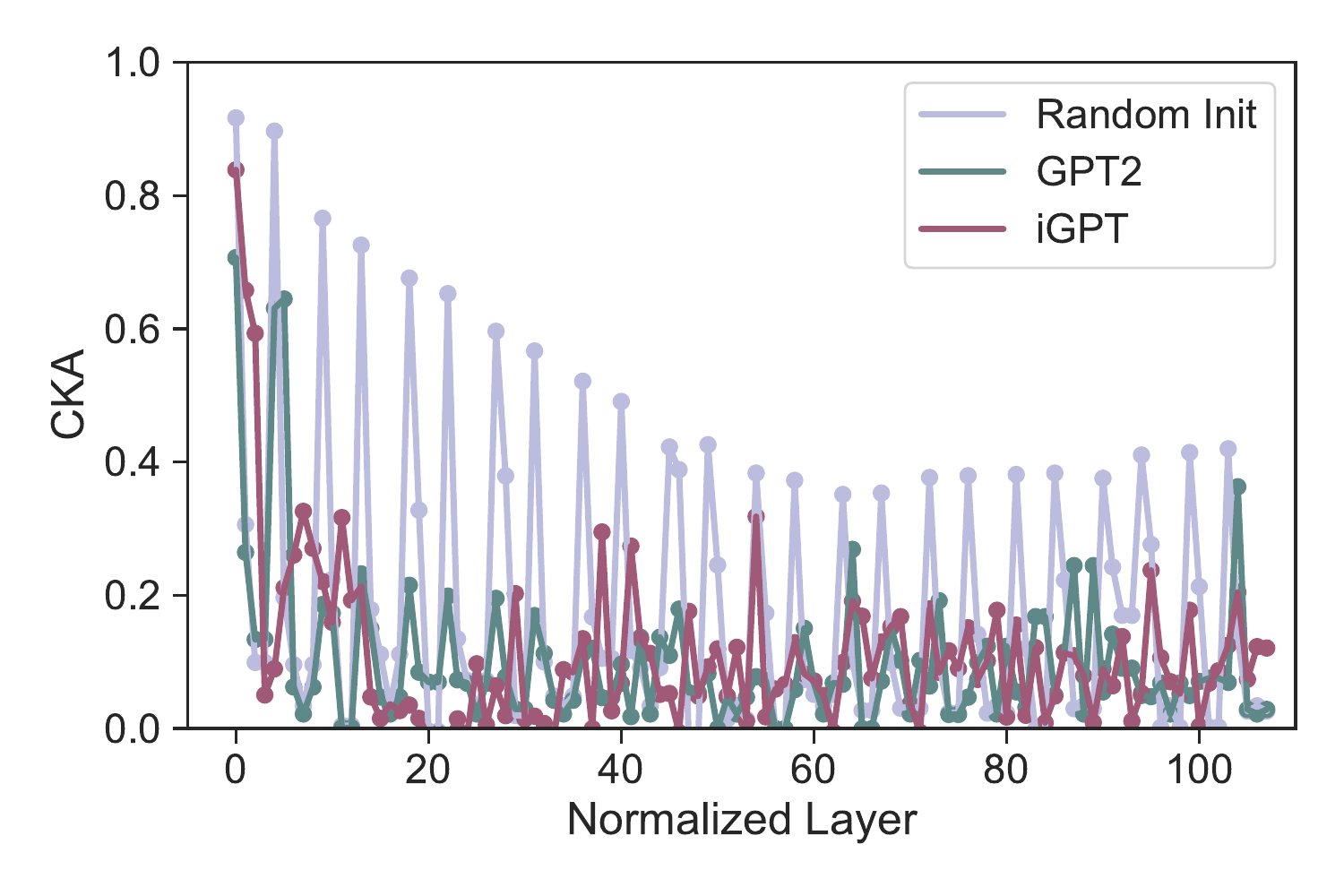}
        \subcaption{Walker2D}
    \end{minipage}
    \caption{CKA similarity of each layer between pre and post-fine-tuning.}
    \label{fig:cka_plot_gpt2_dt}
\end{figure}

Fig. \ref{fig:cka_plot_gpt2_dt} shows the result. The y-axis is each layer's CKA between representation at pre and post-fine-tuning. The x-axis is the normalized layer index: because the number of layers of iGPT (216) is twice the others (108), we average CKA over two adjacent elements starting from the 0th layer for iGPT and align the x-axis for different models for comparison. Higher CKA means higher similarity.

We observe that the CKAs of pre-trained models (GPT2 and iGPT) are uniformly low across layers except for a few layers, indicating that pre-trained models largely change representation. On the other hand, for the randomly initialized model (Random Init), the representations have a higher similarity to the initial state from the middle to the shallow layers. Also, the CKAs of the randomly initialized model seem to be higher than pre-trained models across layers. These observations indicate that the randomly initialized and pre-trained models structure their representations in very different ways. Further support for this conclusion is shown in Appendix \ref{appendix:cka-different-models}.

To identify which layers have relatively higher CKA for the randomly initialized model, we obtain layer names with CKA values above a threshold (dashed line at 0.23 in Fig. \ref{fig:cka_plot_gpt2_dt} (a)). Then, we find that most of them are layer normalization layers (the list of the layer names are put in Appendix \ref{appendix:layer-name-cka}). Therefore, we can speculate that the difference in CKA similarity of layer normalization can be related to the effect of pre-training. This possibility could be supported by previous research \cite{lu2021pretrained} that reports that layer norm parameters are the most important to fine-tune.

In uni-modal cases, where the statistical nature of the input data does not change significantly, previous studies have a consensus that the shallow layer acquires a more universal representation and the deep layer acquires a task-specific representation for a wide range of architectures and tasks \cite{Raghu17,Morcos_nips18,Morcos_iclr18,Raghu19,voita-etal-2019-bottom,kornblith2019similarity,Merchant20,Neyshabur20,ramasesh2021anatomy}. The finding that the language-pre-trained Transformer performs as well as or better than the baseline but largely changes the representation suggests that in multi-modal cases, something different may be happening in forming the representation than in prior studies.

\subsection{Mutual Information Between Hidden Representation and Data}
\label{section:mutual-information-between-hidden-representation-and-input-and-label}
In the previous section, we observe that pre-trained models drastically change their internal representation during fine-tuning, attaining largely different representations. One possible consequence of this observation is that the pre-trained model may have adapted better to the data in the downstream tasks. To explore this possibility, we calculate estimated values of mutual information between hidden representation and data and compare them among different models. Mutual information is a well-known measure of the mutual dependence between the two random variables. Many previous studies have used this metric to investigate the extent to which neural representation encodes input and label information and provided profound insight into how neural networks work \cite{tishby2015deep,shwartz2017opening,hafez2019information,goldfeld2020information,geiger2021information}.

In particular, we calculate the estimated mutual information $\hat{I}(X; T)$ between some layer representation $T$ and input data $X$, and that $\hat{I}(Y; T)$ between $T$ and label $Y$. We employ Mutual Information Neural Estimation (MINE) \cite{pmlr-v80-belghazi18a} to estimate mutual information since this method can estimate mutual information even when random vectors are continuous and have different dimensions \footnote{We use the following code for mutual information estimation: \href{https://github.com/gtegner/mine-pytorch}{https://github.com/gtegner/mine-pytorch}}. Since Decision Transformers are trained to minimize the gap between predicted action $\hat{\bm{a}}_t$ and actual action $\bm{a}_t$, we define the label as $Y = \bm{a}_t$. Unlike non-sequential models, the causal Transformer exploits past context information. Thus, denoting the input token except for $\bm{a}_K$ by $\bm{x}_t$ and the internal activation of Transformer block $l$ of the input by $T_l(\bm{x}_t)$, we compute $\hat{I}(\bm{x}_t; T_l(\bm{s}_K))$ as $\hat{I}(X; T)$ and $\hat{I}(\bm{a}_K; T_l(\bm{x}_t))$ as $\hat{I}(Y; T)$ for all token positions, $t = 0, ..., K$. We calculate these values for the shallow, middle, and deep Transformer blocks, respectively. We show the result of the middle block here and leave the remaining in Appendix \ref{appendix:results-for-other-conditions-mutual-information}. The other details are described in Appendix \ref{appendix:detail-of-experiments-mutual-information}.

\begin{figure}[h]
    \centering
    \begin{minipage}[b]{0.4\linewidth}
    \includegraphics[width=\linewidth]{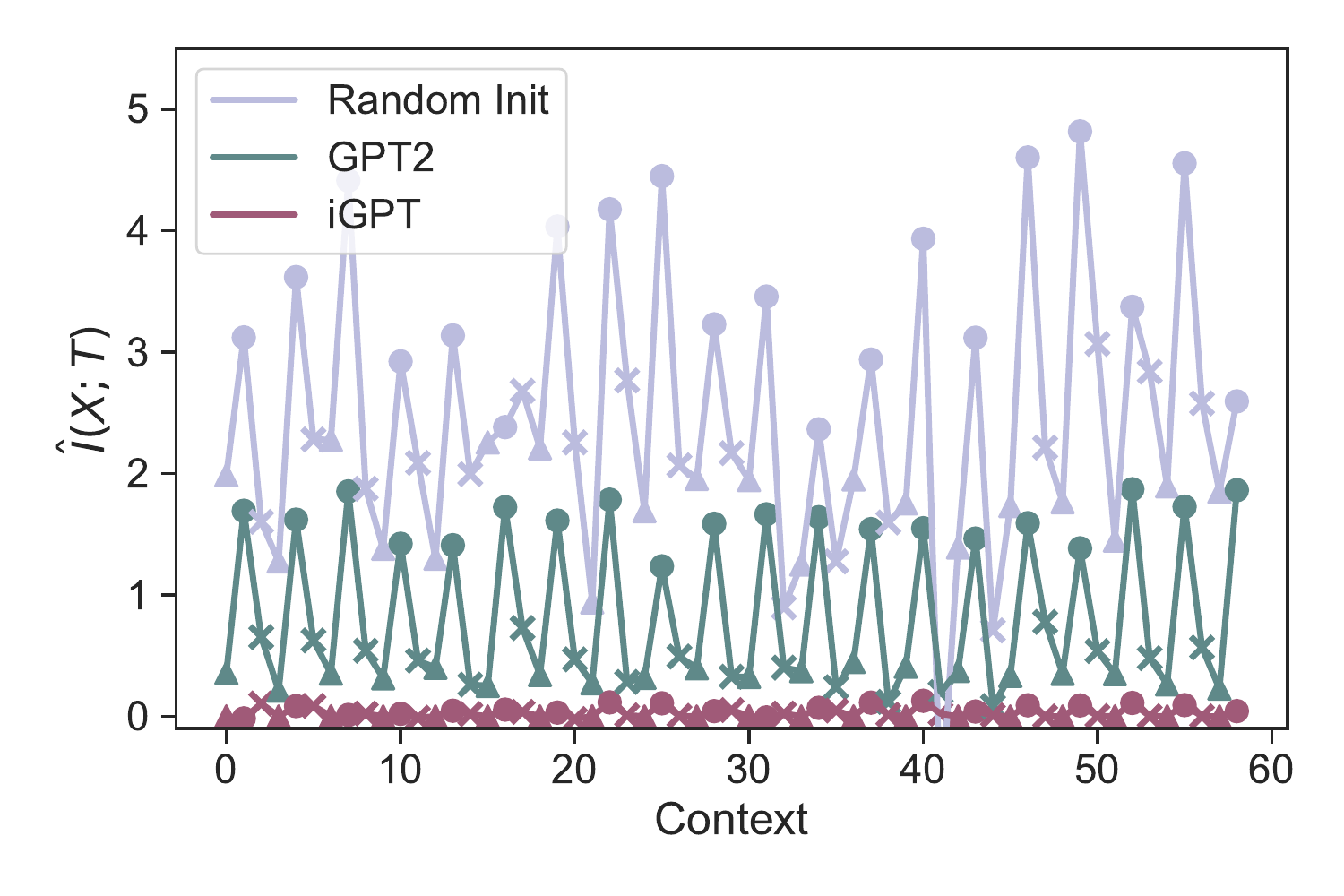}
    \subcaption{$\hat{I}(X; T)$}
    \end{minipage}
    \begin{minipage}[b]{0.4\linewidth}
    \includegraphics[width=\linewidth]{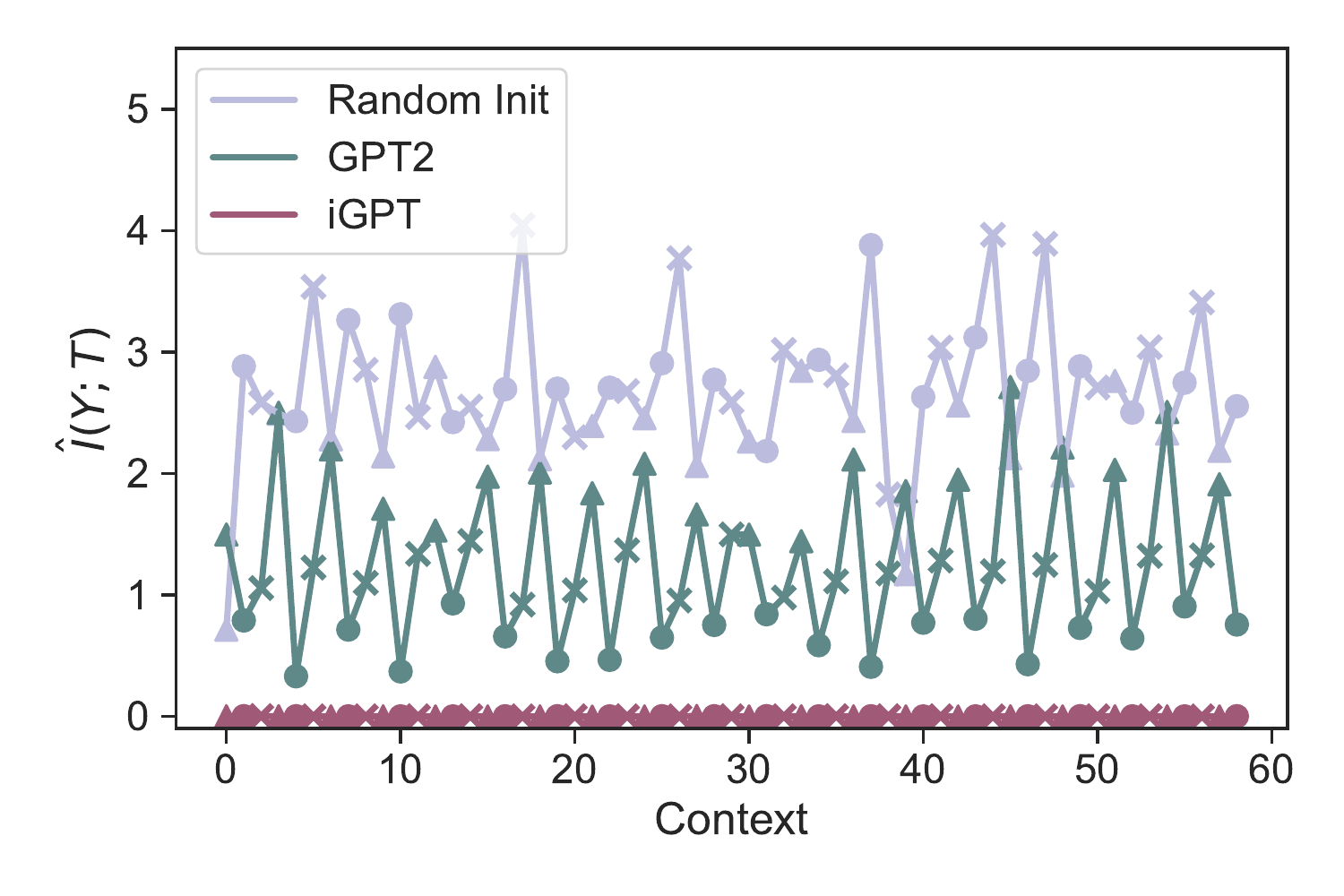}
    \subcaption{$\hat{I}(Y; T)$}
    \end{minipage}
    \caption{Estimated mutual information between data and hidden representation.}
    \label{fig:mutual_information_context}
\end{figure}

Fig. \ref{fig:mutual_information_context} is the result, where Fig. \ref{fig:mutual_information_context} (a) and (b) are results of $\hat{I}(X; T)$ and $\hat{I}(Y; T)$, respectively. The markers \textit{triangle}, \textit{circle}, and \textit{cross} represent return-to-go, state, and action, respectively. The x-axis is the position of tokens in the context and the y-axis is the estimated value of mutual information. 

We observe that the randomly initialized model consistently has higher mutual information both for input and label than pre-trained models. Given the catastrophic performance of iGPT, it is understandable that iGPT representations have almost no information. The result comparing GPT2 to random initialization indicates that the changed representation observed in Section \ref{section:activation-similarity} does not encode more information about the input or the label than the randomly initialized model. That is, the language-pre-trained model does well probably not because it acquires as much or more information about the data as the random initialization.
This result suggests that although the representation has changed, there remains something unchanged, and the language-pre-trained model may leverage it to solve the offline RL task. Note that this might be affected by limitations of mutual information.

We also notice that for label the language-pre-trained model has consistently higher mutual information for a specific token type (return-to-go), while that is not always the case for the randomly initialized model. This result implies that the language-pre-trained model may process the information of a specific token type at each layer for prediction. 

\subsection{Parameter Similarity}
\label{section:parameter-similarity}
In the previous section, we discuss the possibility that the language-pre-trained Transformer exploits some pre-acquired information. To investigate this issue on a finer scale, we turn our eye from representation to parameter. We conduct parameter-level similarity analysis and investigate to what extent parameter change between pre and post-fine-tuning. In concrete, we compute $l2$ distance and cosine similarity of parameters between pre and post-fine-tuning. The details are in Appendix \ref{appendix:detail-of-experiments-parameter-similarity}.

\begin{figure}[ht]
    \centering
        \includegraphics[width=\linewidth]{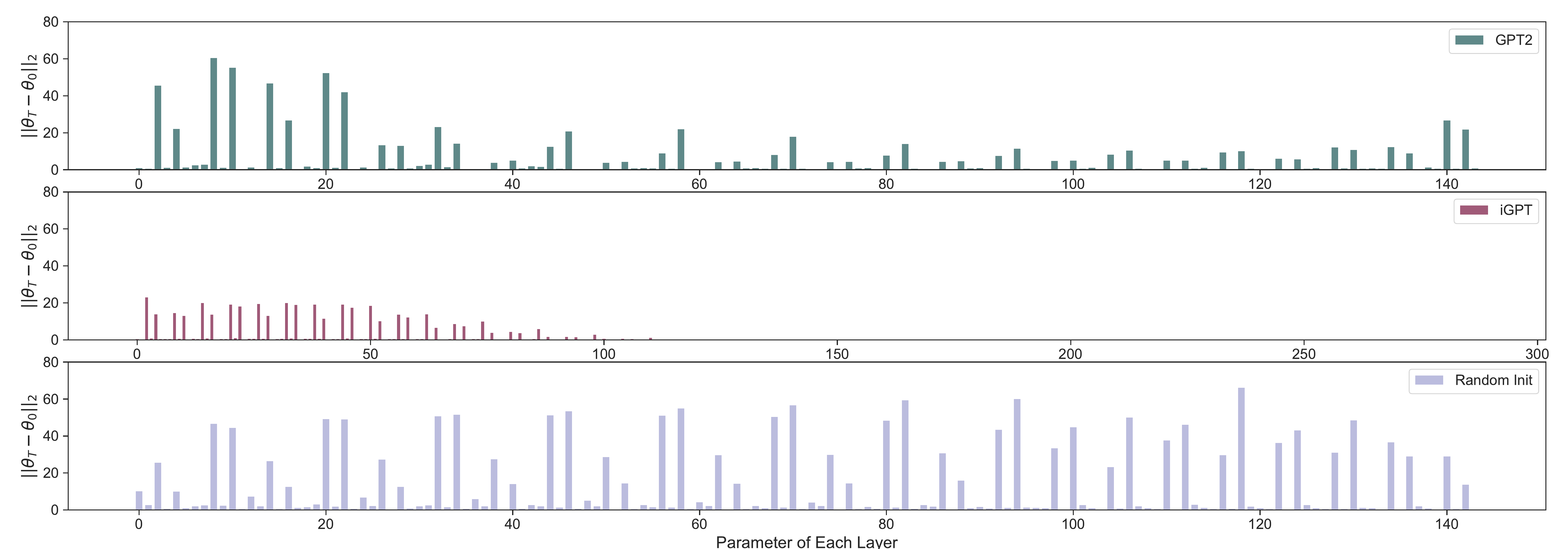}
    \caption{$l2$ distance of each parameter between pre and post-fine-tuning.}
    \label{fig:param-dist}
\end{figure}

\begin{figure}[ht]
    \centering
        \includegraphics[width=\linewidth]{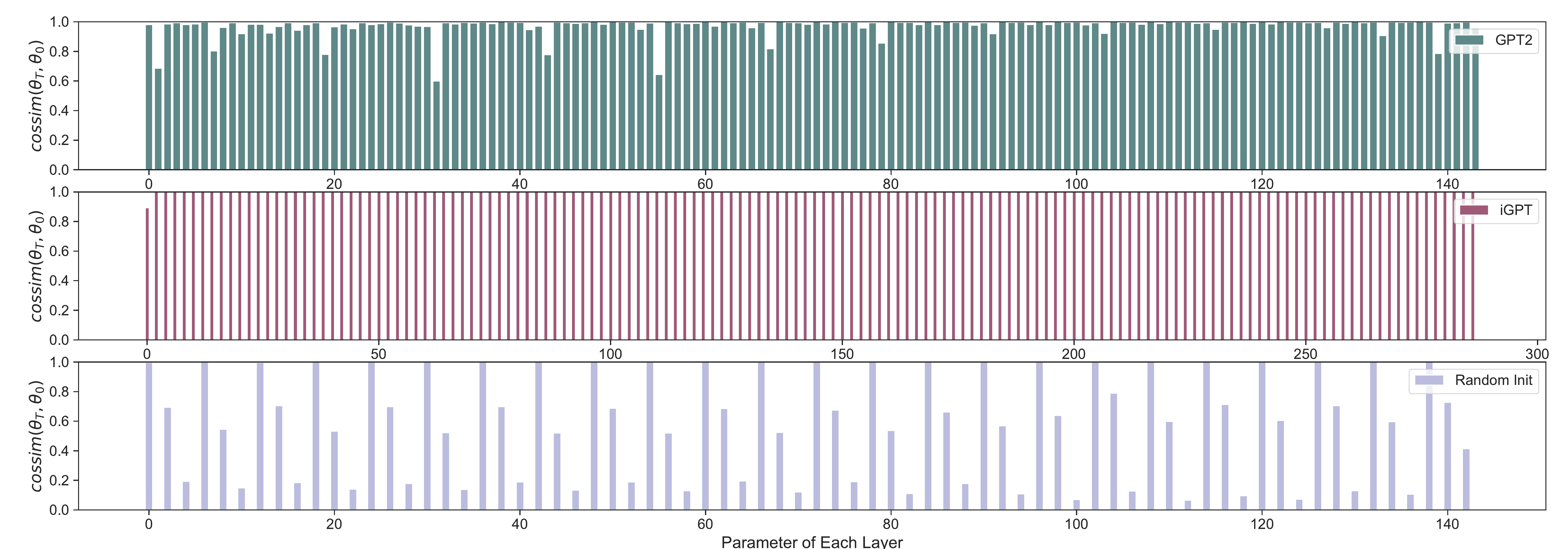}
    \caption{Cosine similarity of each parameter between pre and post-fine-tuning.}
    \label{fig:param-cos}
\end{figure}

The result for $l2$ distance is shown in Fig. \ref{fig:param-dist} and for cosine similarity in Fig. \ref{fig:param-cos}, where the top is the result for GPT2, the middle is for iGPT, and the bottom is for random initialization. The x-axis is the parameter set index, where, from left to right, the ticks correspond to the shallowest to deepest layer parameters. For visibility, we do not show the labels for ticks and put them in Appendix \ref{appendix:parameter-name}. 

Overall, We observe that $l2$ distance is small and cosine similarity is large for pre-trained models, meaning that the parameters of pre-trained models do not change that much. At first glance, this appears to contradict the results of Section \ref{section:activation-similarity}. However, prior research has reported that most of the information in Transformers propagates through skip connections \cite{raghu2021vision}. This means that, as long as the parameters of the shallow layers change, the representation of each layer may change significantly without changing the parameters of each layer. We indeed observe that the shallow layer $l2$ distance of the pre-trained model shows relatively large changes. The existence of unchanged parameters means that knowledge acquired through pre-training could be stored in those parameters.

\subsection{Gradient Analysis}
\label{section:gradient-analysis}
In Section \ref{section:parameter-similarity}, we observe that pre-trained models' parameters don't change that much. We dig into this issue,  investigating why GPT2 and iGPT do not change parameters. To this end, we compare the gradient norm and \textit{gradient confusion} \cite{sankararaman2020impact} at the early phase of the training ($1$st epoch), discussing the training difficulty. When parameter gradients of loss for two different samples are negatively correlated, we say that there is gradient confusion. Empirically, we can measure this by the minimum gradient cosine similarity among all pairs of gradients: when the 
value is close to 0, gradient confusion is low. The previous study showed that training is easier when gradient confusion is low \cite{sankararaman2020impact}. Because gradient clipping is used in the previous work \cite{reid2022can} and this study (the maximum norm is $0.25$), we compute the values for clipped gradients. The details are described in Appendix \ref{appendix:detail-of-experiments-gradient-analysis}.

\begin{figure}[ht]
    \centering
    \begin{minipage}[b]{0.32\linewidth}
        \includegraphics[width=\linewidth]{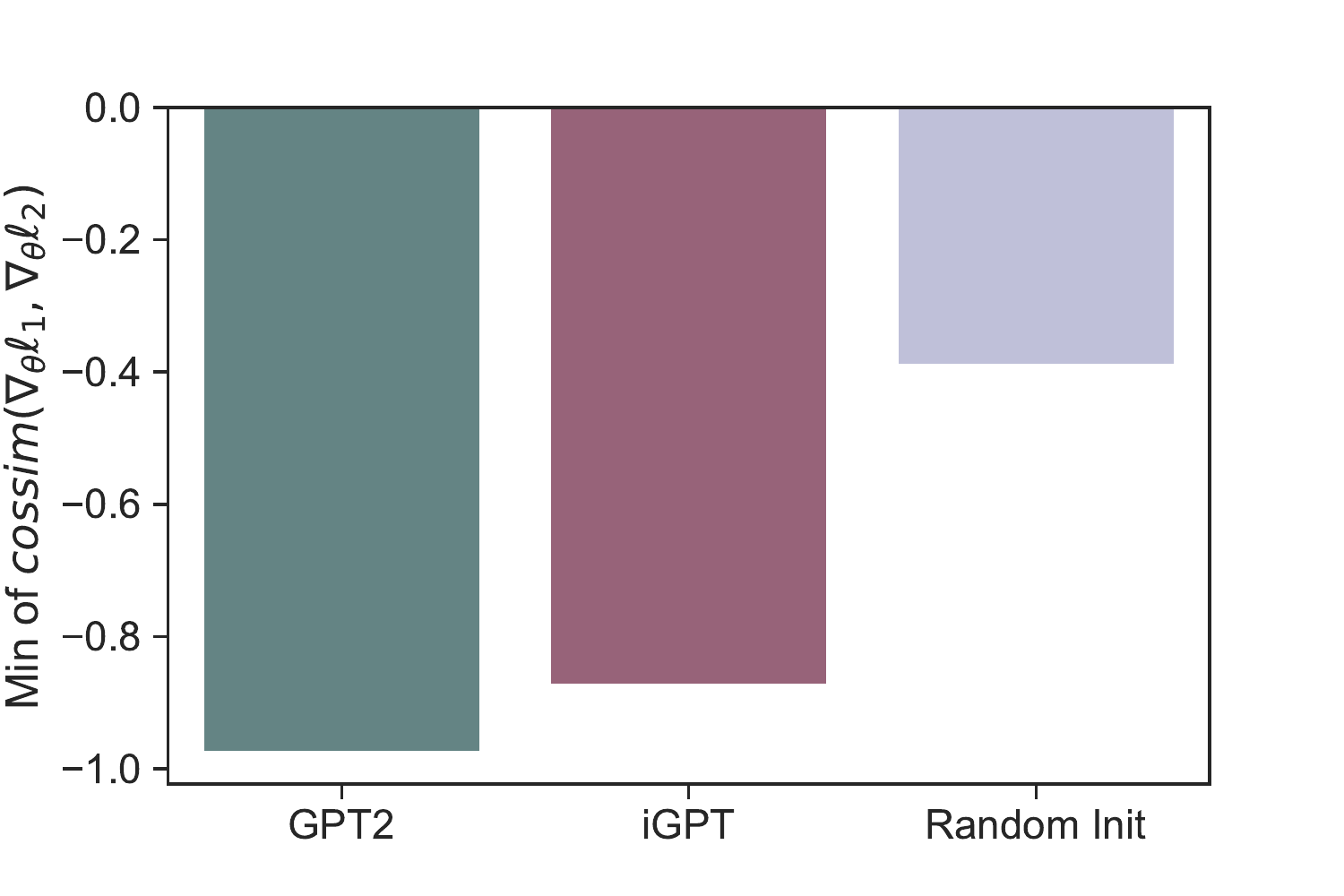}
        \caption{Grad. confusion.}
        \label{fig:grad-confusion}
    \end{minipage}
    \begin{minipage}[b]{0.32\linewidth}
        \includegraphics[width=\linewidth]{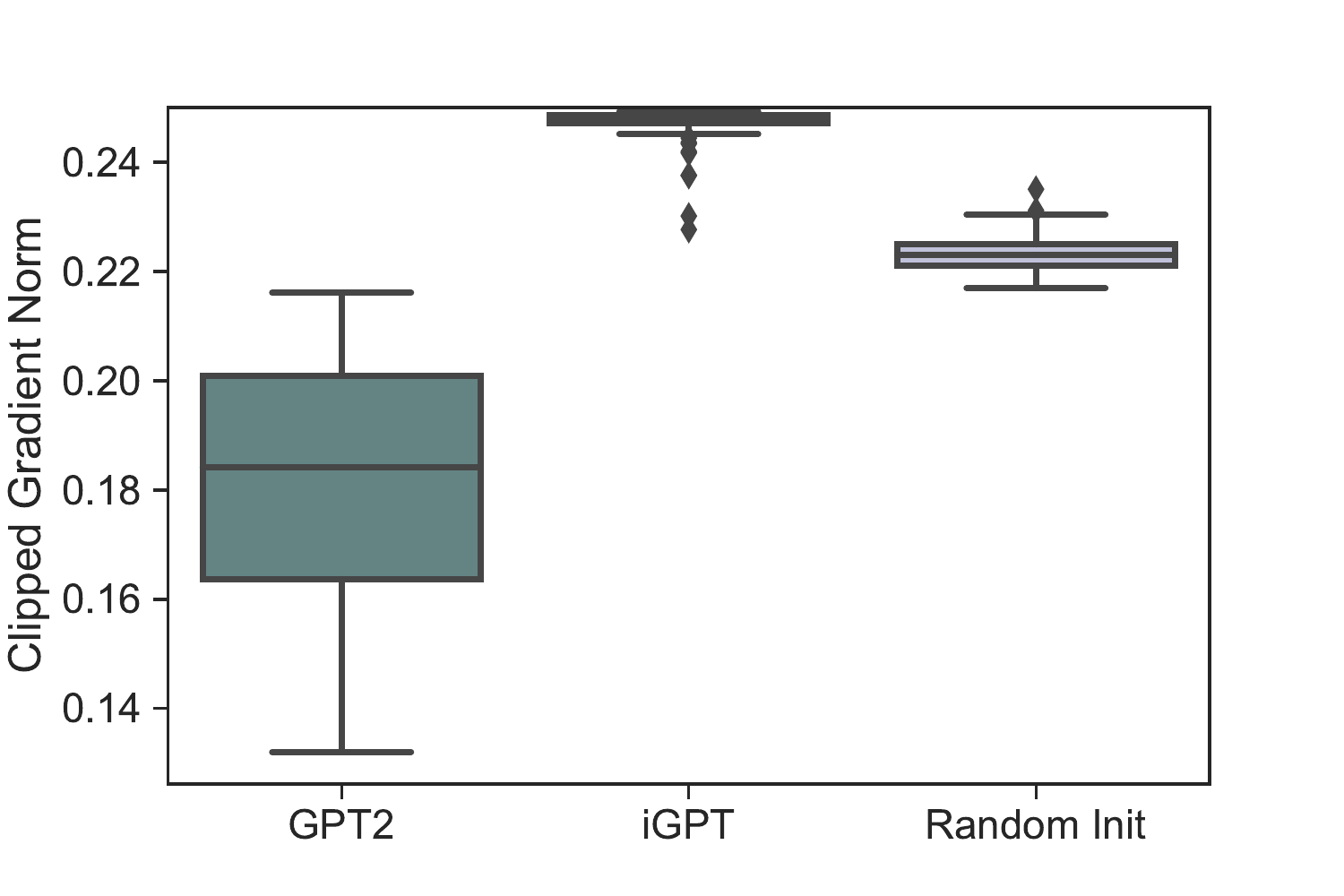}
        \caption{Grad. norm.}
        \label{fig:grad-norm}
    \end{minipage}
    \begin{minipage}[b]{0.32\linewidth}
        \includegraphics[width=\linewidth]{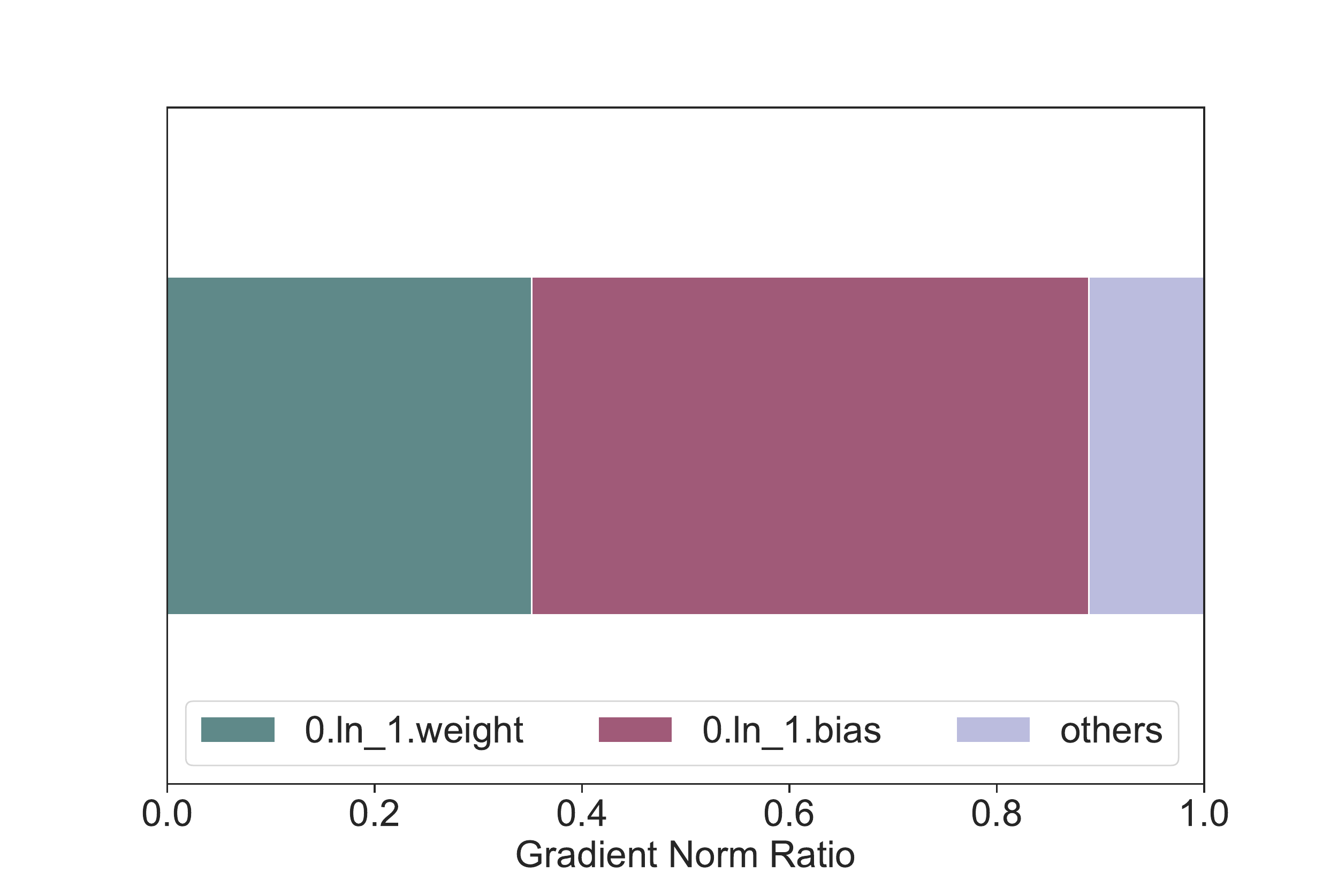}
        \caption{iGPT's grad. norm ratio.}
        \label{fig:grad-norm-param-ratio}
    \end{minipage}
\end{figure}

Fig. \ref{fig:grad-confusion} is the result for the gradient confusion and Fig. \ref{fig:grad-norm} is that for gradient norm. Fig. \ref{fig:grad-confusion} shows that the minimum cosine similarity of iGPT and GPT2 is similarly smaller than the randomly initialized model. This high gradient confusion probably makes pre-trained models harder to train, which explains why they change parameters less than the randomly initialized model. Fig. \ref{fig:grad-norm},
on the other hand, shows that the gradient norm of iGPT is concentrated around the clipping threshold (0.25), while that of the others is below the bound. That is, the differences in the magnitude of the gradient are collapsed for iGPT, reducing the informational value of the gradients. 

To identify which parameter of iGPT dominates the gradient norm, we compare the gradient norm of each parameter set. We calculate the gradient norm ratio of weights and biases of all parameters in all Transformer blocks. The result is shown in Fig. \ref{fig:grad-norm-param-ratio}: for visibility, we show the parameters of the first layer norm module (weights \lstinline{0.ln_1.weight} and biases \lstinline{0.ln_1.bias}) and lump the remaining results together as \lstinline{others}. The full result is shown in Appendix \ref{appendix:gradient-norm-for-each-parameter}. We observe that weights and biases of the first layer norm module dominate the norm. This result implies that the difficulty of training iGPT might partially come from the use of gradient clipping on large gradients dominated by few parameters. A complementary analysis is described in Appendix \ref{appendix:gradient-clipping}.

\subsection{Fine-Tuning with No Context Information}
\label{section:dependency-on-context-informaiton}
In the sections up to this point, we find the possibility that the language-pre-trained model may exploit some pre-acquired information. Then what information does the language-pre-trained model exploit to solve the tasks? One possibility is the ability to handle context because Transformer is skilled at handling the relations of tokens, which enables it to perform well in a variety of 
NLP tasks \cite{tenney2018what}. Therefore, we train the language-pre-trained and the randomly initialized model with no access to the context ($K=1$) to dig into this possibility. If a model learns efficiently without context, the model is already likely to have context-equivalent information. Details of experiments are in Appendix \ref{appendix:detail-of-experiments-dependency-on-context-informaiton}.

\begin{figure}[h]
    \centering
    \begin{minipage}[b]{0.32\linewidth}
        \includegraphics[width=\linewidth]{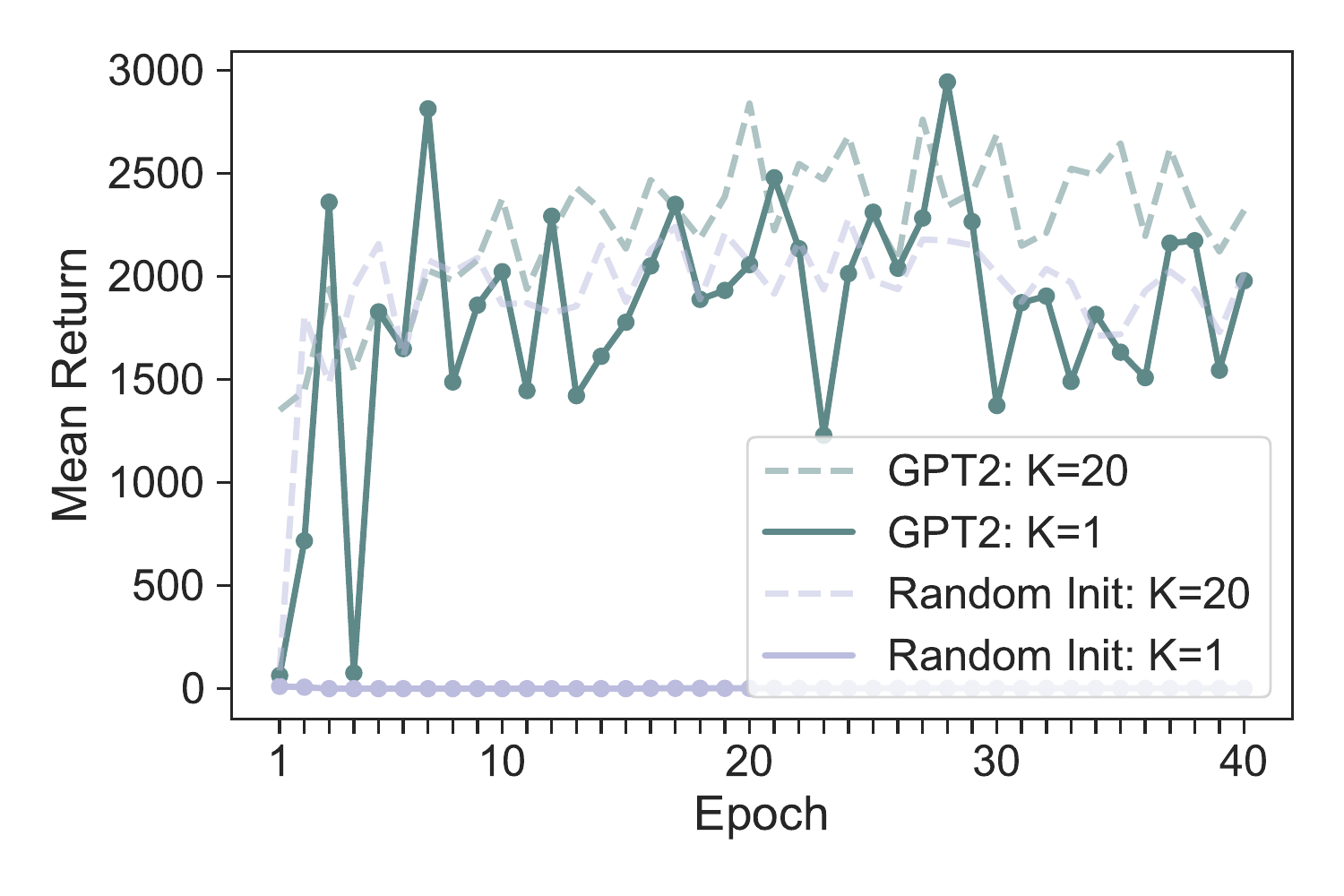}
        \subcaption{Hopper}
    \end{minipage}
    \begin{minipage}[b]{0.32\linewidth}
        \includegraphics[width=\linewidth]{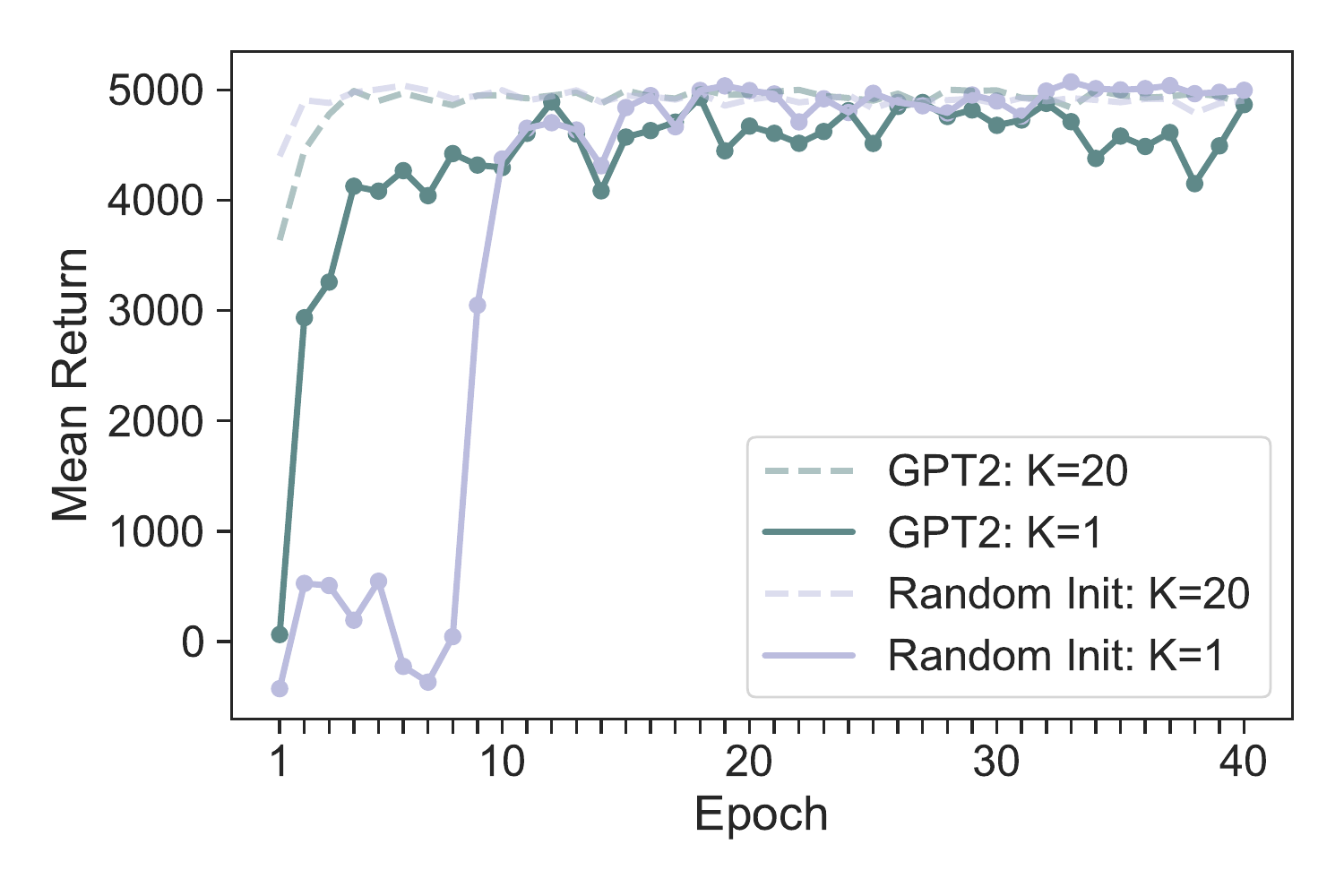}
        \subcaption{HalfCheetah}
    \end{minipage}
    \begin{minipage}[b]{0.32\linewidth}
        \includegraphics[width=\linewidth]{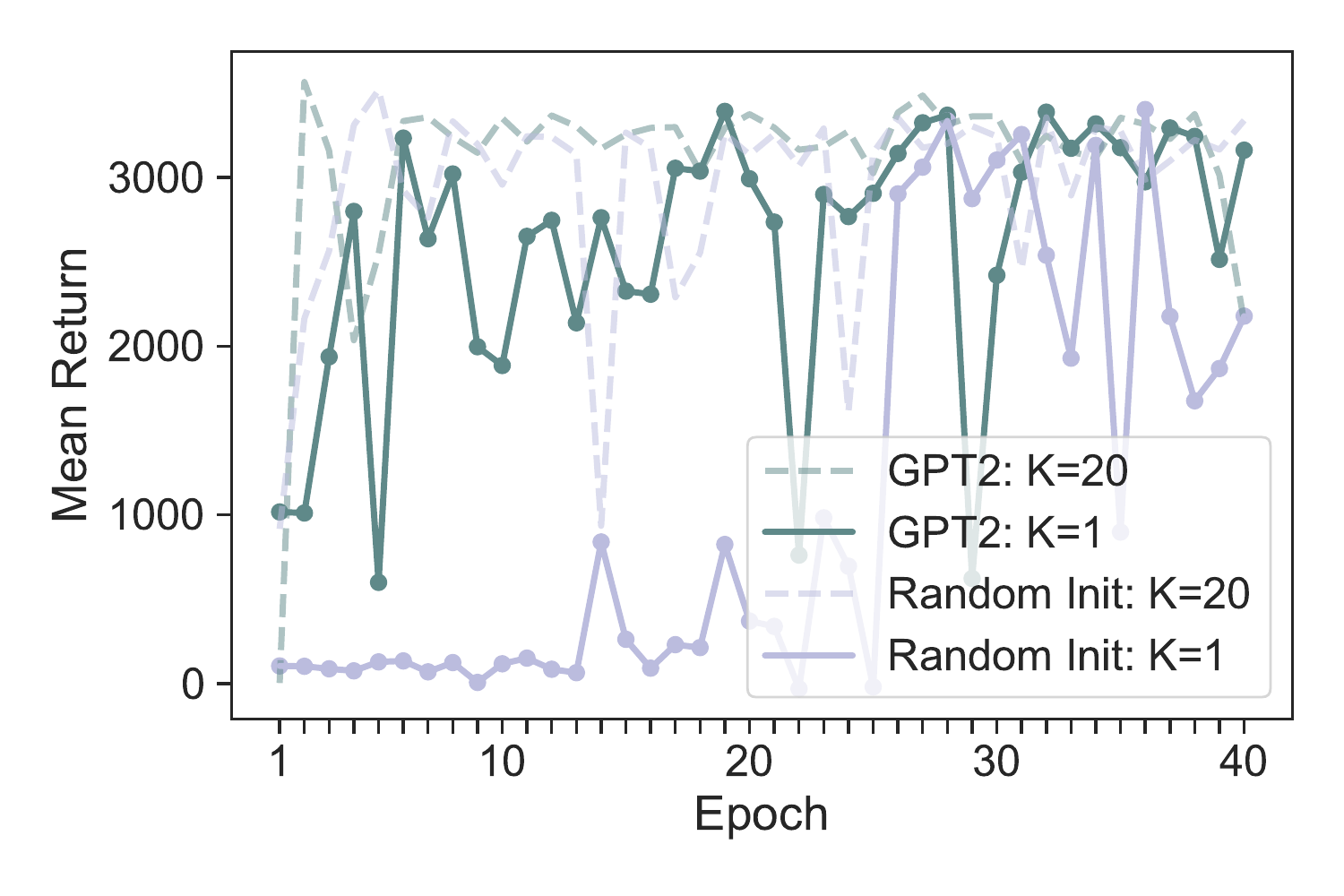}
        \subcaption{Walker2D}
    \end{minipage}
    \caption{Mean return throughout fine-tuning when access to the context information is prohibited.}
    \label{fig:return-mean}
\end{figure}

\begin{wraptable}{r}{0.45\textwidth}
  \caption{Normalized return of $K = 1$.}
  \label{table:k=1}
  \centering
  \scalebox{0.8}{
  \begin{tabular}{llll}
    \toprule
    \cmidrule(r){1-4}
    Dataset     & Environment & GPT2 & Random Init \\
    \midrule
    \multirow{3}{*}{Medium} & Hopper & {$83.73$} & {$-0.24$} \\
    & HalfCheetah     & {$47.7$} & {$49.0$} \\
    & Walker 2D     & {$69.0$} & {$69.1$} \\
    \bottomrule
  \end{tabular}
  }
\end{wraptable}

The mean return throughout training time is shown in Fig. \ref{fig:return-mean} and the normalized scores of return 
\footnote{
Following the previous studies \cite{fu2020d4rl,chen2021decision,reid2022can}, we report normalized scores: $100 \times \frac{\text{score} - \text{random score}}{\text{expert score} - \text{random score}}$.
} are shown in Table \ref{table:k=1}. The solid lines in Fig. \ref{fig:return-mean} are results without context and the dashed lines for reference are those with context. From Fig. \ref{fig:return-mean}, we observe that when the context is not provided ($K=1$), the GPT2 reaches the high mean return much faster than the randomly initialized model. Furthermore, Table \ref{table:k=1} reveals that the randomly initialized model fails to learn the trajectory for a dataset (\textit{Hopper}), while the language-pre-trained model consistently achieves a comparable performance with that of the baseline result ($K=20$). That is, we observe that the language-pre-trained Transformer consistently performs more efficiently when no context is provided. This is somewhat surprising, given that previous studies have pointed out that Decision Transformer performs significantly worse with no context \cite{chen2021decision}. The result that the language-pre-trained Transformer learns efficiently as well without context as it does with context suggests that the model may have already acquired some kind of context-like information through pre-training and is utilizing it to solve the task. 

\subsection{More In-Depth Analysis of Context Dependence}
\label{section:internal-analysis-to-see-the-dependence-on-context}

\subsubsection{Replacement by the Pre-Trained Block}
\label{section:replacement}
In the previous sections, we find that the language-pre-trained Transformer exploits context-related information to solve the task. To identify which block contains the useful information, we conduct a block replacement experiment, where we replace a Transformer block of the randomly initialized model with that of the language-pre-trained model and then fine-tune the model without context. If replacement of a block does not help training, the block probably did not learn effective information in pre-training. Because we observe in Section \ref{section:dependency-on-context-informaiton} that the early phase characterizes the difference of the models, we focus on the epochs up to 10th epoch. We explain the other details in Appendix \ref{appendix:detail-of-experiments-replacement}.

\begin{figure}[H]
    \centering
    \begin{minipage}[b]{0.4\linewidth}
        \includegraphics[width=\linewidth]{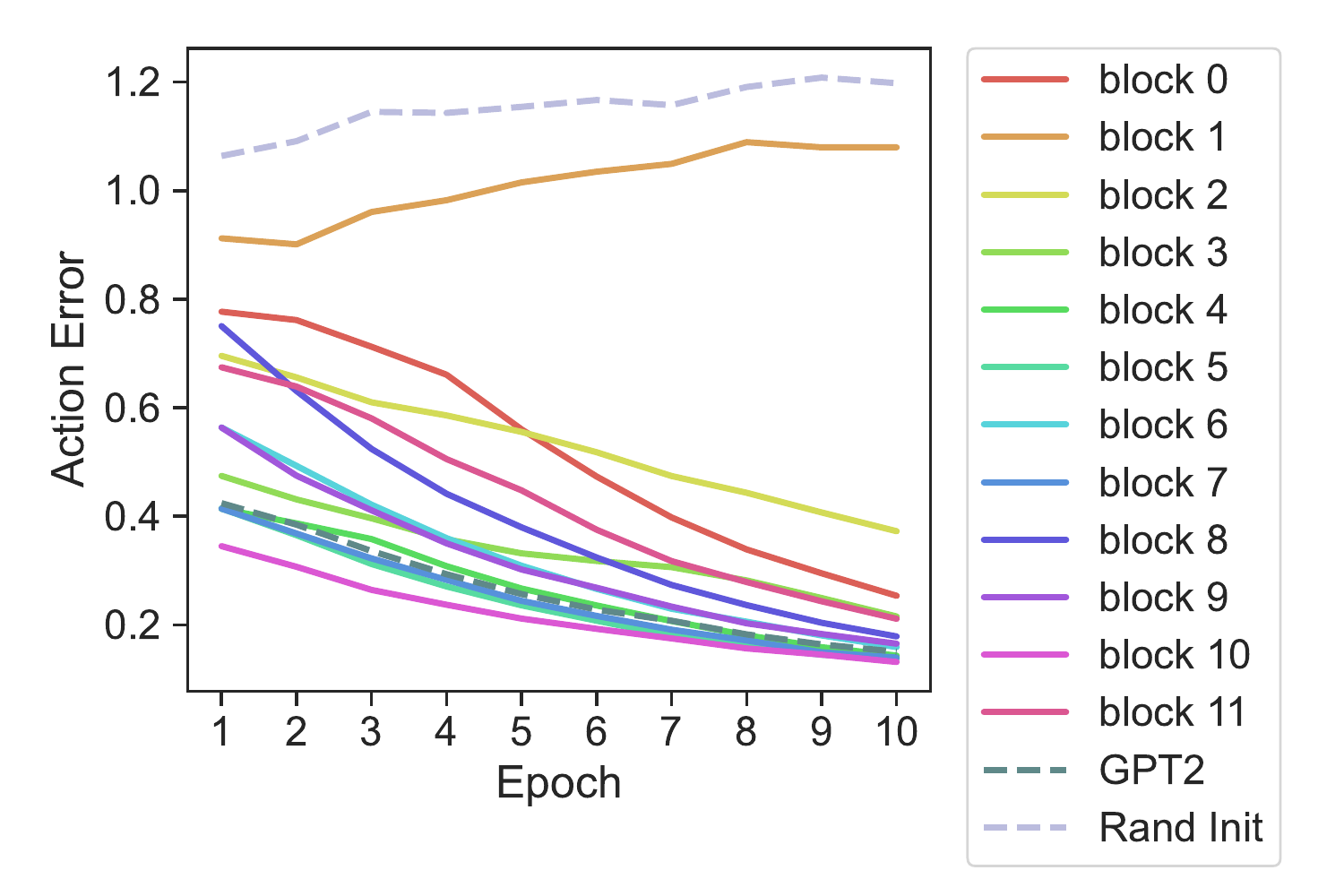}
        \subcaption{Action error}
    \end{minipage}
    \begin{minipage}[b]{0.4\linewidth}
        \includegraphics[width=\linewidth]{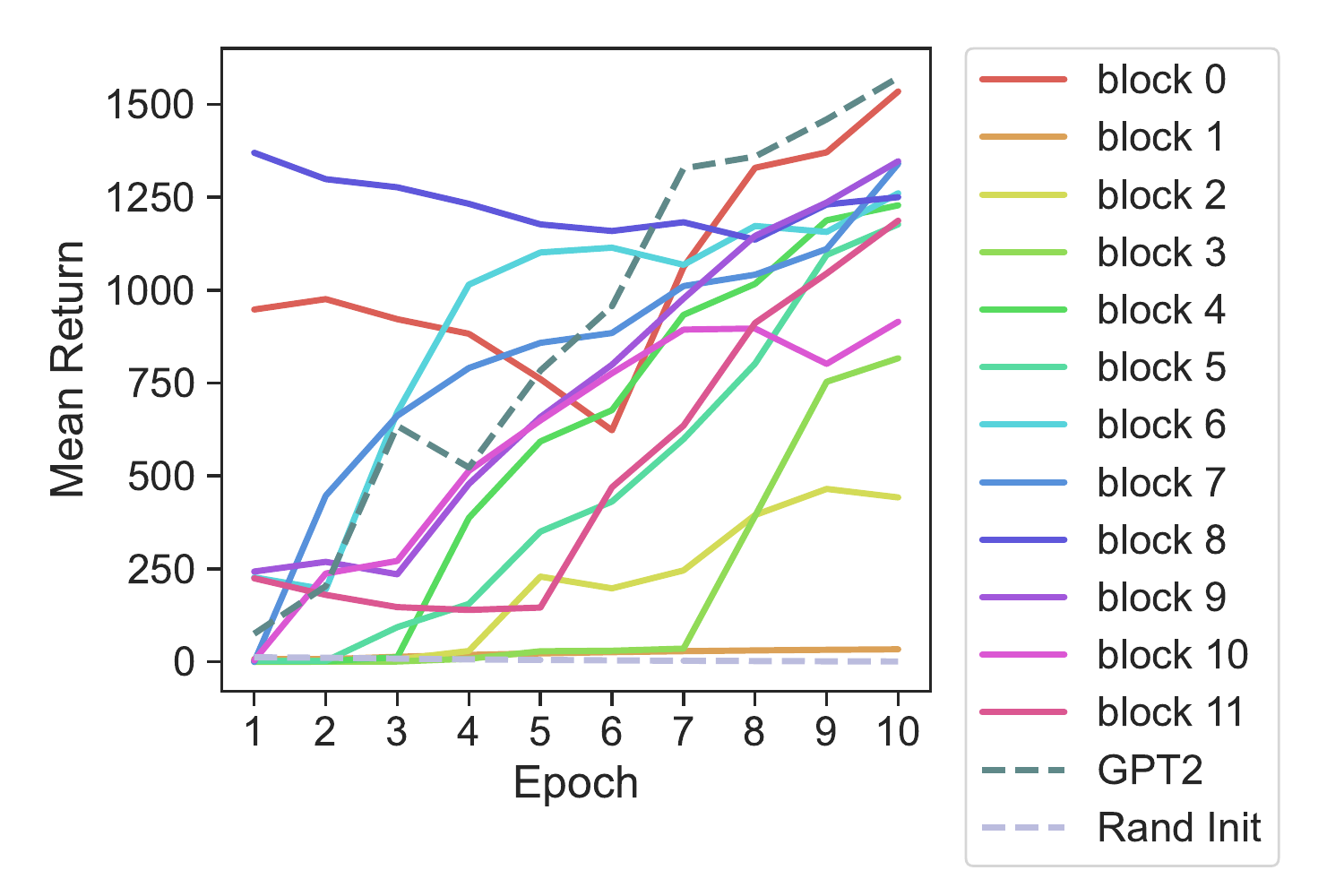}
        \subcaption{Mean return}
    \end{minipage}
    \caption{Learning curve when a single Transformer block is replaced with a pre-trained one.}
    \label{fig:learning_curve}
\end{figure}

The result is shown in Fig. \ref{fig:learning_curve}, where (a) is the prediction error of action and (b) is the mean return 
\footnote{
Note that since the medium dataset we use is corrected by a sub-optimal policy, a better action error does not always result in a better return. We explain this more in detail in Appendix \ref{appendix:note-on-why-better-action-prediction-not-always-result-in-better-return}.
}. 
To get a rough trend, we take the exponential moving average. We observe that only when block 1 is replaced with the pre-trained block, does the action error (Fig. \ref{fig:learning_curve} (a)) and the mean return (Fig. \ref{fig:learning_curve} (b)) behave similarly to the randomly initialized model. Thus, we conclude that all blocks other than block 1 contain context-equivalent useful information to solve tasks. The findings that replacing a single block with the pre-trained one improves learning efficiency and that effective information is distributed across layers alone are surprising enough. This could be related to the dominant role of skip connections for information propagation \cite{raghu2021vision} or the nature of non-monotonic change in the content each layer processes \cite{liu-etal-2019-linguistic} in Transformers. We show the result for Hopper since the trend explained above is most pronounced. For other environments, although the similarity between the random model and block1 is much weaker, the observation that block 1 is not good is consistent. Results for other environments are in Appendix \ref{appendix:results-for-other-conditions-replacement}.

\subsubsection{Attention Distance Analysis}
\label{section:attention-distance}
In previous sections, we find that the information improving learning efficiency is preserved in all blocks only in the varying amount and that is probably context-related. To further confirm that the context-like information is indeed preserved, we investigate how the model processes context by calculating the attention distance \cite{dosovitskiy2020image} of each Transformer block for the result of Section \ref{section:dependency-on-context-informaiton}. Attention distance is the average distance between key and query weighted by attention weight. Previous studies successfully used this measure to reveal how locally Transformers process context \cite{raghu2021vision,dosovitskiy2020image}. Attention distance will elucidate how far away each model processes the tokens of input.

In particular, we compute the gap ($|d_{att}(epoch) - d_{att}(0)|$) between attention distance at the initial state ($d_{att}(0)$) and that at an epoch ($d_{att}(epoch)$) to study to what extent the model preserves the way to utilize context. If the gap is smaller, it means that the attention distance is more preserved. To clarify the difference between the language-pre-trained model and the randomly initialized model, we compare the gap between epoch 0 and 4 since at epoch 4 for all environments the randomly initialized model still produces a low return, while the GPT2 achieves a high return (Fig. \ref{fig:return-mean}). The detail of the setup and results for other epochs are shown in Appendices \ref{appendix:detail-of-experiments-attention-distance} and \ref{appendix:results-for-other-conditions-attention-distance}. 

\begin{figure}[h]
    \centering
    \begin{minipage}[b]{0.32\linewidth}
        \includegraphics[width=\linewidth]{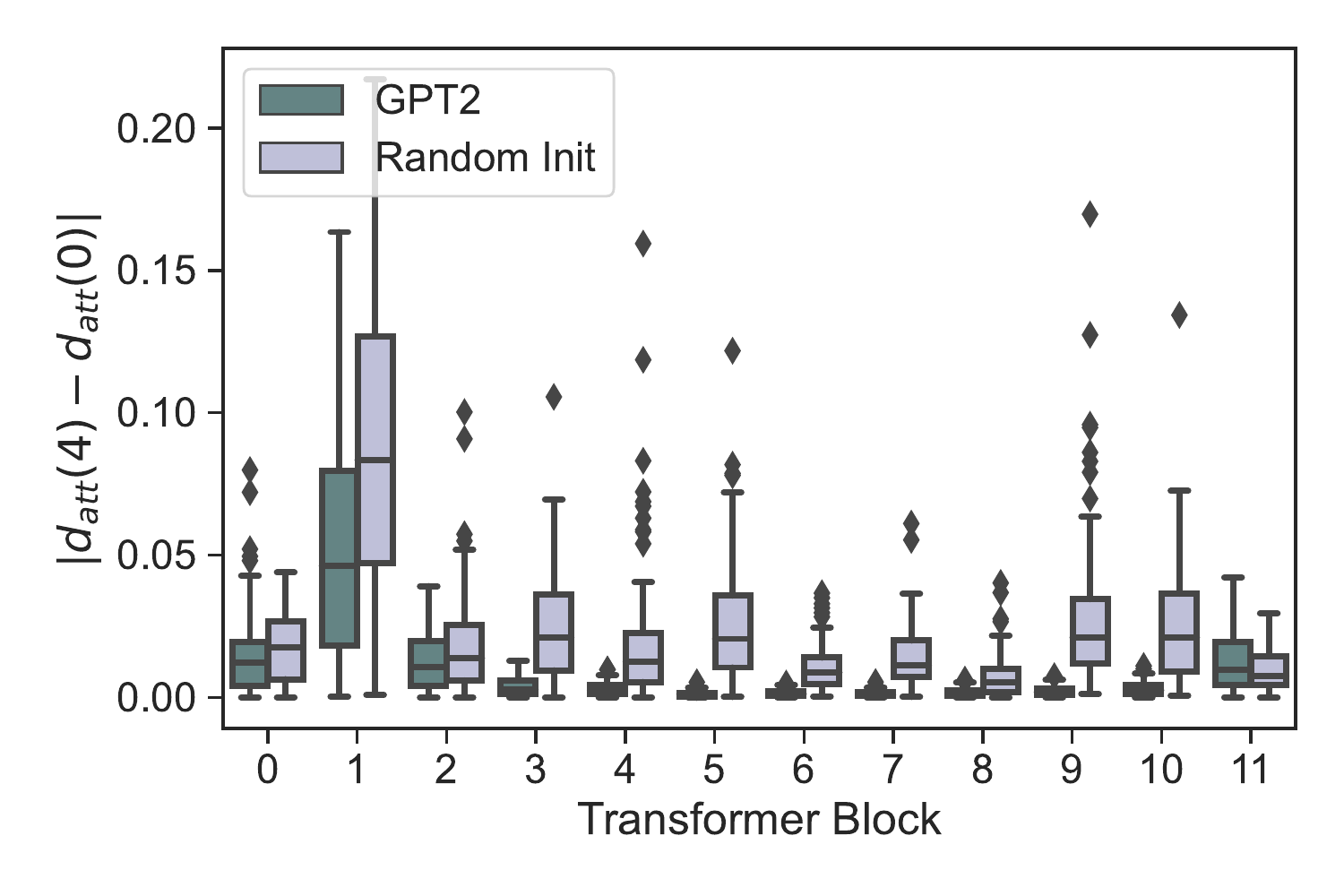}
        \subcaption{Hopper}
    \end{minipage}
    \begin{minipage}[b]{0.32\linewidth}
        \includegraphics[width=\linewidth]{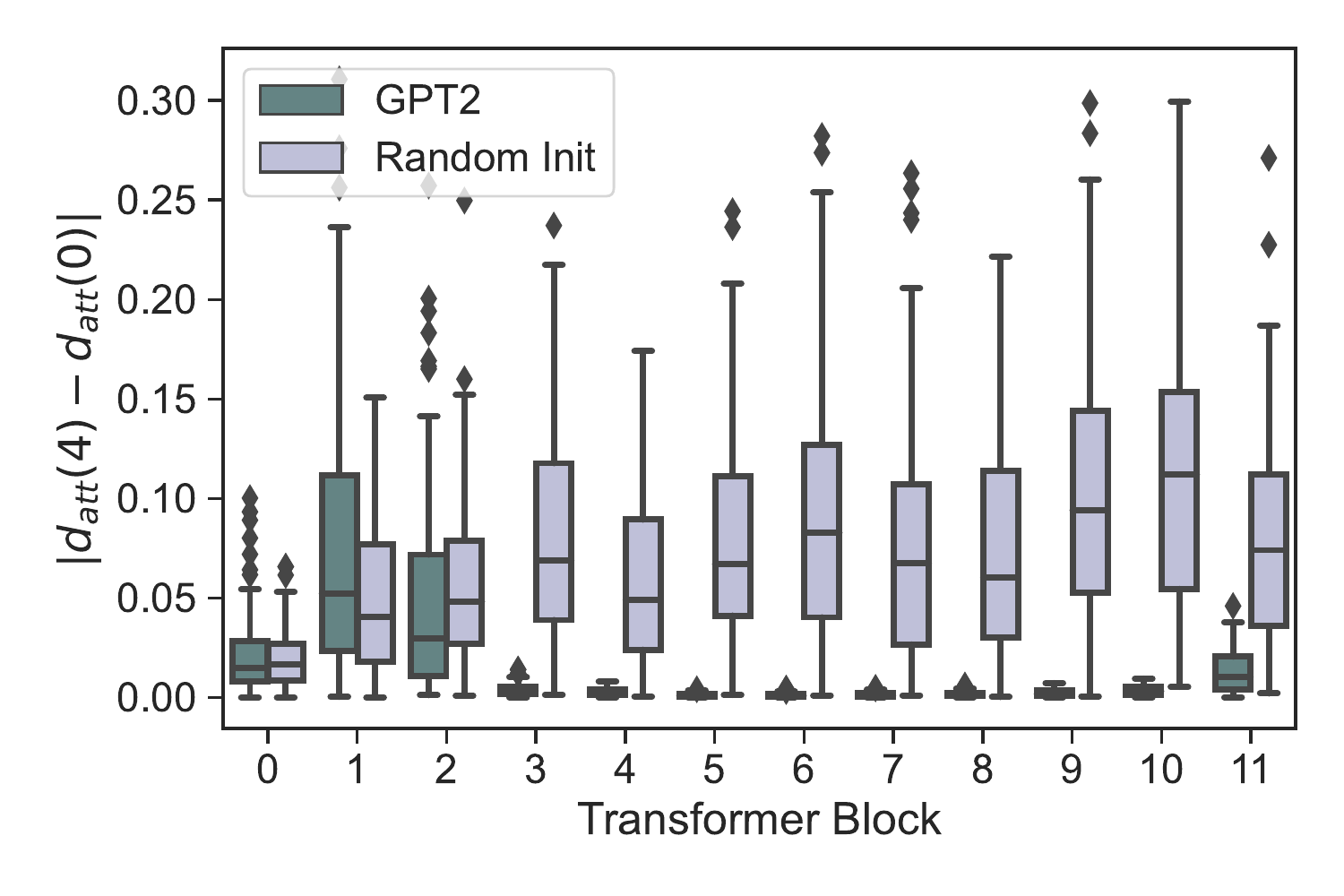}
        \subcaption{HalfCheetah}
    \end{minipage}
    \begin{minipage}[b]{0.32\linewidth}
        \includegraphics[width=\linewidth]{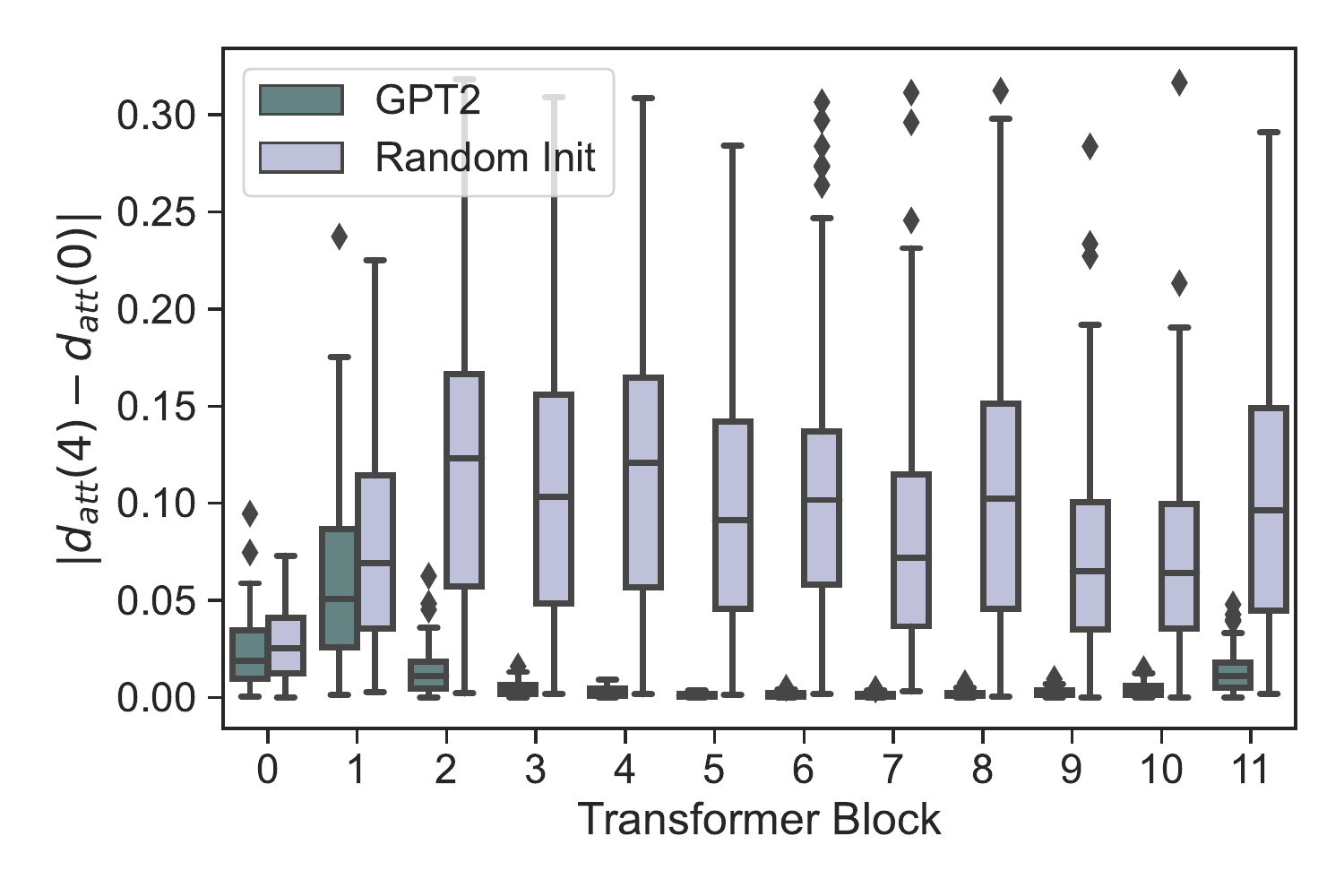}
        \subcaption{Walker2D}
    \end{minipage}
    \caption{Attention distance gap between epoch 0 and epoch 4.}
    \label{fig:attention_distance_before_after_epoch_4}
\end{figure}

Results are shown in Fig. \ref{fig:attention_distance_before_after_epoch_4}, where the x-axis is the Transformer block index and the y-axis is the gap between attention distances. Fig. \ref{fig:attention_distance_before_after_epoch_4} (a), shows that the attention gaps of GPT2 except for block 1 are small, indicating that attention distances are kept from the pre-training in these blocks. Taken together with the finding of Section \ref{section:replacement} that blocks other than block 1 improve learning efficiency, these preserved attention distances are likely to positively affect fine-tuning. This result further supports the hypothesis that language-pre-training brings the model context-like information and the use of this information allows efficient learning.

We also notice the outstanding contrast that gaps of language-pre-trained models are almost zero except for block 1 (and block 2 for HalfCheetah), while those for the randomly initialized model are large. This trend is clearer from middle to deep blocks (3 to 10) for all environments. Turning an eye to Fig. \ref{fig:learning_curve} (a), we notice that the action error of these blocks decreases faster than the other block (0 to 2 and 11). Thus, we conclude that the pre-trained context-like information could help the training by enabling the model to predict action more accurately.

We summarize findings of Section \ref{section:results-and-analysis} with particular attention to the implication for performance:
\begin{itemize}
    \item 5.1: Re-using representation is not the cause of the good performance of GPT2.
    \item 5.2: Fitting to data better is not the cause of the good performance of GPT2.
    \item 5.3: The good performance of GPT2 might come from some unchanged parameters.
    \item 5.4: Gradient clipping and large gradient might be a cause of the bad performance of iGPT.
    \item 5.5: GPT2 can learn efficiently even without the context provided.
    \item 5.6.1: Even a single pre-trained Transformer block makes training more efficient.
    \item 5.6.2: A cause of the good performance of GPT2 is a pre-acquired way to use context.
\end{itemize}

\section{Discussion}
\label{section:discussion}
\paragraph{Conclusion}
We have examined how pre-training on data of different modalities influences fine-tuning to offline RL. Internal representation analysis shows that pre-trained Transformer largely changes its representation while attaining less knowledge of the downstream task. By the analysis of change in parameters, we find that pre-trained Transformers do not change parameters that much and that the bad performance of the image-pre-trained model might partially come from gradient clipping on the large gradient dominated by a few parameters. Fine-tuning models with no context, we find that the language-pre-trained model can efficiently solve offline RL tasks even without context. Follow-up analysis supports the hypothesis that language pre-training probably gives the Transformer context-like information and the model exploits it to tackle the offline RL tasks.

\paragraph{Discussion}
The finding in Section \ref{section:attention-distance} of the usefulness of middle blocks is interesting, given that the contextual model's middle layers contain syntactic information \cite{hewitt-manning-2019-structural,goldberg2019assessing} and are most transferable \cite{liu-etal-2019-linguistic}. This implies that the syntactic capability of Transformers could be related to the performance on offline RL. The role of layer norm is also noteworthy as we observe that it may characterize the behavior in fine-tuning (Sections \ref{section:activation-similarity} and \ref{section:gradient-analysis}). Studying the pre-training on other RL tasks is also critical to understanding the benefit of pre-training in RL. Many studies of human evolution and development point to a close relationship between language and behavior \cite{lashley1951,greenfield1991language}. Hence, studying the relationship between these modalities matters to create human-like intelligence that handles language. A more detailed analysis of this relationship is a promising research direction.

\paragraph{Limitation and Impact}
We consider only a random seed and a few datasets from OpenAI Gym for analysis. 
Thus, using diverse data is important and checking if the claim holds with the average results with many more seeds is crucial.
Also, the validity of the metrics we used should be critically examined since different results could be obtained by using more carefully designed ones. Because this work is fundamental research, no immediate serious negative societal impact is expected.

\begin{ack}
We thank DEEPCORE Inc. for providing the computational resource.
\end{ack}

\bibliographystyle{unsrt}
\bibliography{main}

\begin{thebibliography}{10}

\bibitem{vaswani2017attention}
Ashish Vaswani, Noam Shazeer, Niki Parmar, Jakob Uszkoreit, Llion Jones,
  Aidan~N Gomez, \L~ukasz Kaiser, and Illia Polosukhin.
\newblock Attention is all you need.
\newblock In {\em Advances in Neural Information Processing Systems}, 2017.

\bibitem{devlin2018bert}
Jacob Devlin, Ming-Wei Chang, Kenton Lee, and Kristina Toutanova.
\newblock Bert: Pre-training of deep bidirectional transformers for language
  understanding.
\newblock In {\em Proceedings of the 2019 Conference of the North {A}merican
  Chapter of the Association for Computational Linguistics}, 2019.

\bibitem{radford2019language}
Alec Radford, Jeffrey Wu, Rewon Child, David Luan, Dario Amodei, Ilya
  Sutskever, et~al.
\newblock Language models are unsupervised multitask learners.
\newblock {\em OpenAI blog}, 2019.

\bibitem{brown2020language}
Tom Brown, Benjamin Mann, Nick Ryder, Melanie Subbiah, Jared~D Kaplan, Prafulla
  Dhariwal, Arvind Neelakantan, Pranav Shyam, Girish Sastry, Amanda Askell,
  Sandhini Agarwal, Ariel Herbert-Voss, Gretchen Krueger, Tom Henighan, Rewon
  Child, Aditya Ramesh, Daniel Ziegler, Jeffrey Wu, Clemens Winter, Chris
  Hesse, Mark Chen, Eric Sigler, Mateusz Litwin, Scott Gray, Benjamin Chess,
  Jack Clark, Christopher Berner, Sam McCandlish, Alec Radford, Ilya Sutskever,
  and Dario Amodei.
\newblock Language models are few-shot learners.
\newblock In {\em Advances in Neural Information Processing Systems}, 2020.

\bibitem{lu2021pretrained}
Kevin Lu, Aditya Grover, Pieter Abbeel, and Igor Mordatch.
\newblock Pretrained transformers as universal computation engines.
\newblock {\em arXiv preprint arXiv:2103.05247}, 2021.

\bibitem{noorbakhsh2021pretrained}
Kimia Noorbakhsh, Modar Sulaiman, Mahdi Sharifi, Kallol Roy, and Pooyan
  Jamshidi.
\newblock Pretrained language models are symbolic mathematics solvers too!
\newblock {\em arXiv preprint arXiv:2110.03501}, 2021.

\bibitem{li2022pre}
Shuang Li, Xavier Puig, Chris Paxton, Yilun Du, Clinton Wang, Linxi Fan, Tao
  Chen, De{-}An Huang, Ekin Aky{\"{u}}rek, Anima Anandkumar, Jacob Andreas,
  Igor Mordatch, Antonio Torralba, and Yuke Zhu.
\newblock Pre-trained language models for interactive decision-making.
\newblock {\em arXiv preprint arXiv:2202.01771}, 2022.

\bibitem{huang2022language}
Wenlong Huang, Pieter Abbeel, Deepak Pathak, and Igor Mordatch.
\newblock Language models as zero-shot planners: Extracting actionable
  knowledge for embodied agents.
\newblock In {\em Proceedings of the 39th International Conference on Machine
  Learning}, 2022.

\bibitem{reid2022can}
Machel Reid, Yutaro Yamada, and Shixiang~Shane Gu.
\newblock Can wikipedia help offline reinforcement learning?
\newblock {\em arXiv preprint arXiv:2201.12122}, 2022.

\bibitem{tam2022semantic}
Allison~C Tam, Neil~C Rabinowitz, Andrew~K Lampinen, Nicholas~A Roy,
  Stephanie~CY Chan, DJ~Strouse, Jane~X Wang, Andrea Banino, and Felix Hill.
\newblock Semantic exploration from language abstractions and pretrained
  representations.
\newblock {\em arXiv preprint arXiv:2204.05080}, 2022.

\bibitem{singh2020parrot}
Avi Singh, Huihan Liu, Gaoyue Zhou, Albert Yu, Nicholas Rhinehart, and Sergey
  Levine.
\newblock Parrot: Data-driven behavioral priors for reinforcement learning.
\newblock In {\em International Conference on Learning Representations}, 2021.

\bibitem{yang2021representation}
Mengjiao Yang and Ofir Nachum.
\newblock Representation matters: Offline pretraining for sequential decision
  making.
\newblock In {\em Proceedings of the 38th International Conference on Machine
  Learning}, 2021.

\bibitem{stooke2021decoupling}
Adam Stooke, Kimin Lee, Pieter Abbeel, and Michael Laskin.
\newblock Decoupling representation learning from reinforcement learning.
\newblock In {\em Proceedings of the 38th International Conference on Machine
  Learning}, 2021.

\bibitem{todorov2012mujoco}
Emanuel Todorov, Tom Erez, and Yuval Tassa.
\newblock Mujoco: A physics engine for model-based control.
\newblock In {\em 2012 IEEE/RSJ International Conference on Intelligent Robots
  and Systems}, 2012.

\bibitem{yosinski2014transferable}
Jason Yosinski, Jeff Clune, Yoshua Bengio, and Hod Lipson.
\newblock How transferable are features in deep neural networks?
\newblock In {\em Advances in Neural Information Processing Systems}, 2014.

\bibitem{djolonga2021robustness}
Josip Djolonga, Jessica Yung, Michael Tschannen, Rob Romijnders, Lucas Beyer,
  Alexander Kolesnikov, Joan Puigcerver, Matthias Minderer, Alexander D'Amour,
  Dan Moldovan, et~al.
\newblock On robustness and transferability of convolutional neural networks.
\newblock In {\em Proceedings of the IEEE/CVF Conference on Computer Vision and
  Pattern Recognition}, 2021.

\bibitem{ding2021analyzing}
Yifan Ding, Liqiang Wang, and Boqing Gong.
\newblock Analyzing deep neural network's transferability via frechet distance.
\newblock In {\em Proceedings of the IEEE/CVF Winter Conference on Applications
  of Computer Vision}, 2021.

\bibitem{orhand2021quantification}
Romain Orhand, Hiba Khodji, Amarin Hutt, and Anne Jeannin-Girardon.
\newblock Quantification of the transferability of features between deep neural
  networks.
\newblock {\em Procedia Computer Science}, 192:138--147, 2021.

\bibitem{mou2016transferable}
Lili Mou, Zhao Meng, Rui Yan, Ge~Li, Yan Xu, Lu~Zhang, and Zhi Jin.
\newblock How transferable are neural networks in {NLP} applications?
\newblock In {\em Proceedings of the 2016 Conference on Empirical Methods in
  Natural Language Processing}, 2016.

\bibitem{liu-etal-2019-linguistic}
Nelson~F. Liu, Matt Gardner, Yonatan Belinkov, Matthew~E. Peters, and Noah~A.
  Smith.
\newblock Linguistic knowledge and transferability of contextual
  representations.
\newblock In {\em Proceedings of the 2019 Conference of the North {A}merican
  Chapter of the Association for Computational Linguistics}, 2019.

\bibitem{rogers2020primer}
Anna Rogers, Olga Kovaleva, and Anna Rumshisky.
\newblock A primer in bertology: What we know about how bert works.
\newblock {\em Transactions of the Association for Computational Linguistics},
  8:842--866, 2020.

\bibitem{tenney2018what}
Ian Tenney, Patrick Xia, Berlin Chen, Alex Wang, Adam Poliak, R~Thomas McCoy,
  Najoung Kim, Benjamin~Van Durme, Sam Bowman, Dipanjan Das, and Ellie Pavlick.
\newblock What do you learn from context? probing for sentence structure in
  contextualized word representations.
\newblock In {\em International Conference on Learning Representations}, 2019.

\bibitem{ettinger2020bert}
Allyson Ettinger.
\newblock What bert is not: Lessons from a new suite of psycholinguistic
  diagnostics for language models.
\newblock {\em Transactions of the Association for Computational Linguistics},
  8:34--48, 2020.

\bibitem{artetxe-etal-2020-cross}
Mikel Artetxe, Sebastian Ruder, and Dani Yogatama.
\newblock On the cross-lingual transferability of monolingual representations.
\newblock In {\em Proceedings of the 58th Annual Meeting of the Association for
  Computational Linguistics}, July 2020.

\bibitem{petroni-etal-2019-language}
Fabio Petroni, Tim Rockt{\"a}schel, Sebastian Riedel, Patrick Lewis, Anton
  Bakhtin, Yuxiang Wu, and Alexander Miller.
\newblock Language models as knowledge bases?
\newblock In {\em Proceedings of the 2019 Conference on Empirical Methods in
  Natural Language Processing and the 9th International Joint Conference on
  Natural Language Processing (EMNLP-IJCNLP)}, 2019.

\bibitem{ri2022pretraining}
Ryokan Ri and Yoshimasa Tsuruoka.
\newblock Pretraining with artificial language: Studying transferable knowledge
  in language models.
\newblock In {\em Proceedings of the 60th Annual Meeting of the Association for
  Computational Linguistics (Volume 1: Long Papers)}, 2022.

\bibitem{chiang2021transferability}
Cheng-Han Chiang and Hung-yi Lee.
\newblock On the transferability of pre-trained language models: A study from
  artificial datasets.
\newblock In {\em Proceedings of the AAAI Conference on Artificial
  Intelligence}, 2022.

\bibitem{chi-etal-2020-finding}
Ethan~A. Chi, John Hewitt, and Christopher~D. Manning.
\newblock Finding universal grammatical relations in multilingual {BERT}.
\newblock In {\em Proceedings of the 58th Annual Meeting of the Association for
  Computational Linguistics}, 2020.

\bibitem{yao2020keep}
Shunyu Yao, Rohan Rao, Matthew Hausknecht, and Karthik Narasimhan.
\newblock Keep {CALM} and explore: Language models for action generation in
  text-based games.
\newblock In {\em Proceedings of the 2020 Conference on Empirical Methods in
  Natural Language Processing (EMNLP)}, 2020.

\bibitem{he2016deep}
Kaiming He, Xiangyu Zhang, Shaoqing Ren, and Jian Sun.
\newblock Deep residual learning for image recognition.
\newblock In {\em Proceedings of the IEEE Conference on Computer Vision and
  Pattern Recognition}, 2016.

\bibitem{ba2016layer}
Jimmy~Lei Ba, Jamie~Ryan Kiros, and Geoffrey~E Hinton.
\newblock Layer normalization.
\newblock {\em arXiv preprint arXiv:1607.06450}, 2016.

\bibitem{radford2018improving}
Alec Radford, Karthik Narasimhan, Tim Salimans, and Ilya Sutskever.
\newblock Improving language understanding by generative pre-training.
\newblock {\em OpenAI blog}, 2018.

\bibitem{levine2020offline}
Sergey Levine, Aviral Kumar, George Tucker, and Justin Fu.
\newblock Offline reinforcement learning: Tutorial, review, and perspectives on
  open problems.
\newblock {\em arXiv preprint arXiv:2005.01643}, 2020.

\bibitem{janner2021offline}
Michael Janner, Qiyang Li, and Sergey Levine.
\newblock Offline reinforcement learning as one big sequence modeling problem.
\newblock In {\em Advances in Neural Information Processing Systems}, 2021.

\bibitem{chen2021decision}
Lili Chen, Kevin Lu, Aravind Rajeswaran, Kimin Lee, Aditya Grover, Michael
  Laskin, Pieter Abbeel, Aravind Srinivas, and Igor Mordatch.
\newblock Decision transformer: Reinforcement learning via sequence modeling.
\newblock In {\em Advances in Neural Information Processing Systems}, 2021.

\bibitem{wolf-etal-2020-transformers}
Thomas Wolf, Lysandre Debut, Victor Sanh, Julien Chaumond, Clement Delangue,
  Anthony Moi, Pierric Cistac, Tim Rault, Remi Louf, Morgan Funtowicz, Joe
  Davison, Sam Shleifer, Patrick von Platen, Clara Ma, Yacine Jernite, Julien
  Plu, Canwen Xu, Teven Le~Scao, Sylvain Gugger, Mariama Drame, Quentin Lhoest,
  and Alexander Rush.
\newblock Transformers: State-of-the-art natural language processing.
\newblock In {\em Proceedings of the 2020 Conference on Empirical Methods in
  Natural Language Processing: System Demonstrations}, 2020.

\bibitem{brockman2016openai}
Greg Brockman, Vicki Cheung, Ludwig Pettersson, Jonas Schneider, John Schulman,
  Jie Tang, and Wojciech Zaremba.
\newblock Openai gym.
\newblock {\em arXiv preprint arXiv:1606.01540}, 2016.

\bibitem{fu2020d4rl}
Justin Fu, Aviral Kumar, Ofir Nachum, George Tucker, and Sergey Levine.
\newblock D4rl: Datasets for deep data-driven reinforcement learning.
\newblock {\em arXiv preprint arXiv:2004.07219}, 2020.

\bibitem{kornblith2019similarity}
Simon Kornblith, Mohammad Norouzi, Honglak Lee, and Geoffrey Hinton.
\newblock Similarity of neural network representations revisited.
\newblock In {\em Proceedings of the 36th International Conference on Machine
  Learning}, 2019.

\bibitem{Raghu2020Rapid}
Aniruddh Raghu, Maithra Raghu, Samy Bengio, and Oriol Vinyals.
\newblock Rapid learning or feature reuse? towards understanding the
  effectiveness of maml.
\newblock In {\em International Conference on Learning Representations}, 2020.

\bibitem{wu-etal-2020-similarity}
John Wu, Yonatan Belinkov, Hassan Sajjad, Nadir Durrani, Fahim Dalvi, and James
  Glass.
\newblock Similarity analysis of contextual word representation models.
\newblock In {\em Proceedings of the 58th Annual Meeting of the Association for
  Computational Linguistics}, 2020.

\bibitem{Neyshabur20}
Behnam Neyshabur, Hanie Sedghi, and Chiyuan Zhang.
\newblock What is being transferred in transfer learning?
\newblock In {\em Advances in Neural Information Processing Systems}, 2020.

\bibitem{raghu2021vision}
Maithra Raghu, Thomas Unterthiner, Simon Kornblith, Chiyuan Zhang, and Alexey
  Dosovitskiy.
\newblock Do vision transformers see like convolutional neural networks?
\newblock In {\em Advances in Neural Information Processing Systems}, 2021.

\bibitem{ramasesh2021anatomy}
Vinay~Venkatesh Ramasesh, Ethan Dyer, and Maithra Raghu.
\newblock Anatomy of catastrophic forgetting: Hidden representations and task
  semantics.
\newblock In {\em Proceedings of the 38th International Conference on Learning
  Representations}, 2021.

\bibitem{nguyen2021do}
Thao Nguyen, Maithra Raghu, and Simon Kornblith.
\newblock Do wide and deep networks learn the same things? uncovering how
  neural network representations vary with width and depth.
\newblock In {\em International Conference on Learning Representations}, 2021.

\bibitem{Raghu17}
Maithra Raghu, Justin Gilmer, Jason Yosinski, and Jascha Sohl-Dickstein.
\newblock Svcca: Singular vector canonical correlation analysis for deep
  learning dynamics and interpretability.
\newblock In {\em Advances in Neural Information Processing Systems}, 2017.

\bibitem{Morcos_nips18}
Ari~S. Morcos, Maithra Raghu, and Samy Bengio.
\newblock Insights on representational similarity in neural networks with
  canonical correlation.
\newblock In {\em Advances in Neural Information Processing Systems}, 2018.

\bibitem{Morcos_iclr18}
Ari~S. Morcos, David~G.T. Barrett, Neil~C. Rabinowitz, and Matthew Botvinick.
\newblock On the importance of single directions for generalization.
\newblock In {\em International Conference on Learning Representation}, 2018.

\bibitem{Raghu19}
Maithra Raghu, Chiyuan Zhang, Jon Kleinberg, and Samy Bengio.
\newblock Transfusion: Understanding transfer learning for medical imaging.
\newblock In {\em Advances in Neural Information Processing Systems}, 2019.

\bibitem{voita-etal-2019-bottom}
Elena Voita, Rico Sennrich, and Ivan Titov.
\newblock The bottom-up evolution of representations in the transformer: A
  study with machine translation and language modeling objectives.
\newblock In {\em Proceedings of the 2019 Conference on Empirical Methods in
  Natural Language Processing and the 9th International Joint Conference on
  Natural Language Processing (EMNLP-IJCNLP)}, 2019.

\bibitem{Merchant20}
Amil Merchant, Elahe Rahimtoroghi, Ellie Pavlick, and Ian Tenney.
\newblock What happens to {BERT} embeddings during fine-tuning?
\newblock In {\em Proceedings of the Third BlackboxNLP Workshop on Analyzing
  and Interpreting Neural Networks for NLP}, 2020.

\bibitem{tishby2015deep}
Naftali Tishby and Noga Zaslavsky.
\newblock Deep learning and the information bottleneck principle.
\newblock In {\em 2015 IEEE Information Theory Workshop (ITW)}, 2015.

\bibitem{shwartz2017opening}
Ravid Shwartz-Ziv and Naftali Tishby.
\newblock Opening the black box of deep neural networks via information.
\newblock {\em arXiv preprint arXiv:1703.00810}, 2017.

\bibitem{hafez2019information}
Hassan Hafez-Kolahi and Shohreh Kasaei.
\newblock Information bottleneck and its applications in deep learning.
\newblock {\em arXiv preprint arXiv:1904.03743}, 2019.

\bibitem{goldfeld2020information}
Ziv Goldfeld and Yury Polyanskiy.
\newblock The information bottleneck problem and its applications in machine
  learning.
\newblock {\em IEEE Journal on Selected Areas in Information Theory},
  1(1):19--38, 2020.

\bibitem{geiger2021information}
Bernhard~C Geiger.
\newblock On information plane analyses of neural network classifiers--a
  review.
\newblock {\em IEEE Transactions on Neural Networks and Learning Systems},
  2021.

\bibitem{pmlr-v80-belghazi18a}
Mohamed~Ishmael Belghazi, Aristide Baratin, Sai Rajeshwar, Sherjil Ozair,
  Yoshua Bengio, Aaron Courville, and Devon Hjelm.
\newblock Mutual information neural estimation.
\newblock In {\em Proceedings of the 35th International Conference on Machine
  Learning}, 2018.

\bibitem{sankararaman2020impact}
Karthik~Abinav Sankararaman, Soham De, Zheng Xu, W.~Ronny Huang, and Tom
  Goldstein.
\newblock The impact of neural network overparameterization on gradient
  confusion and stochastic gradient descent.
\newblock In {\em Proceedings of the 37th International Conference on Machine
  Learning}, 2020.

\bibitem{dosovitskiy2020image}
Alexey Dosovitskiy, Lucas Beyer, Alexander Kolesnikov, Dirk Weissenborn,
  Xiaohua Zhai, Thomas Unterthiner, Mostafa Dehghani, Matthias Minderer, Georg
  Heigold, Sylvain Gelly, Jakob Uszkoreit, and Neil Houlsby.
\newblock An image is worth 16x16 words: Transformers for image recognition at
  scale.
\newblock In {\em International Conference on Learning Representations}, 2021.

\bibitem{hewitt-manning-2019-structural}
John Hewitt and Christopher~D. Manning.
\newblock A structural probe for finding syntax in word representations.
\newblock In {\em Proceedings of the 2019 Conference of the North {A}merican
  Chapter of the Association for Computational Linguistics: Human Language
  Technologies, Volume 1 (Long and Short Papers)}, 2019.

\bibitem{goldberg2019assessing}
Yoav Goldberg.
\newblock Assessing bert's syntactic abilities.
\newblock {\em arXiv preprint arXiv:1901.05287}, 2019.

\bibitem{lashley1951}
K.~S. Lashley.
\newblock The problem of serial order in behavior.
\newblock {\em Cerebral Mechanisms in Behavior; The Hixon Symposium}, pages
  112–--146, 1951.

\bibitem{greenfield1991language}
Patricia~M Greenfield.
\newblock Language, tools and brain: The ontogeny and phylogeny of
  hierarchically organized sequential behavior.
\newblock {\em Behavioral and Brain Sciences}, 14(4):531--551, 1991.

\bibitem{haarnoja2018soft}
Tuomas Haarnoja, Aurick Zhou, Pieter Abbeel, and Sergey Levine.
\newblock Soft actor-critic: Off-policy maximum entropy deep reinforcement
  learning with a stochastic actor.
\newblock In {\em Proceedings of the 35th International Conference on Machine
  Learning}, 2018.

\bibitem{KingmaB14}
Diederik~P. Kingma and Jimmy Ba.
\newblock Adam: A method for stochastic optimization.
\newblock In {\em International Conference on Learning Representations}, 2015.

\bibitem{gretton2007kernel}
Arthur Gretton, Kenji Fukumizu, Choon Teo, Le~Song, Bernhard Sch\"{o}lkopf, and
  Alex Smola.
\newblock A kernel statistical test of independence.
\newblock In {\em Advances in Neural Information Processing Systems}, 2007.

\bibitem{Paszke19}
Adam Paszke, Sam Gross, Francisco Massa, Adam Lerer, James Bradbury, Gregory
  Chanan, Trevor Killeen, Zeming Lin, Natalia Gimelshein, Luca Antiga, Alban
  Desmaison, Andreas Kopf, Edward Yang, Zachary DeVito, Martin Raison, Alykhan
  Tejani, Sasank Chilamkurthy, Benoit Steiner, Lu~Fang, Junjie Bai, and Soumith
  Chintala.
\newblock Pytorch: An imperative style, high-performance deep learning library.
\newblock In {\em Advances in Neural Information Processing Systems}, 2019.

\end{thebibliography}

\newpage
\section*{Checklist}
\begin{enumerate}

\item For all authors...
\begin{enumerate}
  \item Do the main claims made in the abstract and introduction accurately reflect the paper's contributions and scope?
    \answerYes{}
  \item Did you describe the limitations of your work?
    \answerYes{Section \ref{section:discussion}}
  \item Did you discuss any potential negative societal impacts of your work?
    \answerYes{Section \ref{section:discussion}}
  \item Have you read the ethics review guidelines and ensured that your paper conforms to them?
    \answerYes{}
\end{enumerate}

\item If you are including theoretical results...
\begin{enumerate}
  \item Did you state the full set of assumptions of all theoretical results?
    \answerNA{}
        \item Did you include complete proofs of all theoretical results?
    \answerNA{}
\end{enumerate}

\item If you ran experiments...
\begin{enumerate}
  \item Did you include the code, data, and instructions needed to reproduce the main experimental results (either in the supplemental material or as a URL)?
    \answerYes{\href{https://github.com/t46/pre-training-different-modality-offline-rl}{https://github.com/t46/pre-training-different-modality-offline-rl}}
  \item Did you specify all the training details (e.g., data splits, hyperparameters, how they were chosen)?
    \answerYes{Appendix \ref{appendix:training-detail}}
        \item Did you report error bars (e.g., with respect to the random seed after running experiments multiple times)?
    \answerYes{}
        \item Did you include the total amount of compute and the type of resources used (e.g., type of GPUs, internal cluster, or cloud provider)?
    \answerYes{Appendix \ref{appendix:details-for-computation}}
\end{enumerate}

\item If you are using existing assets (e.g., code, data, models) or curating/releasing new assets...
\begin{enumerate}
  \item If your work uses existing assets, did you cite the creators?
    \answerYes{All existing assets we used are mentioned in footnote or References.}
  \item Did you mention the license of the assets?
    \answerYes{Appendix \ref{appendix:licence}}
  \item Did you include any new assets either in the supplemental material or as a URL?
    \answerNA{}
  \item Did you discuss whether and how consent was obtained from people whose data you're using/curating?
    \answerYes{Appendix \ref{appendix:licence}}
  \item Did you discuss whether the data you are using/curating contains personally identifiable information or offensive content?
    \answerNA{}
\end{enumerate}

\item If you used crowdsourcing or conducted research with human subjects...
\begin{enumerate}
  \item Did you include the full text of instructions given to participants and screenshots, if applicable?
    \answerNA{}
  \item Did you describe any potential participant risks, with links to Institutional Review Board (IRB) approvals, if applicable?
    \answerNA{}
  \item Did you include the estimated hourly wage paid to participants and the total amount spent on participant compensation?
    \answerNA{}
\end{enumerate}

\end{enumerate}


\newpage
\appendix
\addcontentsline{toc}{section}{Appendix} 
\part{Appendix} 
\parttoc 

\section{Fine-Tuning Result for Sanity Check}
\label{appendix:fine-tuning-result}
We compare our performance results to those of previous studies to ensure that they are not  far off from the results of the prior study \citep{reid2022can}, the result of which is shown in Table \ref{table:sanity-check}. Following the previous studies \citep{fu2020d4rl,chen2021decision,reid2022can}, we report the normalized score of mean return: $100 \times \frac{\text{score} - \text{random score}}{\text{expert score} - \text{random score}}$, where \textit{random score} is mean return generated by a random policy and \textit{expert score} is generated by a policy trained by Soft Actor-Critic \citep{haarnoja2018soft}. Mean return is the sum of rewards averaged over trajectory.
For further details of the datasets and metrics, please refer to the paper that proposes D4RL \citep{fu2020d4rl}. Although the previous work \citep{reid2022can} used several techniques to improve the performance, e.g. language model co-training, the extension of positional embedding, and similarity encouragement, we do not employ any of these techniques so that we study the pure effect of pre-training. The training details are described in Appendix \ref{appendix:training-detail}

The result for the previous work is the average and standard deviation of three random seeds, while our result is those of two random seeds. The aim of this comparison is just to confirm that our result is not too pathological, checking soundness with two seeds would be valid enough. For reference, we also include the results of the Decision Transformer (DT) since the randomly initialized model (Random Init) is a large Decision Transformer. Following the previous study \citep{reid2022can}, which says that it conducted early stopping, we show the result of the best checkpoint in 40 epoch checkpoints. The result shows that our result is not too far away from the previous work, ensuring that the models of the subject of our analysis are valid enough.

\section{Training Details}
\label{appendix:training-detail}
Hyperparameters are determined following previous study \citep{reid2022can}. The hyperparameters and other setups for fine-tuning the model are summarized in Table \ref{table:training-configuration}. The number of layers is 12 for GPT2 and randomly initialized model and 24 for iGPT. The other setup not specified, including $\beta$ of Adam optimizer \citep{KingmaB14} and model initialization scheme follows Pytorch default ones or the default values of optional arguments in the scripts that our experiment is based on \footnote{\lstinline{experiment.py} in the following repository:\\
\href{https://github.com/machelreid/can-wikipedia-help-offline-rl/blob/main/code/experiment.py}{https://github.com/machelreid/can-wikipedia-help-offline-rl/blob/main/code/experiment.py}}. We do not do any pre-training ourselves, but use publicly available pre-trained models explained in Section \ref{section:background}. 

Each model is trained to minimize the mean squared loss between its predicted actions $\hat{\bm{a}}_t$ and the true actions $\bm{a}_t$. To get the big picture of how fine-tuning of GPT-based models to offline RL data is conducted, please refer to the pseudo-code of the previous work \citep{chen2021decision}.

The D4RL mujoco task used as offline RL data for our analysis has \textit{expert}, \textit{medium}, \textit{medium-replay}, \textit{medium-expert}, and \textit{random} datasets. The \textit{expert} dataset is the one trained by Soft Actor-Critic \citep{haarnoja2018soft}, \textit{medium} is the one partially trained by Soft Actor-Critic and was stopped early, \textit{medium-replay} is the one accumulated in the replay buffer before the model reached \textit{medium}'s level, and \textit{medium-expert} is a mixture of \textit{medium} and \textit{expert} results. The \textit{random} dataset is the trajectory collected by the random policy. We conducted our experiments on the \textit{medium} data set since it was used by the previous study \citep{reid2022can} and our analysis is based on the observations of this previous study.
The \textit{medium} dataset we used is 1 million time steps collected by the policy explained above. For the detail of the dataset, please refer to the original paper \citep{fu2020d4rl}.

\begin{table}[H]
\caption{Training configuration.}
\label{table:training-configuration}
\centering
\begin{tabular}{ll}
\toprule
\cmidrule(r){1-2}
\# Layers & 12 \\
\midrule
Emb. Dim. & 768 \\
\midrule
\# Attention Heads & 1 \\
\midrule
Batch size & 64 \\
\midrule
Context & 20 \\
\midrule
Return-to-go conditioning & 6000 HalfCheetah \\
& 3600 Hopper \\
& 5000 Walker \\
\midrule
Dropout & $0.2$ \\
\midrule
Learning rate & $1e-4$ \\
\midrule
LR Warmup & 5000 steps \\
\midrule
Epoch & 40 \\
\midrule
\# Steps per Epoch & 2500 \\
\midrule
Optimizer & Adam \\
\midrule
Weight Decay & $1e-4$ \\
\bottomrule
\end{tabular}
\end{table}

\section{Model Architecture and Module Names}
In the following sections, we use the module names of our models (\lstinline{gpt2} and \lstinline{openai/imagegpt-small}) to describe the experimental details. Thus, for the reference, we show the model architecture we use and the names of the modules of the model. We show that of \lstinline{gpt2} but the configuration is the same for \lstinline{openai/imagegpt-small} except for the number of Transformer blocks. For the detailed configurations of pre-trained models, please refer to the following links:
\begin{itemize}
    \item \lstinline{gpt2}: \href{https://huggingface.co/gpt2}{https://huggingface.co/gpt2}
    \item \lstinline{openai/imagegpt-small}: \href{https://huggingface.co/openai/imagegpt-small}{https://huggingface.co/openai/imagegpt-small}
\end{itemize}
We show the model architecture below, following the Pytorch format:  

\begin{lstlisting}[language=Python, caption=Model architecture.]
DecisionTransformer(
  (transformer): GPT2Model(
    (wte): Embedding(50257, 768)
    (wpe): Embedding(1024, 768)
    (drop): Dropout(p=0.1, inplace=False)
    (h): ModuleList(
      (0): GPT2Block(
        (ln_1): LayerNorm((768,), eps=1e-05, elementwise_affine=True)
        (attn): GPT2Attention(
          (c_attn): Conv1D()
          (c_proj): Conv1D()
          (attn_dropout): Dropout(p=0.1, inplace=False)
          (resid_dropout): Dropout(p=0.2, inplace=False)
        )
        (ln_2): LayerNorm((768,), eps=1e-05, elementwise_affine=True)
        (mlp): GPT2MLP(
          (c_fc): Conv1D()
          (c_proj): Conv1D()
          (act): NewGELUActivation()
          (dropout): Dropout(p=0.2, inplace=False)
        )
      )
      (1): GPT2Block(
        (ln_1): LayerNorm((768,), eps=1e-05, elementwise_affine=True)
        (attn): GPT2Attention(
          (c_attn): Conv1D()
          (c_proj): Conv1D()
          (attn_dropout): Dropout(p=0.1, inplace=False)
          (resid_dropout): Dropout(p=0.2, inplace=False)
        )
        (ln_2): LayerNorm((768,), eps=1e-05, elementwise_affine=True)
        (mlp): GPT2MLP(
          (c_fc): Conv1D()
          (c_proj): Conv1D()
          (act): NewGELUActivation()
          (dropout): Dropout(p=0.2, inplace=False)
        )
      )
      (2): GPT2Block(
      .
      .
      .
      )
      (11): GPT2Block(
        (ln_1): LayerNorm((768,), eps=1e-05, elementwise_affine=True)
        (attn): GPT2Attention(
          (c_attn): Conv1D()
          (c_proj): Conv1D()
          (attn_dropout): Dropout(p=0.1, inplace=False)
          (resid_dropout): Dropout(p=0.2, inplace=False)
        )
        (ln_2): LayerNorm((768,), eps=1e-05, elementwise_affine=True)
        (mlp): GPT2MLP(
          (c_fc): Conv1D()
          (c_proj): Conv1D()
          (act): NewGELUActivation()
          (dropout): Dropout(p=0.2, inplace=False)
        )
      )
    )
    (ln_f): LayerNorm((768,), eps=1e-05, elementwise_affine=True)
  )
  (embed_timestep): Embedding(1000, 768)
  (embed_return): Linear(in_features=1, out_features=768, bias=True)
  (embed_state): Linear(in_features=11, out_features=768, bias=True)
  (embed_action): Linear(in_features=3, out_features=768, bias=True)
  (embed_ln): LayerNorm((768,), eps=1e-05, elementwise_affine=True)
)
\end{lstlisting}

Among the above modules, for our analysis, we focus on the modules in \lstinline{DecisionTransformer.transformer.h}. We summarize the module-related notations used in this paper below for reference:
\begin{itemize}
    \item \textit{Transformer blocks}: modules directly under
    \lstinline{DecisionTransformer.transformer.h}
        \begin{itemize}
            \item e.g. \lstinline{DecisionTransformer.transformer.h[0]}.
        \end{itemize}
    \item \textit{Outputs of blocks}: outputs of \lstinline{mlp.dropout} of blocks.
        \begin{itemize}
            \item e.g. \lstinline{DecisionTransformer.transformer.h[0].mlp.dropout}.
        \end{itemize}
    \item \textit{Layers}: modules under Transformer blocks.
        \begin{itemize}
            \item e.g. \lstinline{DecisionTransformer.transformer.h[0].attn.c_attn}.
        \end{itemize}
    \item \textit{Parameter set}: parameters of each layer.
        \begin{itemize}
            \item e.g. \lstinline{DecisionTransformer.transformer.h[0].ln_1.weight}.
        \end{itemize}
\end{itemize}
Note that these notations are sometimes used interchangeably as long as it doesn't significantly deteriorate the readability. 

\section{Activation Similarity}
\label{appendix:activation-similarity}

\subsection{Details of Experiments}
\label{appendix:detail-of-experiments-activation-similarity}

\begin{figure}[H]
    \centering
    \includegraphics[width=0.5\linewidth]{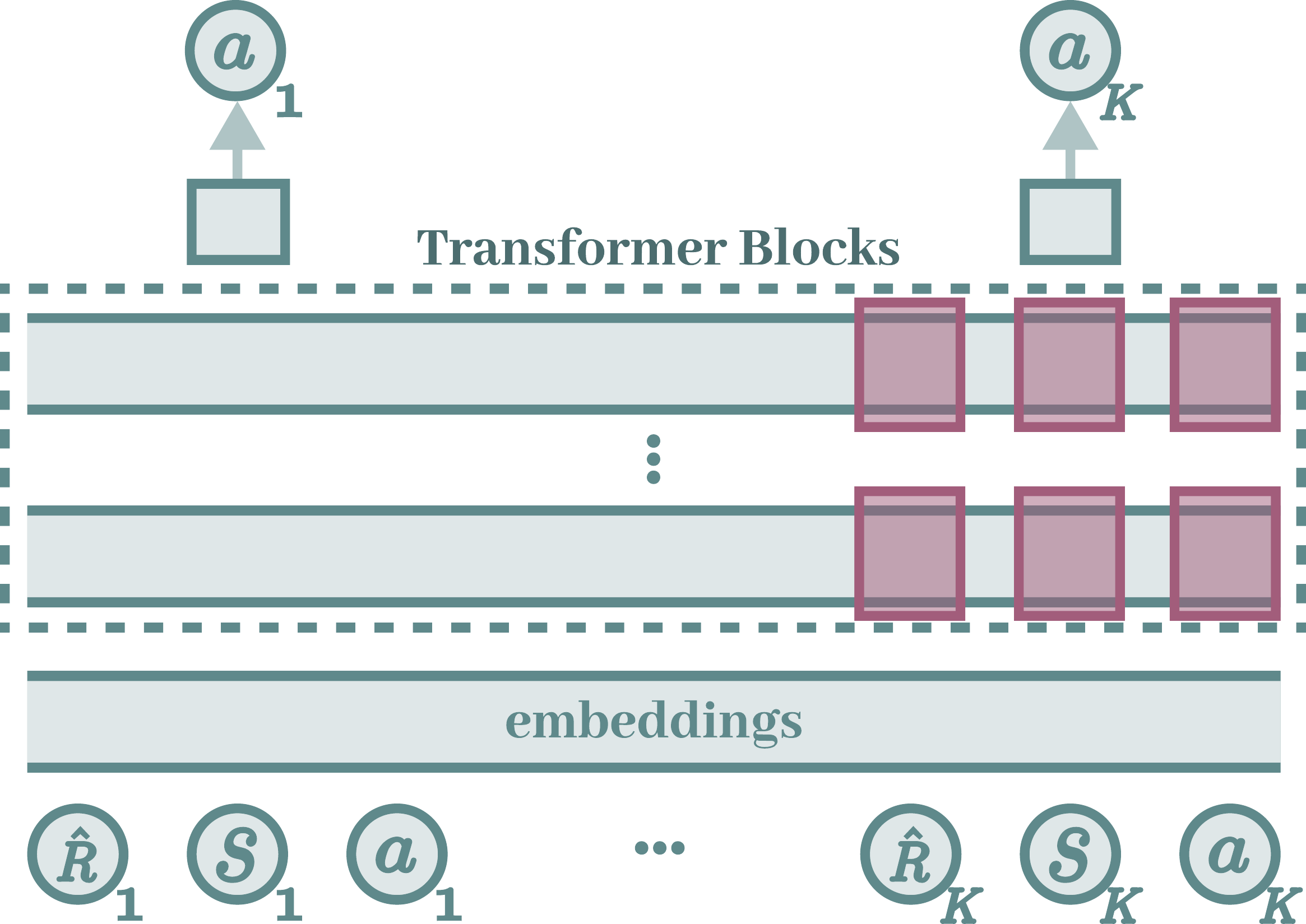}
    \caption{Activation we consider to compute CKA.}
    \label{fig:diagram_cka_activation}
\end{figure}

We randomly sample 100 samples and compute unbiased estimators \citep{kornblith2019similarity} of linear CKA for these 100 activation vectors. The activation to be analyzed are outputs from all \textit{layers}.
In the Decision Transformer, the effective total input length is context length $K$ times the number of token types $(\hat{R}, \bm{s}, \bm{a})$. As a result, the shape of the activation obtained at each layer is \lstinline{(batch size, 3x context length, embedding dimension)}. In this study, we compute the CKA for the activity of shape \lstinline{(batch size, embedding dimension)} that corresponds to the last time step in the context of return-to-go, state, and action, respectively. In other words, noting activation as in the form of python NumPy array \lstinline{activation}, we obtain \lstinline{activation[:, -1, :]}, \lstinline{activation[:, -2, :]}, and \lstinline{activation[:, -3, :]}. 
In the previous work analyzing the internal representation of Transformer \citep{wu-etal-2020-similarity}, the dimension of context and representation seem to be concatenated:  \lstinline{activation.contiguous().view(activation.shape[0] -1)}. Since the representation obtained in this way is hard to interpret, we decide to use the way we described above to obtain activation. A descriptive diagram for activation that we compute CKA about is shown in Fig. \ref{fig:diagram_cka_activation}. The design of the diagram is based on a previous study \citep{chen2021decision}. 

Each rectangle colored in red is an activation we consider. The left-most rectangles are the activation with return-to-go as input, the middle rectangles are the that with status as input, and the right-most rectangles are that with action as input. Since we consider only the last in the context, the rectangles are drawn only where the tokens of the K-th step are received as input. In Section \ref{section:activation-similarity}, we show the results of activation that $\bm{s}_K$ is fed in the diagram.

\subsection{Centered Kernel Alignment (CKA)}
\label{appendix:cka}
We show the definition of Centered Kernel Alignment (CKA). Given two activation vectors $\bm{X} \in \mathbb{R}^{m \times p_1}$ and $\bm{Y} \in \mathbb{R}^{m \times p_1}$, where $m$ is data size and $p_1$ and $p_2$ are dimensions of hidden layers, the CKA of the two representations is as follows:
 \begin{equation}
     \text{CKA} = \frac{\text{HSCI}(\bm{K}, \bm{L})}{\sqrt{\text{HSCI}(\bm{K}, \bm{K})\text{HSCI}(\bm{L}, \bm{L})}},
 \end{equation}
 where HSCI is the Hilbert-Schmidt independence criterion \citep{gretton2007kernel} and $\bm{K}_{ij} = k(\bm{x}_i, \bm{x}_j)$ and $\bm{L}_{ij} = l(\bm{x}_i, \bm{x}_j)$ are kernels. Particularly, linear CKA is defined as follows:
  \begin{equation}
     \text{linear CKA} = \frac{||\bm{Y}^{\top} \bm{X}||_F^2}{||\bm{X}^{\top} \bm{X}||_F||\bm{Y}^{\top}\bm{Y}||_F},
 \end{equation}
 where $||\cdot||_F$ is the Frobenius norm. For further details, please refer to the previous work \citep{kornblith2019similarity}.
 
\subsection{Layer Names Whose CKA is Above a Threshold}
\label{appendix:layer-name-cka}
In Section \ref{section:activation-similarity}, we say we observe that most of the high CKAs of Fig. \ref{fig:cka_plot_gpt2_dt} (b), which are greater than 0.38, are those of layer normalization. The reason we set this seemingly arbitrary threshold is that we observe higher layers in a block than other layers in Fig. \ref{fig:cka_plot_gpt2_dt} and want to identify these layers. By eye inspection, we find the threshold over which all these points are included in Fig. \ref{fig:cka_plot_gpt2_dt} is around 0.38 and so we select this value as a threshold. 
Here, we show the raw result of the observation. The layer name list is shown below:
\begin{lstlisting}[language=Python, caption=Layer names whose CKA is above a threshold]
0.ln_1, 0.attn.c_attn, 0.ln_2, 0.mlp.c_fc, 0.mlp.c_proj, 0.mlp.dropout
1.ln_1, 1.attn.c_attn, 1.ln_2
2.ln_1,                2.ln_2
3.ln_1,                3.ln_2, 3.mlp.c_fc
4.ln_1, 4.attn.c_attn, 4.ln_2
5.ln_1, 5.attn.c_attn, 5.ln_2
6.ln_1, 6.attn.c_attn, 6.ln_2
7.ln_1, 7.ln_2
8.ln_1,                8.ln_2
9.ln_1, 9.ln_2
10.ln_1, 10.attn.c_attn, 10.ln_2
11.ln_1.                 11.ln_2
\end{lstlisting}
Note that, for example, \lstinline{0.ln_1} is the first layer normalization layer (\lstinline{ln_1}) of the first transformer block (\lstinline{0}). We observe that 24 of 34 layers are layer norm modules (\lstinline{ln_1} and \lstinline{ln_2}).

\subsection{CKA Between Different Models}
\label{appendix:cka-different-models}
\begin{figure}[h]
    \centering
    \begin{minipage}[b]{0.32\linewidth}
        \includegraphics[width=\linewidth]{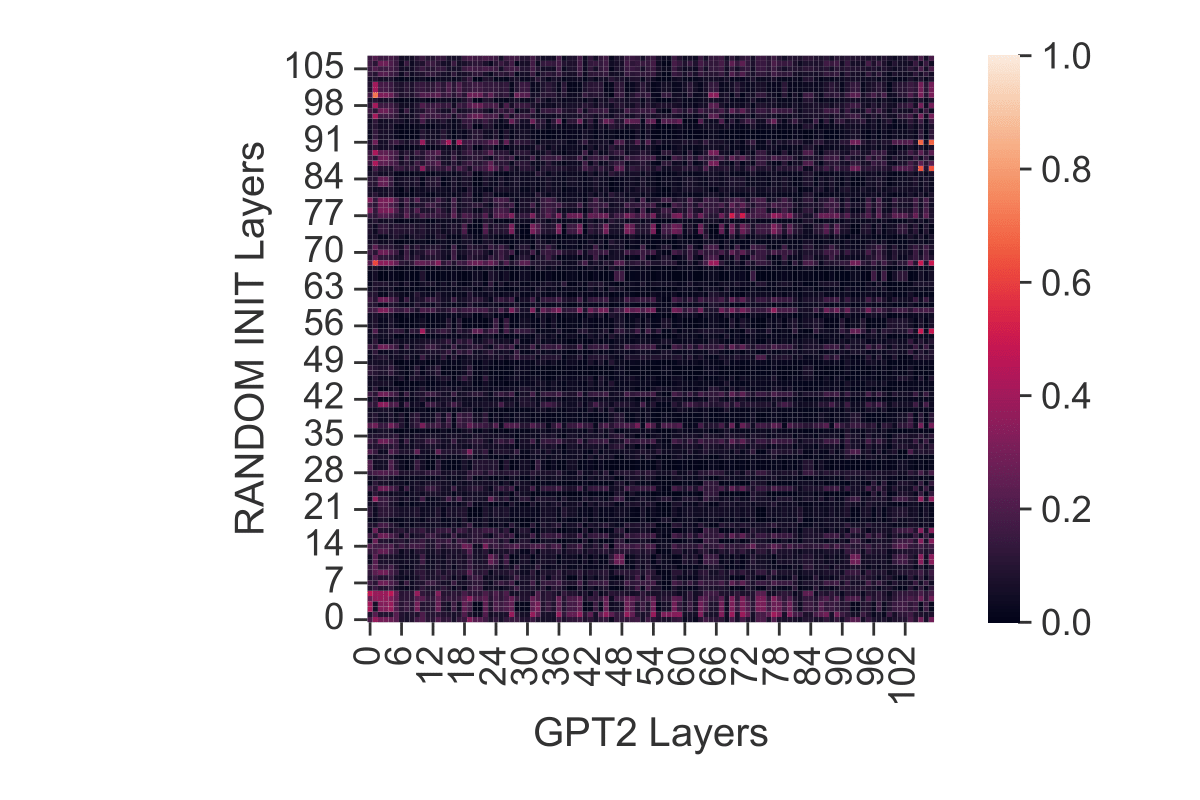}
        \subcaption{Random Init. vs GPT2}
    \end{minipage}
    \begin{minipage}[b]{0.32\linewidth}
        \includegraphics[width=\linewidth]{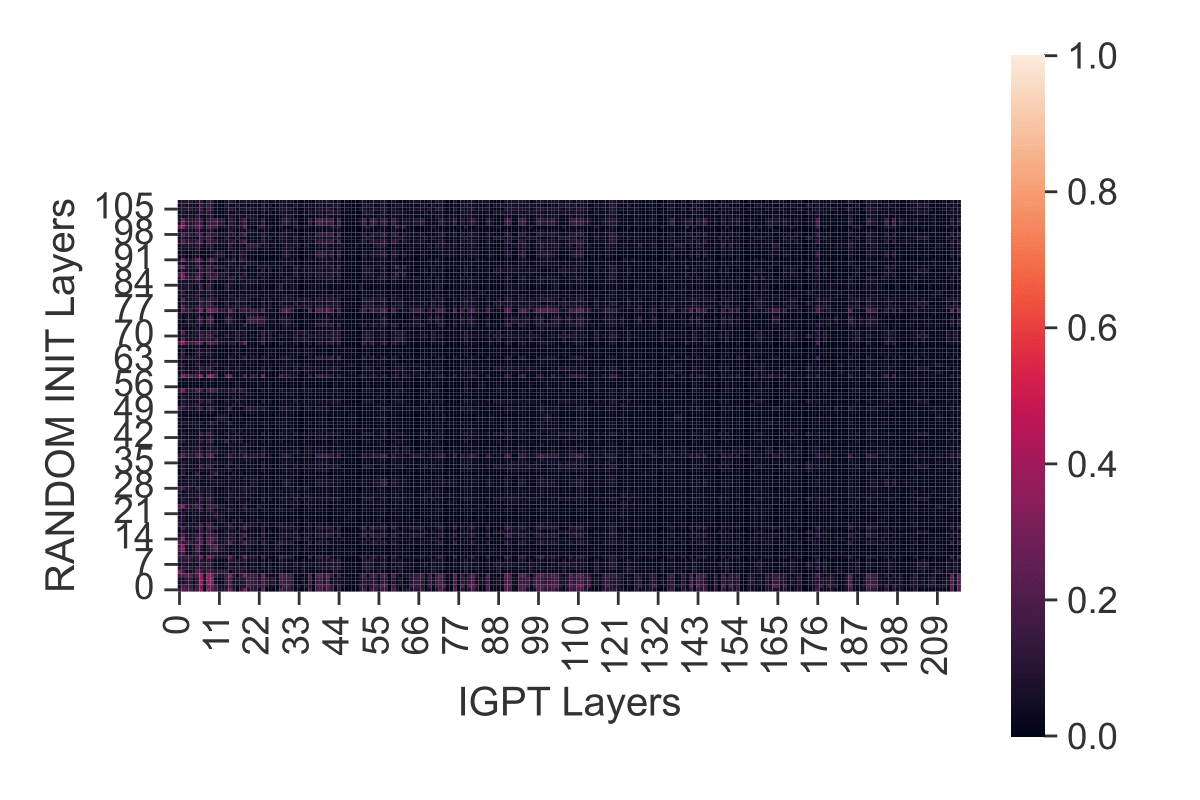}
        \subcaption{Random Init. vs iGPT}
    \end{minipage}
    \begin{minipage}[b]{0.32\linewidth}
        \includegraphics[width=\linewidth]{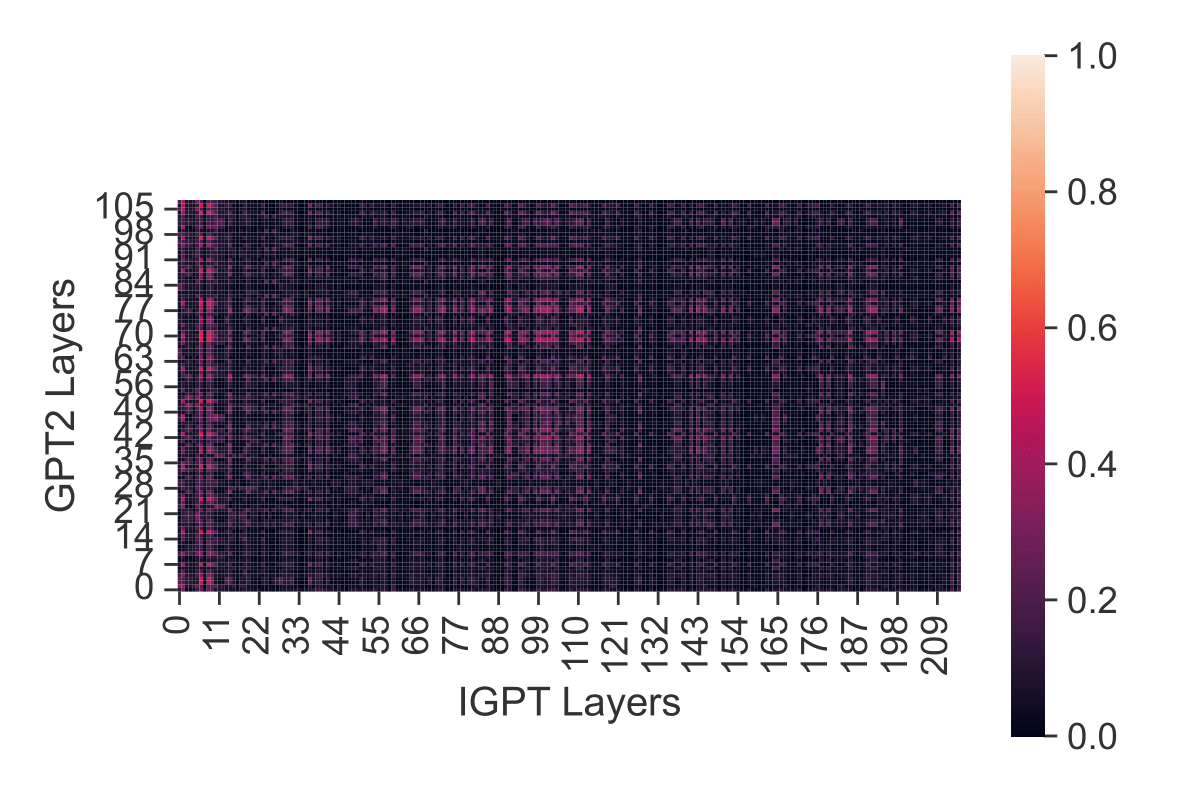}
        \subcaption{GPT2 vs iGPT}
    \end{minipage}
    \caption{CKA similarity across layers between different fine-tuned models.}
    \label{fig:cka_gpt2_dt_igpt}
\end{figure}

To check if pre-trained models and the randomly initialized model converge to a similar representation or not, we compute the CKA between representations of fine-tuned these models. The result is shown in Fig. \ref{fig:cka_gpt2_dt_igpt}, where (a) is the CKA between GPT2 and iGPT, (b) is that between GPT2 and random initialization, and (c) is that between iGPT and random initialization. We observe that CKA values in these heat maps are small, confirming that different models learn different representations. 

\subsection{CKA Between Different Layers in a Model}
\subsubsection{Post-Fine-Tuning}
In Appendix \ref{appendix:cka-different-models}, we find that each layer of the pre-trained model and randomly initialized model learn representation differently. In this section, we delve into what representation these models acquire. For that purpose, we compute CKA values of the different layers in the same model after fine-tuning. The result for the state input token with the \textit{Hopper-medium} dataset is shown in Fig. \ref{fig:cka_40_gpt2_igpt_dt}. The results for other environments and input tokens are in Appendix \ref{appendix:results-for-other-conditions-activation-similarity}.

\begin{figure}[H]
    \centering
    \begin{minipage}[b]{0.32\linewidth}
        \includegraphics[width=\linewidth]{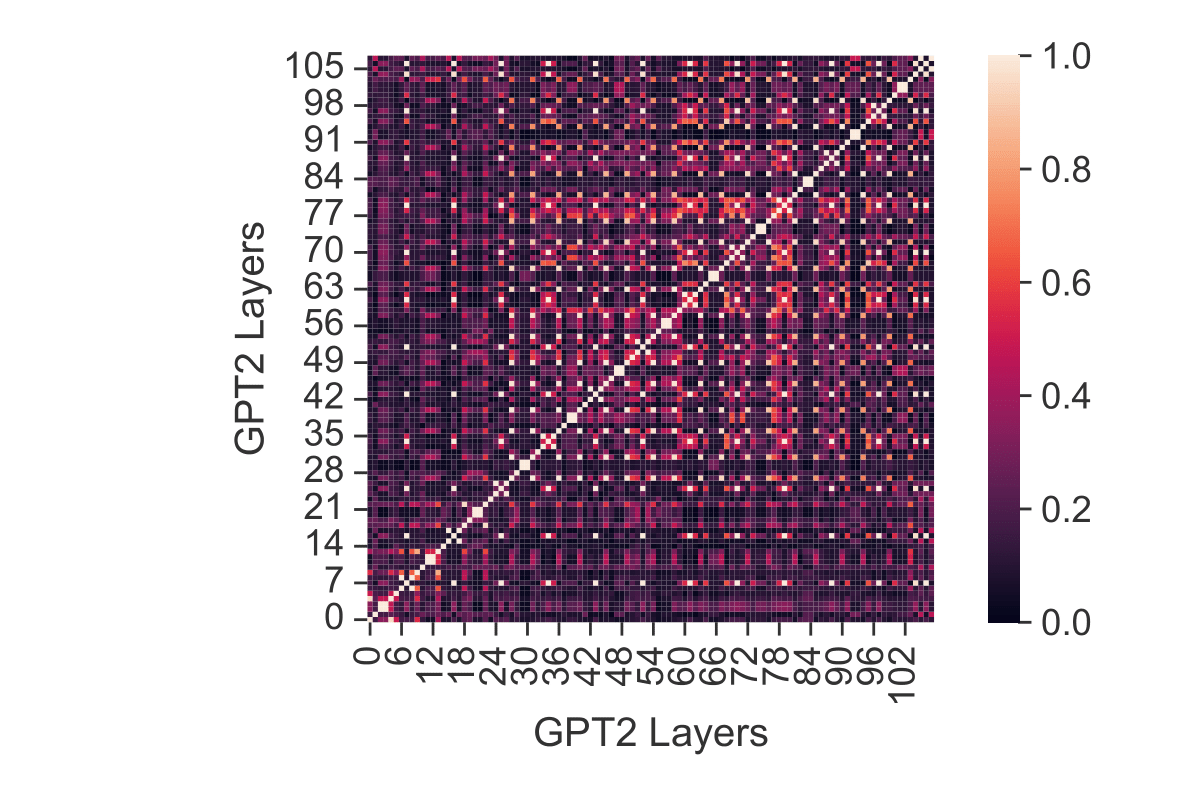}
        \subcaption{GPT2}
    \end{minipage}
    \begin{minipage}[b]{0.32\linewidth}
        \includegraphics[width=\linewidth]{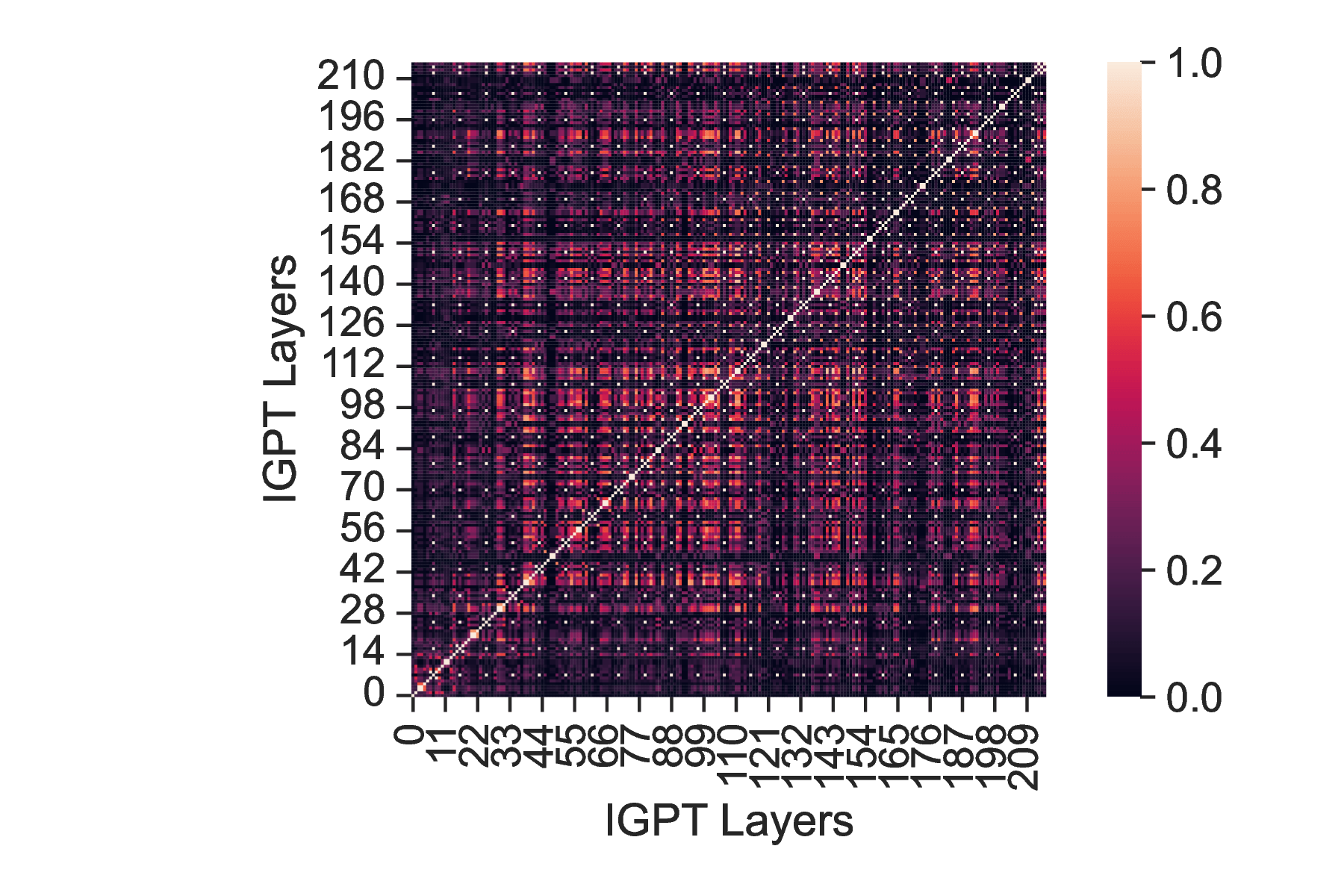}
        \subcaption{iGPT}
    \end{minipage}
    \begin{minipage}[b]{0.32\linewidth}
        \includegraphics[width=\linewidth]{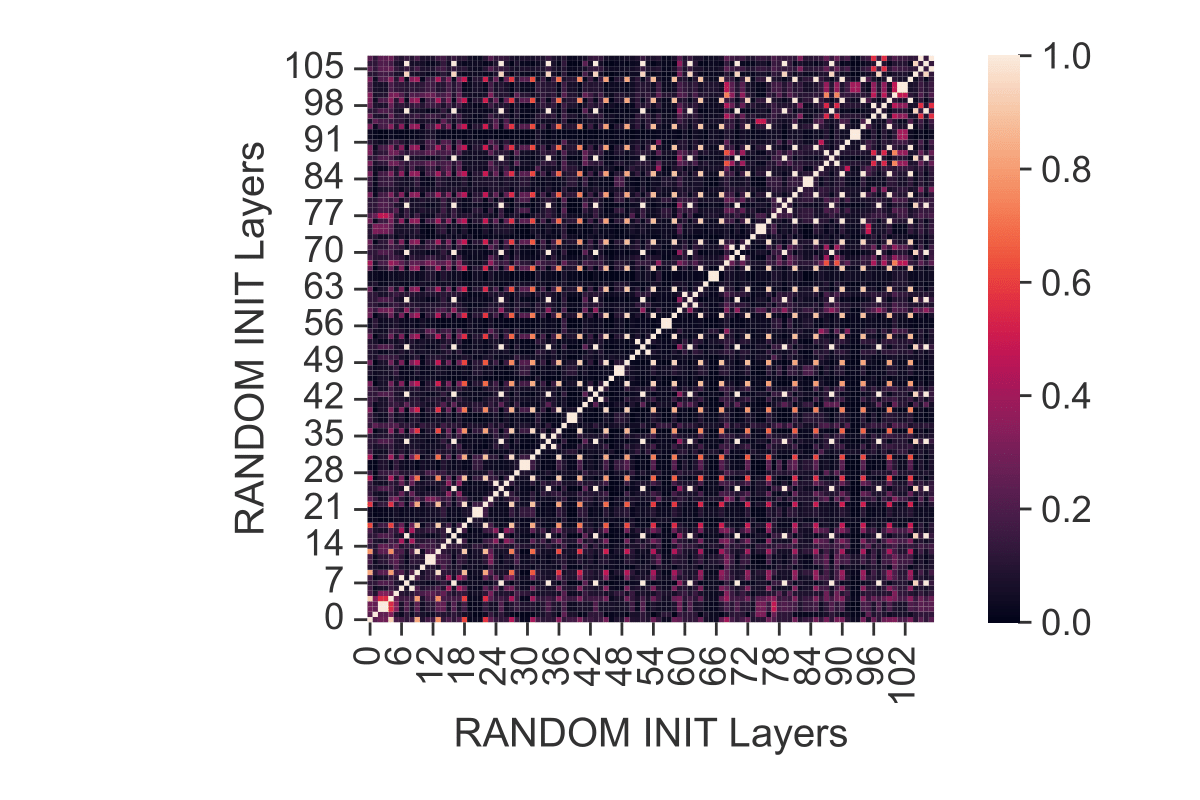}
        \subcaption{Random Initialization}
    \end{minipage}
    \caption{CKA of different layers in the same models (epoch 40).}
    \label{fig:cka_40_gpt2_igpt_dt}
\end{figure}

Comparing the CKA matrix of pre-trained models (Fig. \ref{fig:cka_40_gpt2_igpt_dt} (a) and Fig. \ref{fig:cka_40_gpt2_igpt_dt} (b)) with that of the randomly initialized model (Fig. \ref{fig:cka_40_gpt2_igpt_dt} (c)), we observe that the randomly initialized model has an almost equally divided lattice structure, while pre-trained models have some block-like structure. In particular, the shallow blocks and the middle to top blocks of language-pre-trained model have a bit similar representation, respectively. 
This result indicates that probably layers of random initialization process information separately, while some layers of language-pre-trained models coordinate to represent information.

\subsubsection{Pre-Fine-Tuning}
\begin{figure}[H]
    \centering
    \begin{minipage}[b]{0.32\linewidth}
        \includegraphics[width=\linewidth]{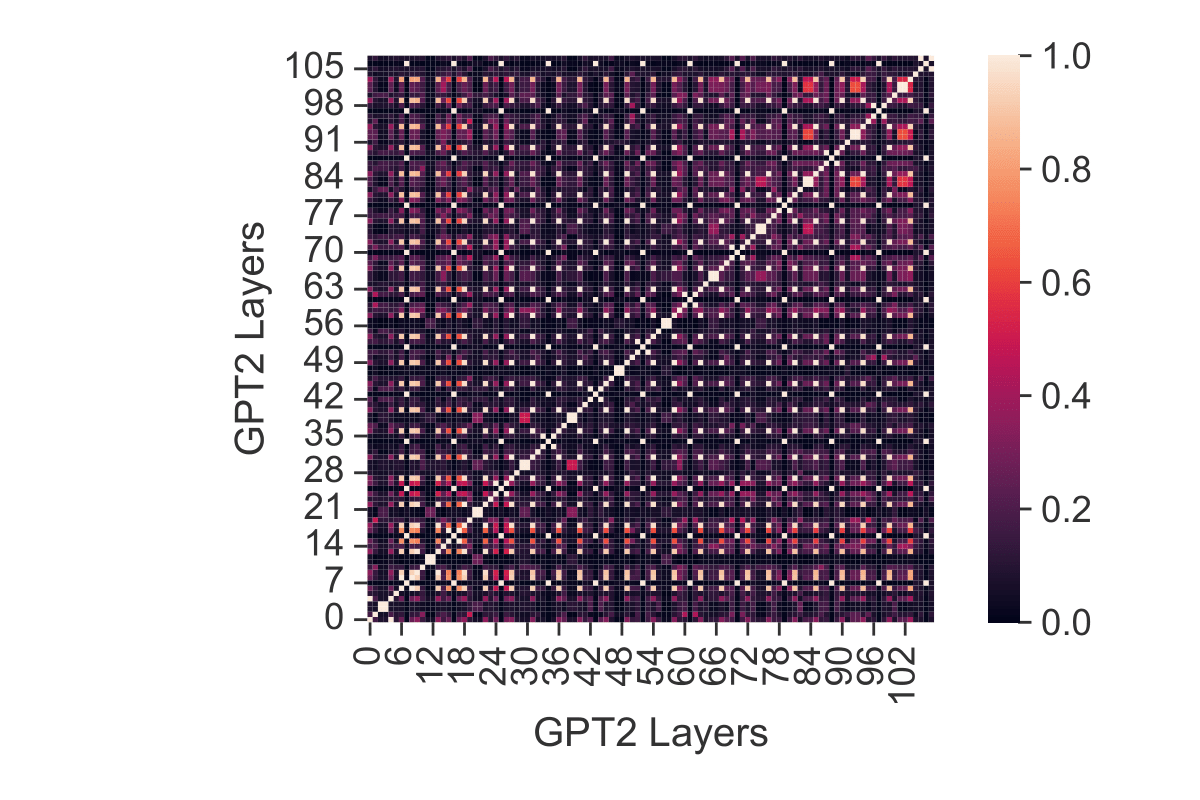}
        \subcaption{GPT2}
    \end{minipage}
    \begin{minipage}[b]{0.32\linewidth}
        \includegraphics[width=\linewidth]{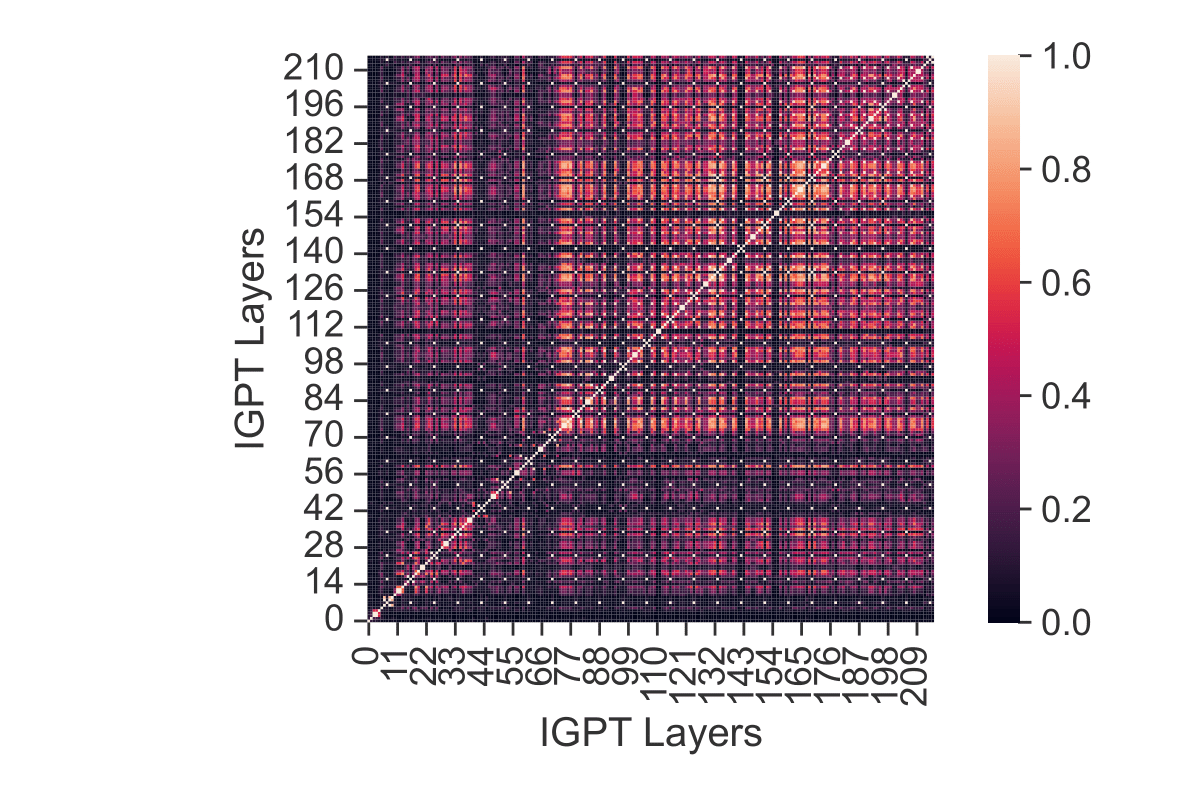}
        \subcaption{iGPT}
    \end{minipage}
    \begin{minipage}[b]{0.32\linewidth}
        \includegraphics[width=\linewidth]{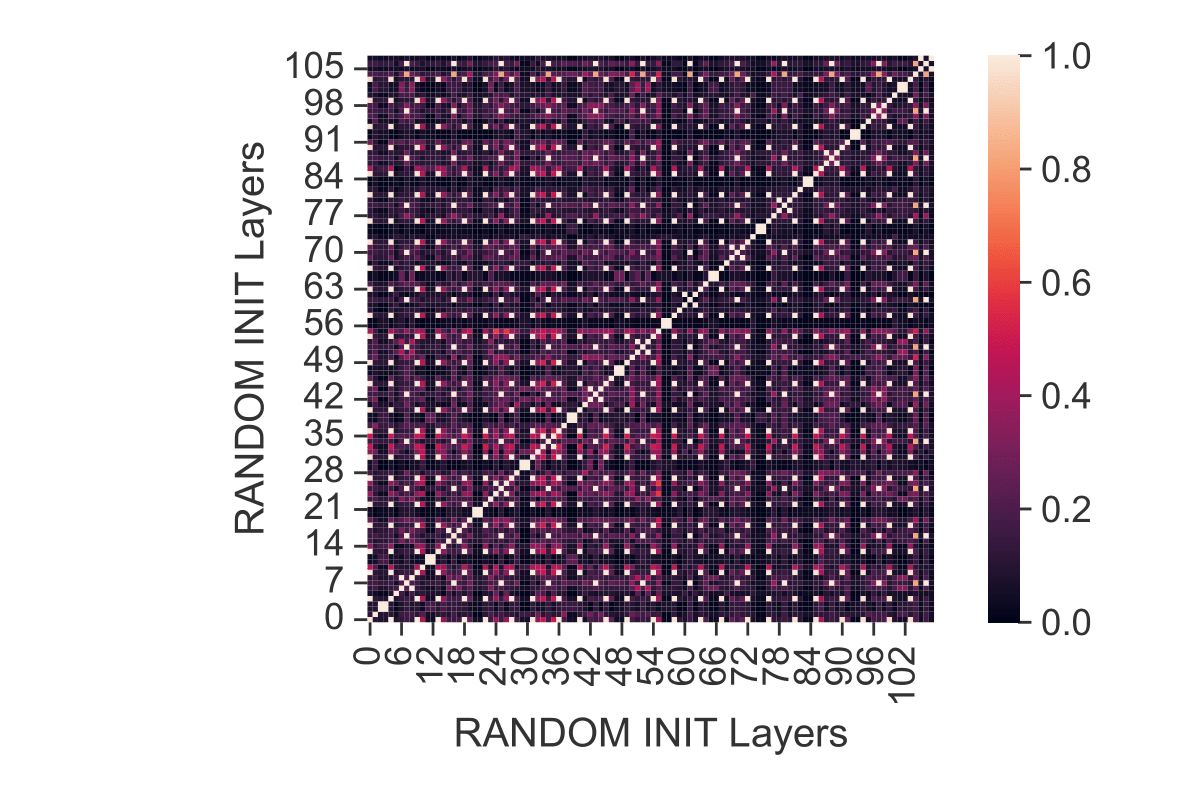}
        \subcaption{Random Initialization}
    \end{minipage}
    \caption{CKA of different layers in the same models (epoch 0).}
    \label{fig:cka_0_gpt2_igpt_dt}
\end{figure}

For just reference, we put the CKA matrix of pre-trained models before fine-tuning as well. The result is shown in Fig. \ref{fig:cka_0_gpt2_igpt_dt}. We observe the grid structure at the initial state as well.

\section{Mutual Information Between Hidden Representation and Data}
\label{appendix:mutual-information-between-hidden-representation-and-input-and-label}

\subsection{Details of Experiments}
\label{appendix:detail-of-experiments-mutual-information}

MINE, the definition of which is explained in Appendix \ref{appendix:definition-mutual-information}, estimates mutual information by training a neural network. The neural network we use is a feed-forward ReLU network with two hidden layers of width 400. We train the neural network for 1000 iterations by Adam optimizer with a learning rate of $1e-4$. Same as in Appendix \ref{appendix:detail-of-experiments-activation-similarity}, we randomly sample 100 trajectories and use them to obtain the activation. Thus, the dataset for mutual information calculation is the pair between these 100 activation vectors and 100 trajectories for $\hat{I}(X; T)$ and 100 last action vectors for $\hat{I}(Y; T)$. If the estimated value is \lstinline{NaN}, we exclude the point from the figure.

\begin{figure}[ht]
    \centering
    \begin{minipage}[b]{0.4\linewidth}
    \includegraphics[width=\linewidth]{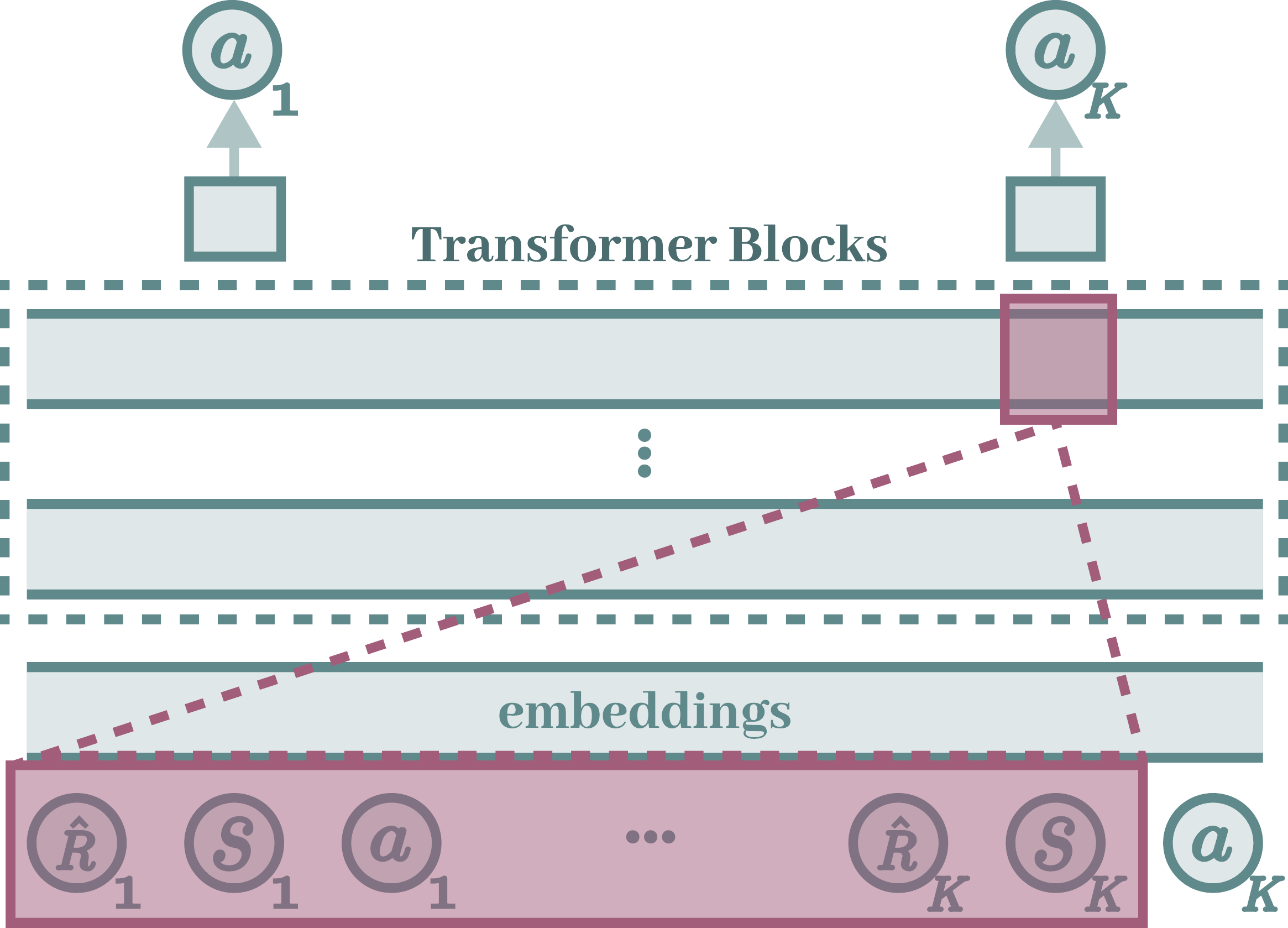}
    \subcaption{$\hat{I}(X; T)$}
    \end{minipage}
    \hspace{1cm}
    \begin{minipage}[b]{0.4\linewidth}
    \includegraphics[width=\linewidth]{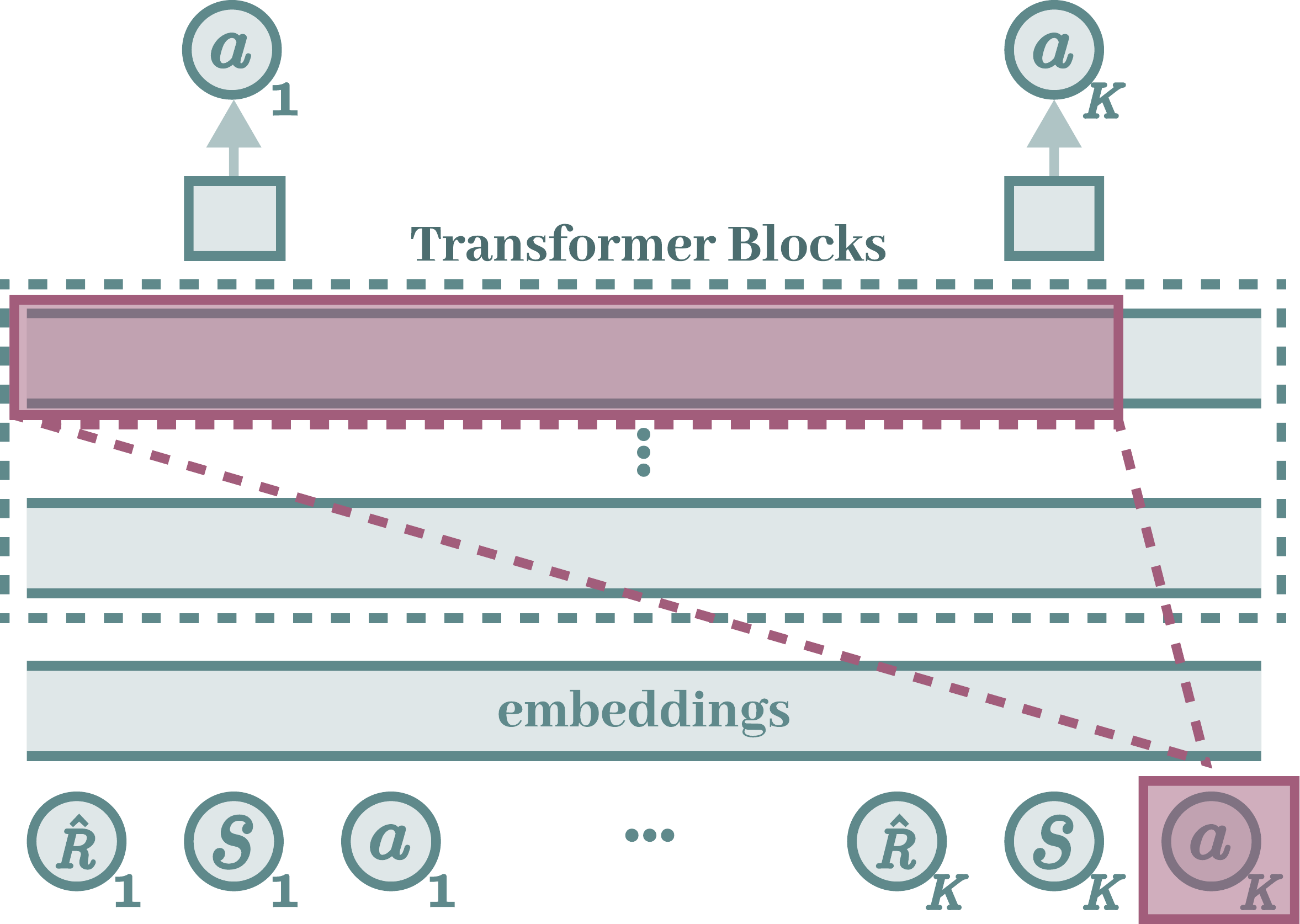}
    \subcaption{$\hat{I}(Y; T)$}
    \end{minipage}
    \caption{Data and activation we consider to compute estimated mutual information.}
    \label{fig:diagram_mutual_information_context}
\end{figure}

The descriptive diagram that shows hidden activation and data we consider to compute estimated mutual information is Fig. \ref{fig:diagram_mutual_information_context}, where (a) is for $\hat{I}(X; T)$ and (b) is for $\hat{I}(Y; T)$. This is an example of mutual information for the deep Transformer block. For the middle block, we use block 6 for GPT2 and random initialization and block 12 for iGPT. Specifically, the outputs of the shallow, middle, and deep Transformer blocks are the outputs of \lstinline{0.mlp.dropout}, \lstinline{6.mlp.dropout}, and \lstinline{11.mlp.dropout} for GPT2 and randomly initialized models and \lstinline{0.mlp.dropout}, \lstinline{12.mlp.dropout}, and \lstinline{23.mlp.dropout} for iGPT.

\subsection{Mutual Information Neural Estimation (MINE)}
\label{appendix:definition-mutual-information}
Given random variables $X$ and $T$, whose joint probability is $\mathbb{P}_{X T}$ and marginal distributions are $\mathbb{P}_{X}=\int_{\mathcal{T}} \mathrm{d} \mathbb{P}_{X T}$ and $\mathbb{P}_{T}=\int_{\mathcal{X}} \mathrm{d} \mathbb{P}_{X T}$, the mutual information of $X$ and $T$ are as follows:
\begin{equation}
    I(X ; T)=\int_{\mathcal{X} \times \mathcal{T}} \log \frac{\mathrm{d} \mathbb{P}_{X T}}{\mathrm{d} \mathbb{P}_{X} \otimes \mathbb{P}_{T}} \mathrm{d} \mathbb{P}_{X T}.
\end{equation}

Denoting neural network with parameters $\bm{\theta} \in \Theta$ by $\text{NN}_{\bm{\theta}}$ and the empirical distribution of $\mathbb{P}$ characterized by $n$ i.i.d. samples by $\hat{\mathbb{P}}^{(n)}$, MINE is defined as follows \citep{pmlr-v80-belghazi18a}:
\begin{equation}
\hat{I}(X ; T)_{n}=\sup _{\theta \in \Theta} \mathbb{E}_{\mathbb{P}_{X T}^{(n)}}\left[\text{NN}_{\theta}\right]-\log \left(\mathbb{E}_{\mathbb{P}_{X}^{(n)} \otimes \hat{\mathbb{P}}_{T}^{(n)}}\left[e^{\text{NN}_{\theta}}\right]\right).
\end{equation}
Please refer to the previous study \citep{pmlr-v80-belghazi18a} for the details.

\subsection{Mutual Information Without Considering Context}
\begin{figure}[h]
    \centering
    \begin{minipage}[b]{0.4\linewidth}
    \includegraphics[width=\linewidth]{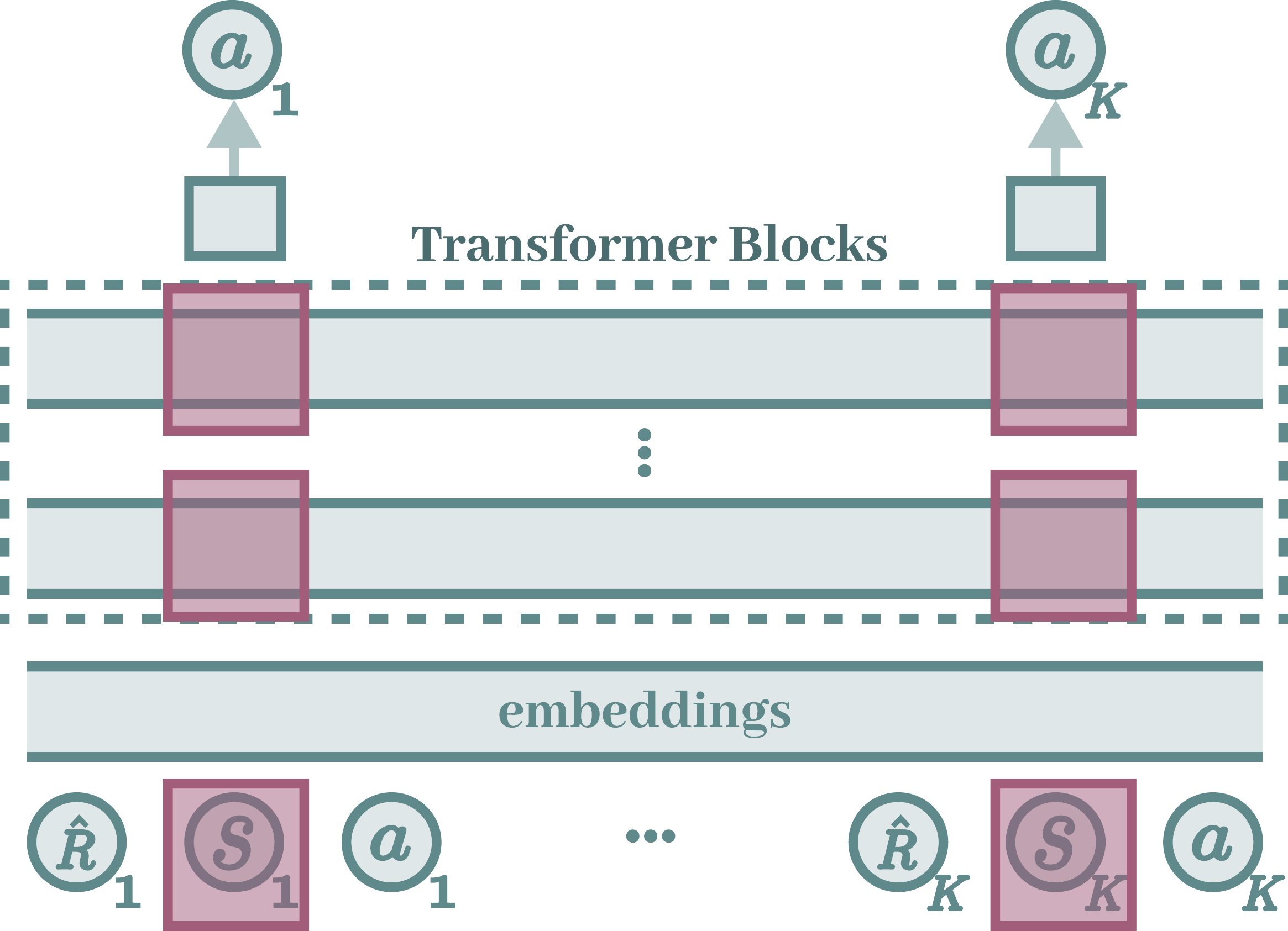}
    \subcaption{$\hat{I}(X; T)$}
    \end{minipage}
    \hspace{1cm}
    \begin{minipage}[b]{0.4\linewidth}
    \includegraphics[width=\linewidth]{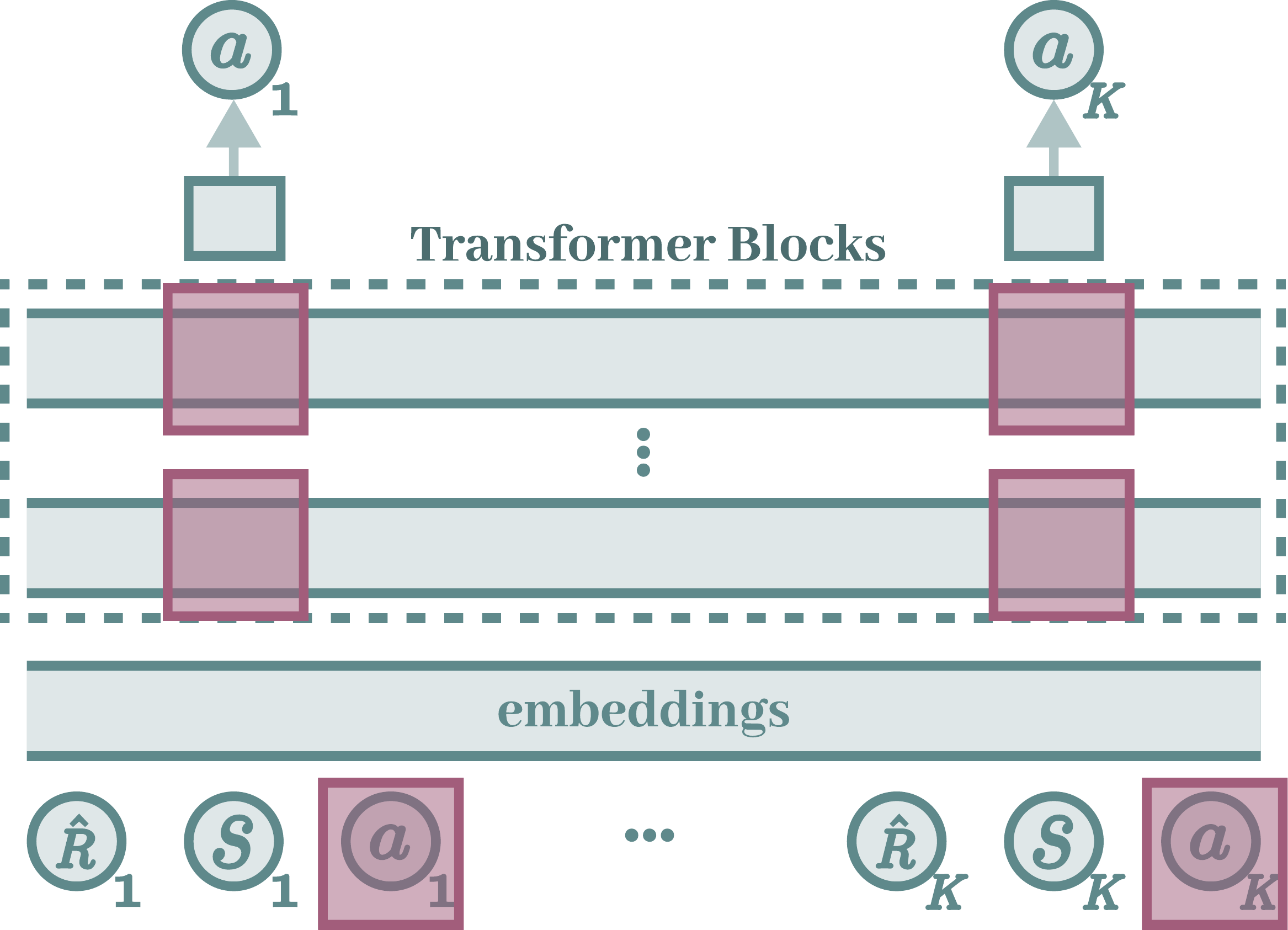}
    \subcaption{$\hat{I}(Y; T)$}
    \end{minipage}
    \caption{Data and activation we consider to compute estimated mutual information.}
    \label{fig:diagram_mutual_information_no_context}
\end{figure}

As a complementary analysis, we also calculate the estimated mutual information between hidden representation and $\bm{s}_t$ and $\bm{a}_t$. In other words, we compute $\hat{I}(\bm{s}_t;  T_l(\bm{s}_t))$ and $\hat{I}(\bm{a}_t;  T_l(\bm{s}_t))$ as $\hat{I}(X; T)$ and $\hat{I}(Y; T)$ for all time steps in the context $t = 1, ..., K$. Then, we take an average of them over the context. The descriptive diagram is shown in Fig. \ref{fig:diagram_mutual_information_no_context}.

\begin{figure}[H]
    \centering
    \begin{minipage}[b]{0.45\linewidth}
    \includegraphics[width=\linewidth]{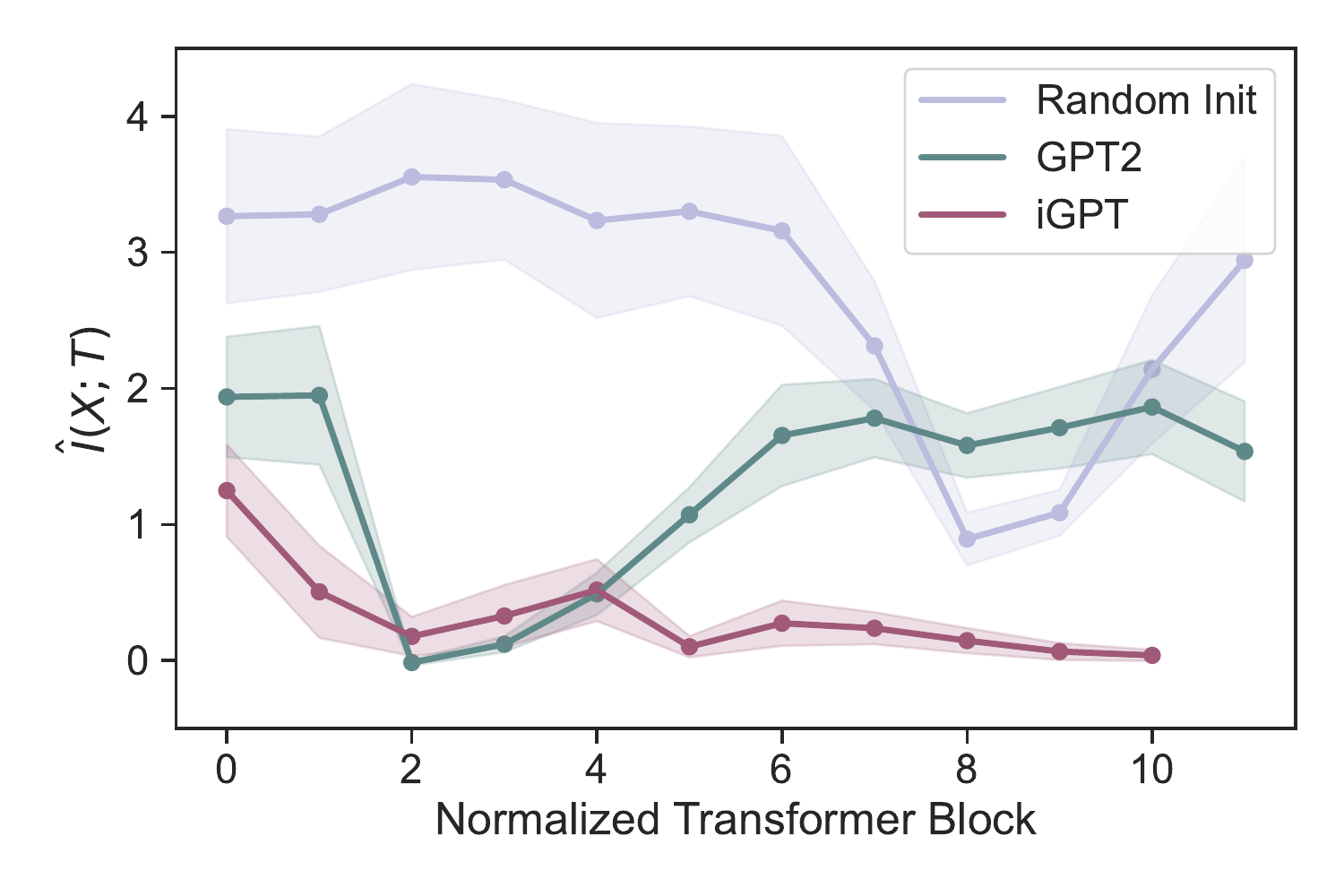}
    \subcaption{$\hat{I}(X; T)$}
    \end{minipage}
    \begin{minipage}[b]{0.45\linewidth}
    \includegraphics[width=\linewidth]{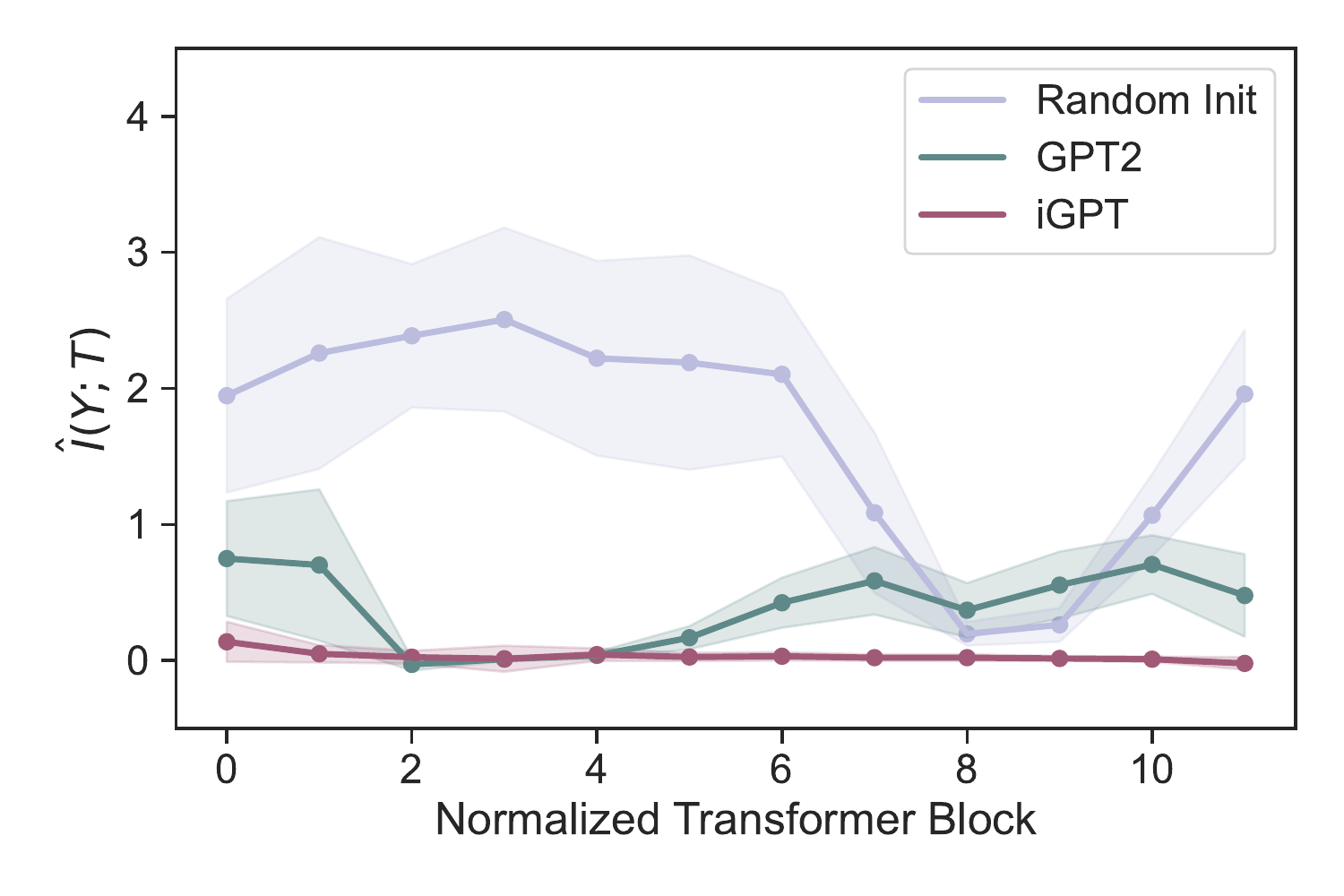}
    \subcaption{$\hat{I}(Y; T)$}
    \end{minipage}
    \caption{Estimated mutual information between hidden representation and state and action.}
    \label{fig:mutual_information_no_context}
\end{figure}

The result for \textit{Hopper-medium} is shown in Fig. \ref{fig:mutual_information_no_context}. Each point indicates the estimated mean mutual information and the shaded area is the standard deviation. Just as we did in Section \ref{section:activation-similarity}, we take an average over two adjacent elements for iGPT for comparison. We again observe that the randomly initialized model generally has more information on input and label, and iGPT has almost nothing about label-related information.

\section{Parameter Similarity}
\label{appendix:parameter-similarity}

\subsection{Details of Experiments}
\label{appendix:detail-of-experiments-parameter-similarity}
The parameters considered in the analysis are those in Transformer blocks (\lstinline{DecisionTransformer.transformer.h}). We concatenate all these parameters into a single vector and compute $l2$ distance and cosine similarity for this vector. For the post-fine-tuning model in Section \ref{section:parameter-similarity}, we use the model trained with the \textit{Hopper-medium} dataset. The results for other environments are in Appendix \ref{appendix:results-for-other-conditions-parameter-similarity}. 

\subsection{Ticks' Labels of Figures \ref{fig:param-dist} and \ref{fig:param-cos}}
\label{appendix:parameter-name}
The labels of ticks of Figs. \ref{fig:param-dist} and \ref{fig:param-cos} in Section \ref{section:parameter-similarity} correspond to all \textit{parameter sets}. For example, one of the parameter sets is \lstinline{0.ln_1.weight}: this is the weight vector of the layer normalization in the first (\lstinline{0}) transformer block (\lstinline{h}). The labels for ticks of GPT2 and the randomly initialized model are summarized below:

\begin{lstlisting}[language=Python, caption=Label list of ticks.]
 '0.ln_1.weight',
 '0.ln_1.bias',
 '0.attn.c_attn.weight',
 '0.attn.c_attn.bias',
 '0.attn.c_proj.weight',
 '0.attn.c_proj.bias',
 '0.ln_2.weight',
 '0.ln_2.bias',
 '0.mlp.c_fc.weight',
 '0.mlp.c_fc.bias',
 '0.mlp.c_proj.weight',
 '0.mlp.c_proj.bias',
 '1.ln_1.weight',
 '1.ln_1.bias',
 '1.attn.c_attn.weight',
 '1.attn.c_attn.bias',
 '1.attn.c_proj.weight',
 '1.attn.c_proj.bias',
 '1.ln_2.weight',
 '1.ln_2.bias',
 '1.mlp.c_fc.weight',
 '1.mlp.c_fc.bias',
 '1.mlp.c_proj.weight',
 '1.mlp.c_proj.bias',
 '2.ln_1.weight',
 '2.ln_1.bias',
 '2.attn.c_attn.weight',
 '2.attn.c_attn.bias',
 '2.attn.c_proj.weight',
 '2.attn.c_proj.bias',
 '2.ln_2.weight',
 '2.ln_2.bias',
 '2.mlp.c_fc.weight',
 '2.mlp.c_fc.bias',
 '2.mlp.c_proj.weight',
 '2.mlp.c_proj.bias',
 '3.ln_1.weight',
 '3.ln_1.bias',
 '3.attn.c_attn.weight',
 '3.attn.c_attn.bias',
 '3.attn.c_proj.weight',
 '3.attn.c_proj.bias',
 '3.ln_2.weight',
 '3.ln_2.bias',
 '3.mlp.c_fc.weight',
 '3.mlp.c_fc.bias',
 '3.mlp.c_proj.weight',
 '3.mlp.c_proj.bias',
 '4.ln_1.weight',
 '4.ln_1.bias',
 '4.attn.c_attn.weight',
 '4.attn.c_attn.bias',
 '4.attn.c_proj.weight',
 '4.attn.c_proj.bias',
 '4.ln_2.weight',
 '4.ln_2.bias',
 '4.mlp.c_fc.weight',
 '4.mlp.c_fc.bias',
 '4.mlp.c_proj.weight',
 '4.mlp.c_proj.bias',
 '5.ln_1.weight',
 '5.ln_1.bias',
 '5.attn.c_attn.weight',
 '5.attn.c_attn.bias',
 '5.attn.c_proj.weight',
 '5.attn.c_proj.bias',
 '5.ln_2.weight',
 '5.ln_2.bias',
 '5.mlp.c_fc.weight',
 '5.mlp.c_fc.bias',
 '5.mlp.c_proj.weight',
 '5.mlp.c_proj.bias',
 '6.ln_1.weight',
 '6.ln_1.bias',
 '6.attn.c_attn.weight',
 '6.attn.c_attn.bias',
 '6.attn.c_proj.weight',
 '6.attn.c_proj.bias',
 '6.ln_2.weight',
 '6.ln_2.bias',
 '6.mlp.c_fc.weight',
 '6.mlp.c_fc.bias',
 '6.mlp.c_proj.weight',
 '6.mlp.c_proj.bias',
 '7.ln_1.weight',
 '7.ln_1.bias',
 '7.attn.c_attn.weight',
 '7.attn.c_attn.bias',
 '7.attn.c_proj.weight',
 '7.attn.c_proj.bias',
 '7.ln_2.weight',
 '7.ln_2.bias',
 '7.mlp.c_fc.weight',
 '7.mlp.c_fc.bias',
 '7.mlp.c_proj.weight',
 '7.mlp.c_proj.bias',
 '8.ln_1.weight',
 '8.ln_1.bias',
 '8.attn.c_attn.weight',
 '8.attn.c_attn.bias',
 '8.attn.c_proj.weight',
 '8.attn.c_proj.bias',
 '8.ln_2.weight',
 '8.ln_2.bias',
 '8.mlp.c_fc.weight',
 '8.mlp.c_fc.bias',
 '8.mlp.c_proj.weight',
 '8.mlp.c_proj.bias',
 '9.ln_1.weight',
 '9.ln_1.bias',
 '9.attn.c_attn.weight',
 '9.attn.c_attn.bias',
 '9.attn.c_proj.weight',
 '9.attn.c_proj.bias',
 '9.ln_2.weight',
 '9.ln_2.bias',
 '9.mlp.c_fc.weight',
 '9.mlp.c_fc.bias',
 '9.mlp.c_proj.weight',
 '9.mlp.c_proj.bias',
 '10.ln_1.weight',
 '10.ln_1.bias',
 '10.attn.c_attn.weight',
 '10.attn.c_attn.bias',
 '10.attn.c_proj.weight',
 '10.attn.c_proj.bias',
 '10.ln_2.weight',
 '10.ln_2.bias',
 '10.mlp.c_fc.weight',
 '10.mlp.c_fc.bias',
 '10.mlp.c_proj.weight',
 '10.mlp.c_proj.bias',
 '11.ln_1.weight',
 '11.ln_1.bias',
 '11.attn.c_attn.weight',
 '11.attn.c_attn.bias',
 '11.attn.c_proj.weight',
 '11.attn.c_proj.bias',
 '11.ln_2.weight',
 '11.ln_2.bias',
 '11.mlp.c_fc.weight',
 '11.mlp.c_fc.bias',
 '11.mlp.c_proj.weight',
 '11.mlp.c_proj.bias'
\end{lstlisting}

For iGPT, the number of transformer block is twice (from 0-11 to 0-23) but the configuration is the same.

\section{Gradient Analysis}
\label{appendix:gradient-analysis}

\subsection{Details of Experiments}
\label{appendix:detail-of-experiments-gradient-analysis}
We randomly sample 100 samples for gradient norm in Fig. \ref{fig:grad-norm} and 50 samples for gradient confusion in Fig. \ref{fig:grad-confusion} and compute the gradient of the loss on each of these samples for all parameters of the models fine-tuned after 1 epoch. The bar plot of Fig. \ref{fig:grad-confusion} is the minimum of the $50 \times 50 = 2500$ cosine similarities. Each point of Fig. \ref{fig:grad-norm} corresponds to the gradient norm of each data sample in 100 gradient norms. The top of the box is the 1st quartile and the bottom of the box is the 3rd quartile of the points. The whiskers extend from the box by 1.5 times the inter-quartile range. For gradient clipping, we use a method in Pytorch \citep{Paszke19} and set the maximum norm to be $0.25$, following the previous work \citep{reid2022can}. 

The process of creating Fig. \ref{fig:grad-norm-param-ratio} is as follows. We first compute gradient norms per parameter set and examine the bar plot that we will explain in Appendix \ref{appendix:gradient-norm-for-each-parameter}. Noticing that two peaks exist in the figure, we check the label of the parameter sets and find that they are \lstinline{0.ln_1.weight} and \lstinline{0.ln_1.bias}. Thus, we lump the remaining as \lstinline{others} and only show the ratio in the main body of the paper to tell readers the main finding from the observation in an easy-to-understand manner. The parameter sets considered to compute the parameter set-wise gradient norms are the same as those in Appendix \ref{appendix:parameter-name}.

\subsection{Gradient Norm for Each Parameter}
\label{appendix:gradient-norm-for-each-parameter}
For visibility, we show only the ratio in Section \ref{section:gradient-analysis} as Fig. \ref{fig:grad-norm-param-ratio}. The full results for gradient norms of each parameter are shown in Fig. \ref{fig:param-norm-per-param}. The labels of the ticks on the x-axis are the same as that in Appendix \ref{appendix:parameter-name}. We immediately notice that the first two parameters (\lstinline{0.ln_1.weight} and \lstinline{0.ln_1.bias}) are much larger than the remains.

\begin{figure}[H]
    \centering
        \includegraphics[width=\linewidth]{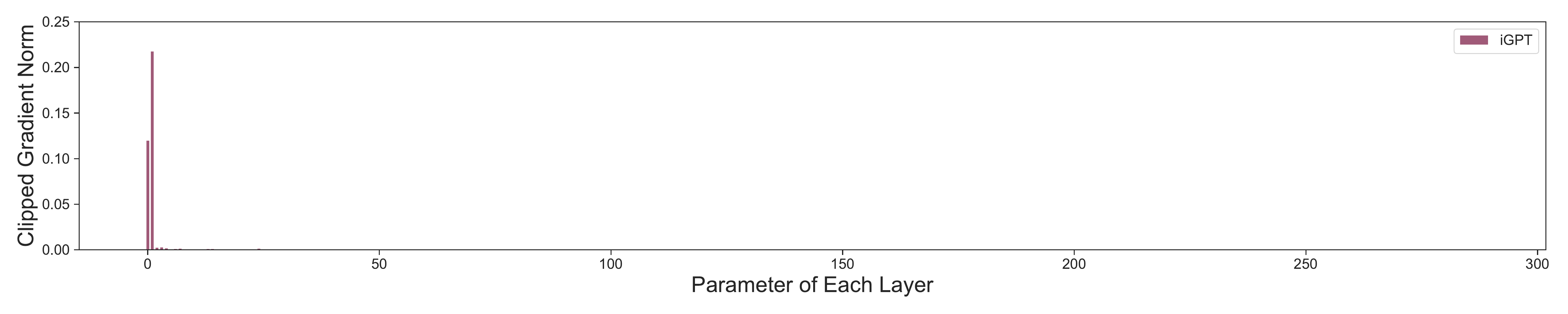}
    \caption{Gradient norm of iGPT's each parameter at epoch 1.}
    \label{fig:param-norm-per-param}
\end{figure}

\subsection{Analysis of the Effect of Gradient Clipping}
\label{appendix:gradient-clipping}
In Section \ref{section:gradient-analysis}, we noted that large gradient and gradient clipping could be a cause of the bad performance of iGPT. To further explore this point, we conducted a simple experiment.

The experiments of the previous study \citep{reid2022can} and our experiments both used a Pytorch function called \lstinline{torch.nn.utils.clip_grad_norm_}
\footnote{
Documentation of \lstinline{torch.nn.utils.clip_grad_norm_} : \\
\href{https://pytorch.org/docs/stable/generated/torch.nn.utils.clip_grad_norm_.html}{https://pytorch.org/docs/stable/generated/torch.nn.utils.clip\_grad\_norm\_.html}
}
for gradient clipping. This function divides each gradient by the total norm of all gradients and multiplies the clipping constant (gradient norm clipping). However, as we pointed out in Section \ref{section:gradient-analysis}, the gradient norm values are dominated by only a few parameters (the layer normalization layer in the first block). This large norm affects the normalization of all gradients, decreasing the informational value of most gradients. Therefore, we hypothesized that one of the possible reasons why iGPT is difficult to learn can be the use of \lstinline{torch.nn.utils.clip_grad_norm_ function}. 

\begin{figure}[h]
    \centering
    \begin{minipage}[b]{0.32\linewidth}
        \includegraphics[width=\linewidth]{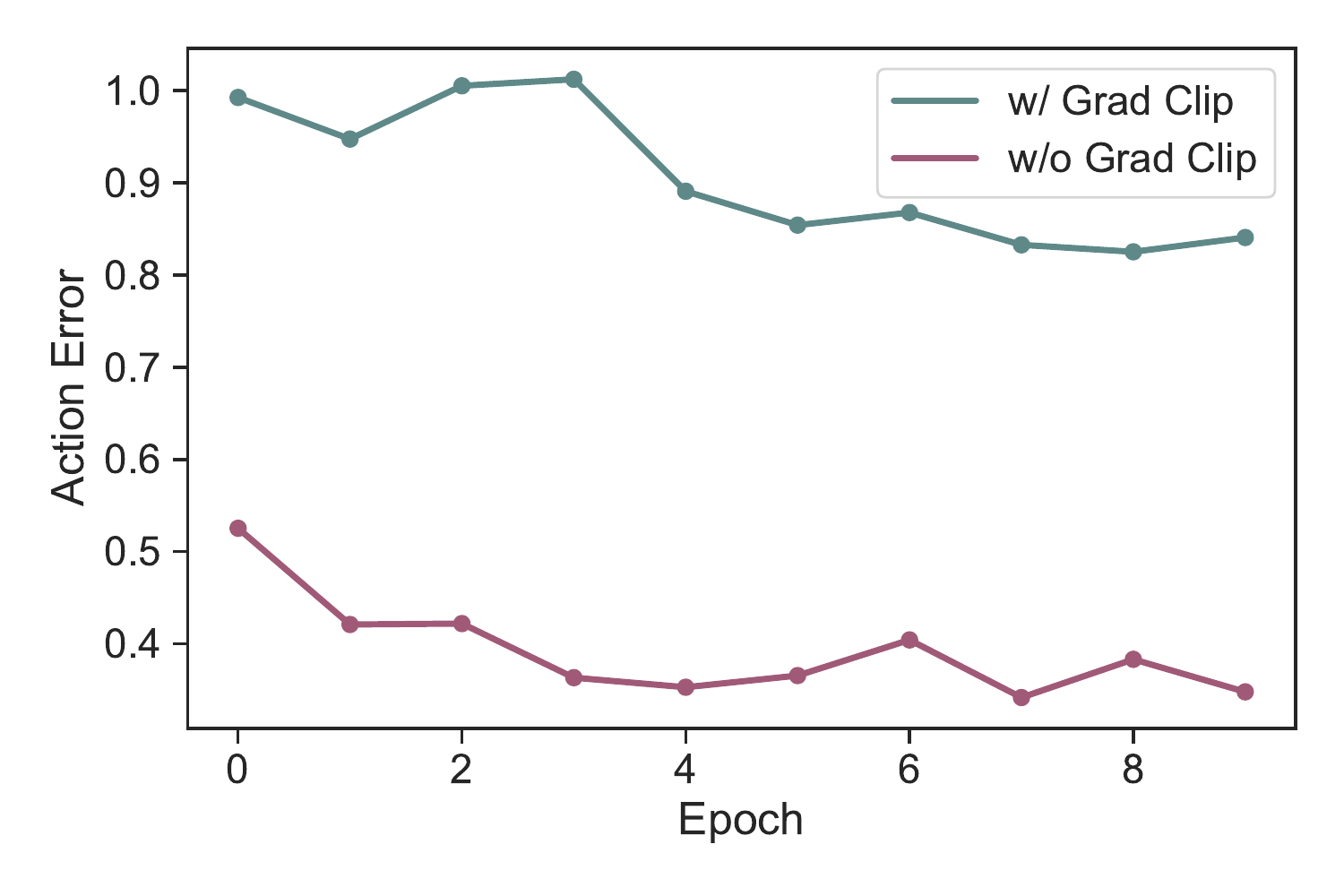}
        \subcaption{Hopper}
    \end{minipage}
    \begin{minipage}[b]{0.32\linewidth}
        \includegraphics[width=\linewidth]{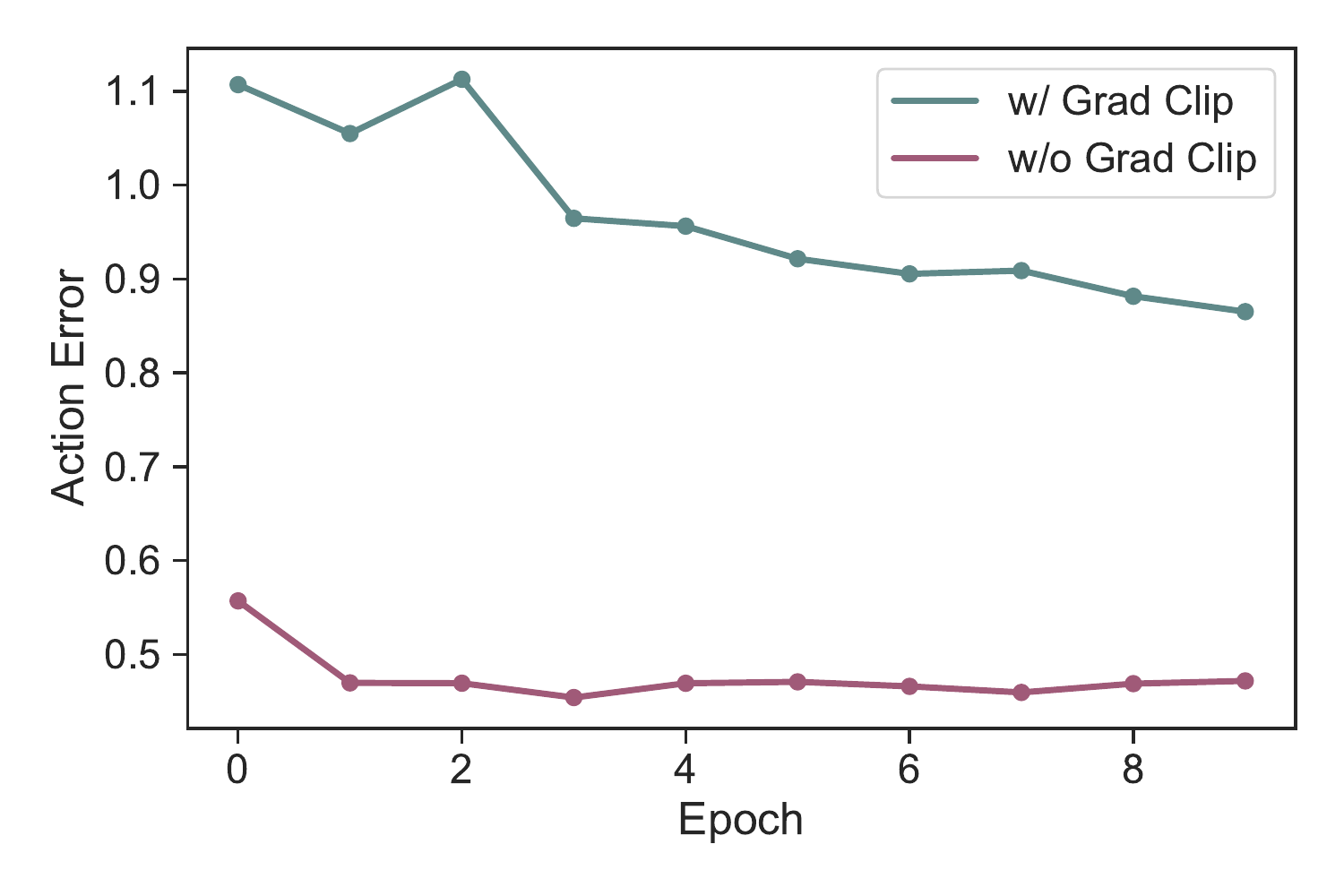}
        \subcaption{HalfCheetah}
    \end{minipage}
    \begin{minipage}[b]{0.32\linewidth}
        \includegraphics[width=\linewidth]{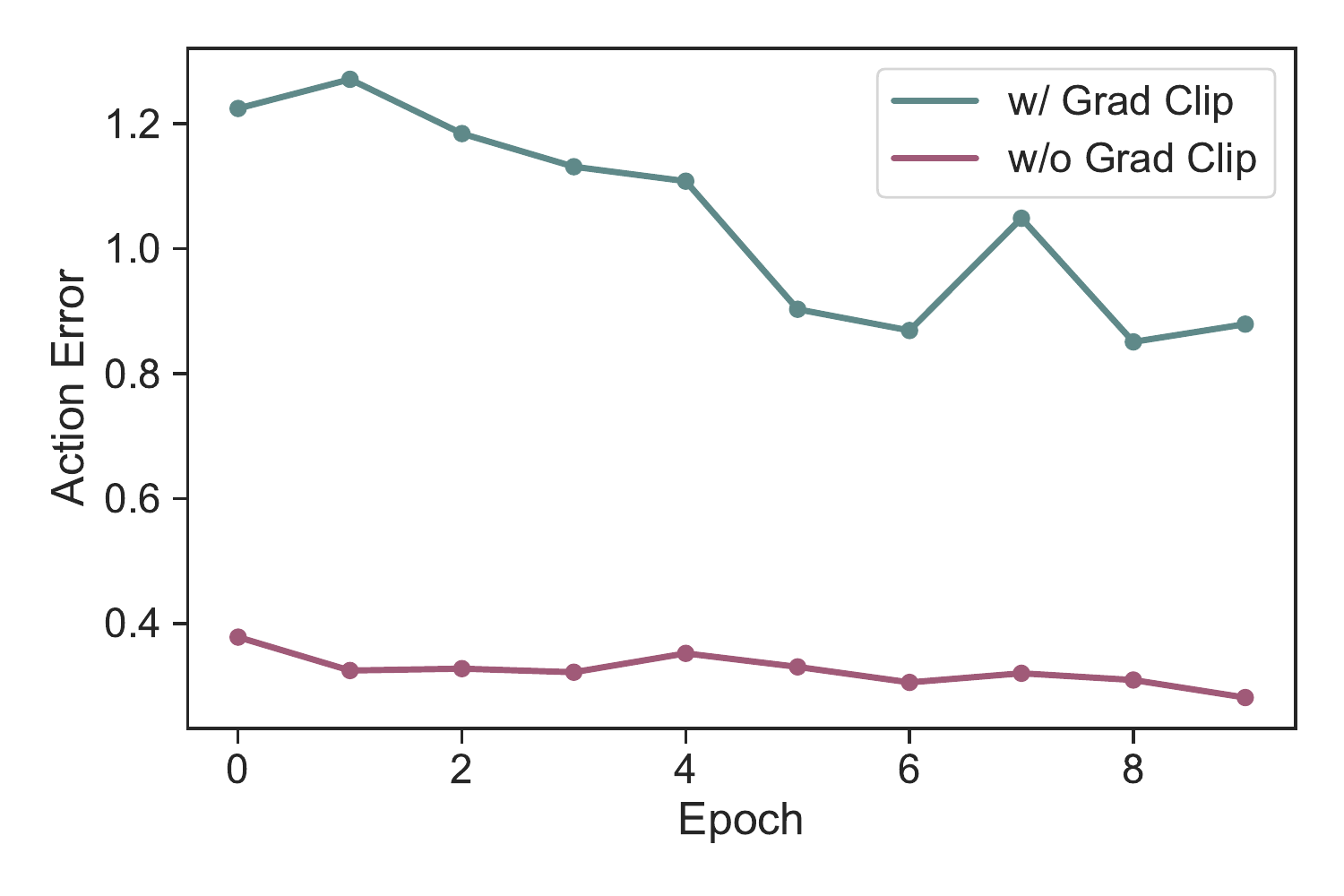}
        \subcaption{Walker2D}
    \end{minipage}
    \caption{Action error: with gradient clipping v.s. without gradient clipping}
    \label{fig:action-error-igpt-no-grad-clip}
\end{figure}

\begin{figure}[h]
    \centering
    \begin{minipage}[b]{0.32\linewidth}
        \includegraphics[width=\linewidth]{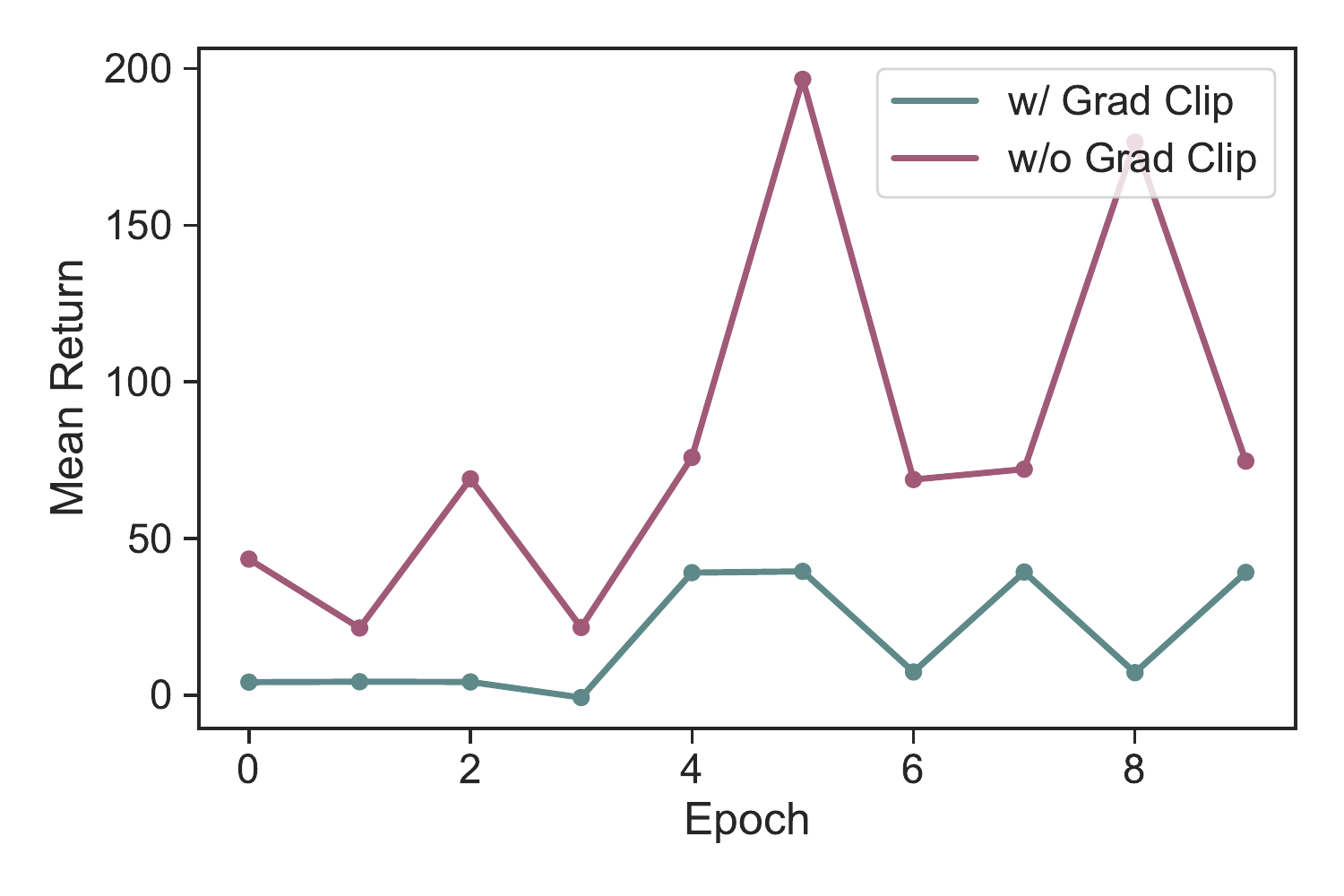}
        \subcaption{Hopper}
    \end{minipage}
    \begin{minipage}[b]{0.32\linewidth}
        \includegraphics[width=\linewidth]{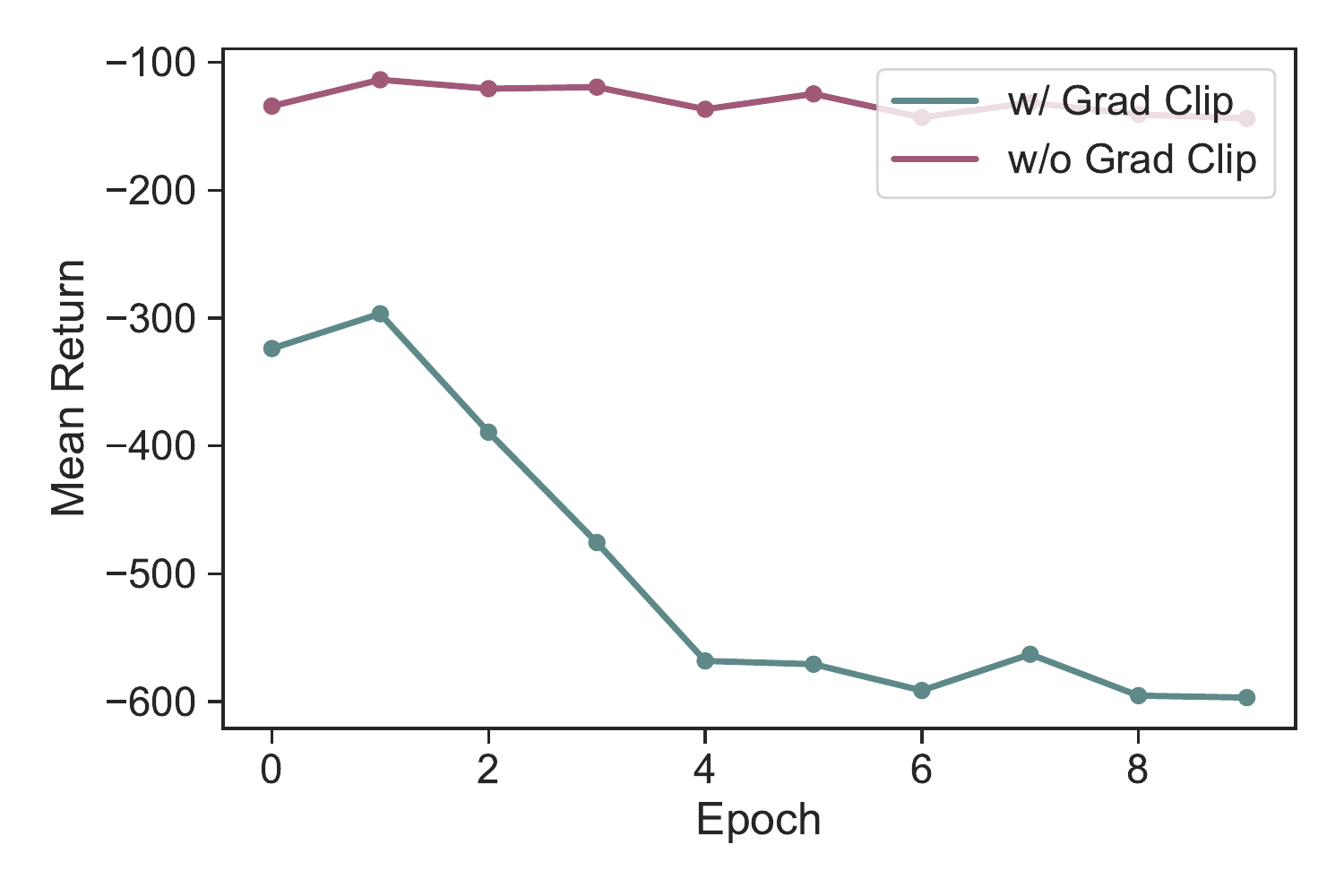}
        \subcaption{HalfCheetah}
    \end{minipage}
    \begin{minipage}[b]{0.32\linewidth}
        \includegraphics[width=\linewidth]{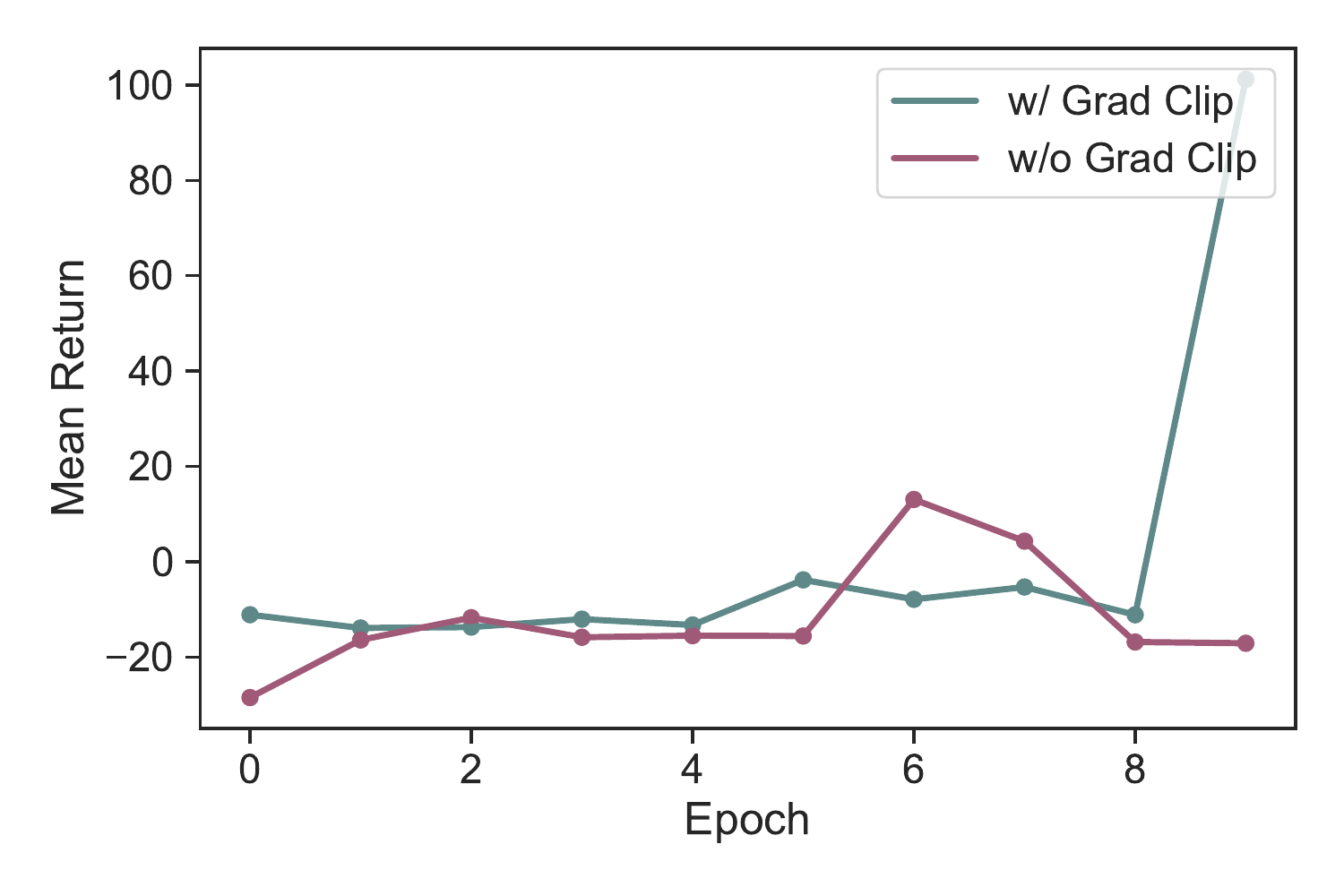}
        \subcaption{Walker2D}
    \end{minipage}
    \caption{Mean return: with gradient clipping v.s. without gradient clipping}
    \label{fig:mean-return-igpt-no-grad-clip}
\end{figure}

Thus we trained the image-pre-trained model without using gradient clipping. The basic experimental setting is the same as that of Section \ref{section:gradient-analysis} except that we trained the model for 10 epochs, instead of 40 epochs. The results are shown in Figs \ref{fig:action-error-igpt-no-grad-clip} and \ref{fig:mean-return-igpt-no-grad-clip} for action error and mean return, respectively. The x-axis is the epoch. These figures show that eliminating gradient clipping does not immediately solve the catastrophic performance at least up to 10 epochs. However, we did find that action error is much smaller for no gradient clipping condition, indicating that the learning process of it seems to be more stable and efficient than the clipping was applied. We also found that the return means seem to improve, albeit very slightly. Thus, we can conclude that gradient clipping is a cause of the bad performance of iGPT, though it does not seem to be critical. Exploring the critical factor that makes iGPT performance catastrophic is left a future work.

\section{Fine-Tuning with No Context Information}
\label{appendix:dependency-on-context-informaiton}

\subsection{Details of Experiments}
\label{appendix:detail-of-experiments-dependency-on-context-informaiton}
The configuration of the training is the same as that described in Appendix \ref{appendix:training-detail} except that the context is $1$ (no context). We evaluate the model per epoch and the normalized score in Table \ref{table:k=1-666-42} is the result of the best checkpoint in the 40 checkpoints. The mean return reported in Fig. \ref{fig:return-mean} and Table \ref{table:k=1-666-42} is the average of the returns over the trajectory. Table \ref{table:k=1-666-42} is the result of one random seed (seed = 666). The mean and standard deviation of the normalized score for the two seeds (seed = 666 \& 42) and the mean return of another seed (seed = 42) are reported in Appendix \ref{apendx:results-for-other-conditions-fine-tuning-with-no-context-information}.

In this paper, we mean \textit{context} by the number of accessible time steps for a model to predict, following the previous study \citep{chen2021decision}. If the context length is $K$, the model can use inputs in past $K$ steps to predict the action at the current step, while the context length is 1 (no context), the model has to predict only from the current time step. 

\begin{figure}[H]
    \centering
    \begin{minipage}[b]{0.4\linewidth}
    \includegraphics[width=\linewidth]{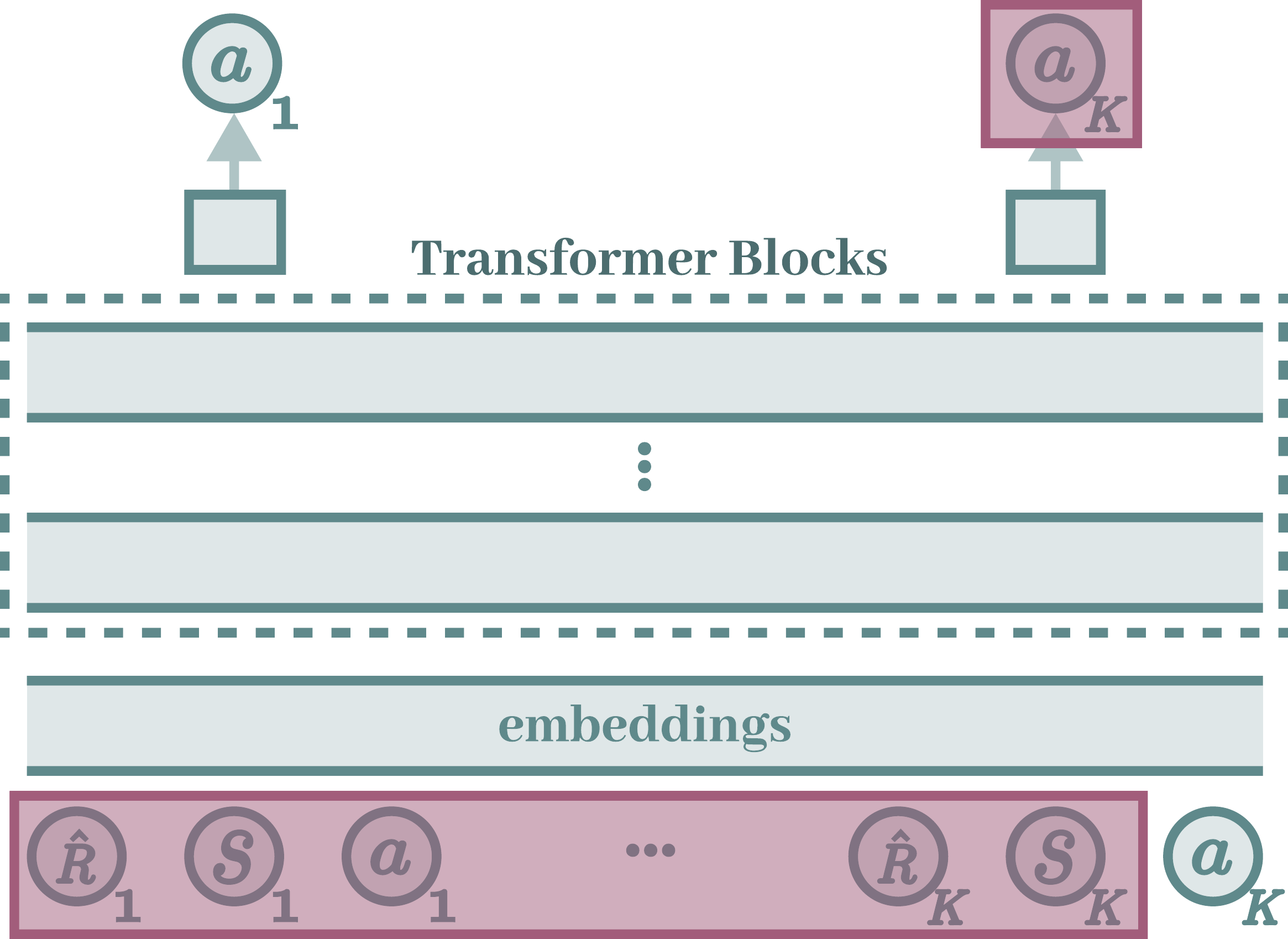}
    \subcaption{Context length is $K$}
    \end{minipage}
    \hspace{1cm}
    \begin{minipage}[b]{0.4\linewidth}
    \includegraphics[width=\linewidth]{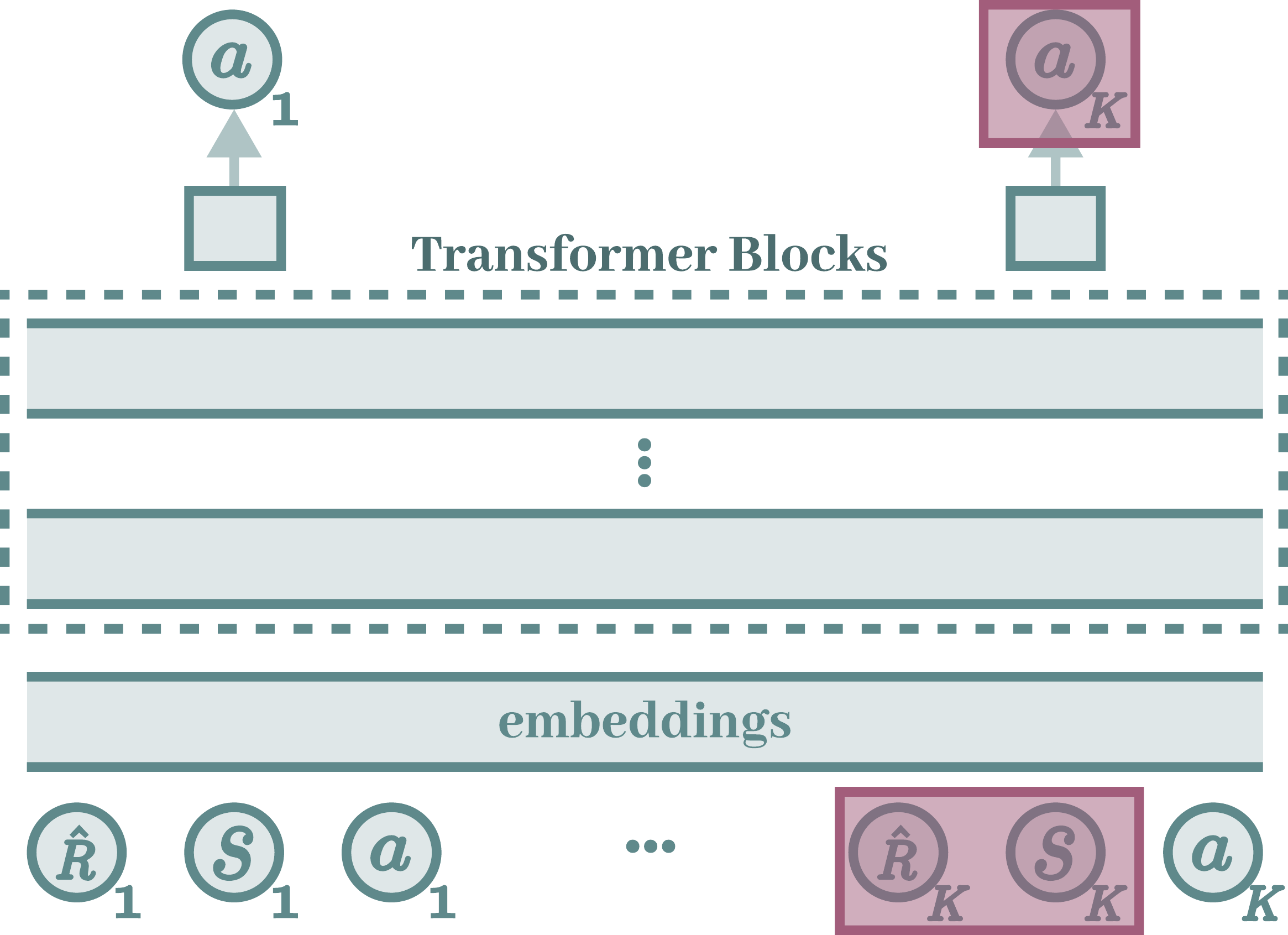}
    \subcaption{Context length is $1$}
    \end{minipage}
    \caption{Descriptive diagram to explain what we mean by \textit{context}.}
    \label{fig:diagram_context}
\end{figure}

\subsection{Analysis of Why Randomly Initialized Model Fails for Hopper with No Context}

In Section \ref{section:dependency-on-context-informaiton}, we observe that the performance of randomly initialized models for the Hopper environment is particularly worse than that for other environments (Table \ref{table:k=1-666-42}). In this Section, we further explore a possible cause of this observation.

We speculate that this may be because \textit{how much of the range of context that needs to be looked at} changes depending on the data set. For example, prior studies on decision transformer reported that the effect of context varied depending on the task \citep{chen2021decision}. Improvement by context means that the context is important information for solving the task. Thus, it could be possible that Hopper is the dataset that requires more information from context than the other two data sets.

As a test of the hypothesis, we randomly sampled a batch sample and calculated the mutual information between action and state or return-to-go at the same time step and compare them between different environments; note that this is the mutual information between the data, not between the data and representation. The higher the value of the mutual information, the higher the mutual dependence between state or return-to-go and action at the same time step. Hence, higher mutual information suggests that the model could predict action better only from the information at the current time step. 

\begin{figure}[h]
    \centering
    \begin{minipage}[b]{0.48\linewidth}
        \includegraphics[width=\linewidth]{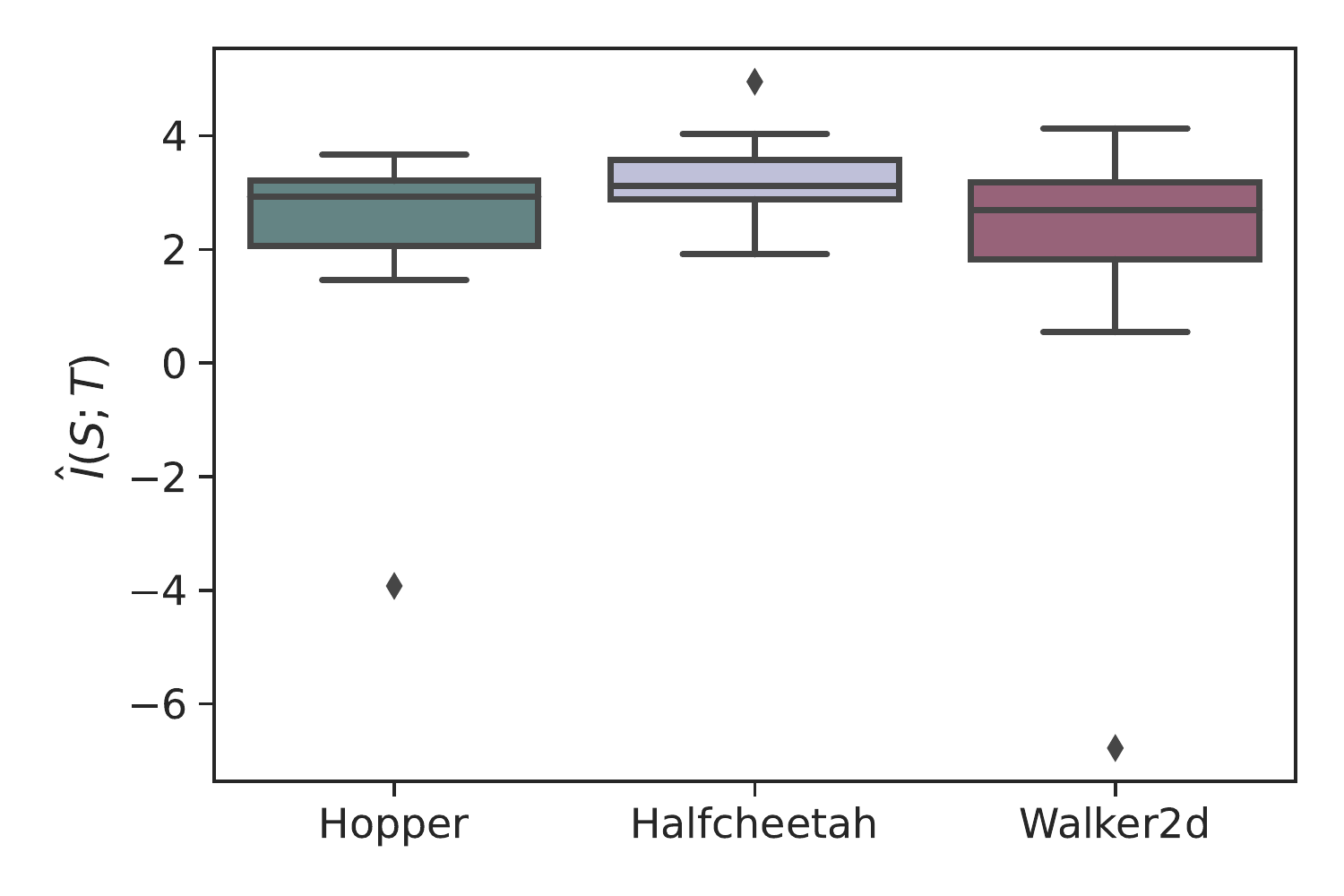}
        \subcaption{$\hat{I}(\bm{s}_t; \bm{a}_t)$}
    \end{minipage}
    \begin{minipage}[b]{0.48\linewidth}
        \includegraphics[width=\linewidth]{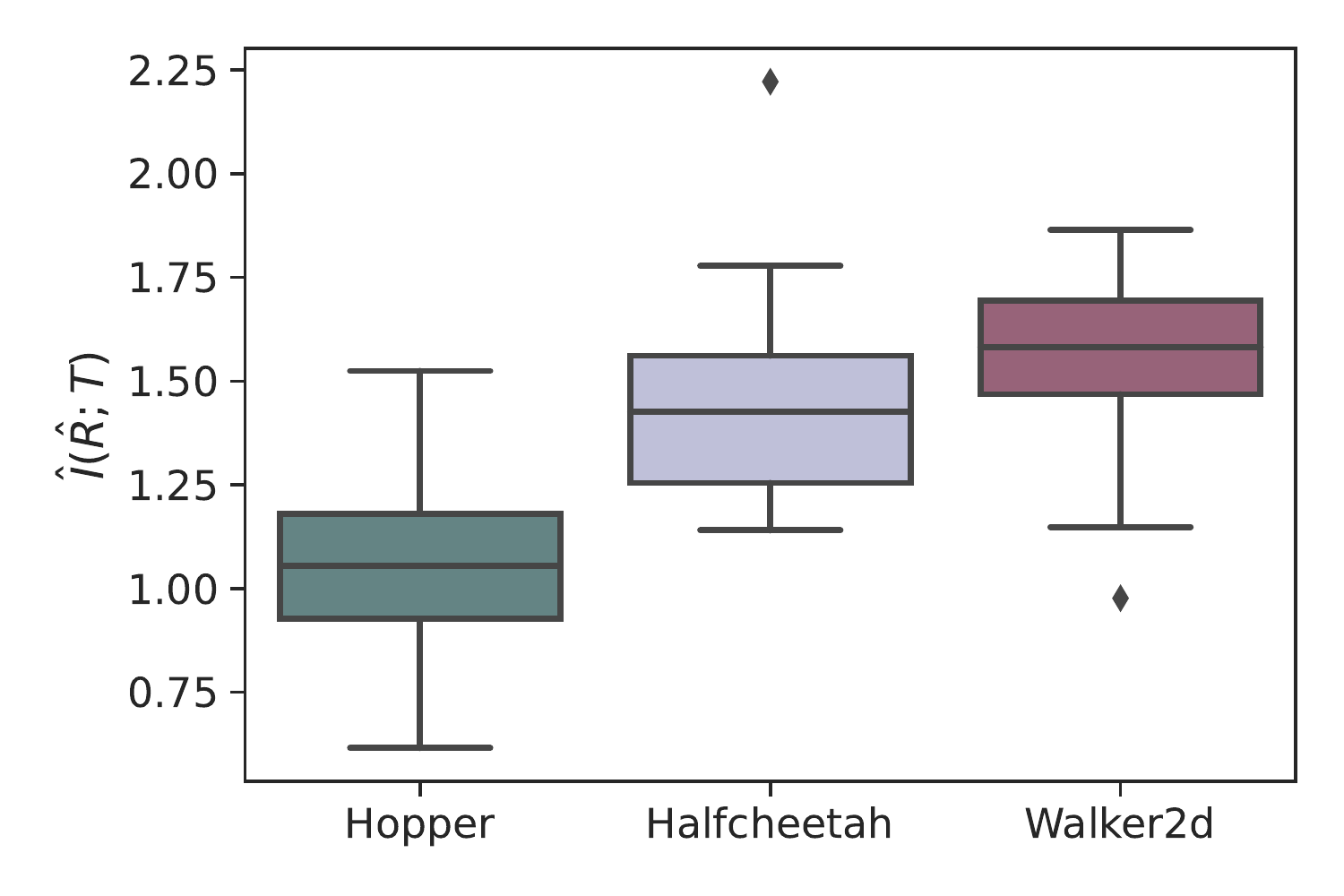}
        \subcaption{$\hat{I}(\hat{R}_t; \bm{a}_t)$}
    \end{minipage}
    \caption{Mutual information between data at the same time step in trajectories}
    \label{fig:mutual-information-data}
\end{figure}

In particular, we sample 100 samples of context length $K = 20$. Then, for all time step $t$, we compute the estimated mutual information $\hat{I}(\bm{s}_t; \bm{a}_t)$ between state $\bm{s}_t$ and action $\bm{a}_t$ and that $\hat{I}(\hat{R}_t; \bm{a}_t)$ between return-to-go $\hat{R}_t$ and action. Other setup for mutual information estimation is the same as that of Section \ref{section:mutual-information-between-hidden-representation-and-input-and-label}. The result is shown in Fig \ref{fig:mutual-information-data}, where (a) is the box plot for the state and action and (b) is that for the return-to-go and action. The top of the box is the 1st quartile and the bottom of the box is the 3rd quartile of the points. The whiskers extend from the box by 1.5 times the inter-quartile range. Each point corresponds to different time step in the context. 

We observe that the estimated mutual information between return-to-go and action is smaller on average for Hopper than for other environments (Fig. \ref{fig:mutual-information-data} (b)), though that between action and state does not differ that much (Fig. \ref{fig:mutual-information-data} (a)). Prior research has indicated that return-to-go information seems to be important for prediction \citep{reid2022can}. Hence, we can say that models have to use more information from the other steps in the context to solve the Hopper task than models do for other environments. This result supports our hypothesis above.

\section{More In-Depth Analysis of Context Dependence}
\label{appendix:internal-analysis-to-see-the-dependence-on-context}

\subsection{Replacement by the Pre-Trained Block}
\label{appendix:replacement}

\subsubsection{Details of Experiments}
\label{appendix:detail-of-experiments-replacement}
The training configuration is the same as that of Section \ref{section:dependency-on-context-informaiton}.
The smoothing factor of the exponential moving average for Fig. \ref{fig:learning_curve} is 0.8. The action error is the mean square loss between the true action $\bm{a}_t$ and the predicted action $\hat{\bm{a}}_t$. Considering that the goal of this experiment is to highlight the effect of pre-training, we train the model for 10 epochs for Hopper and 5 epochs for HalfCheetah and Walker2D because Fig. \ref{fig:return-mean} shows that the difference between the randomly initialized and pre-trained model is evident during these epochs. 

\subsection{Attention Distance Analysis}
\label{appendix:attention-distance}

\subsubsection{Details of Experiments}
\label{appendix:detail-of-experiments-attention-distance}
In the main body of this paper, we mean \textit{attention distance} by \textit{average/mean attention distance} in the previous work \citep{dosovitskiy2020image,raghu2021vision}. Because we use only one attention head in this study, there is a single point per sample in Fig. \ref{fig:attention_distance_before_after_epoch_4}. We randomly sample 100 trajectory samples and compute attention for these trajectories. When calculating attention distance, we allow the model to access the context of $K=20$ length and compute the attention distance up to the length. Each point of the box plot (Fig. \ref{fig:attention_distance_before_after_epoch_4}) corresponds to the attention distance of each sample. The configuration for the box plot is the same as that in Appendix \ref{appendix:gradient-analysis}.

\newpage

\section{Results for Other Conditions}
\label{appendix:results-for-other-conditions}

\subsection{Activation Similarity}
\label{appendix:results-for-other-conditions-activation-similarity}

\subsubsection{CKA Between Pre and Post-Fine-Tuning}
\label{appendix:results-for-other-conditions-cka-between-pre-and-post-fine-tuning}

\begin{figure}[H]
    \centering
    \begin{minipage}[b]{0.32\linewidth}
        \includegraphics[width=\linewidth]{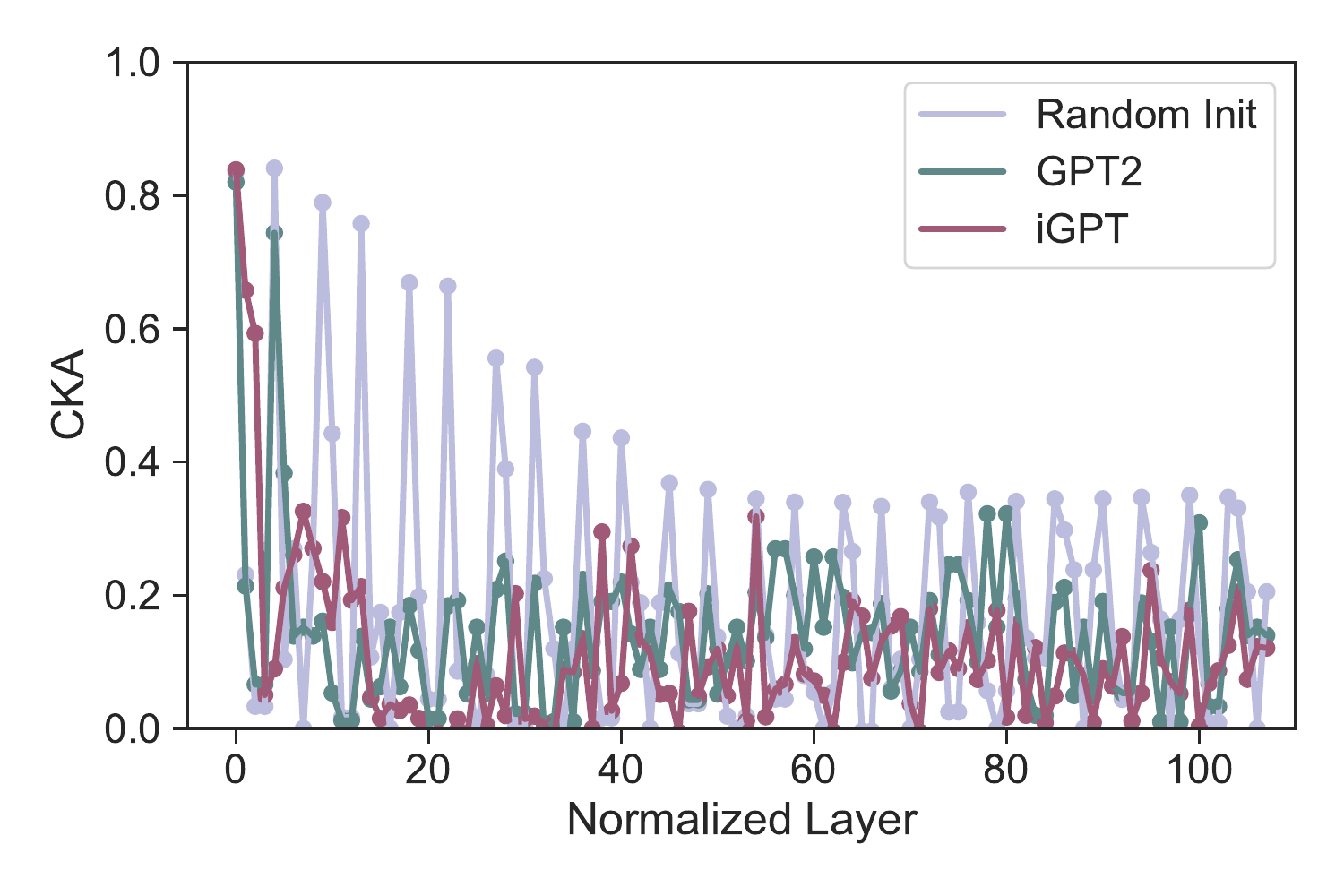}
        \subcaption{Hopper}
    \end{minipage}
    \begin{minipage}[b]{0.32\linewidth}
        \includegraphics[width=\linewidth]{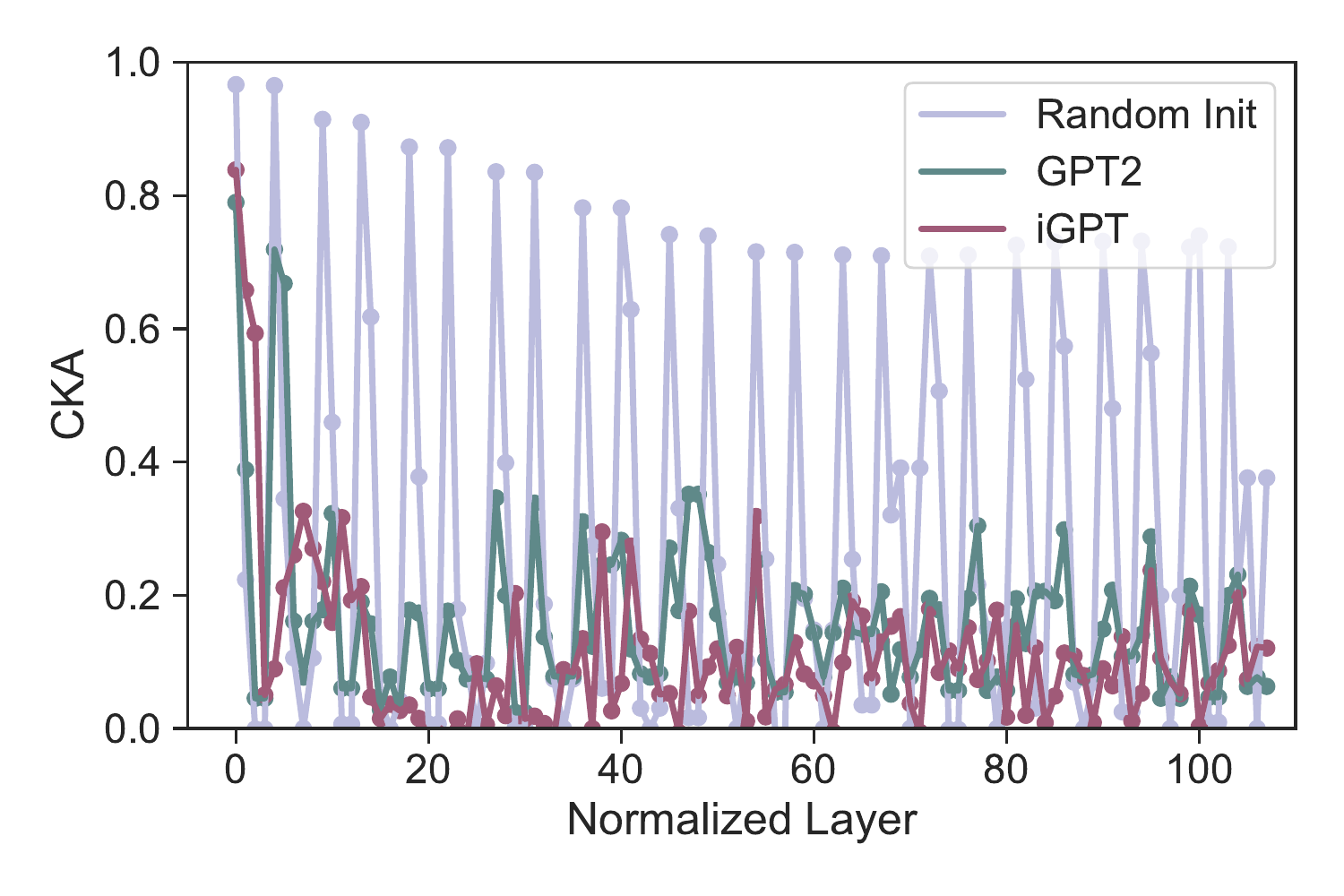}
        \subcaption{HalfCheetah}
    \end{minipage}
    \begin{minipage}[b]{0.32\linewidth}
        \includegraphics[width=\linewidth]{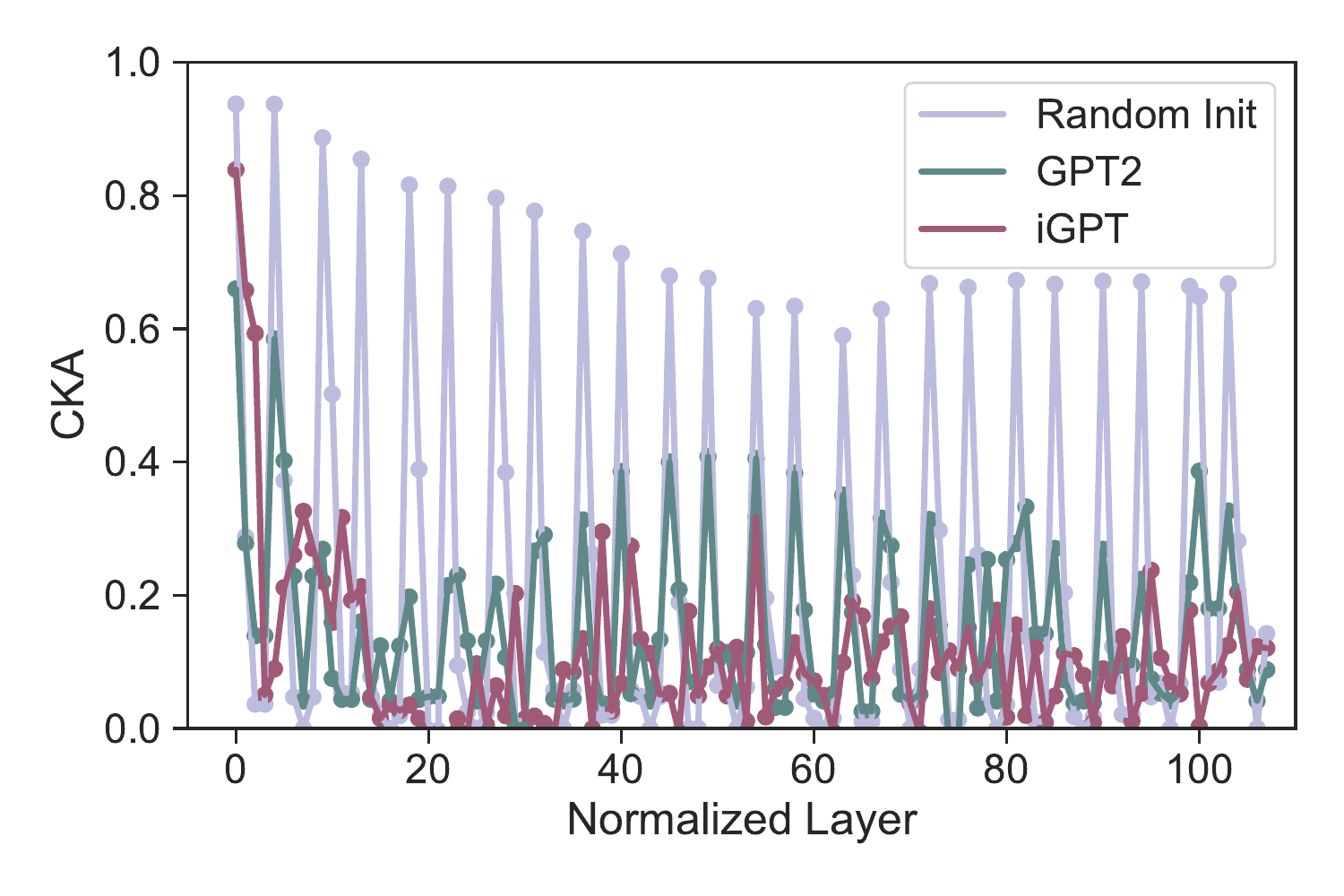}
        \subcaption{Walker2D}
    \end{minipage}
    \caption{CKA similarity of each layer between pre and post-fine-tuning (Action).}
\end{figure}

\begin{figure}[H]
    \centering
    \begin{minipage}[b]{0.32\linewidth}
        \includegraphics[width=\linewidth]{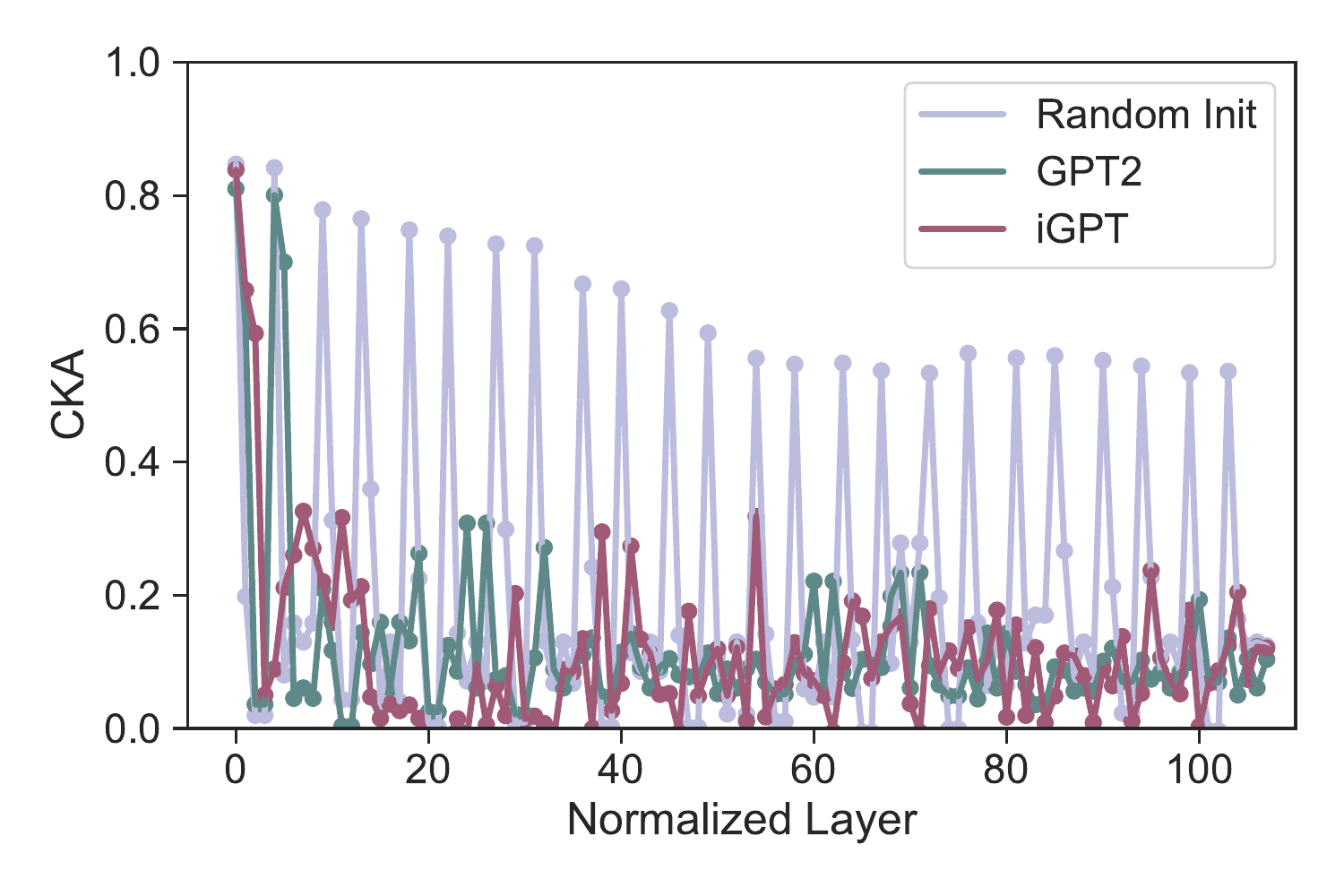}
        \subcaption{Hopper}
    \end{minipage}
    \begin{minipage}[b]{0.32\linewidth}
        \includegraphics[width=\linewidth]{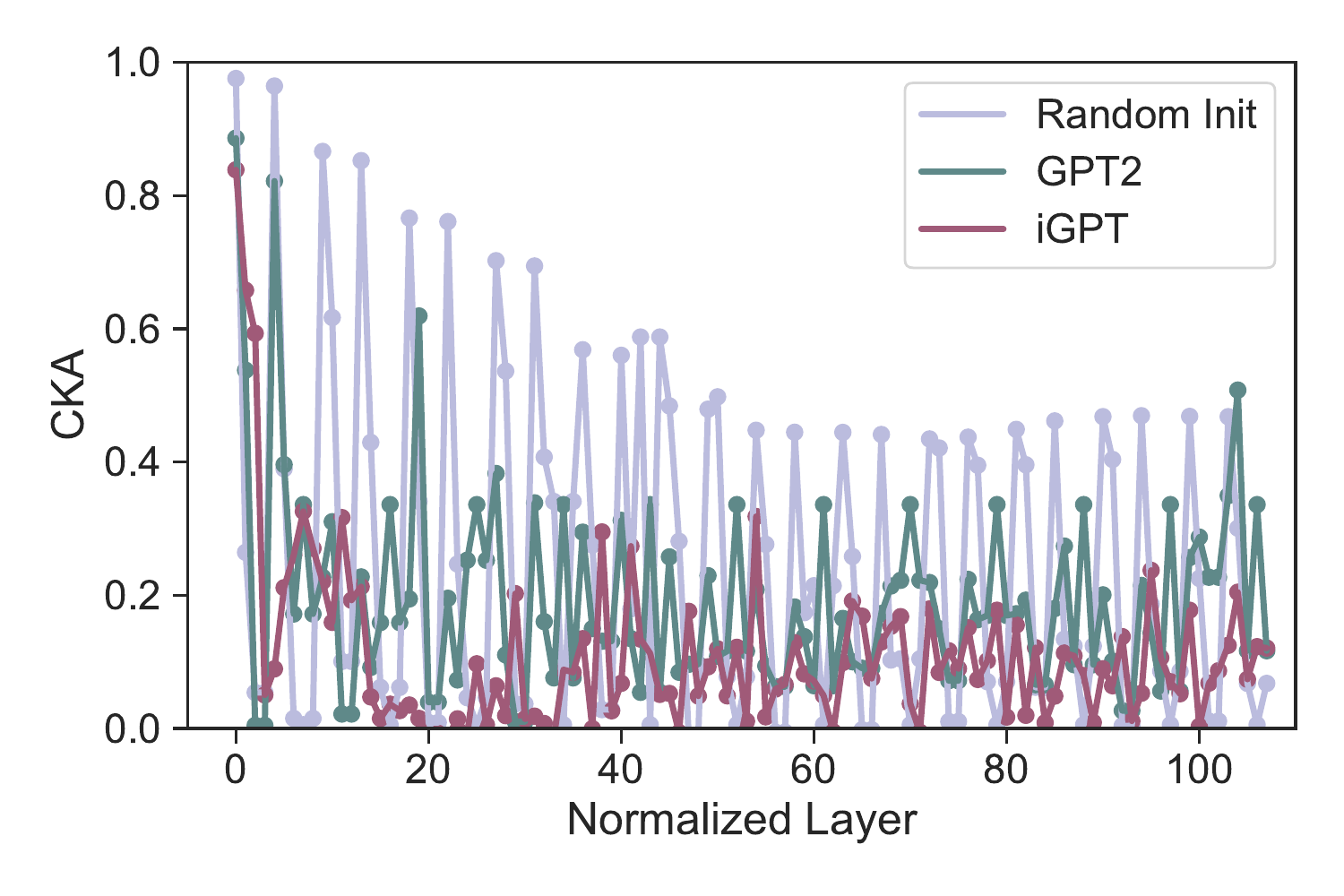}
        \subcaption{HalfCheetah}
    \end{minipage}
    \begin{minipage}[b]{0.32\linewidth}
        \includegraphics[width=\linewidth]{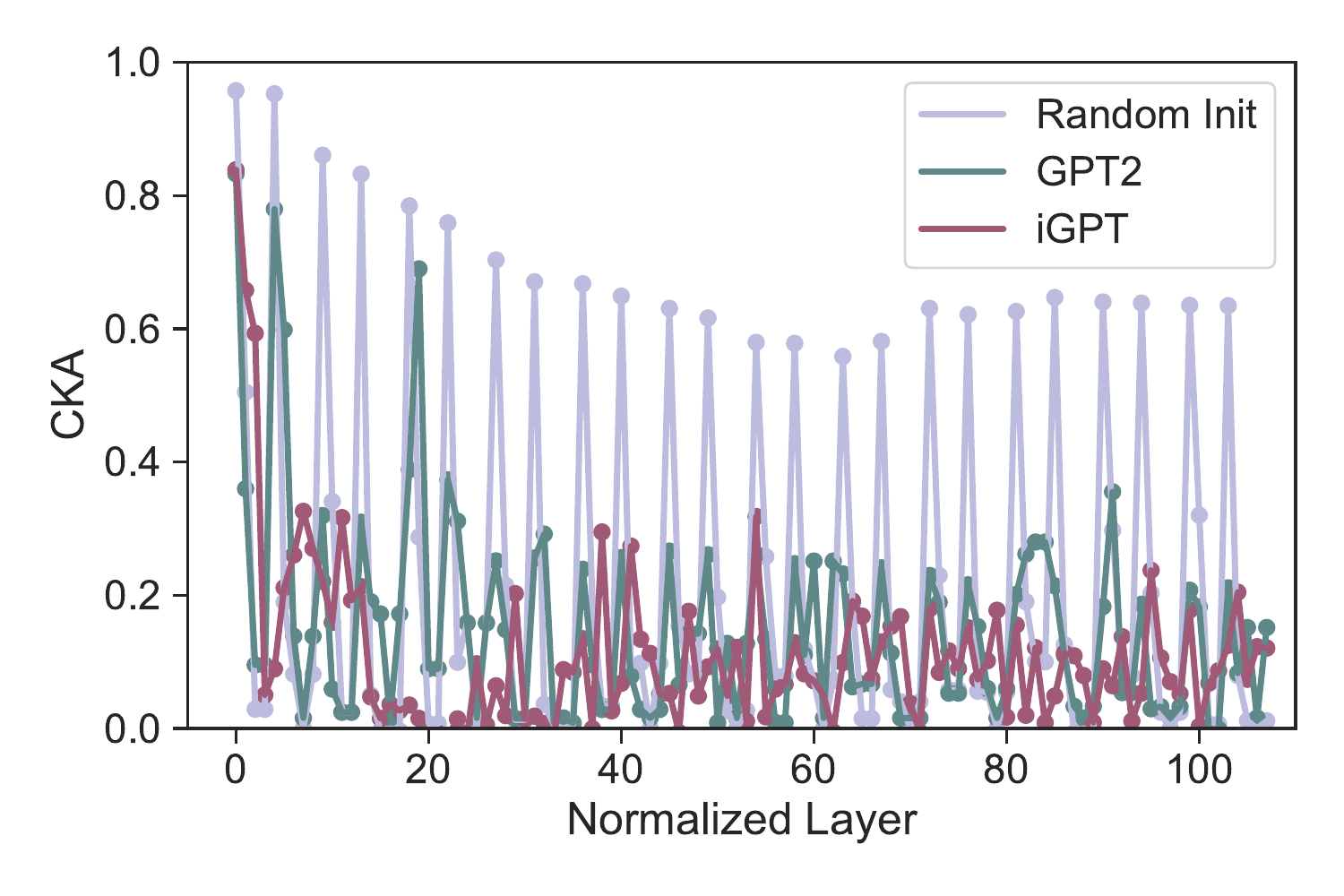}
        \subcaption{Walker2D}
    \end{minipage}
    \caption{CKA similarity of each layer between pre and post-fine-tuning (Return-to-go).}
\end{figure}

\subsubsection{CKA Between Pre and Post-Fine-Tuning (Seed = 42)}

\begin{figure}[H]
    \centering
    \begin{minipage}[b]{0.32\linewidth}
        \includegraphics[width=\linewidth]{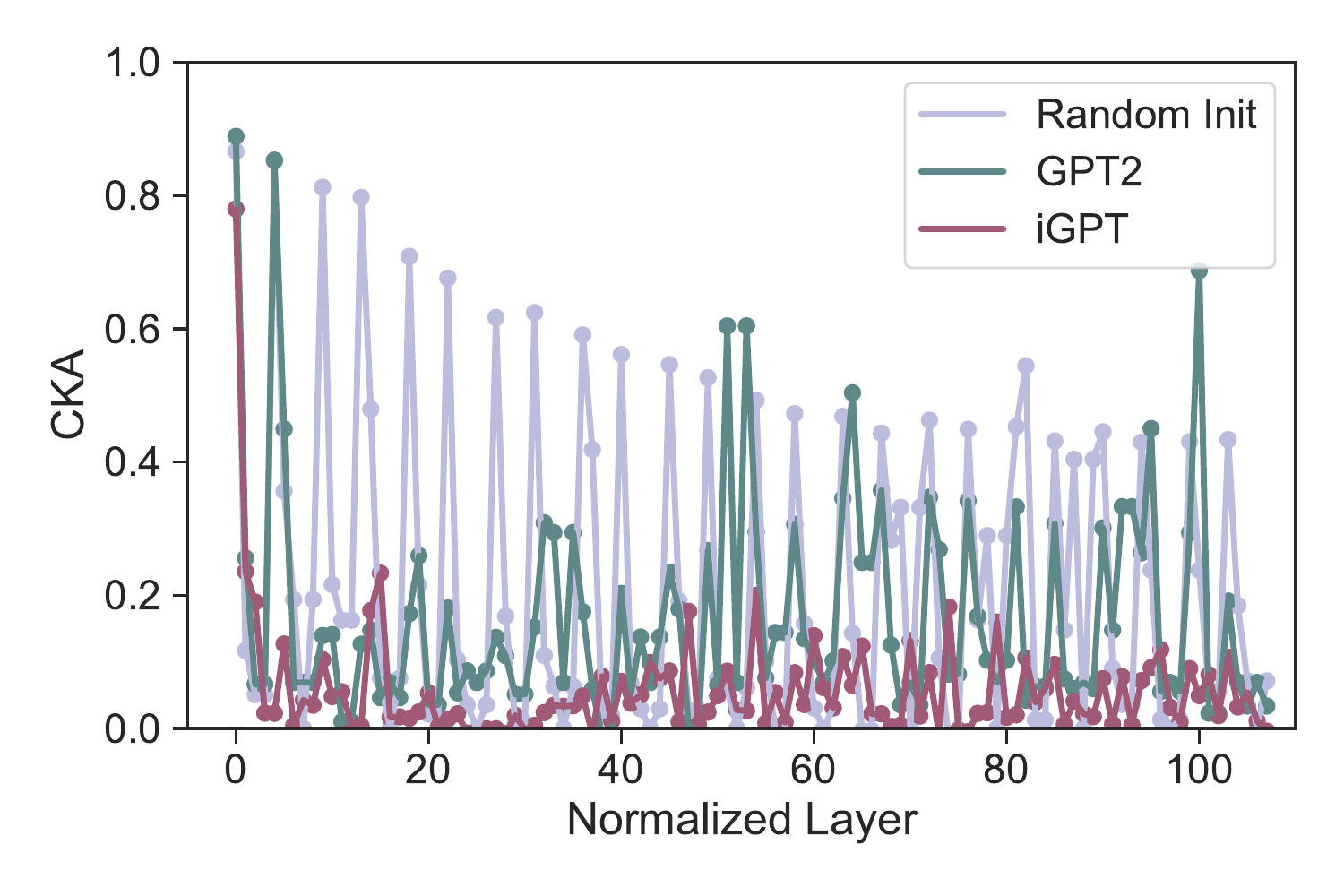}
        \subcaption{Hopper}
    \end{minipage}
    \begin{minipage}[b]{0.32\linewidth}
        \includegraphics[width=\linewidth]{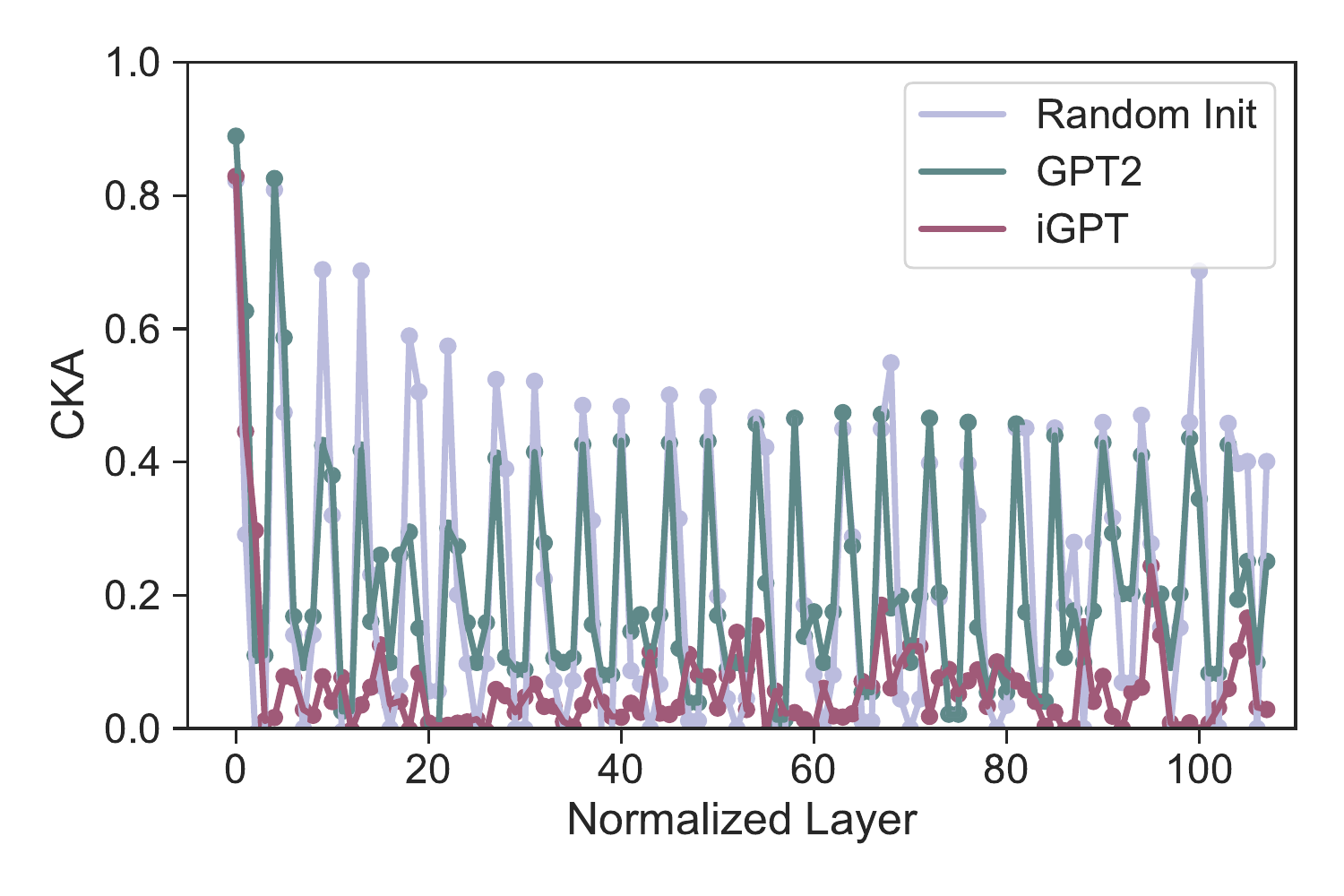}
        \subcaption{HalfCheetah}
    \end{minipage}
    \begin{minipage}[b]{0.32\linewidth}
        \includegraphics[width=\linewidth]{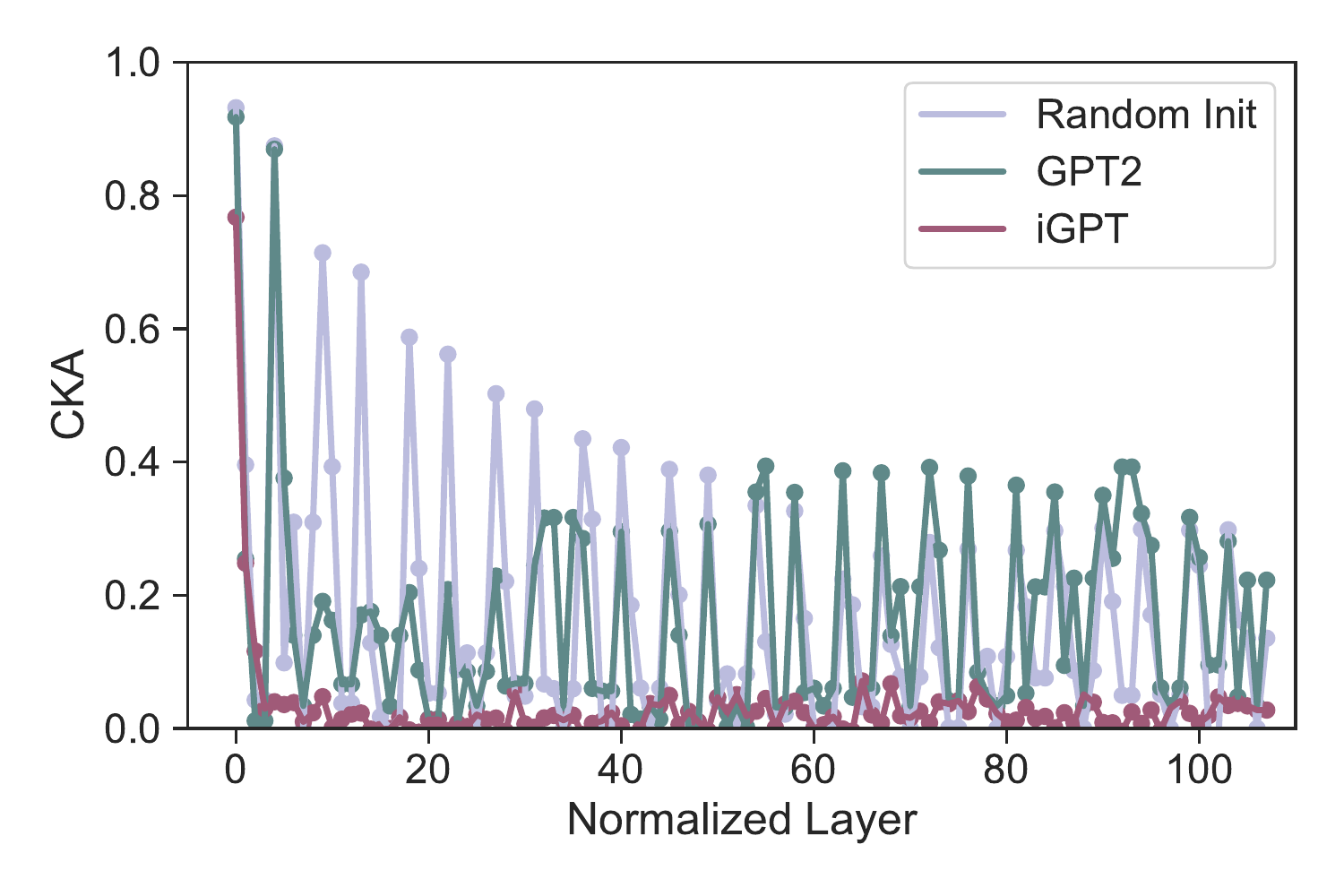}
        \subcaption{Walker2D}
    \end{minipage}
    \caption{CKA similarity of each layer between pre and post-fine-tuning (State, Seed = 42).}
\end{figure}

\begin{figure}[H]
    \centering
    \begin{minipage}[b]{0.32\linewidth}
        \includegraphics[width=\linewidth]{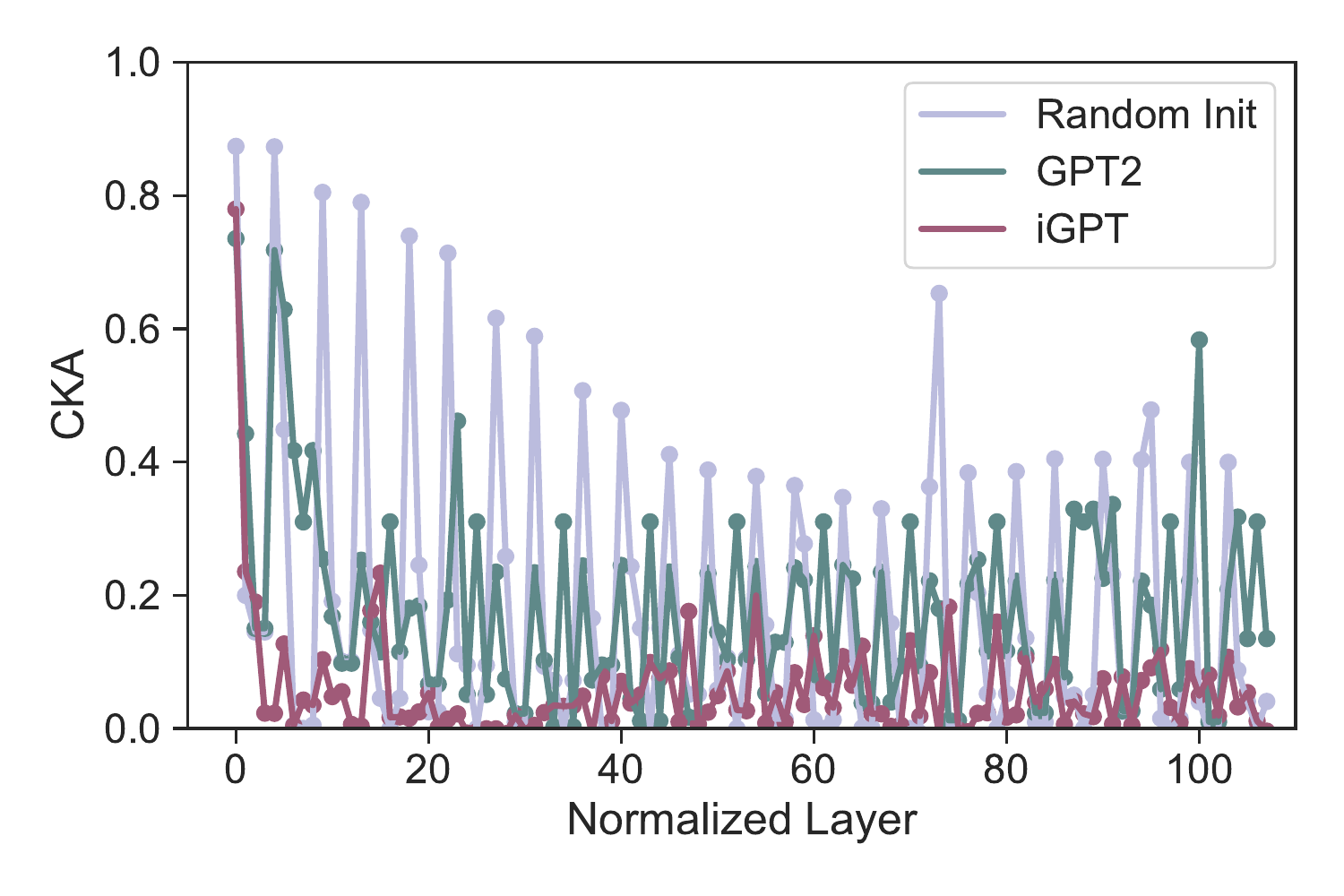}
        \subcaption{Hopper}
    \end{minipage}
    \begin{minipage}[b]{0.32\linewidth}
        \includegraphics[width=\linewidth]{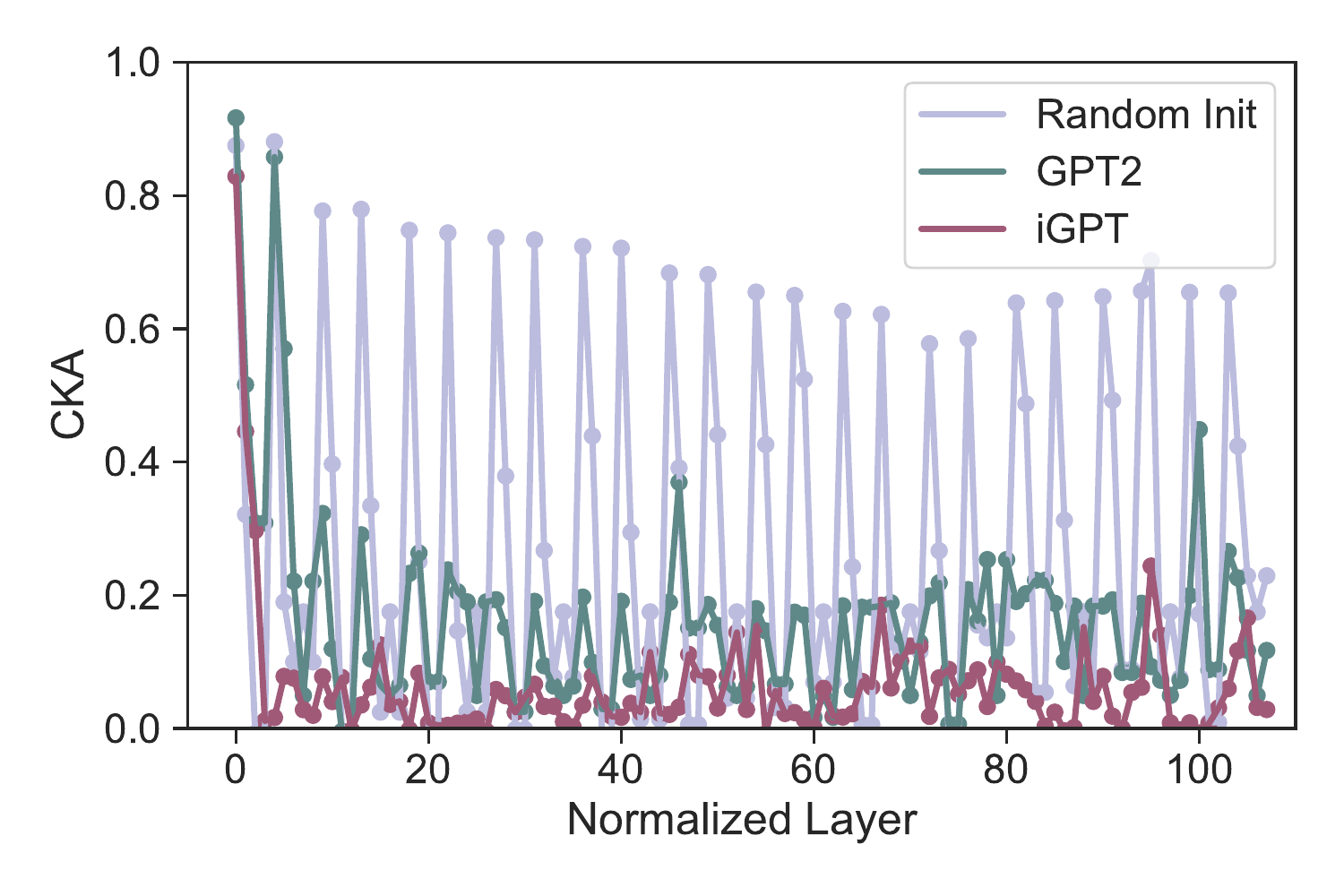}
        \subcaption{HalfCheetah}
    \end{minipage}
    \begin{minipage}[b]{0.32\linewidth}
        \includegraphics[width=\linewidth]{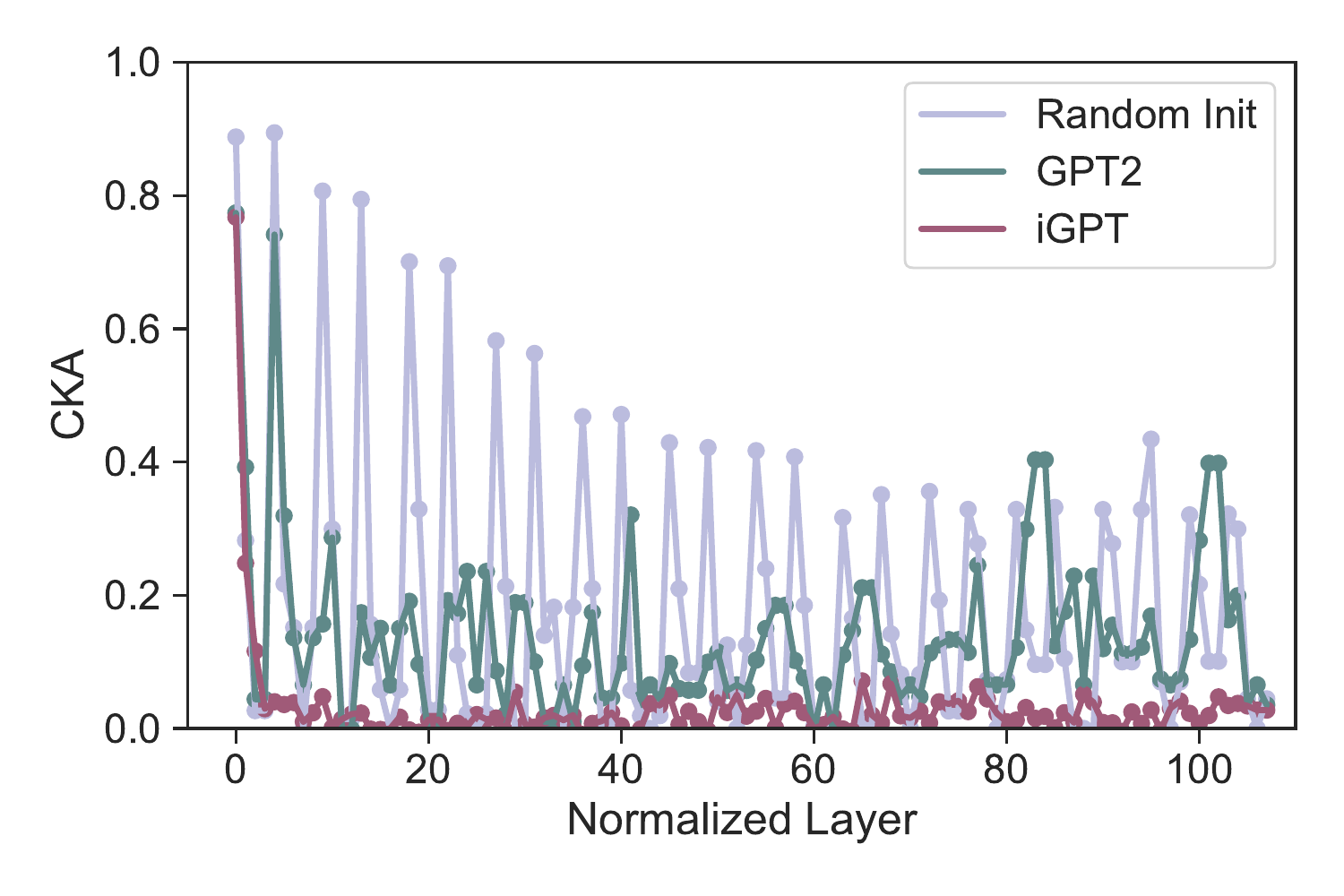}
        \subcaption{Walker2D}
    \end{minipage}
    \caption{CKA similarity of each layer between pre and post-fine-tuning (Action, Seed = 42).}
\end{figure}

\begin{figure}[H]
    \centering
    \begin{minipage}[b]{0.32\linewidth}
        \includegraphics[width=\linewidth]{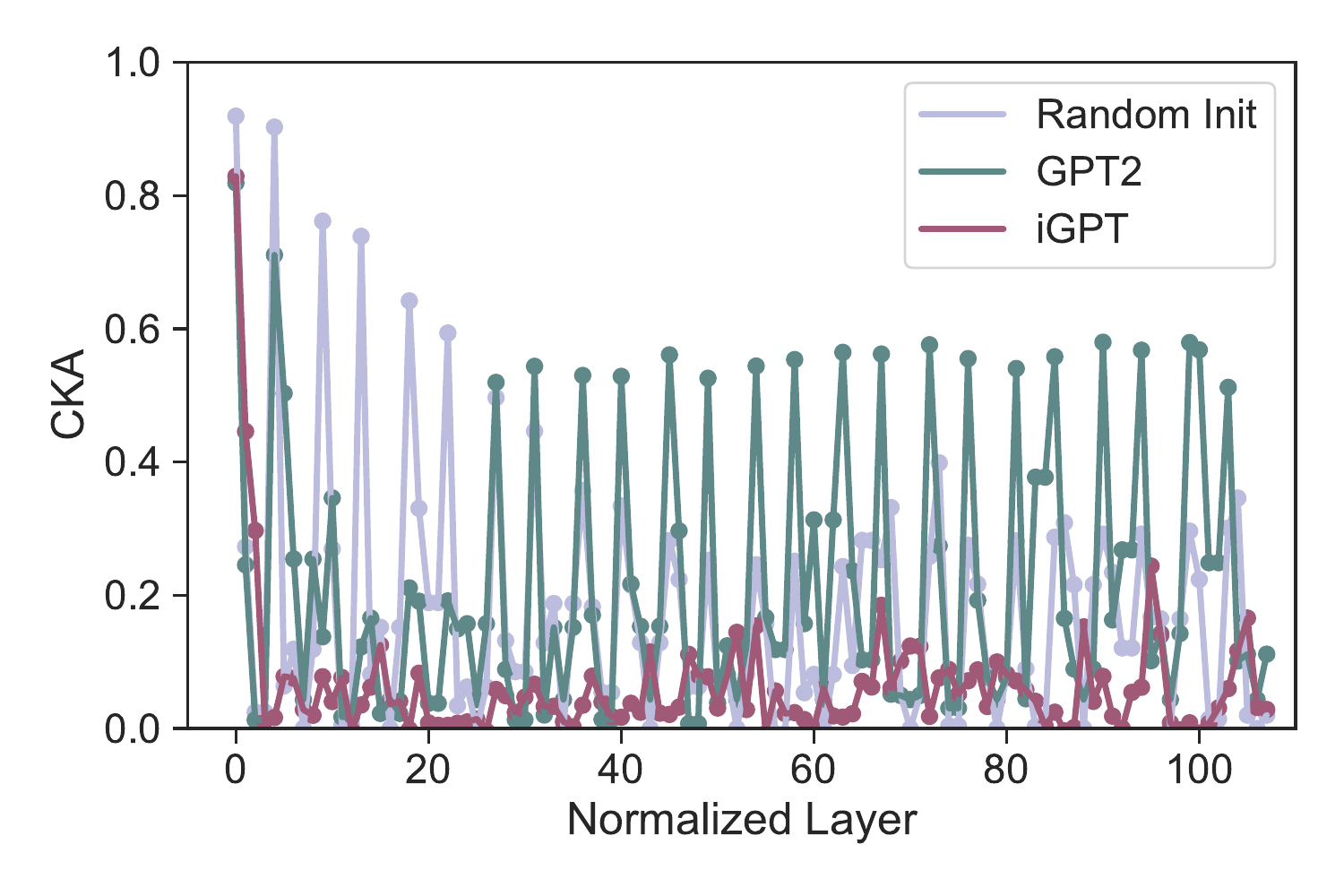}
        \subcaption{Hopper}
    \end{minipage}
    \begin{minipage}[b]{0.32\linewidth}
        \includegraphics[width=\linewidth]{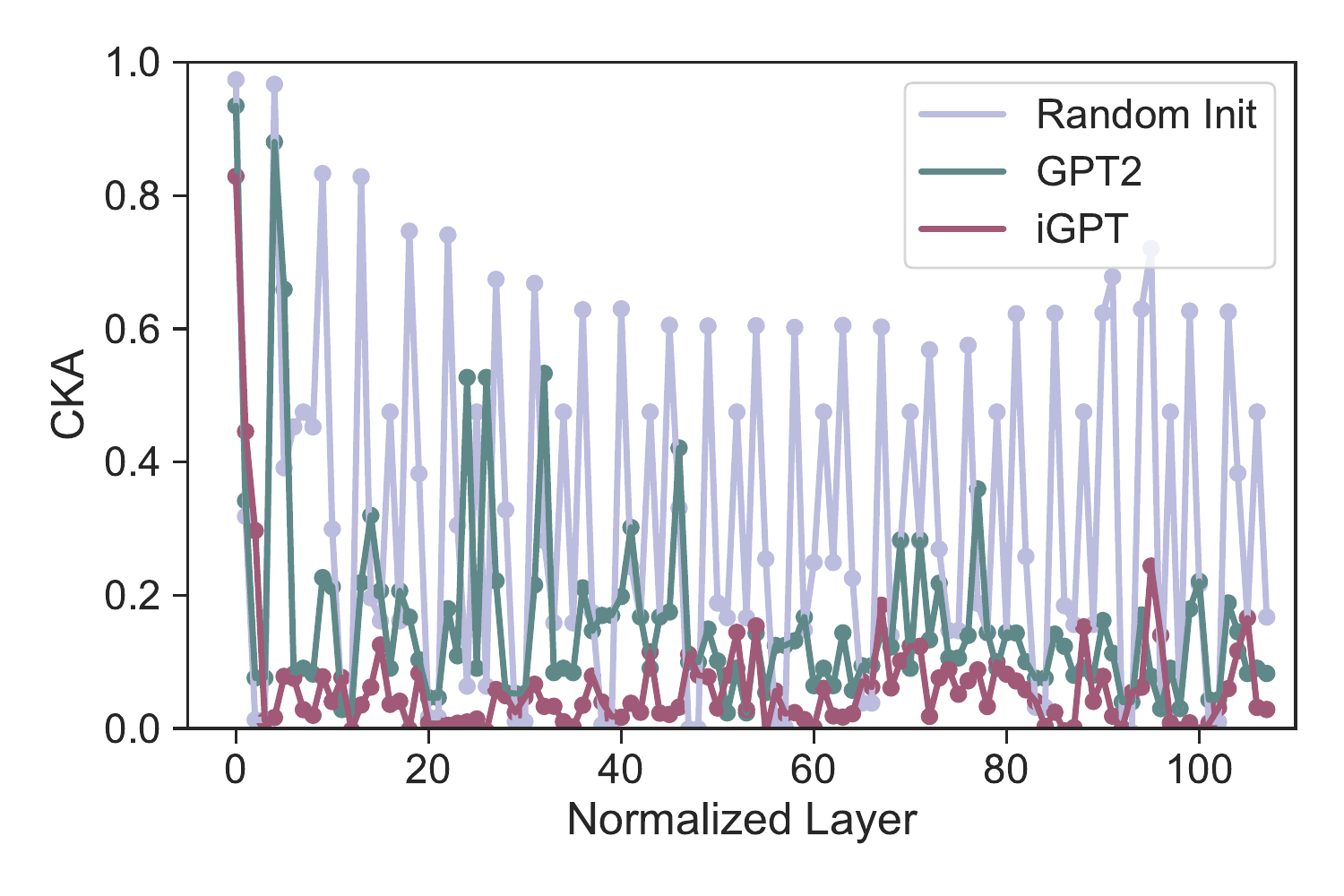}
        \subcaption{HalfCheetah}
    \end{minipage}
    \begin{minipage}[b]{0.32\linewidth}
        \includegraphics[width=\linewidth]{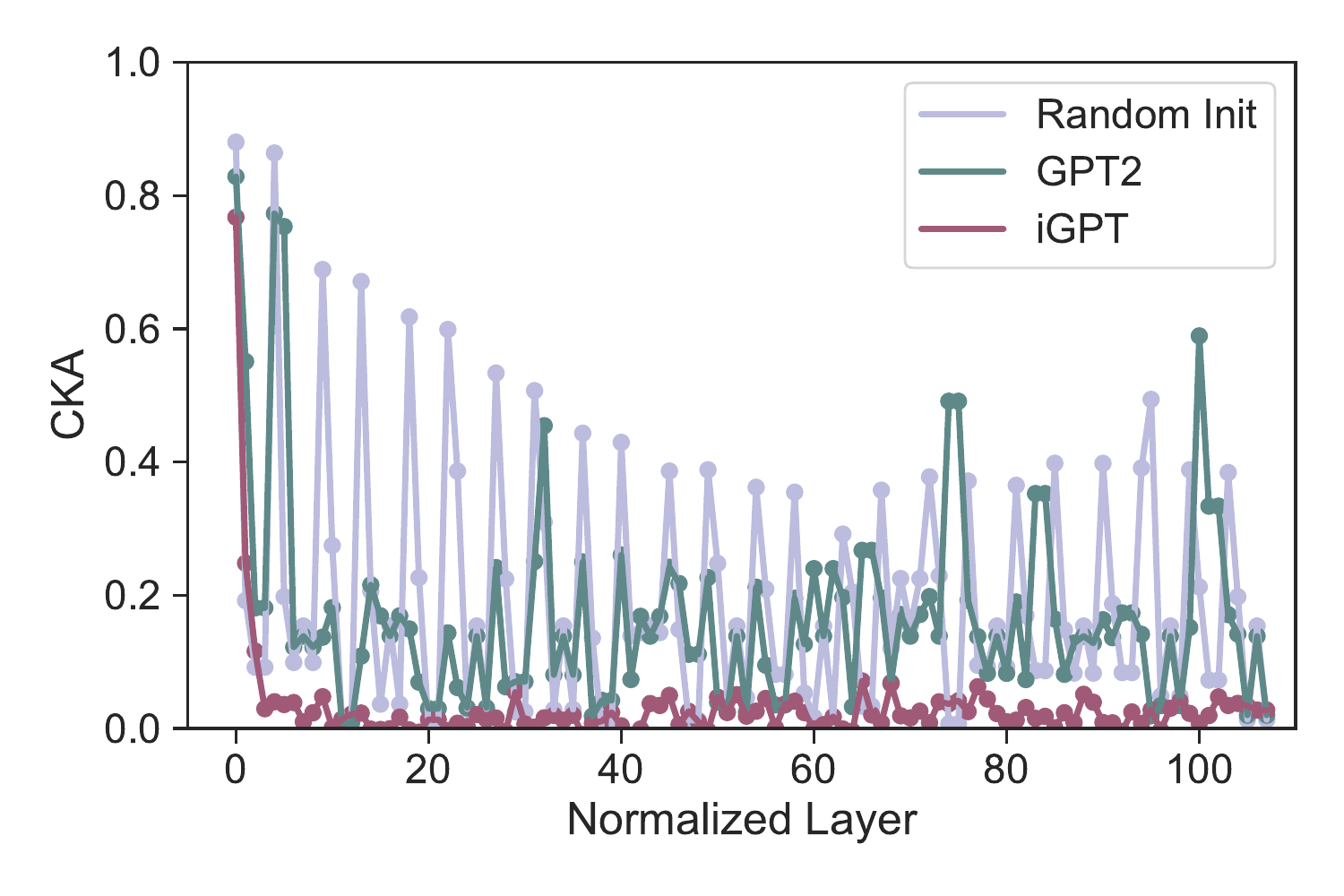}
        \subcaption{Walker2D}
    \end{minipage}
    \caption{CKA similarity of each layer between pre and post-fine-tuning (Return-to-go, Seed = 42).}
\end{figure}

\subsubsection{CKA Between Different Models}
\label{appendix:results-for-other-conditions-cka-between-different-models}

\begin{figure}[H]
    \centering
    \begin{minipage}[b]{0.32\linewidth}
        \includegraphics[width=\linewidth]{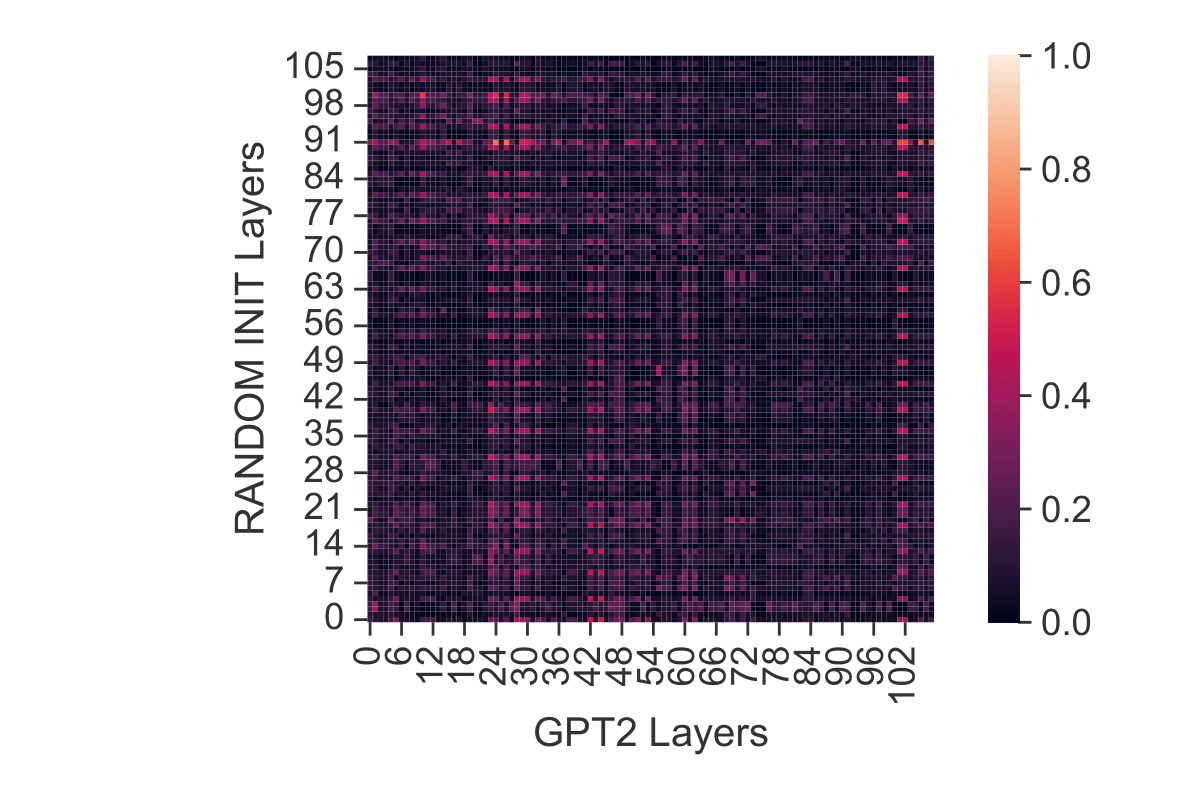}
        \subcaption{Random Init. vs GPT2}
    \end{minipage}
    \begin{minipage}[b]{0.32\linewidth}
        \includegraphics[width=\linewidth]{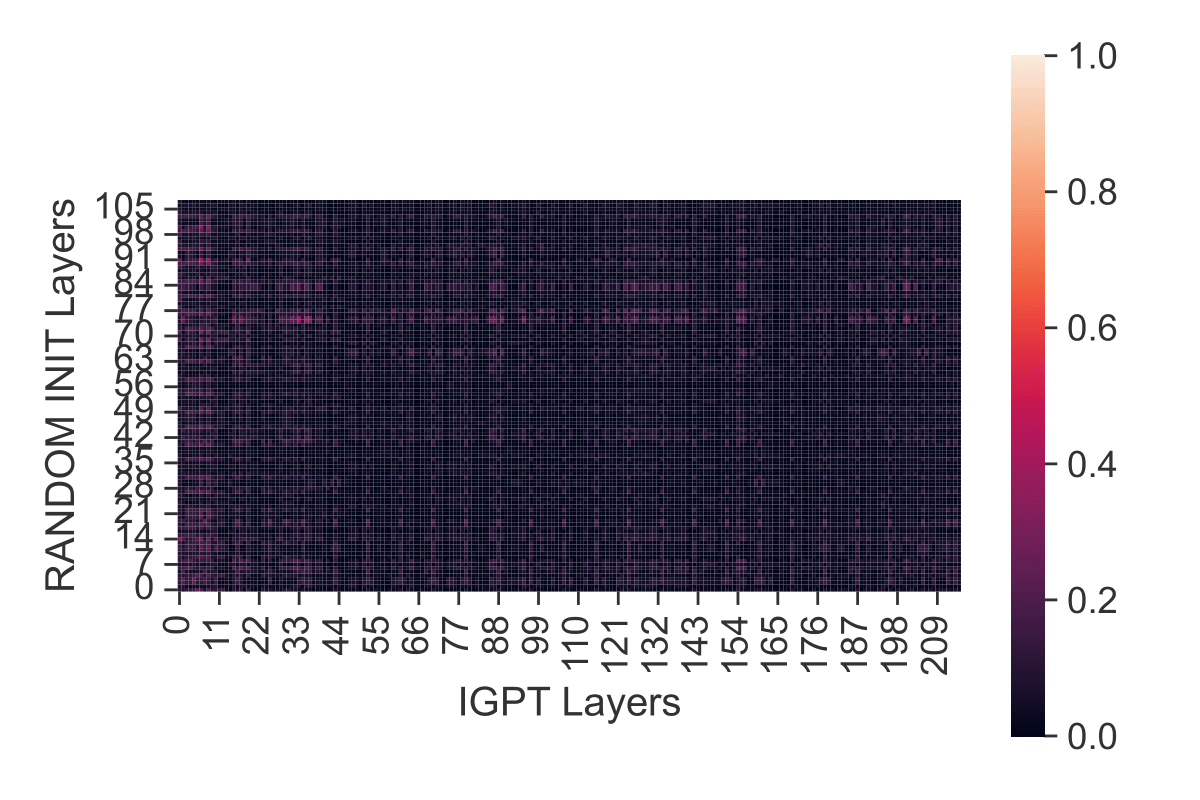}
        \subcaption{Random Init. vs iGPT}
    \end{minipage}
    \begin{minipage}[b]{0.32\linewidth}
        \includegraphics[width=\linewidth]{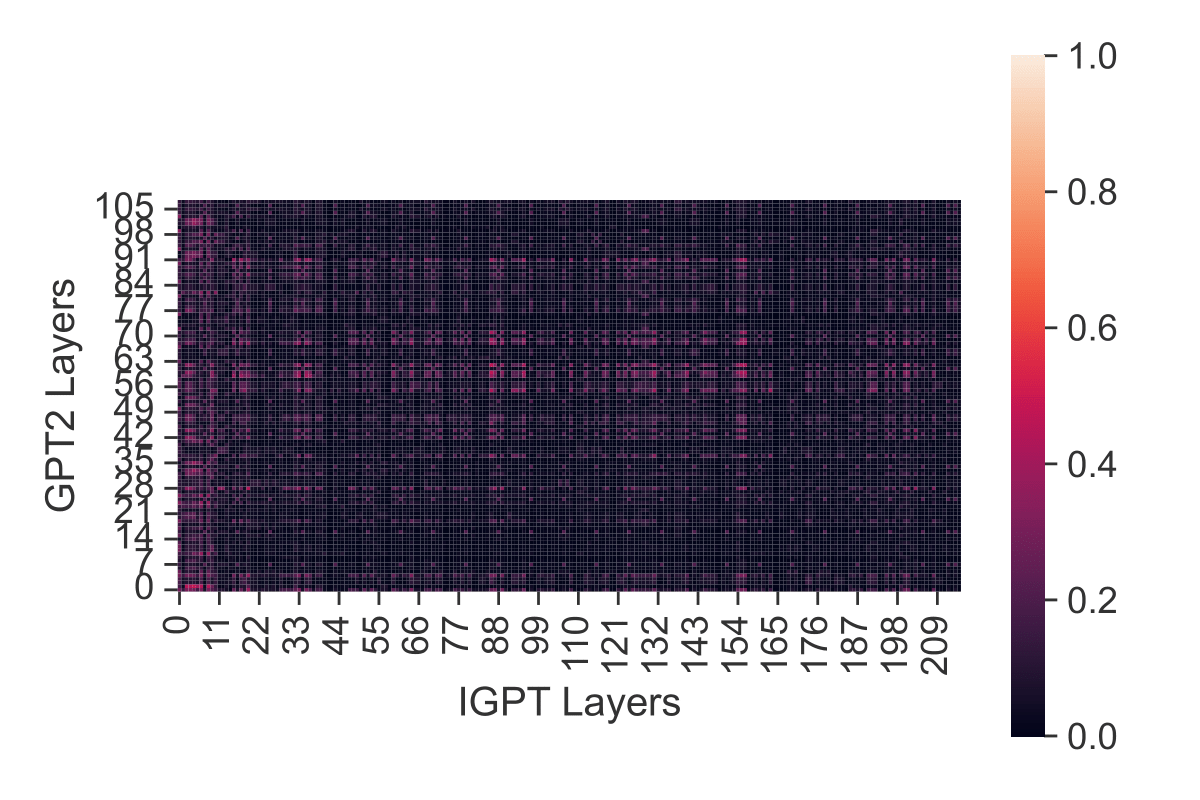}
        \subcaption{GPT2 vs iGPT}
    \end{minipage}
    \caption{CKA between different models (Hopper \& Return-to-go).}
\end{figure}

\begin{figure}[H]
    \centering
    \begin{minipage}[b]{0.32\linewidth}
        \includegraphics[width=\linewidth]{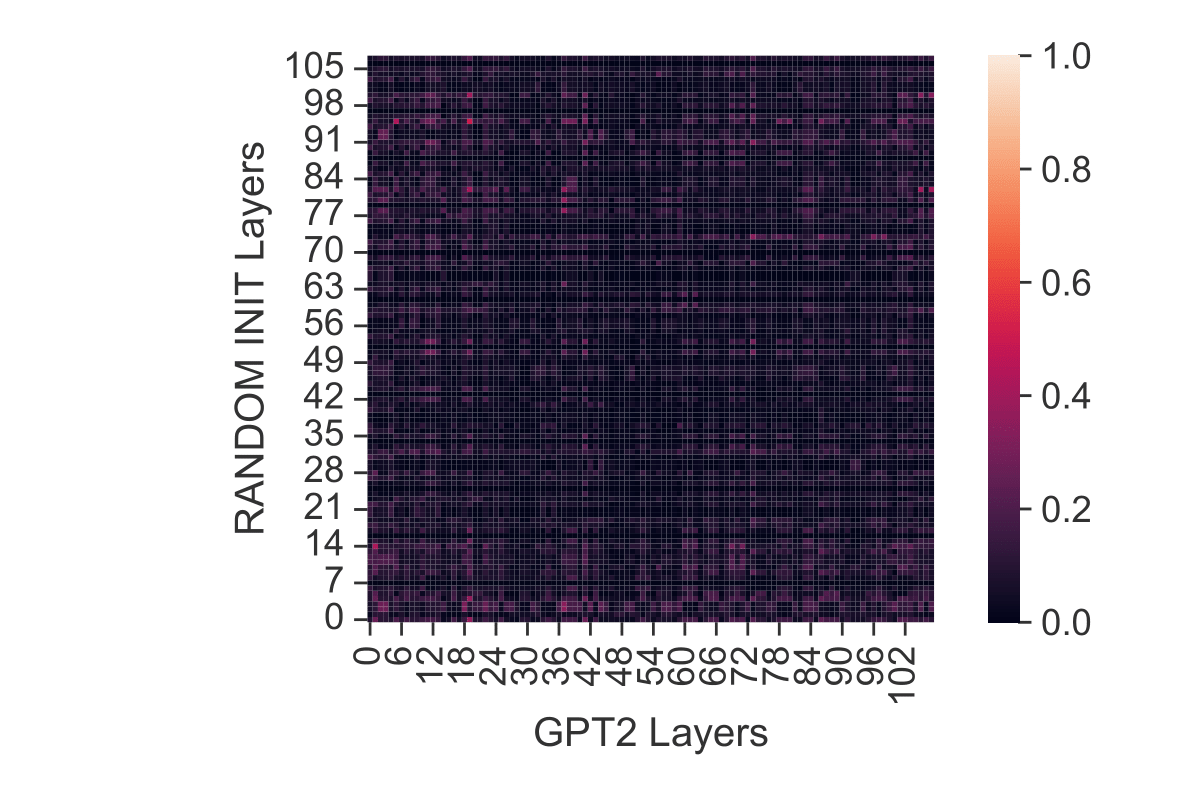}
        \subcaption{Random Init. vs GPT2}
    \end{minipage}
    \begin{minipage}[b]{0.32\linewidth}
        \includegraphics[width=\linewidth]{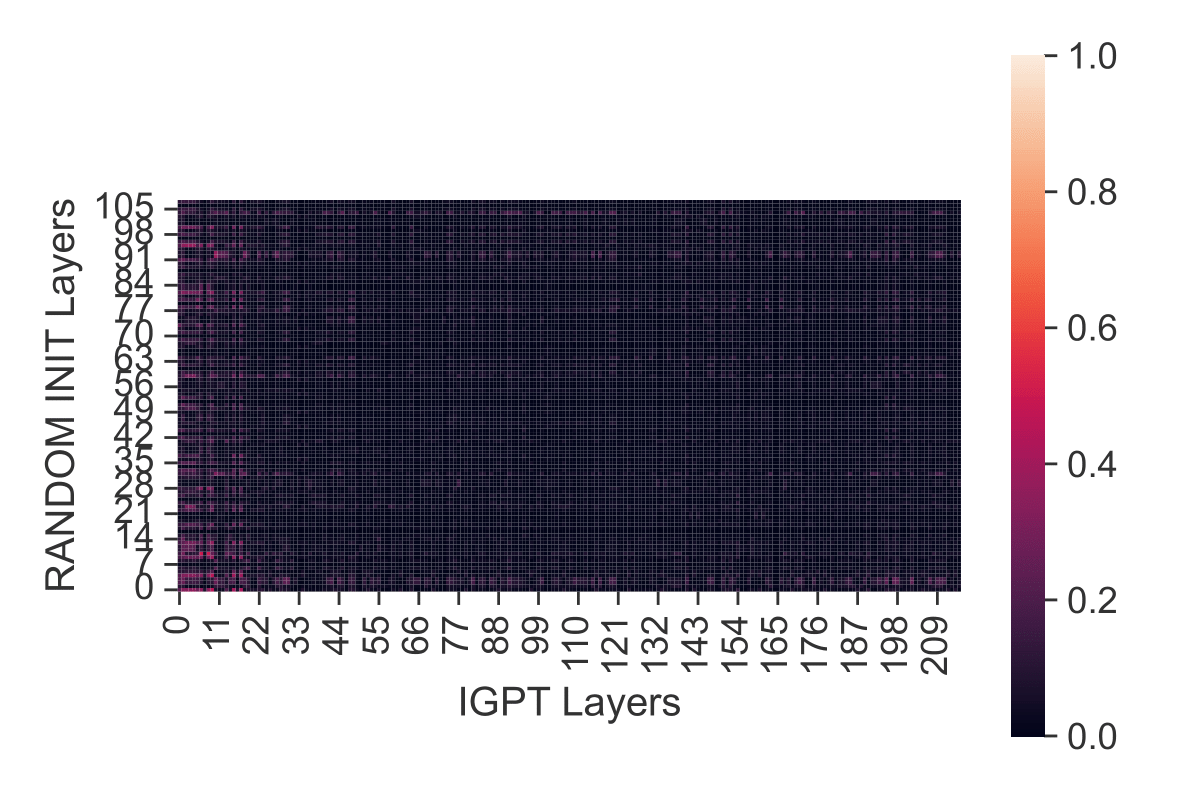}
        \subcaption{Random Init. vs iGPT}
    \end{minipage}
    \begin{minipage}[b]{0.32\linewidth}
        \includegraphics[width=\linewidth]{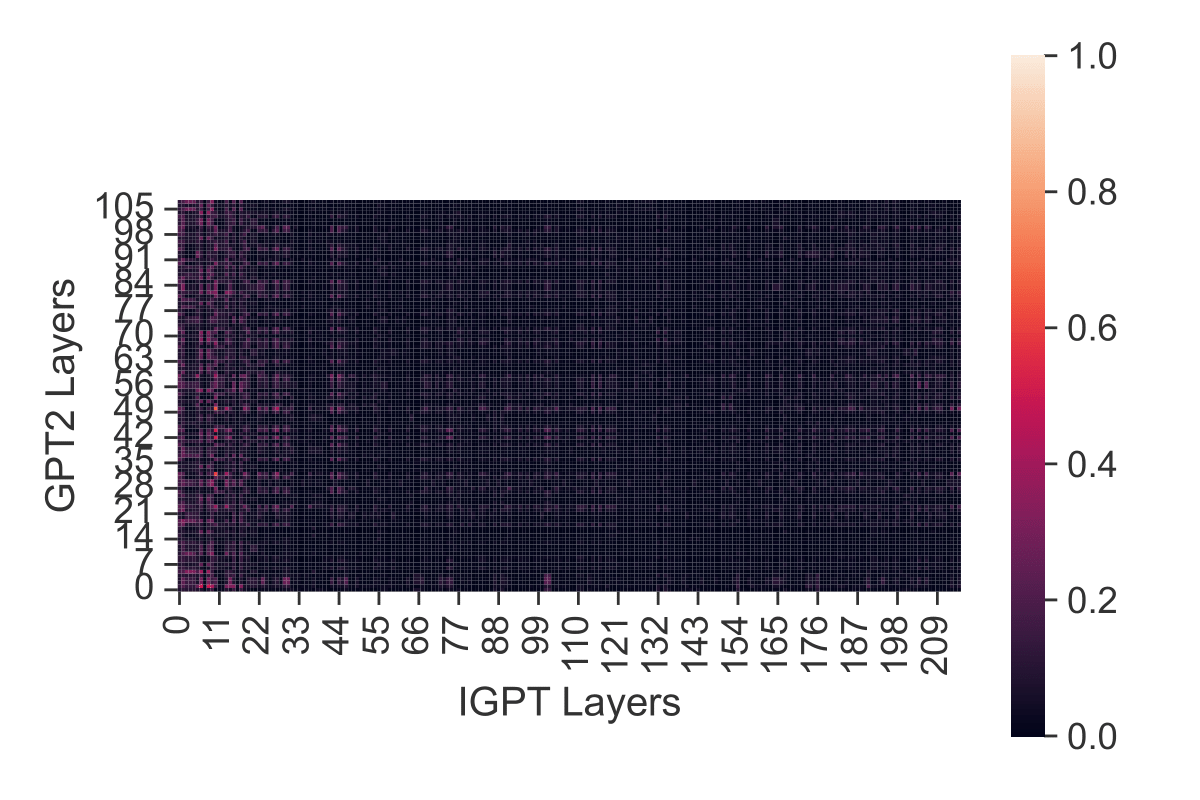}
        \subcaption{GPT2 vs iGPT}
    \end{minipage}
    \caption{CKA between different models (Hopper \& Action).}
\end{figure}

\begin{figure}[H]
    \centering
    \begin{minipage}[b]{0.32\linewidth}
        \includegraphics[width=\linewidth]{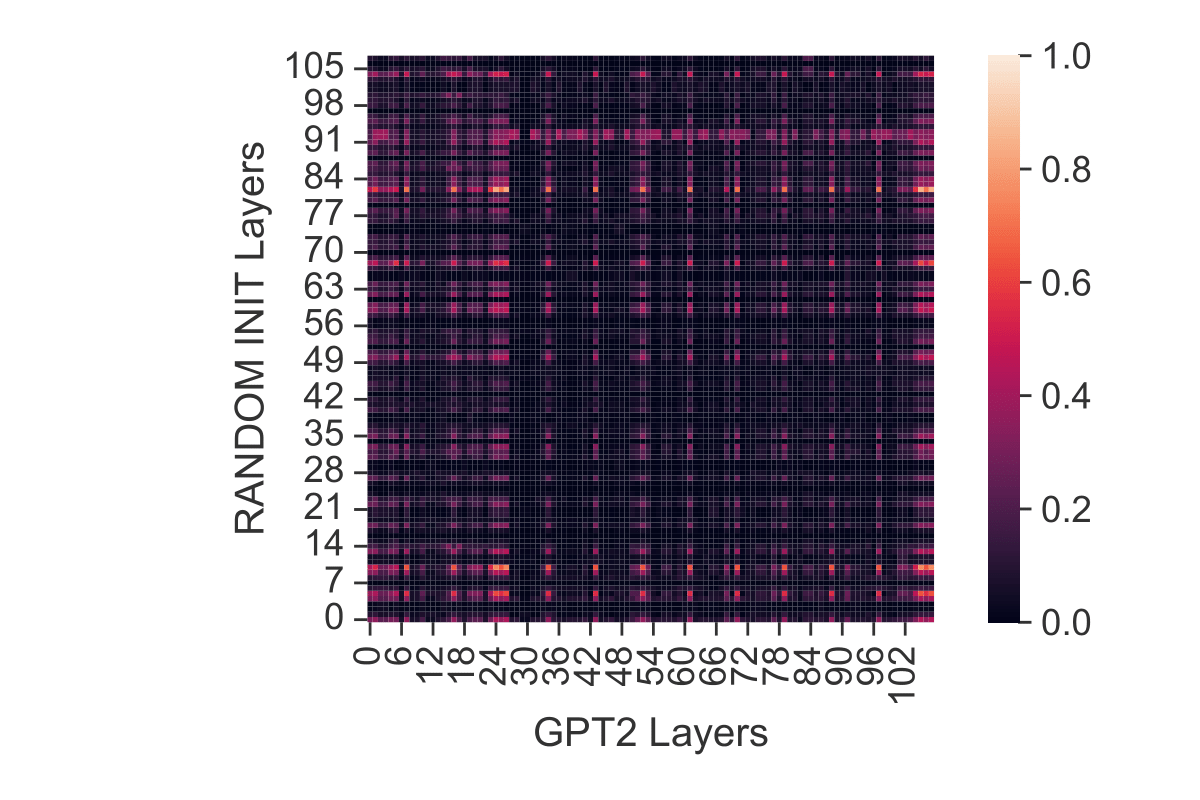}
        \subcaption{Random Init. vs GPT2}
    \end{minipage}
    \begin{minipage}[b]{0.32\linewidth}
        \includegraphics[width=\linewidth]{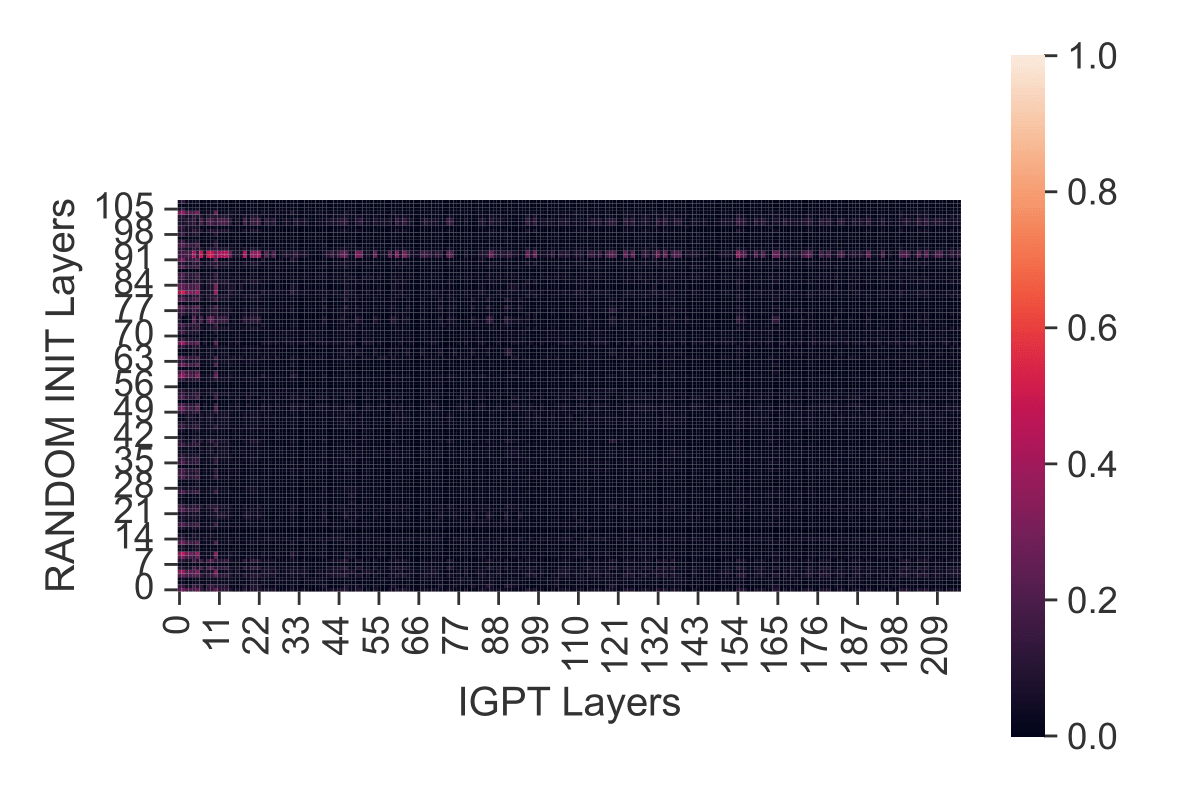}
        \subcaption{Random Init. vs iGPT}
    \end{minipage}
    \begin{minipage}[b]{0.32\linewidth}
        \includegraphics[width=\linewidth]{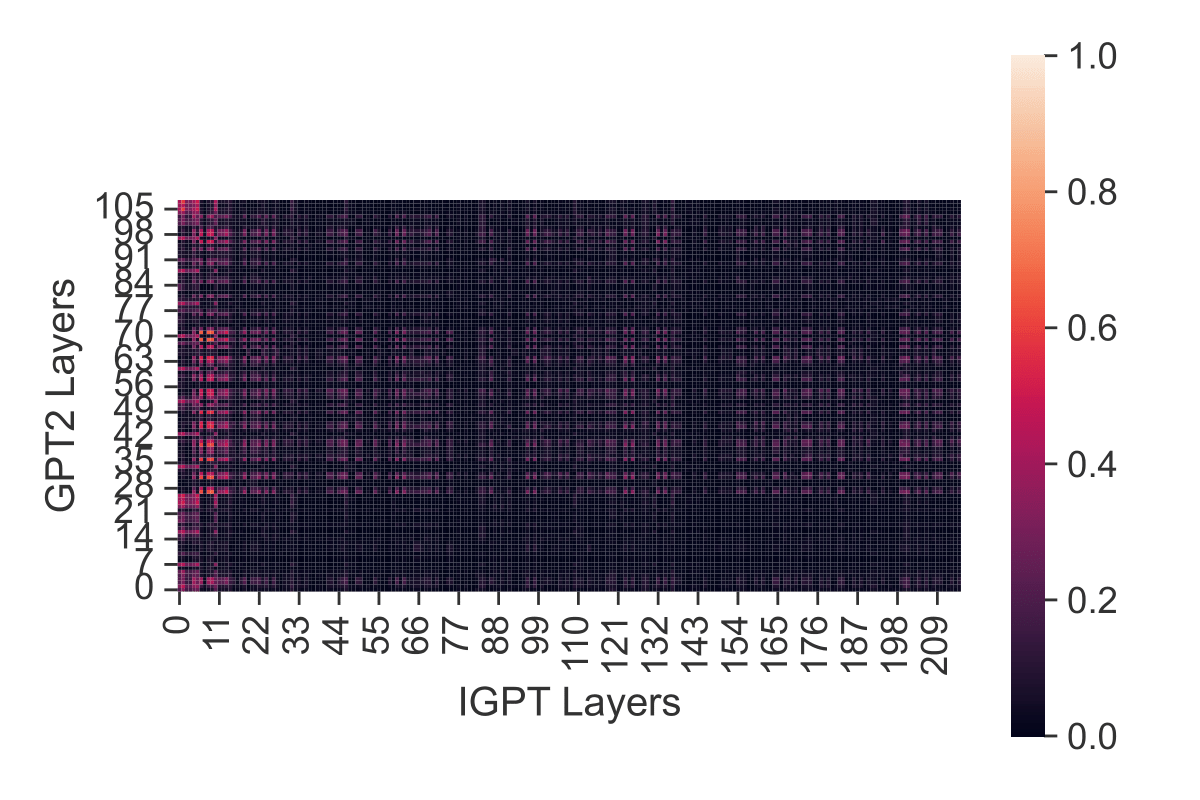}
        \subcaption{GPT2 vs iGPT}
    \end{minipage}
    \caption{CKA between different models (HalfCheetah \& State).}
\end{figure}

\begin{figure}[H]
    \centering
    \begin{minipage}[b]{0.32\linewidth}
        \includegraphics[width=\linewidth]{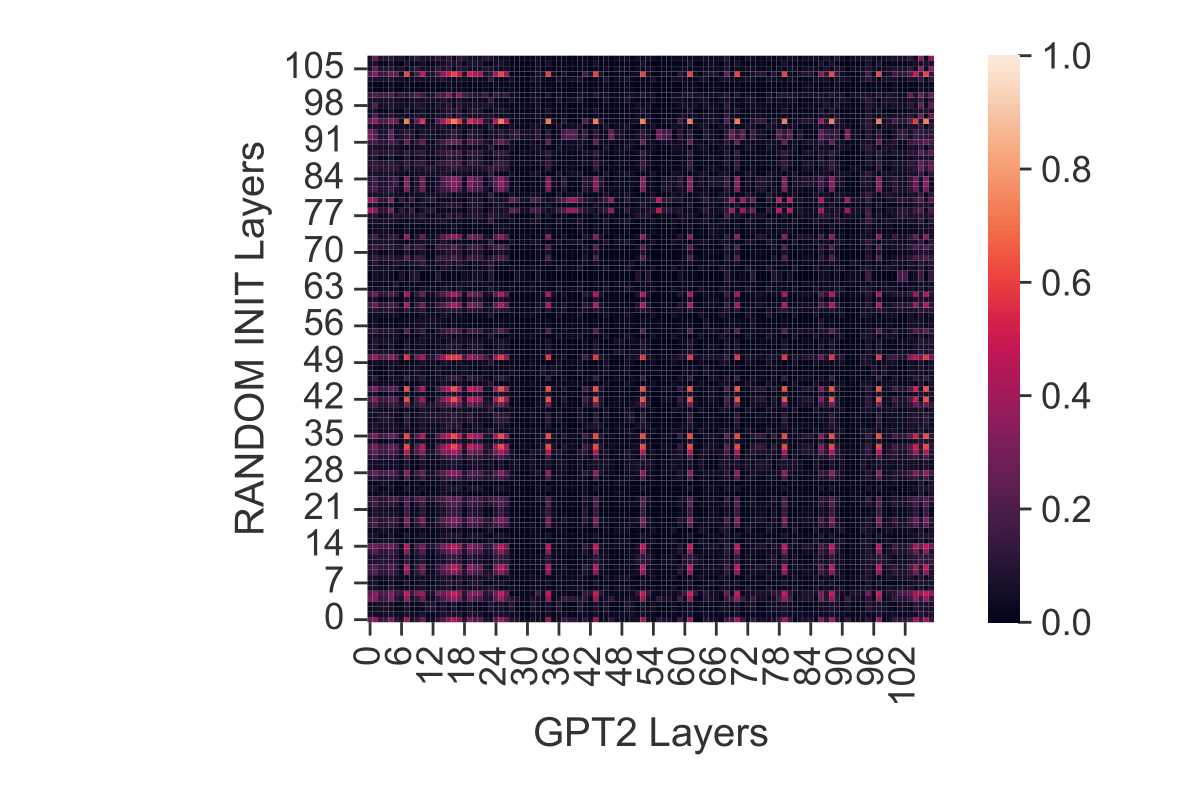}
        \subcaption{Random Init. vs GPT2}
    \end{minipage}
    \begin{minipage}[b]{0.32\linewidth}
        \includegraphics[width=\linewidth]{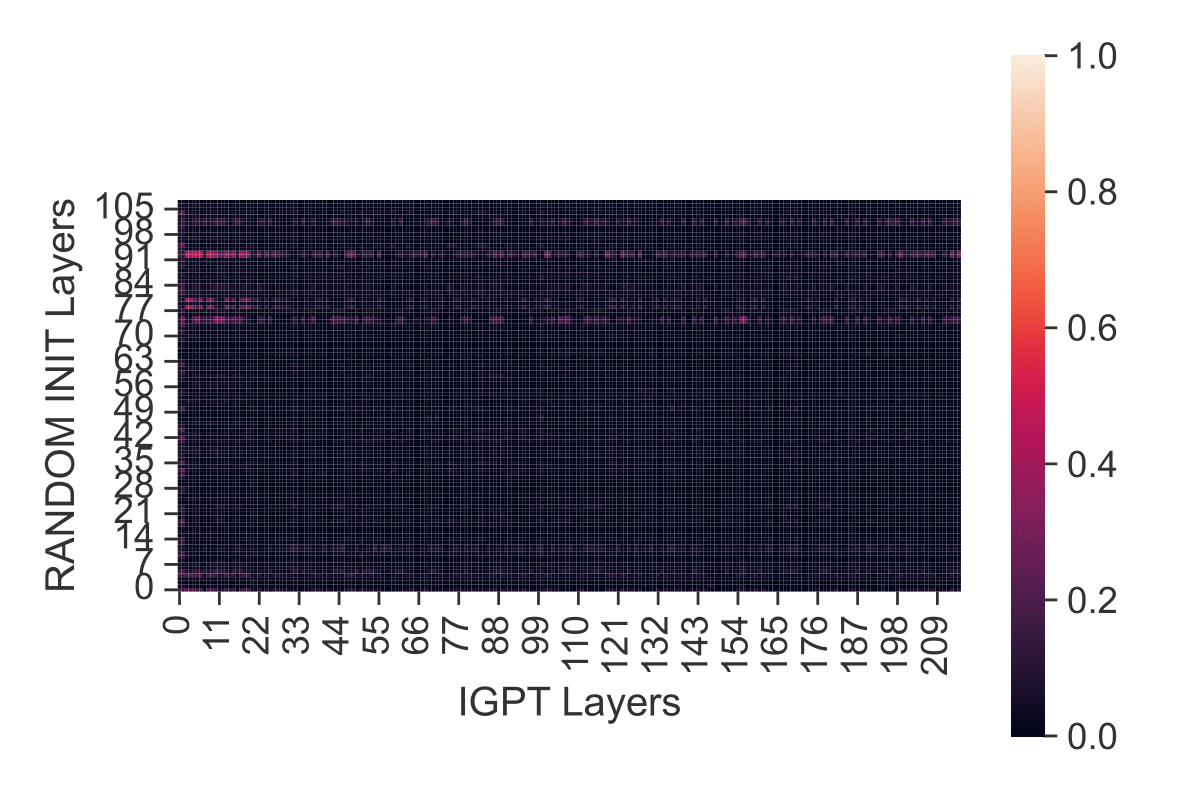}
        \subcaption{Random Init. vs iGPT}
    \end{minipage}
    \begin{minipage}[b]{0.32\linewidth}
        \includegraphics[width=\linewidth]{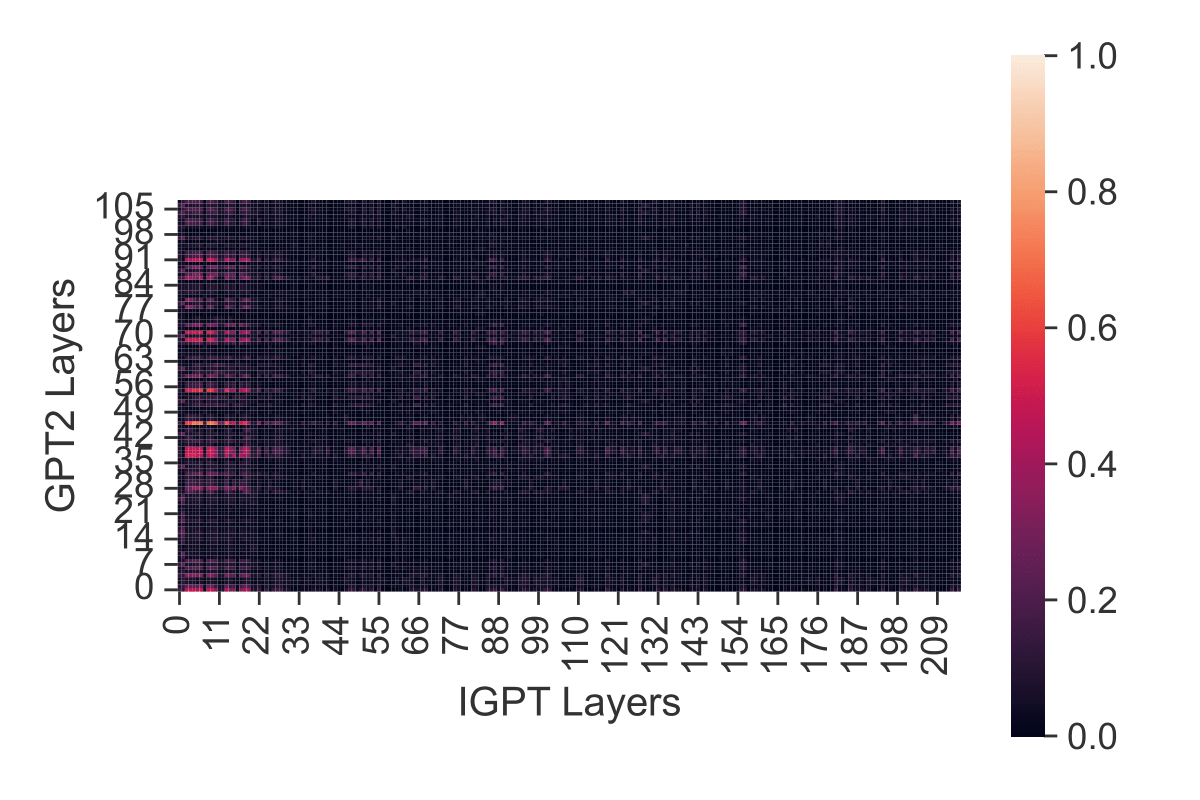}
        \subcaption{GPT2 vs iGPT}
    \end{minipage}
    \caption{CKA between different models (HalfCheetah \& Return-to-go).}
\end{figure}

\begin{figure}[H]
    \centering
    \begin{minipage}[b]{0.32\linewidth}
        \includegraphics[width=\linewidth]{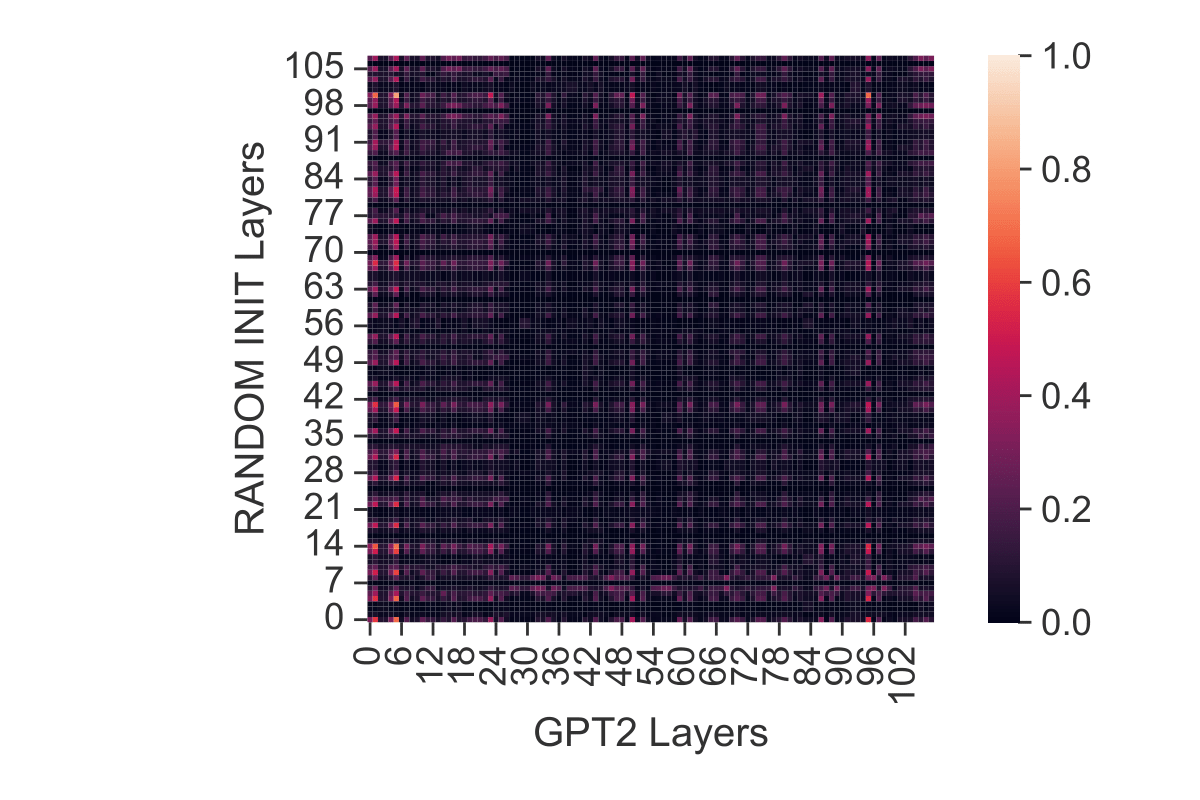}
        \subcaption{Random Init. vs GPT2}
    \end{minipage}
    \begin{minipage}[b]{0.32\linewidth}
        \includegraphics[width=\linewidth]{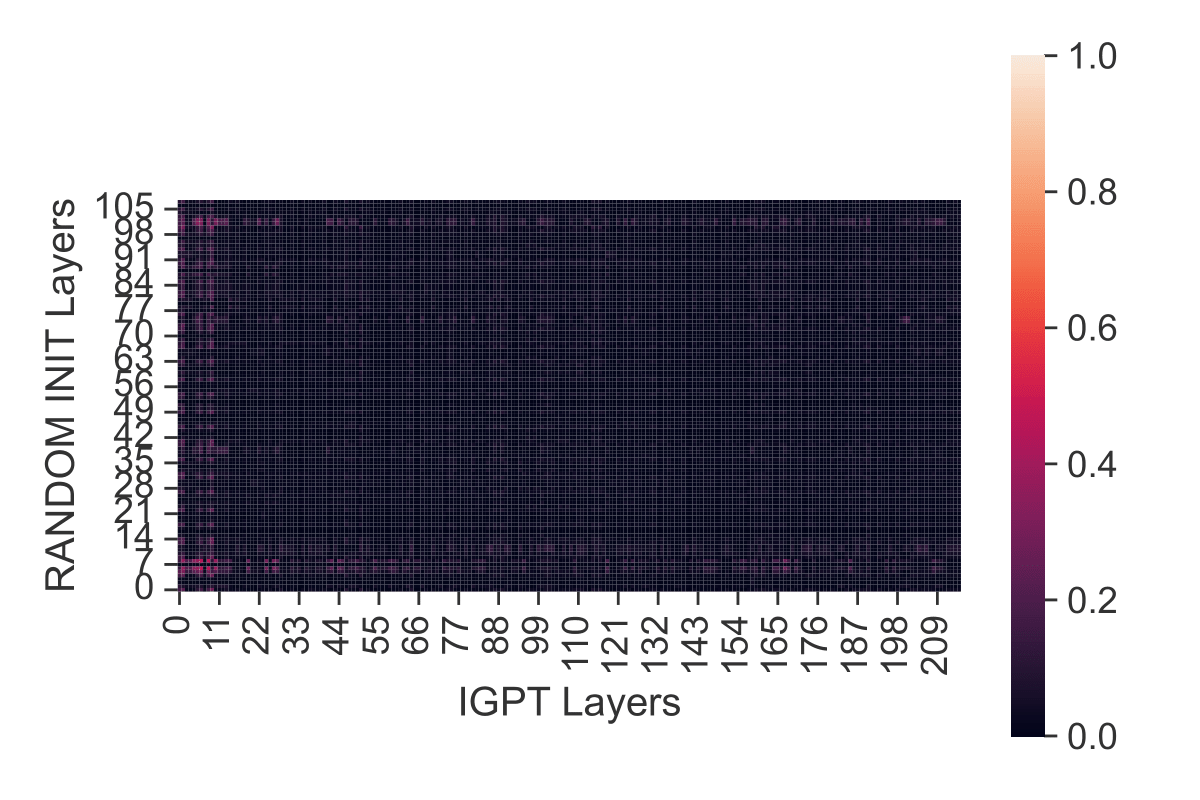}
        \subcaption{Random Init. vs iGPT}
    \end{minipage}
    \begin{minipage}[b]{0.32\linewidth}
        \includegraphics[width=\linewidth]{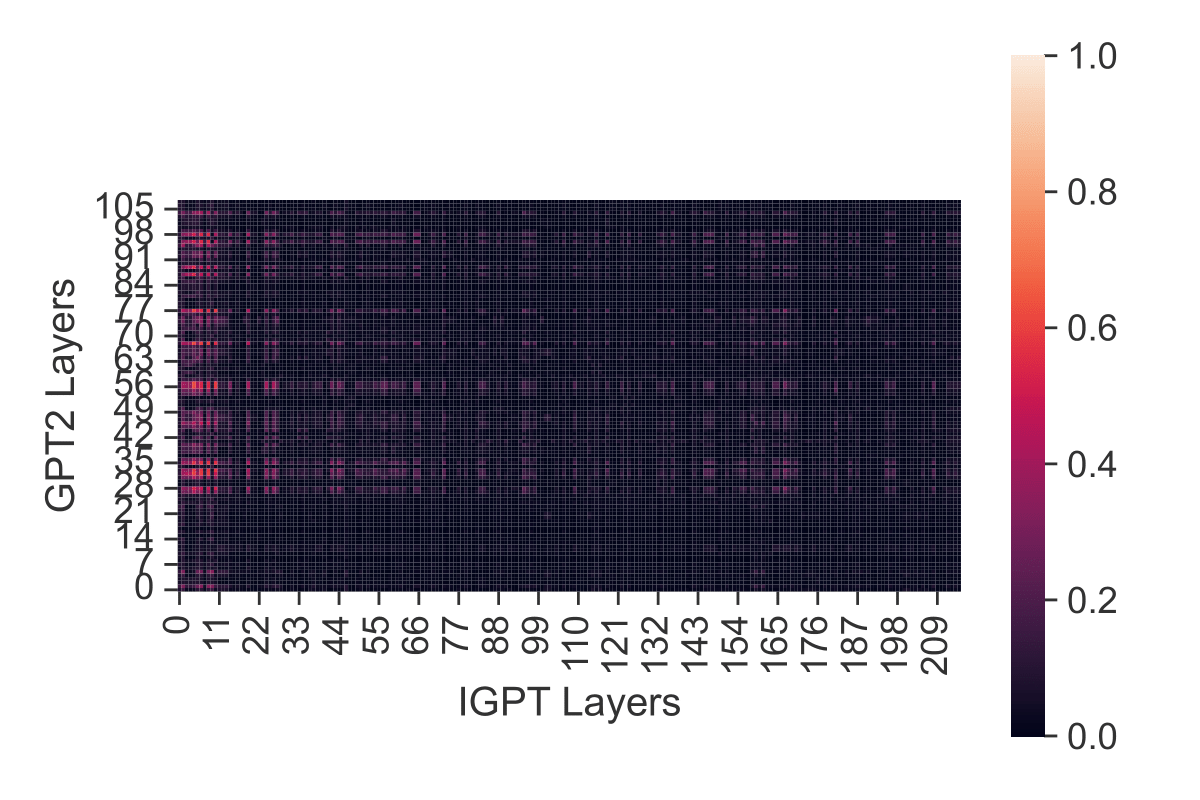}
        \subcaption{GPT2 vs iGPT}
    \end{minipage}
    \caption{CKA between different models (HalfCheetah \& Action).}
\end{figure}

\begin{figure}[H]
    \centering
    \begin{minipage}[b]{0.32\linewidth}
        \includegraphics[width=\linewidth]{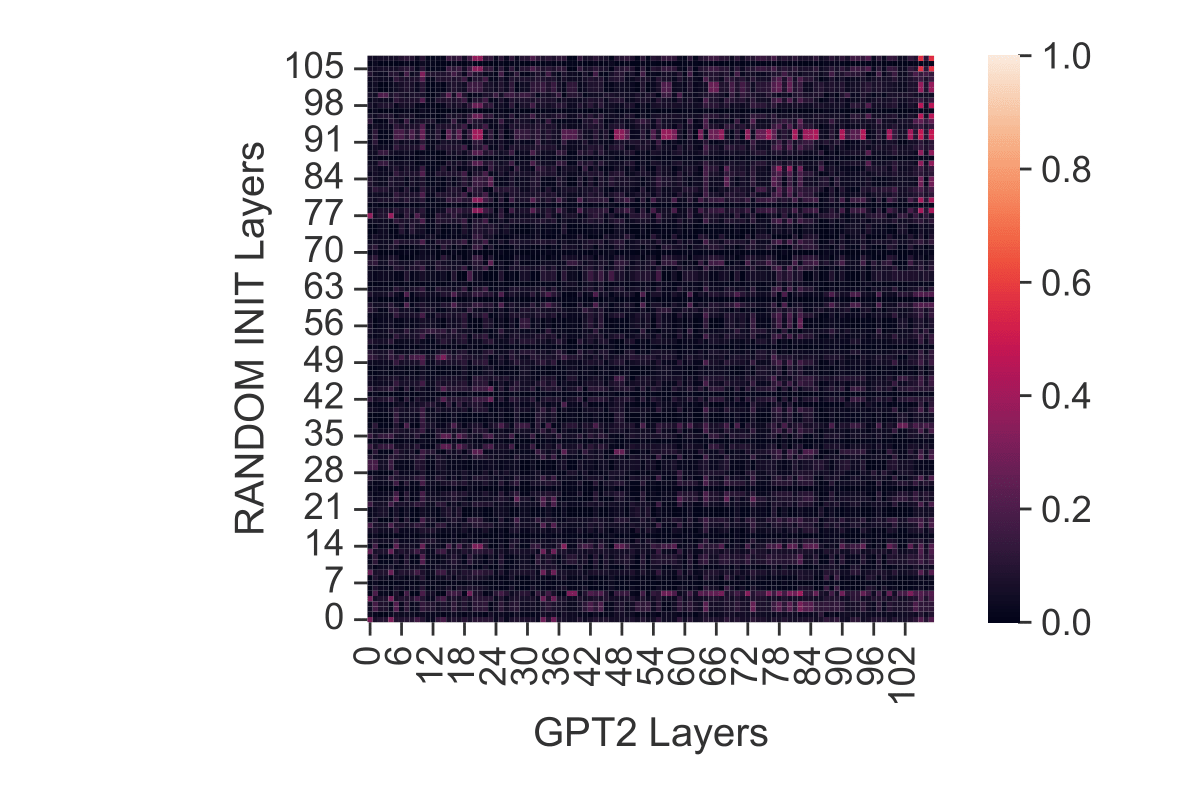}
        \subcaption{Random Init. vs GPT2}
    \end{minipage}
    \begin{minipage}[b]{0.32\linewidth}
        \includegraphics[width=\linewidth]{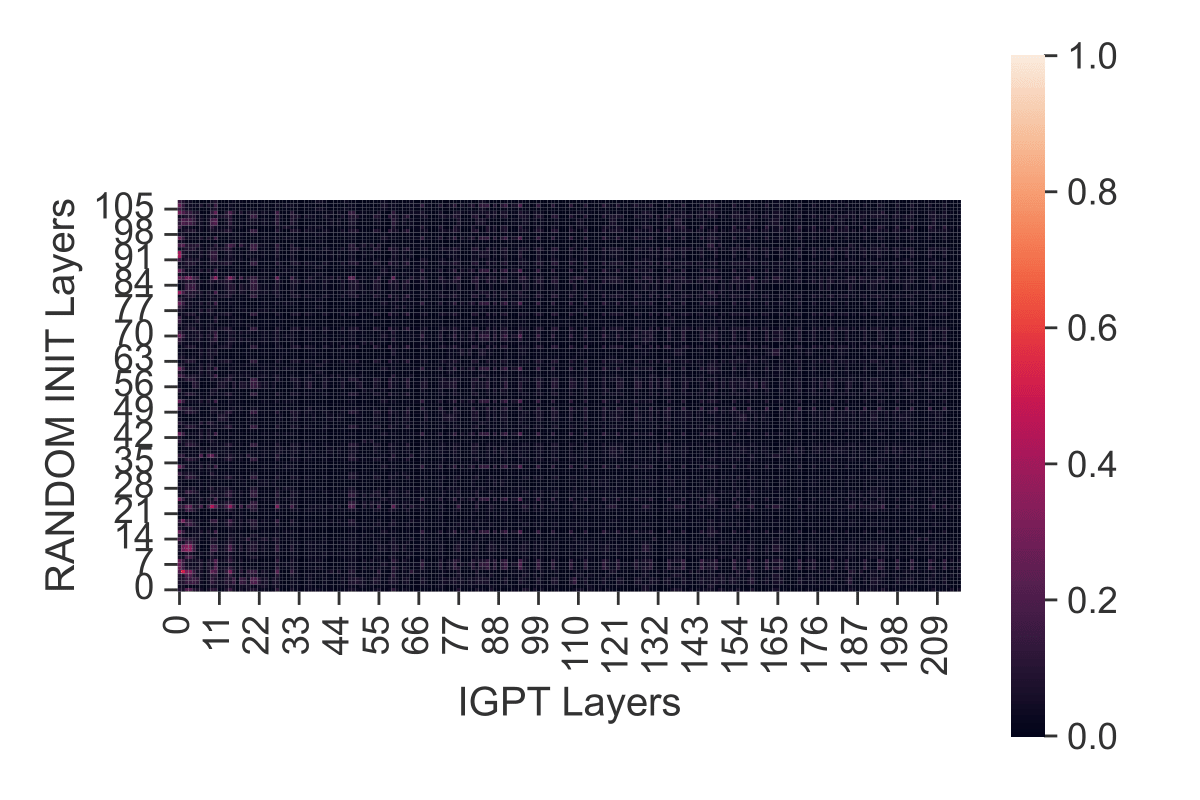}
        \subcaption{Random Init. vs iGPT}
    \end{minipage}
    \begin{minipage}[b]{0.32\linewidth}
        \includegraphics[width=\linewidth]{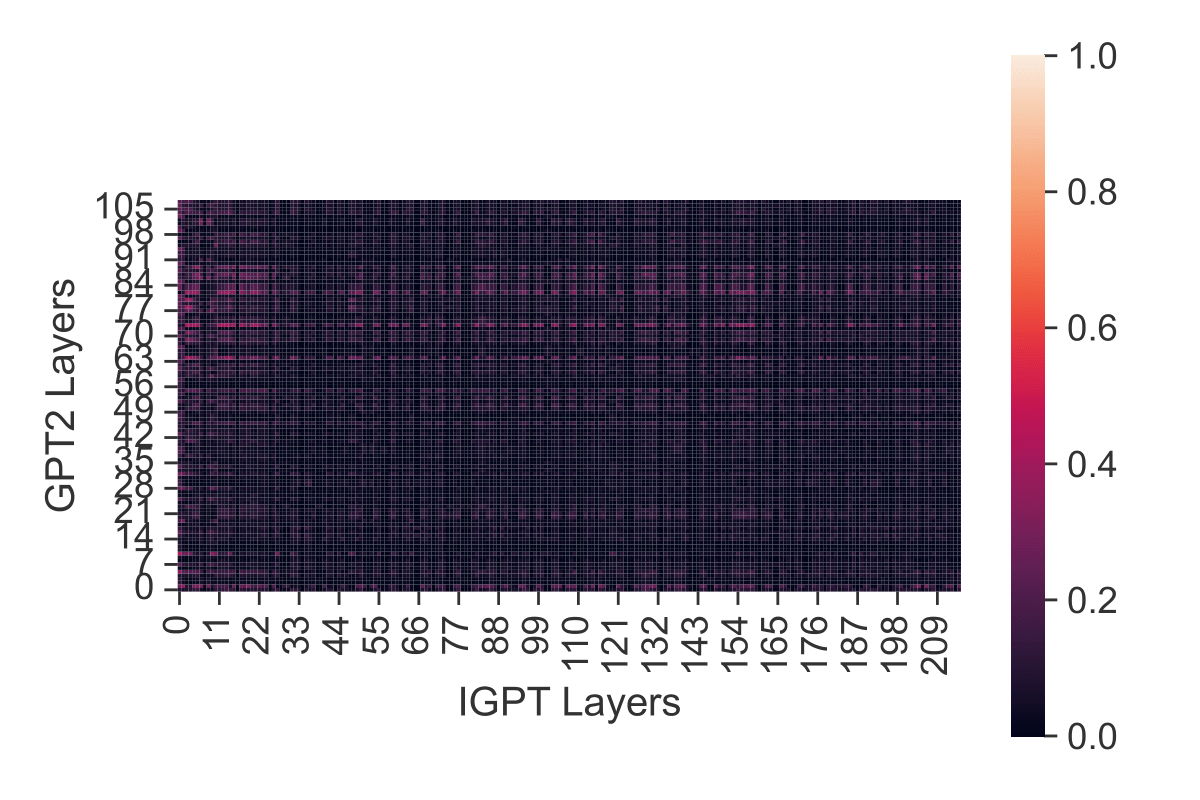}
        \subcaption{GPT2 vs iGPT}
    \end{minipage}
    \caption{CKA between different models (Walker2D \& State).}
\end{figure}

\begin{figure}[H]
    \centering
    \begin{minipage}[b]{0.32\linewidth}
        \includegraphics[width=\linewidth]{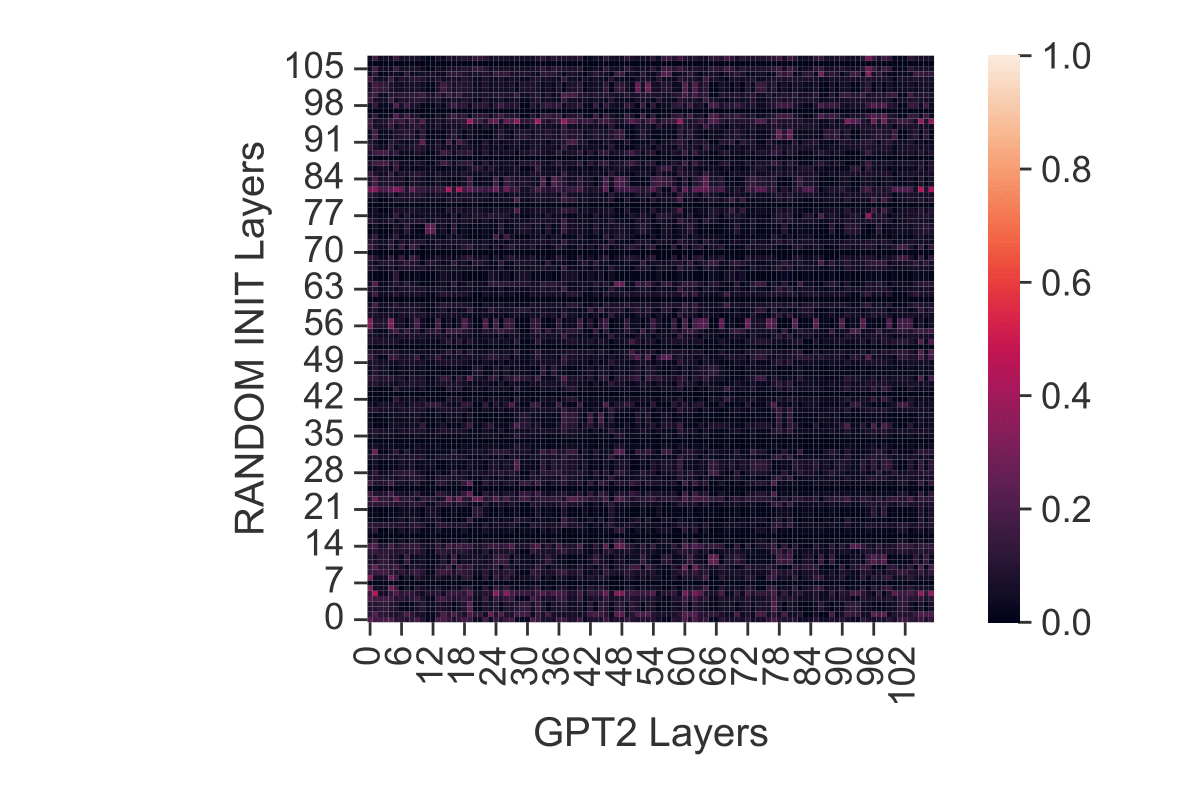}
        \subcaption{Random Init. vs GPT2}
    \end{minipage}
    \begin{minipage}[b]{0.32\linewidth}
        \includegraphics[width=\linewidth]{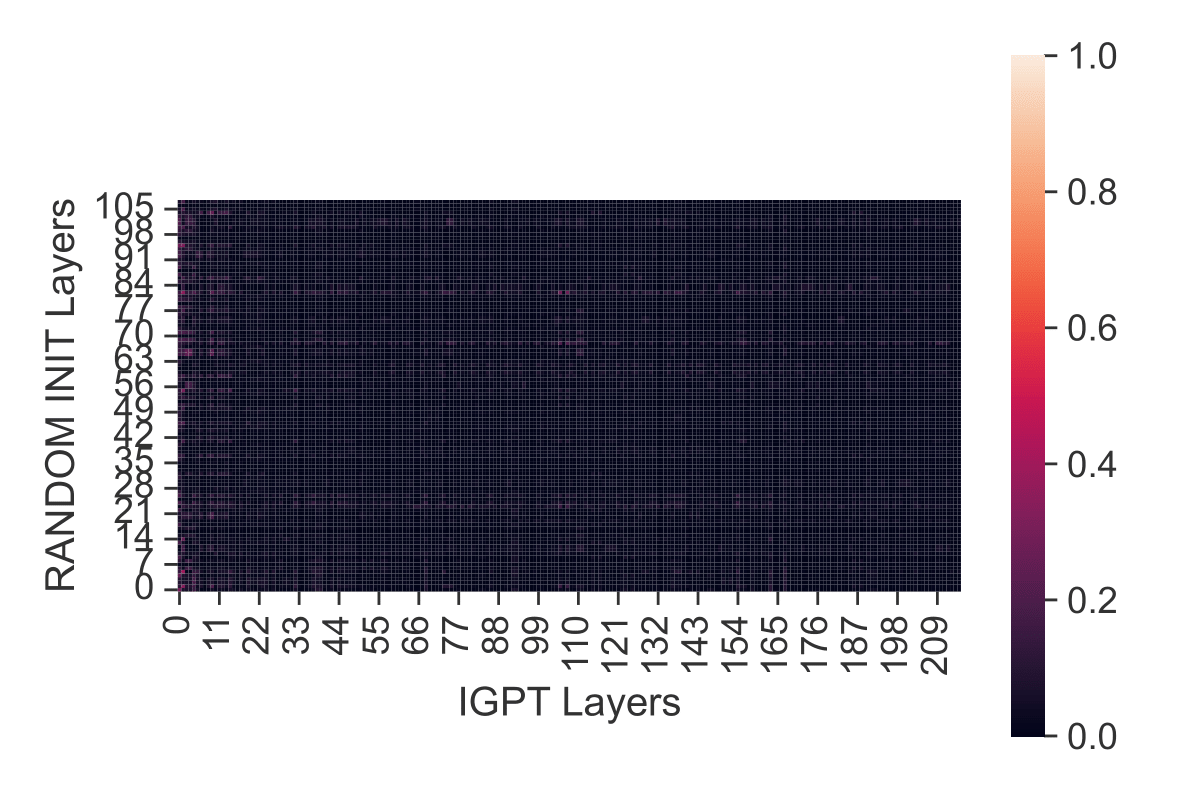}
        \subcaption{Random Init. vs iGPT}
    \end{minipage}
    \begin{minipage}[b]{0.32\linewidth}
        \includegraphics[width=\linewidth]{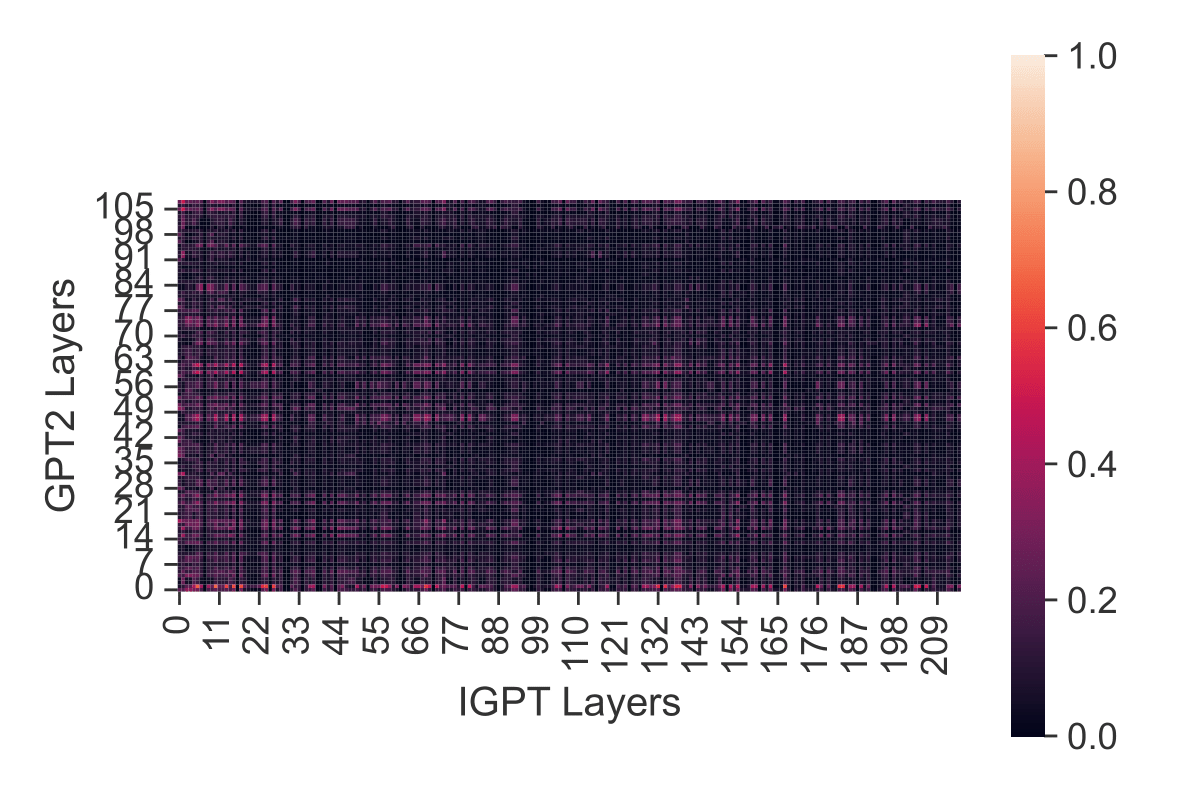}
        \subcaption{GPT2 vs iGPT}
    \end{minipage}
    \caption{CKA between different models (Walker2D \& Return-to-go).}
\end{figure}

\begin{figure}[H]
    \centering
    \begin{minipage}[b]{0.32\linewidth}
        \includegraphics[width=\linewidth]{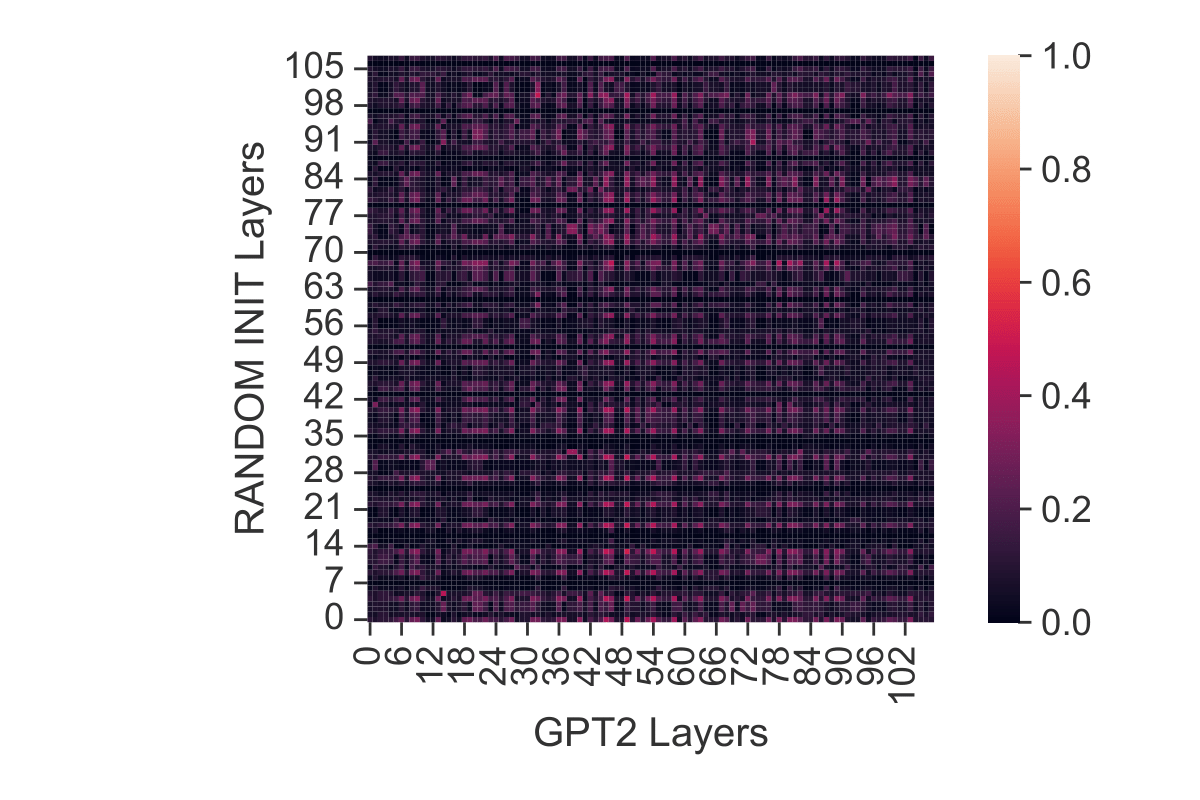}
        \subcaption{Random Init. vs GPT2}
    \end{minipage}
    \begin{minipage}[b]{0.32\linewidth}
        \includegraphics[width=\linewidth]{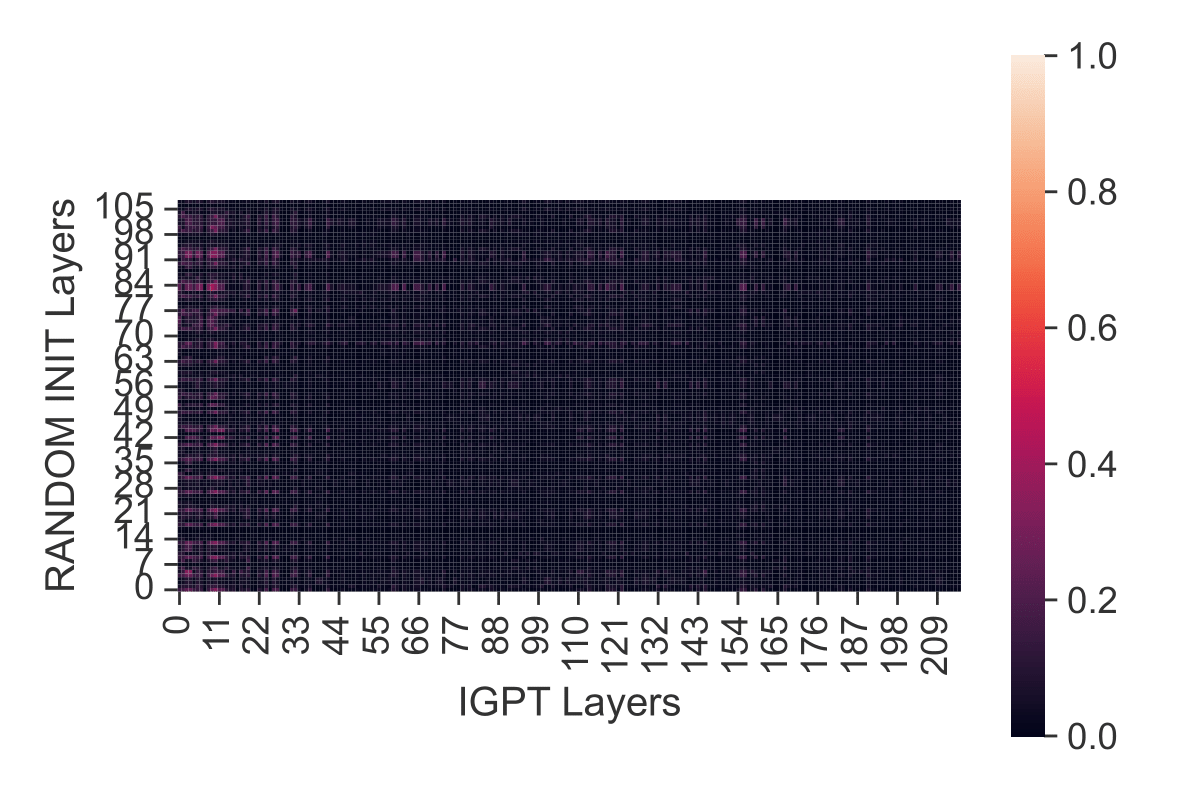}
        \subcaption{Random Init. vs iGPT}
    \end{minipage}
    \begin{minipage}[b]{0.32\linewidth}
        \includegraphics[width=\linewidth]{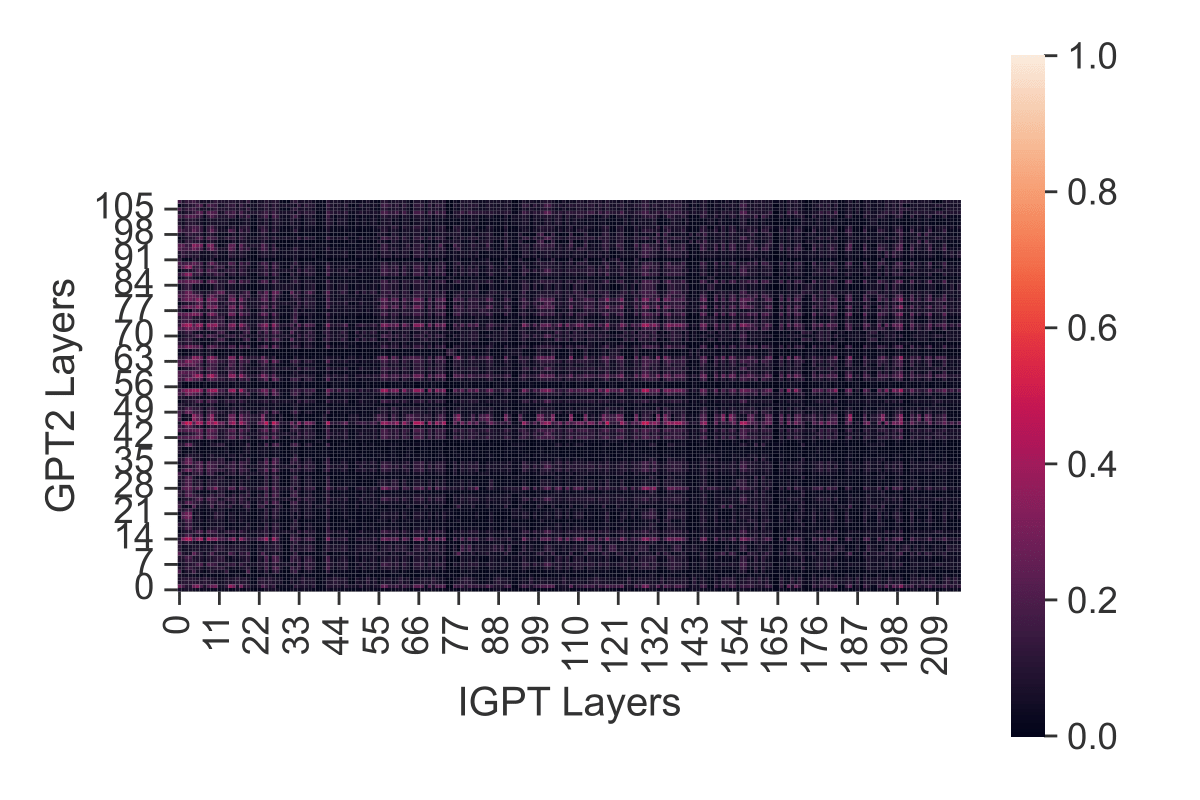}
        \subcaption{GPT2 vs iGPT}
    \end{minipage}
    \caption{CKA between different models (Walker2D \& Action).}
\end{figure}

\subsubsection{CKA Between Different Layers in a Model}

\begin{figure}[H]
    \centering
    \begin{minipage}[b]{0.32\linewidth}
        \includegraphics[width=\linewidth]{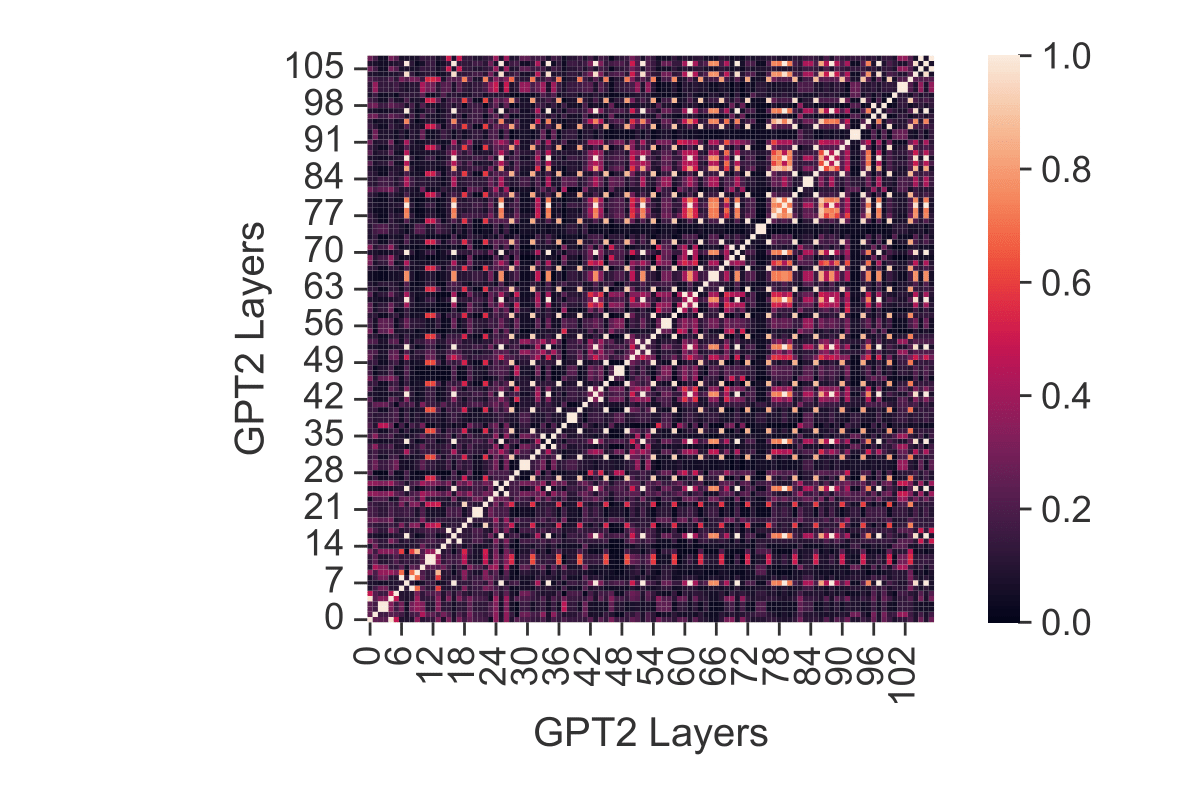}
        \subcaption{GPT2}
    \end{minipage}
    \begin{minipage}[b]{0.32\linewidth}
        \includegraphics[width=\linewidth]{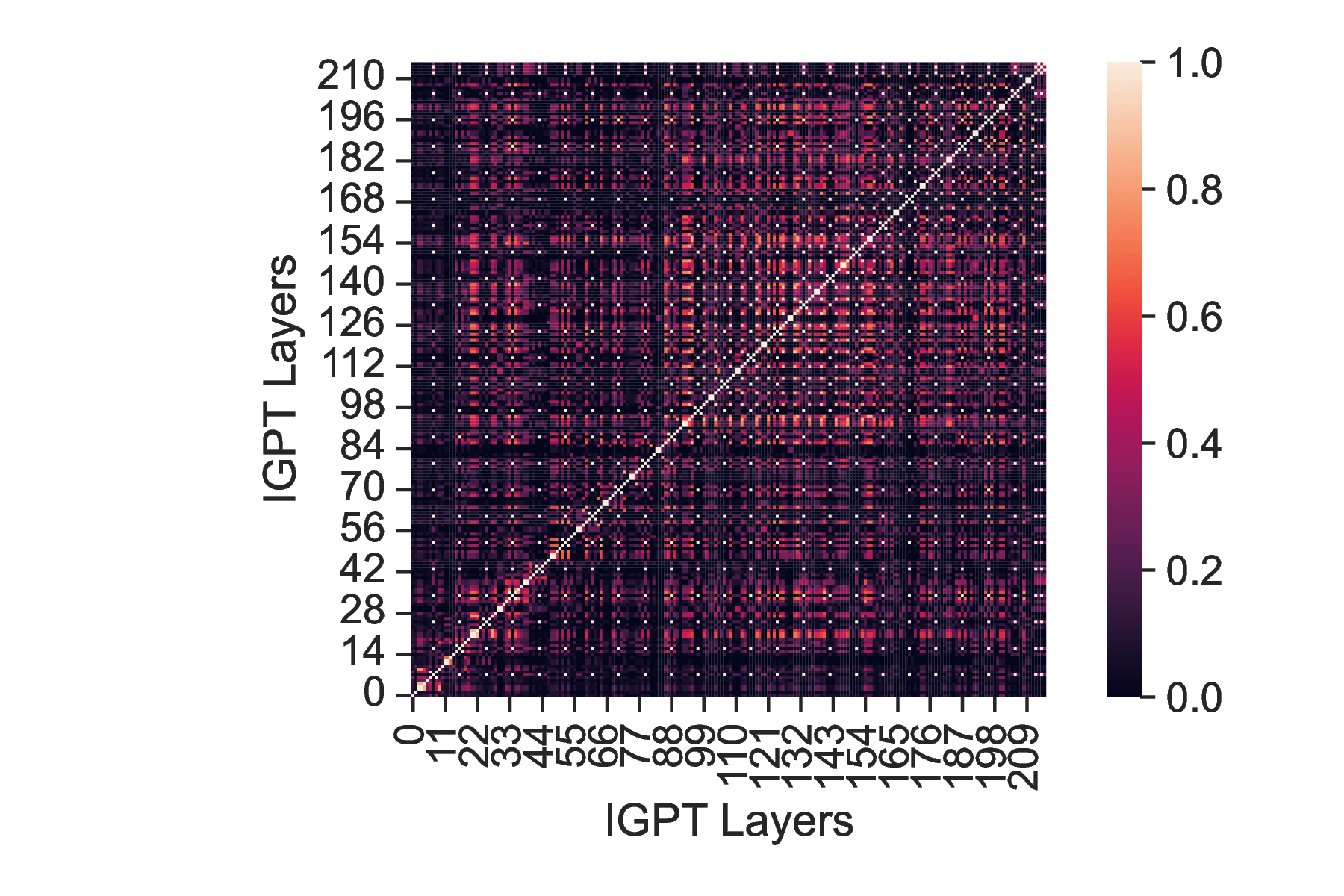}
        \subcaption{iGPT}
    \end{minipage}
    \begin{minipage}[b]{0.32\linewidth}
        \includegraphics[width=\linewidth]{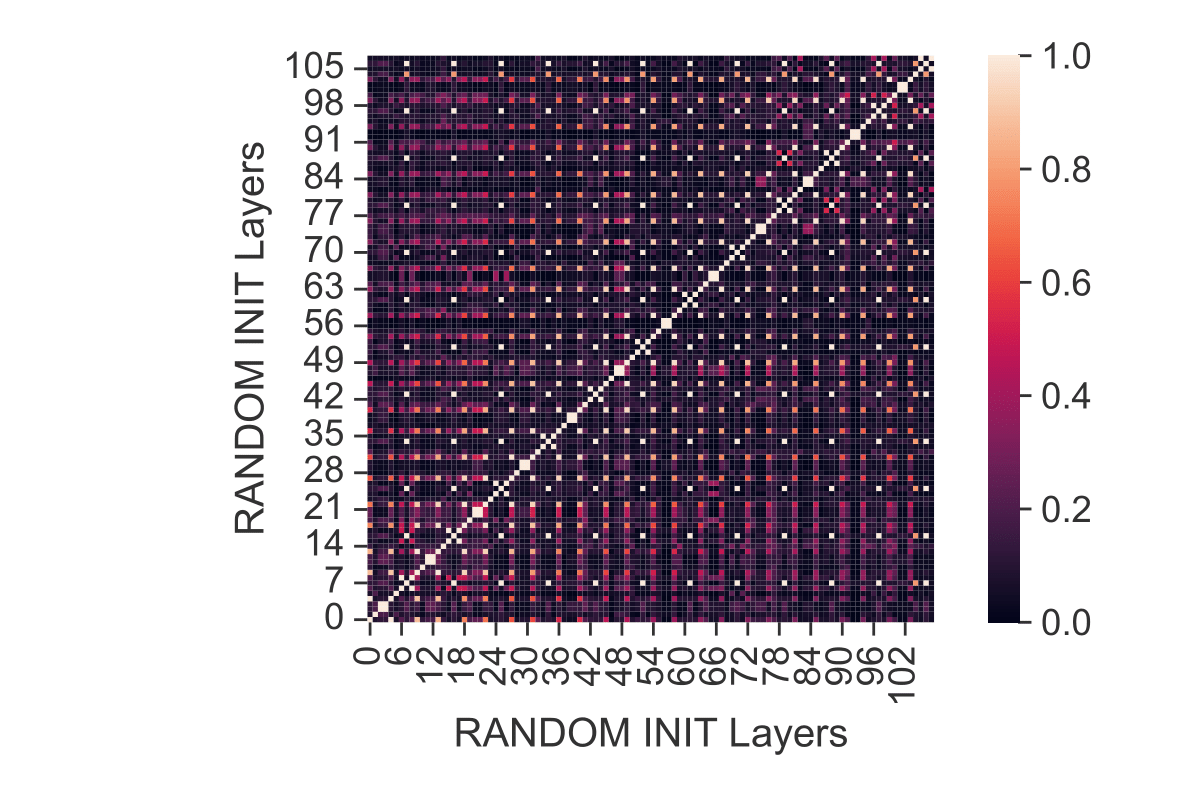}
        \subcaption{Random Initialization}
    \end{minipage}
    \caption{CKA of different layers in the same model (Hopper \& Return-to-go).}
\end{figure}

\begin{figure}[H]
    \centering
    \begin{minipage}[b]{0.32\linewidth}
        \includegraphics[width=\linewidth]{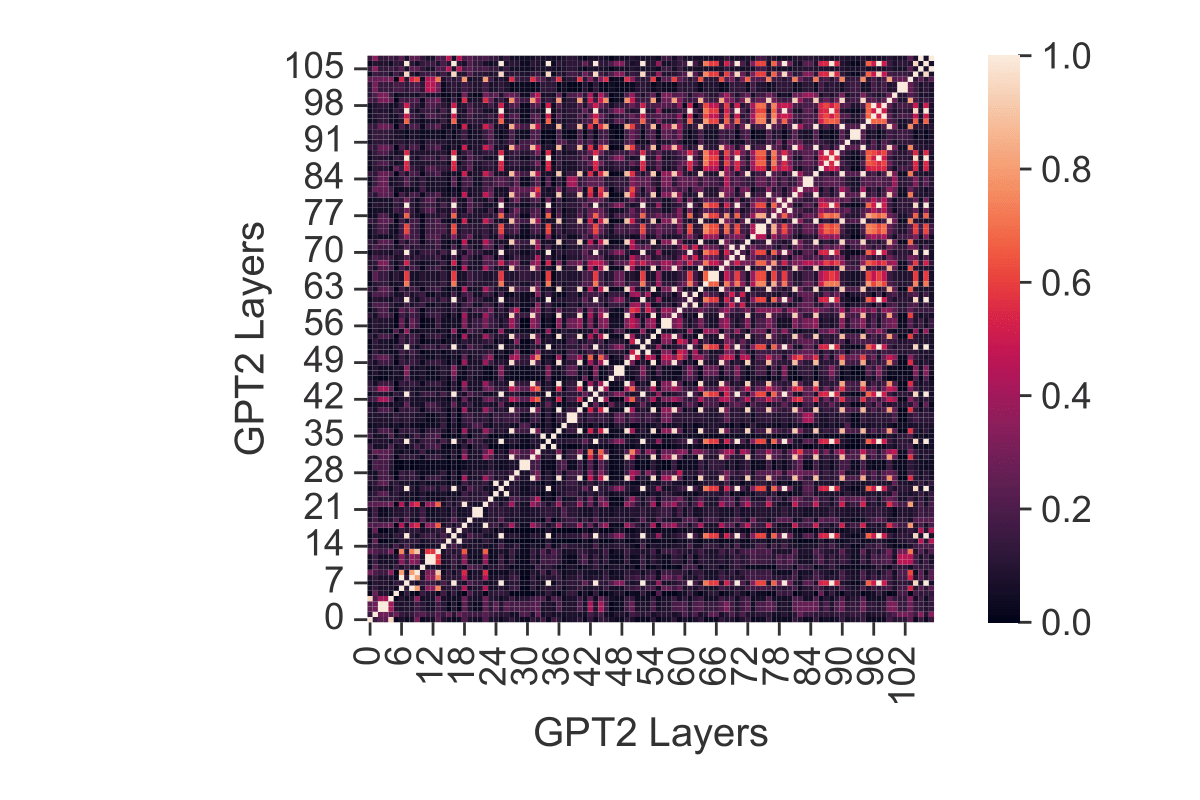}
        \subcaption{GPT2}
    \end{minipage}
    \begin{minipage}[b]{0.32\linewidth}
        \includegraphics[width=\linewidth]{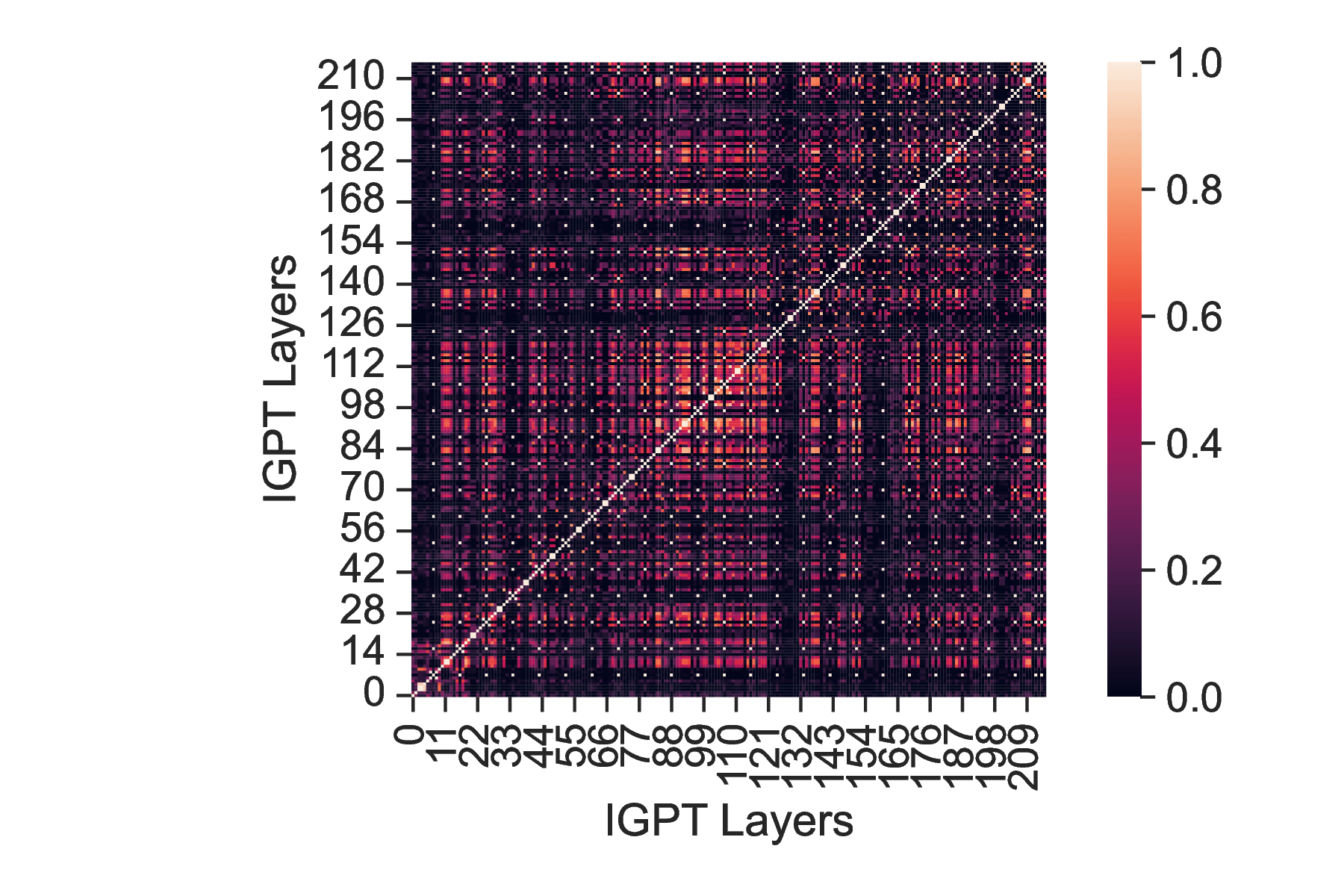}
        \subcaption{iGPT}
    \end{minipage}
    \begin{minipage}[b]{0.32\linewidth}
        \includegraphics[width=\linewidth]{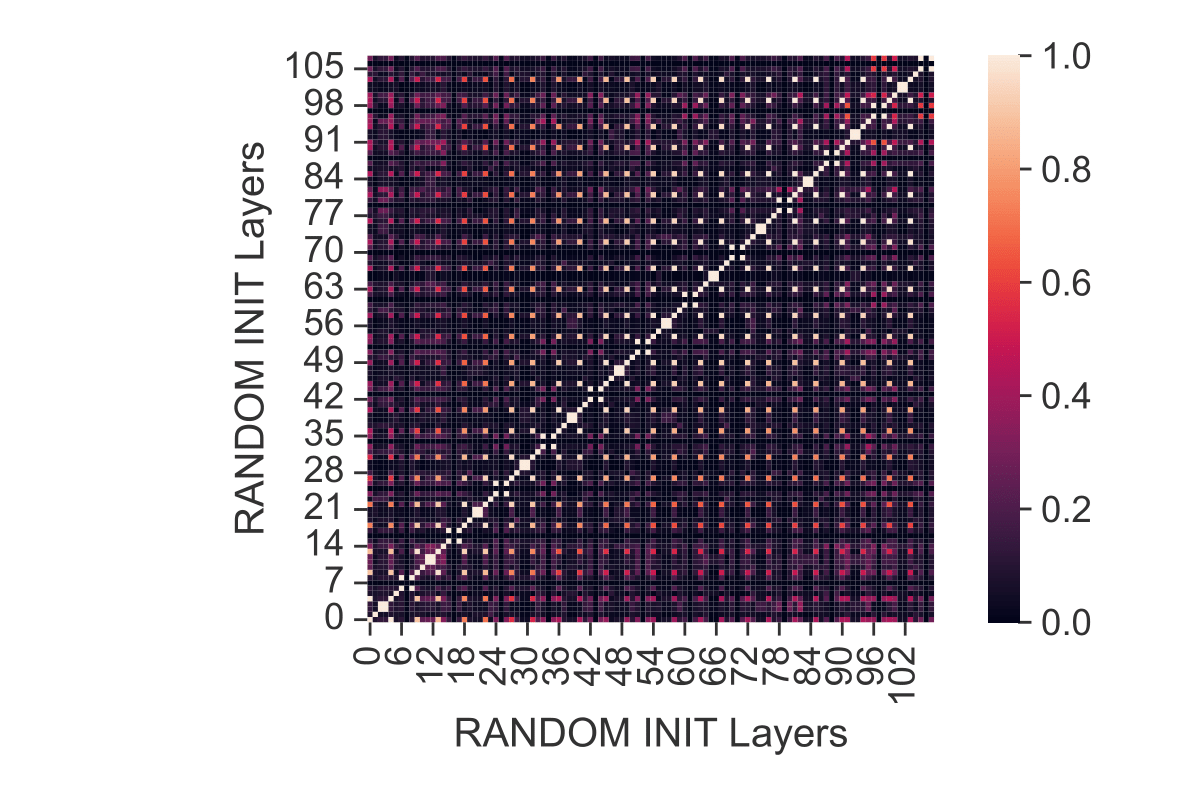}
        \subcaption{Random Initialization}
    \end{minipage}
    \caption{CKA of different layers in the same model (Hopper \& Action).}
\end{figure}

\begin{figure}[H]
    \centering
    \begin{minipage}[b]{0.32\linewidth}
        \includegraphics[width=\linewidth]{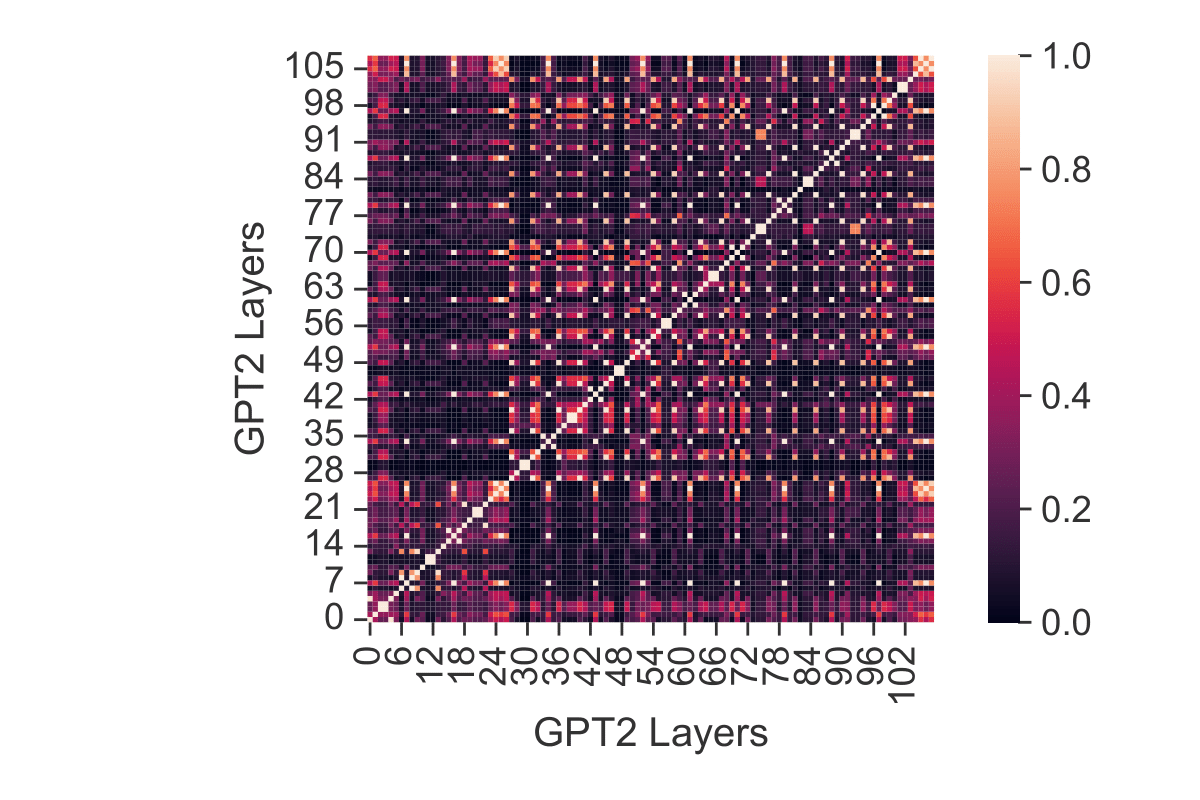}
        \subcaption{GPT2}
    \end{minipage}
    \begin{minipage}[b]{0.32\linewidth}
        \includegraphics[width=\linewidth]{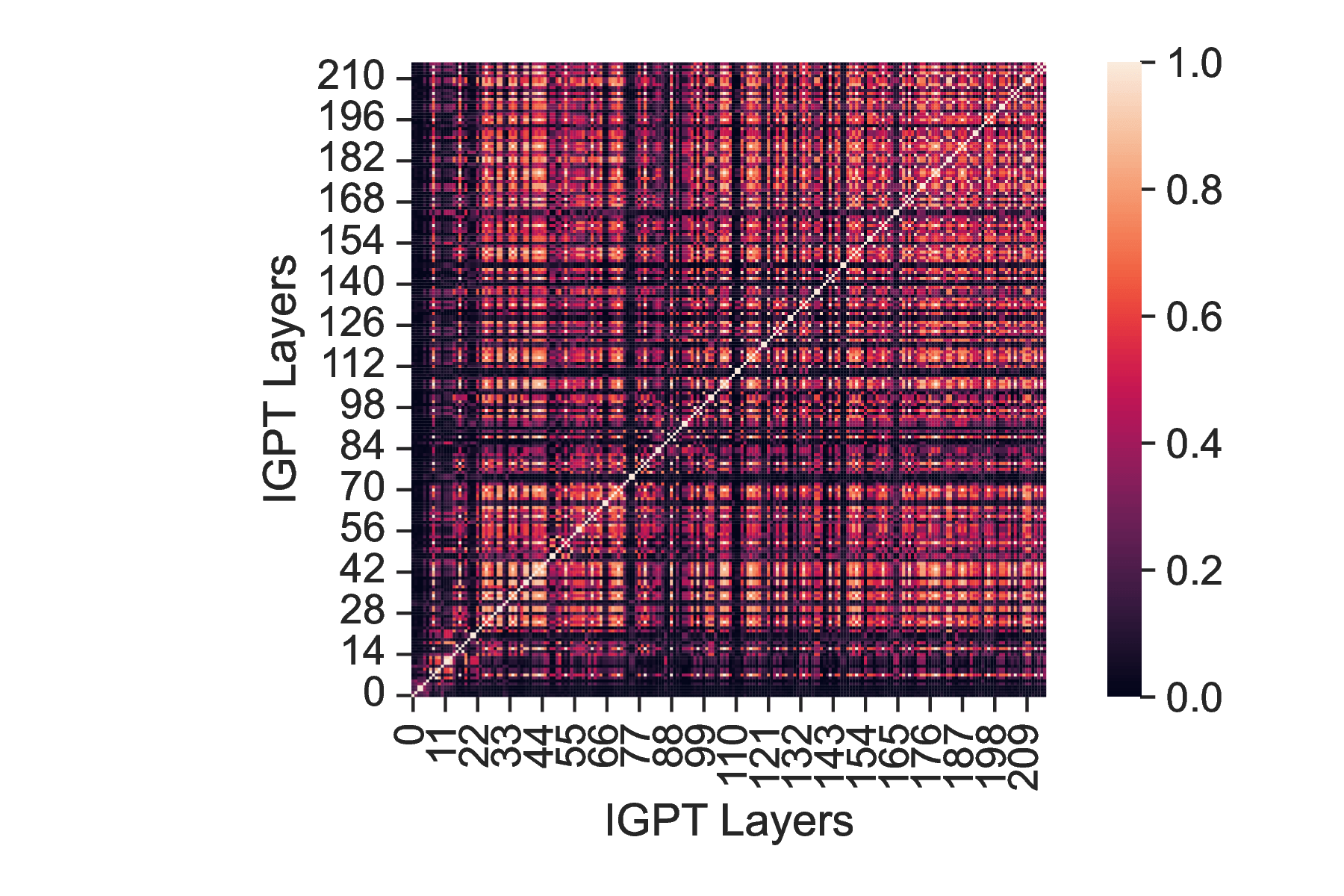}
        \subcaption{iGPT}
    \end{minipage}
    \begin{minipage}[b]{0.32\linewidth}
        \includegraphics[width=\linewidth]{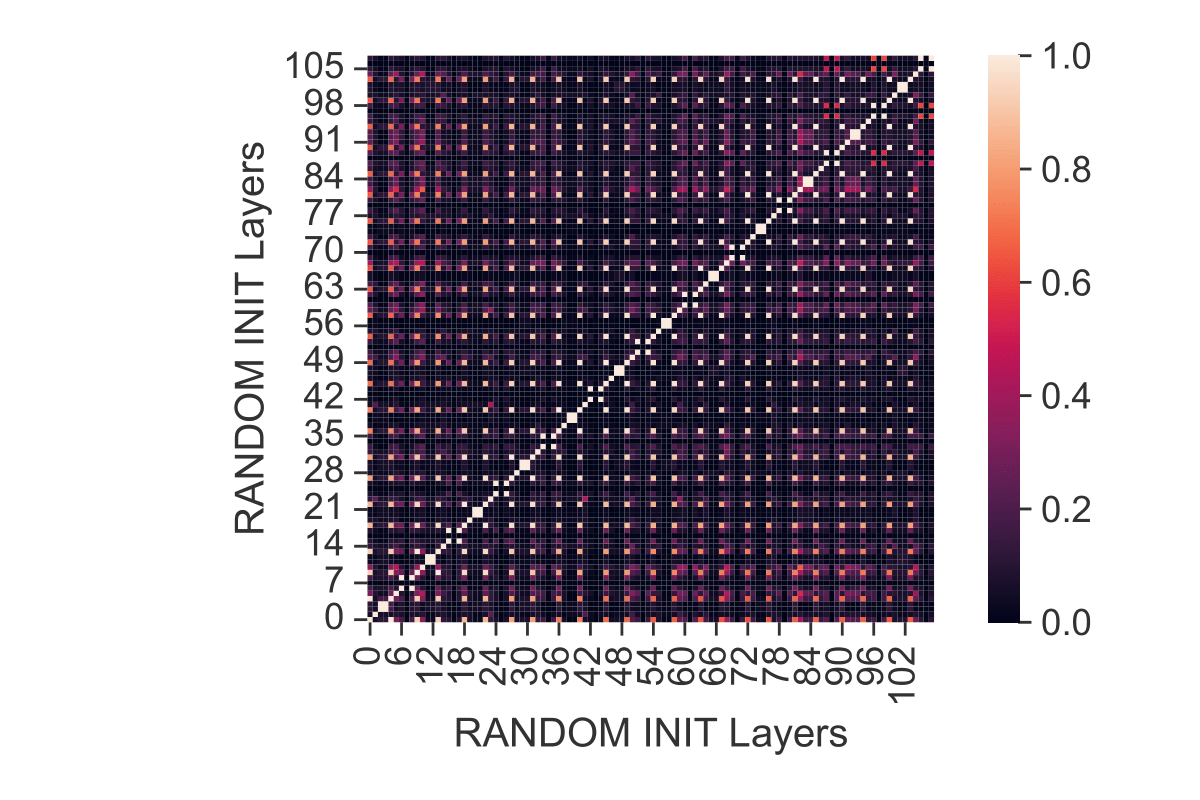}
        \subcaption{Random Initialization}
    \end{minipage}
    \caption{CKA of different layers in the same model (HalfCheetah \& State).}
\end{figure}

\begin{figure}[H]
    \centering
    \begin{minipage}[b]{0.32\linewidth}
        \includegraphics[width=\linewidth]{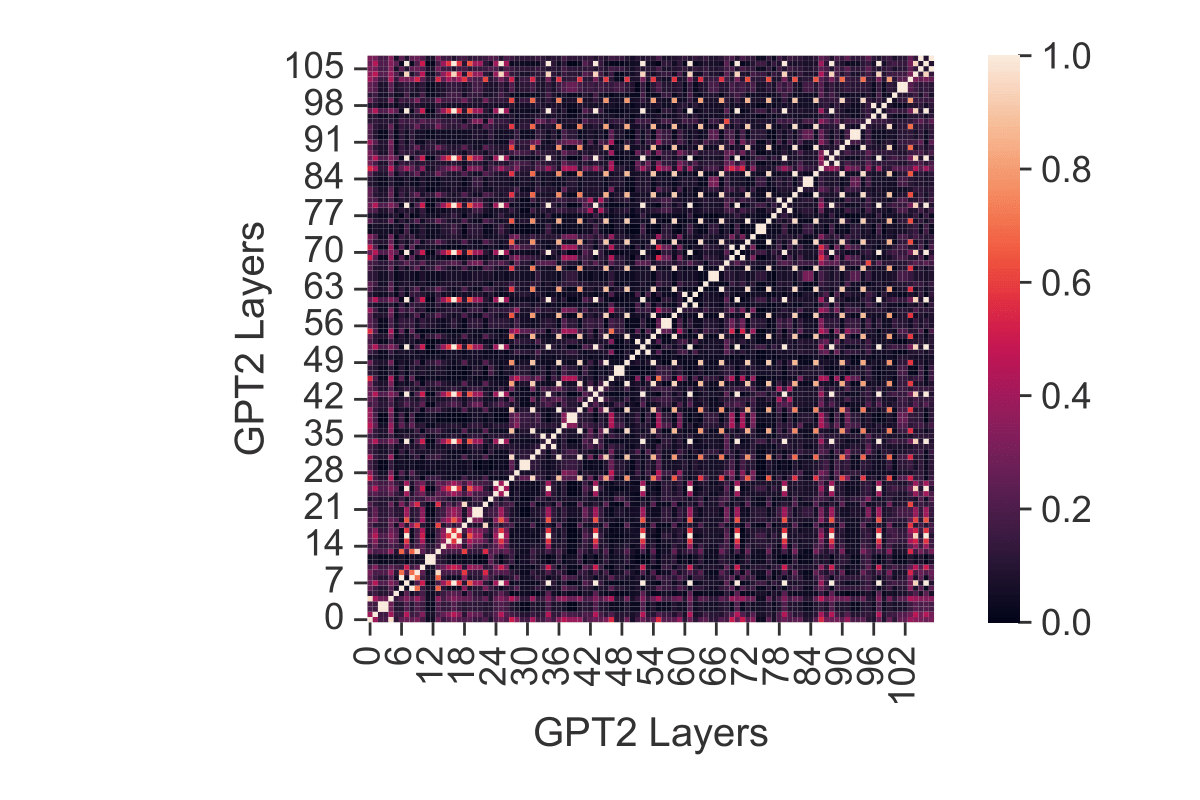}
        \subcaption{GPT2}
    \end{minipage}
    \begin{minipage}[b]{0.32\linewidth}
        \includegraphics[width=\linewidth]{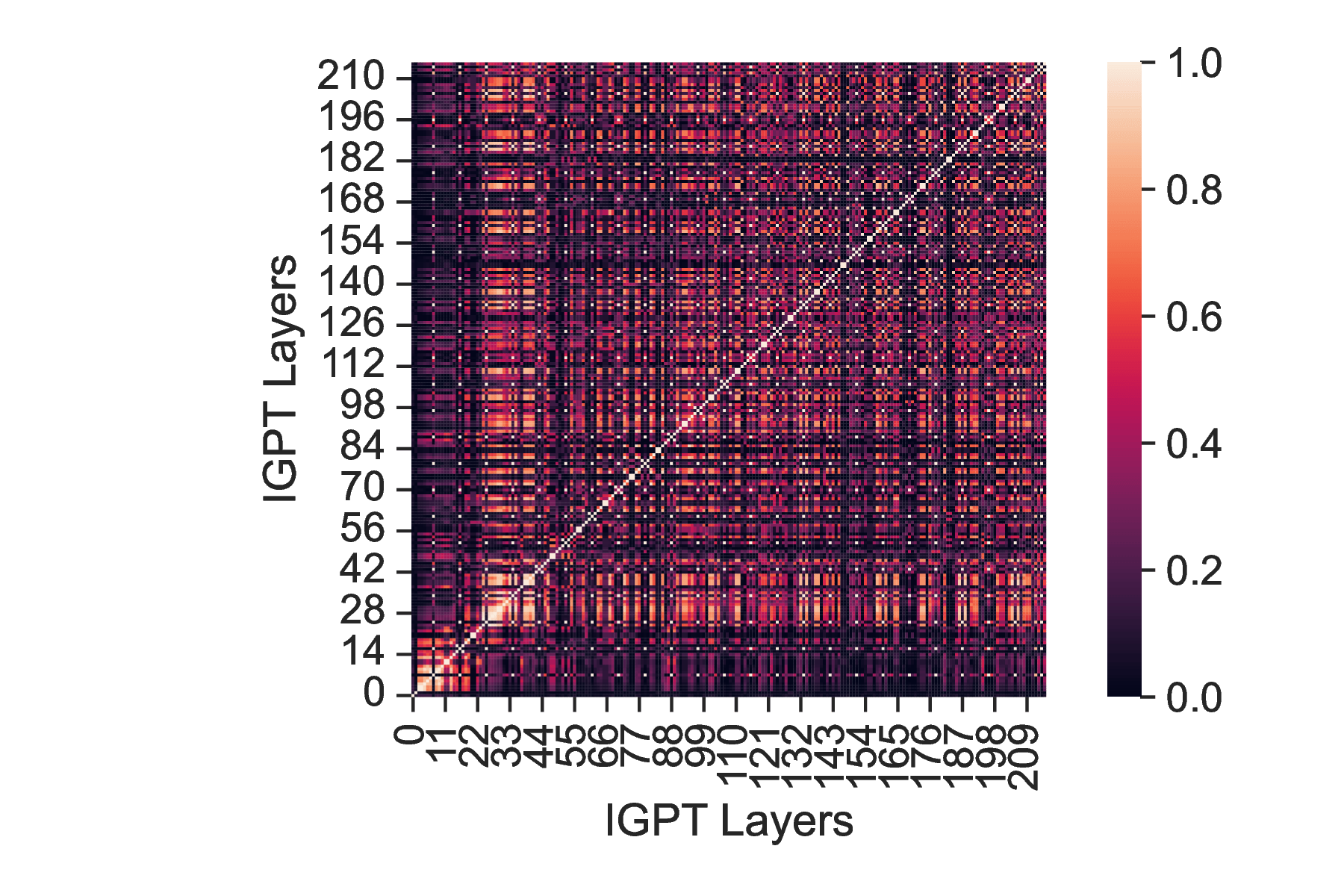}
        \subcaption{iGPT}
    \end{minipage}
    \begin{minipage}[b]{0.32\linewidth}
        \includegraphics[width=\linewidth]{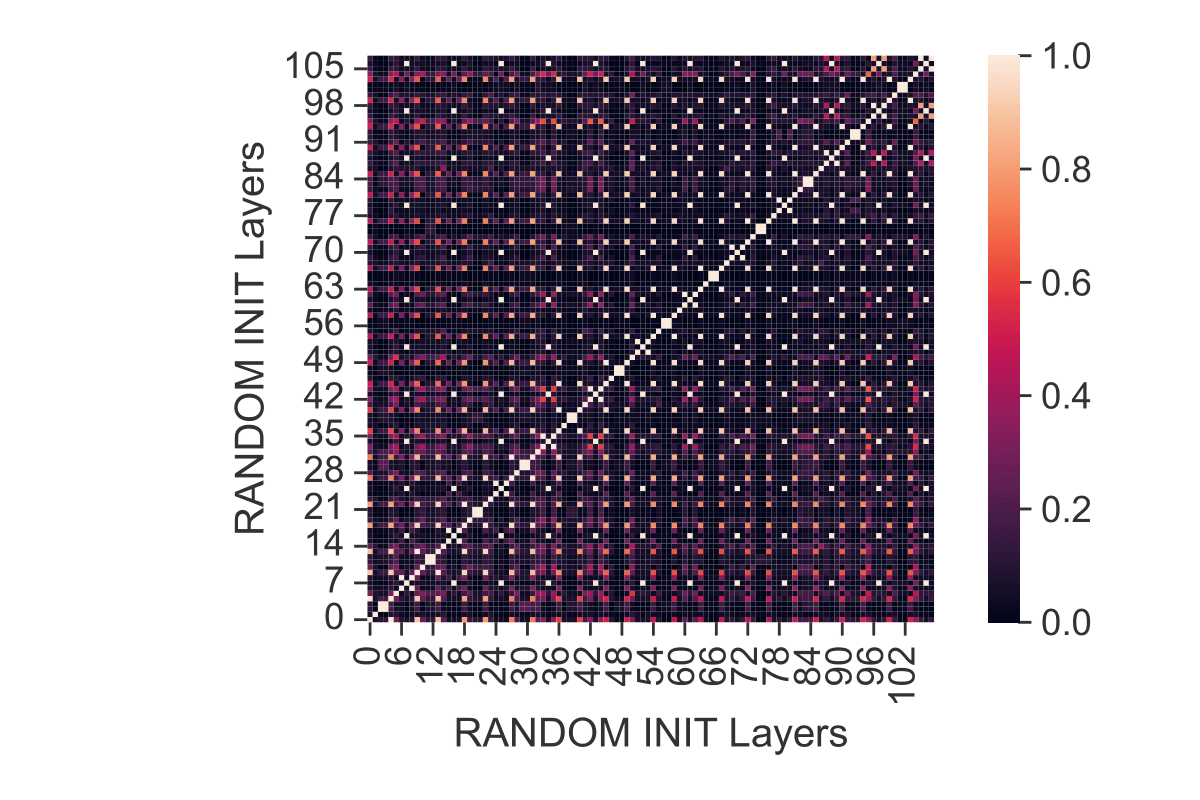}
        \subcaption{Random Initialization}
    \end{minipage}
    \caption{CKA of different layers in the same model (HalfCheetah \& Return-to-go).}
\end{figure}

\begin{figure}[H]
    \centering
    \begin{minipage}[b]{0.32\linewidth}
        \includegraphics[width=\linewidth]{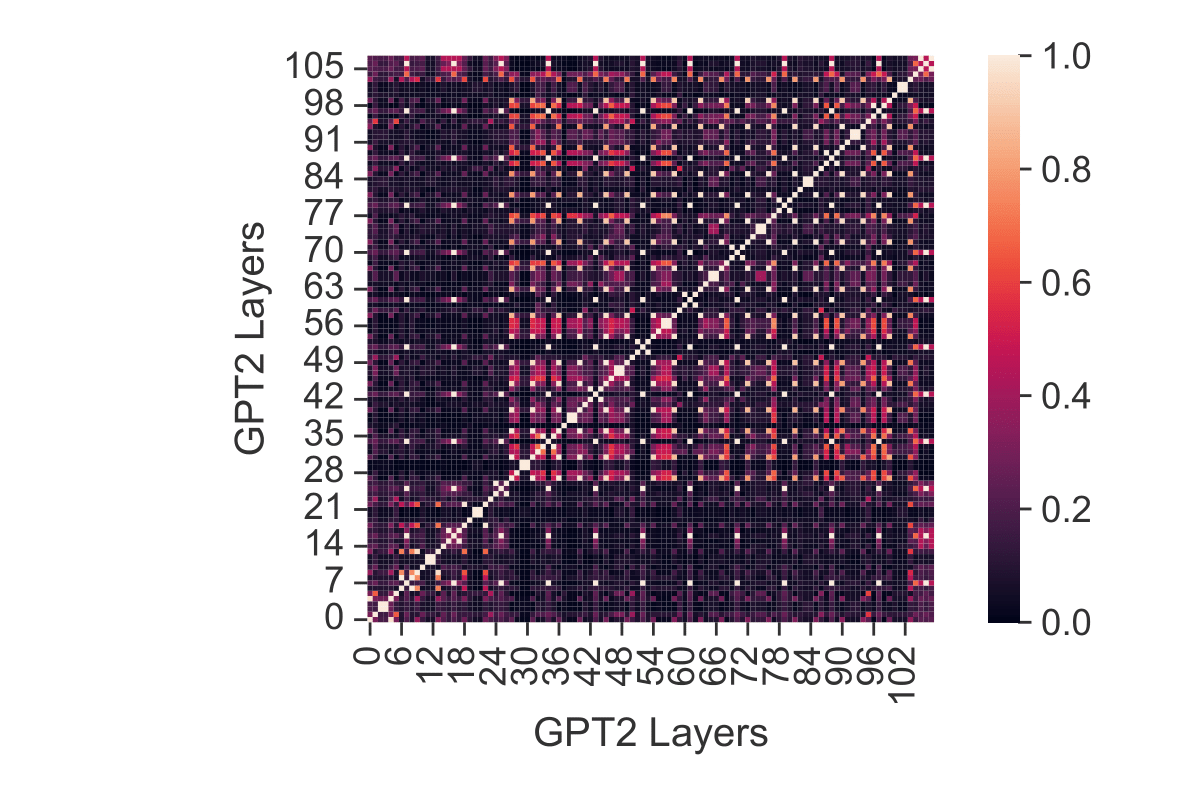}
        \subcaption{GPT2}
    \end{minipage}
    \begin{minipage}[b]{0.32\linewidth}
        \includegraphics[width=\linewidth]{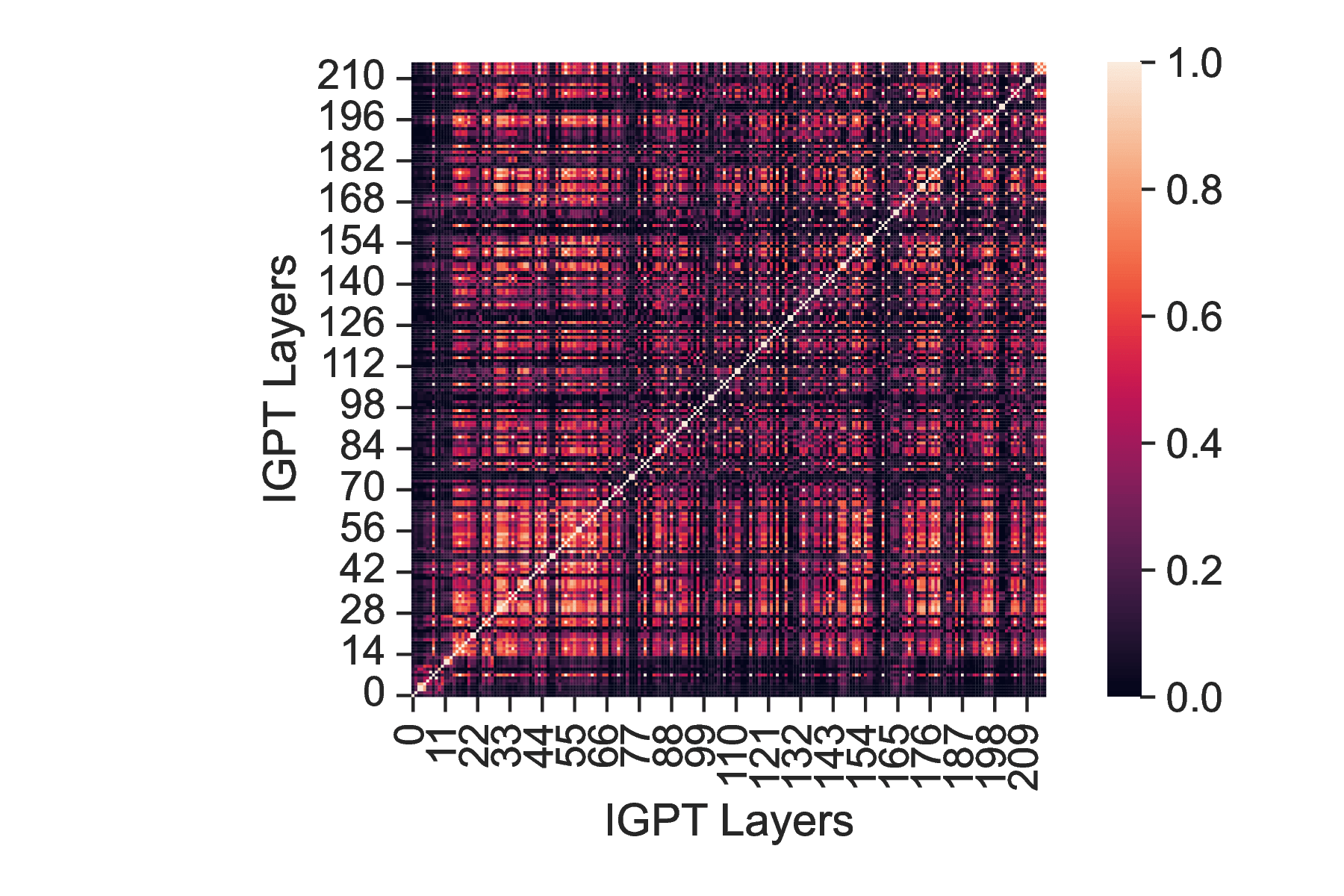}
        \subcaption{iGPT}
    \end{minipage}
    \begin{minipage}[b]{0.32\linewidth}
        \includegraphics[width=\linewidth]{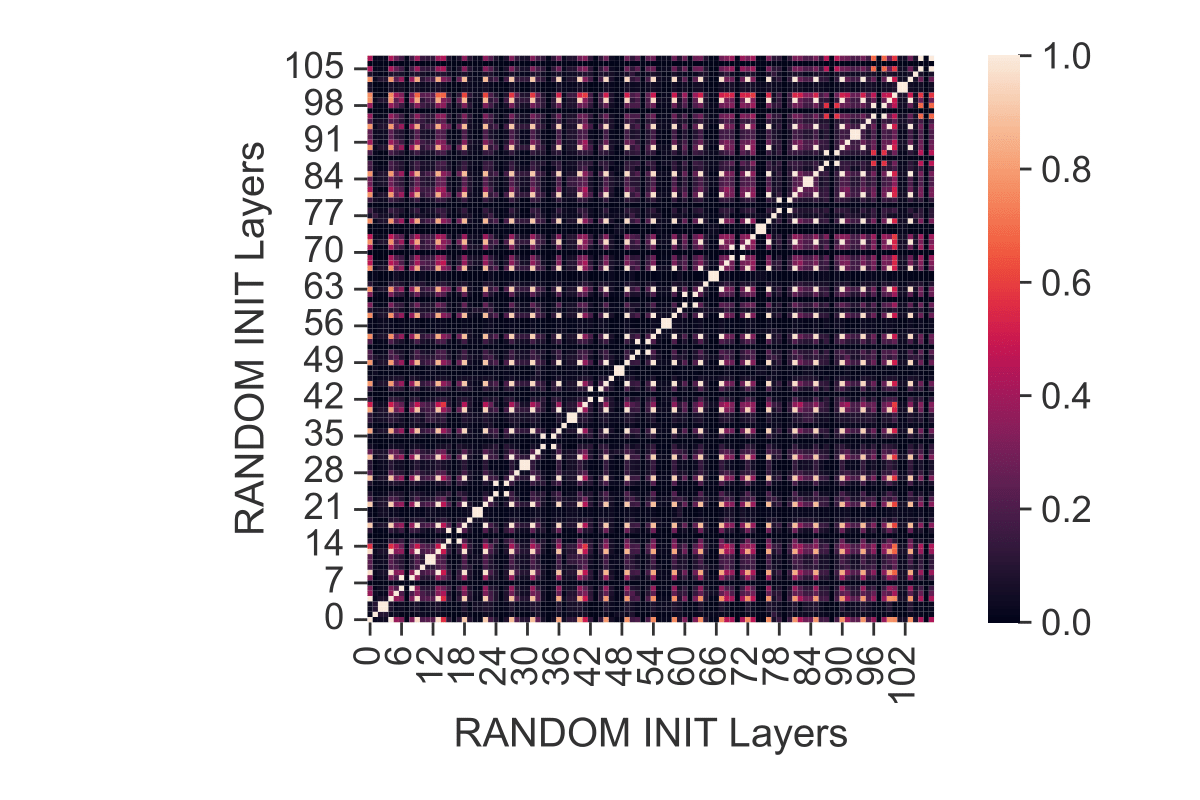}
        \subcaption{Random Initialization}
    \end{minipage}
    \caption{CKA of different layers in the same model (HalfCheetah \& Action).}
\end{figure}

\begin{figure}[H]
    \centering
    \begin{minipage}[b]{0.32\linewidth}
        \includegraphics[width=\linewidth]{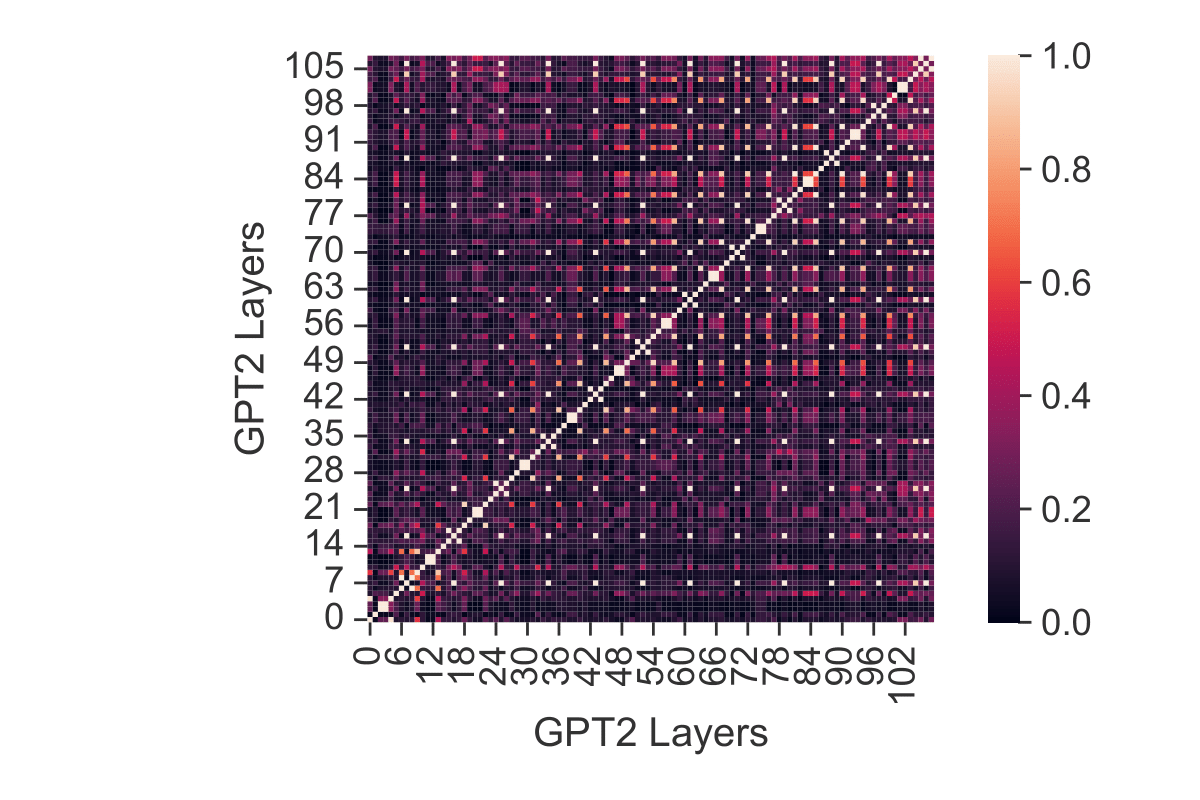}
        \subcaption{GPT2}
    \end{minipage}
    \begin{minipage}[b]{0.32\linewidth}
        \includegraphics[width=\linewidth]{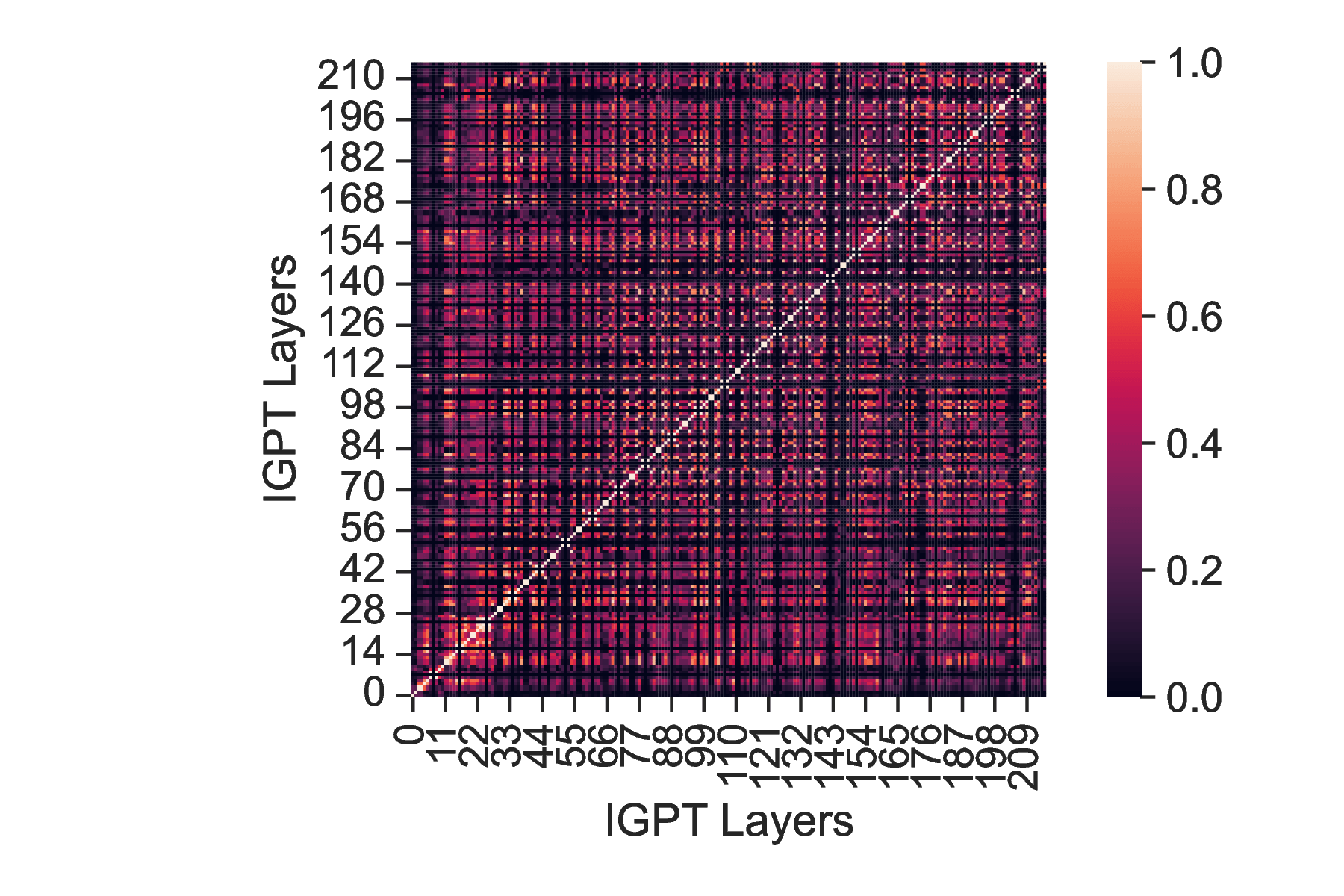}
        \subcaption{iGPT}
    \end{minipage}
    \begin{minipage}[b]{0.32\linewidth}
        \includegraphics[width=\linewidth]{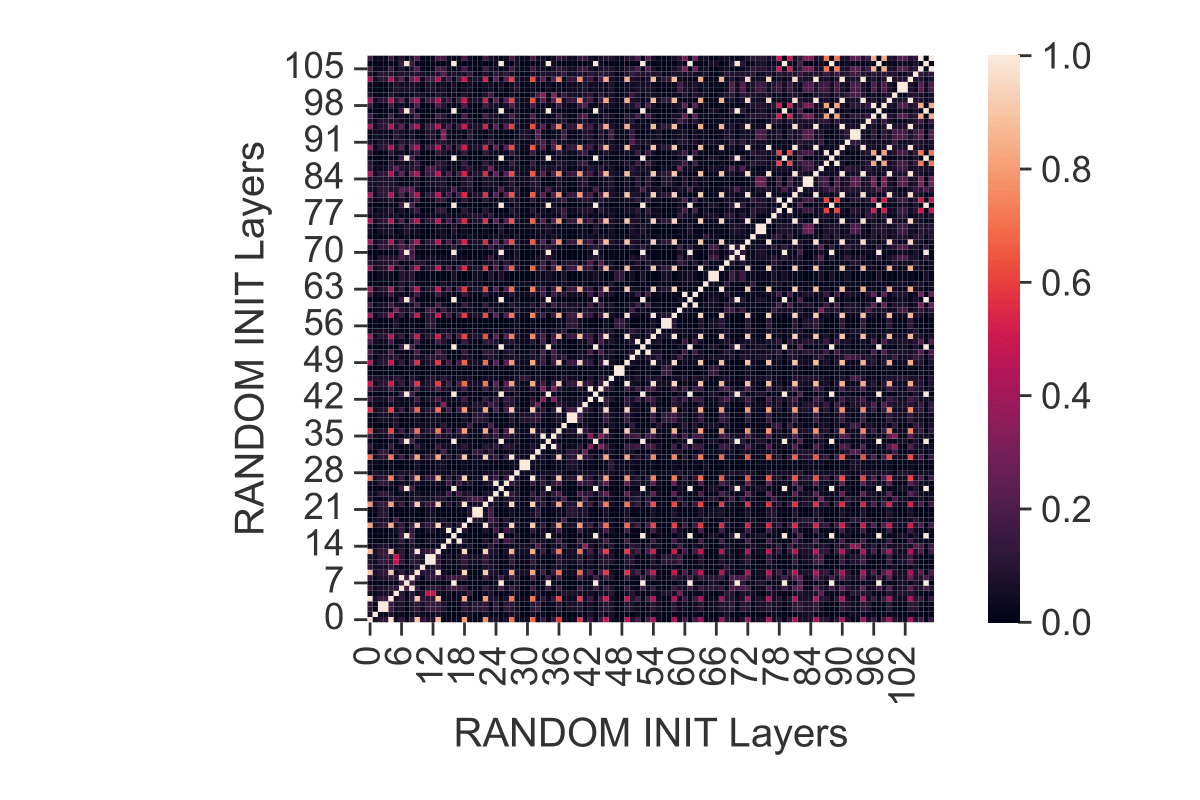}
        \subcaption{Random Initialization}
    \end{minipage}
    \caption{CKA of different layers in the same model (Walker2D \& State).}
\end{figure}

\begin{figure}[H]
    \centering
    \begin{minipage}[b]{0.32\linewidth}
        \includegraphics[width=\linewidth]{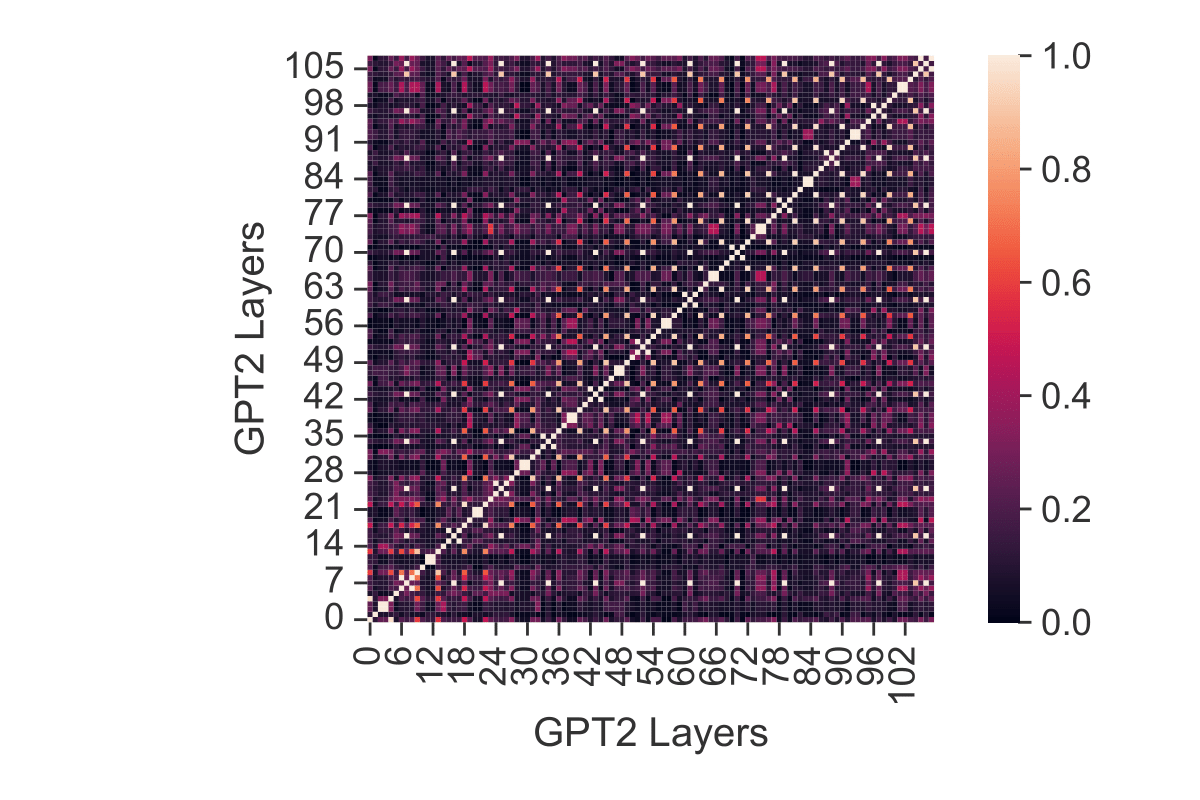}
        \subcaption{GPT2}
    \end{minipage}
    \begin{minipage}[b]{0.32\linewidth}
        \includegraphics[width=\linewidth]{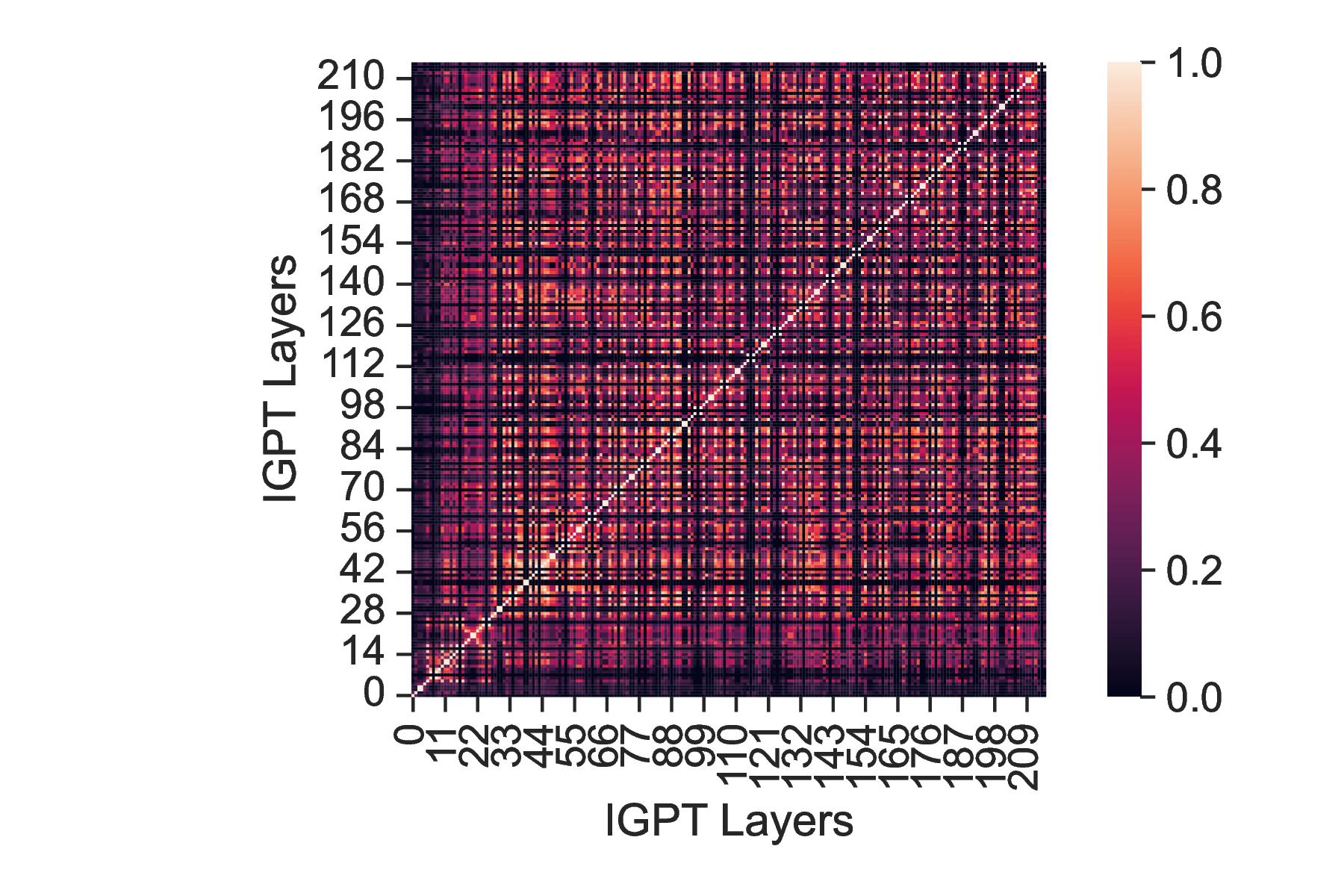}
        \subcaption{iGPT}
    \end{minipage}
    \begin{minipage}[b]{0.32\linewidth}
        \includegraphics[width=\linewidth]{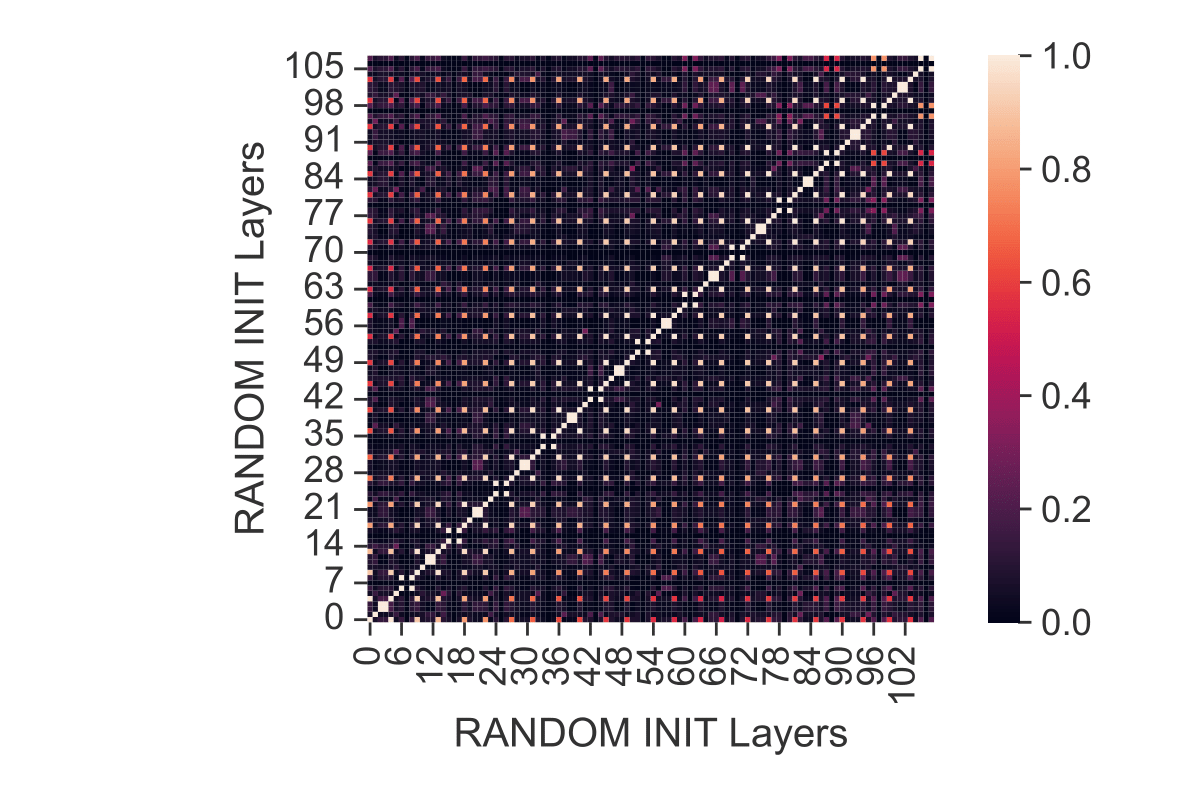}
        \subcaption{Random Initialization}
    \end{minipage}
    \caption{CKA of different layers in the same model (Walker2D \& Return-to-go).}
\end{figure}

\begin{figure}[H]
    \centering
    \begin{minipage}[b]{0.32\linewidth}
        \includegraphics[width=\linewidth]{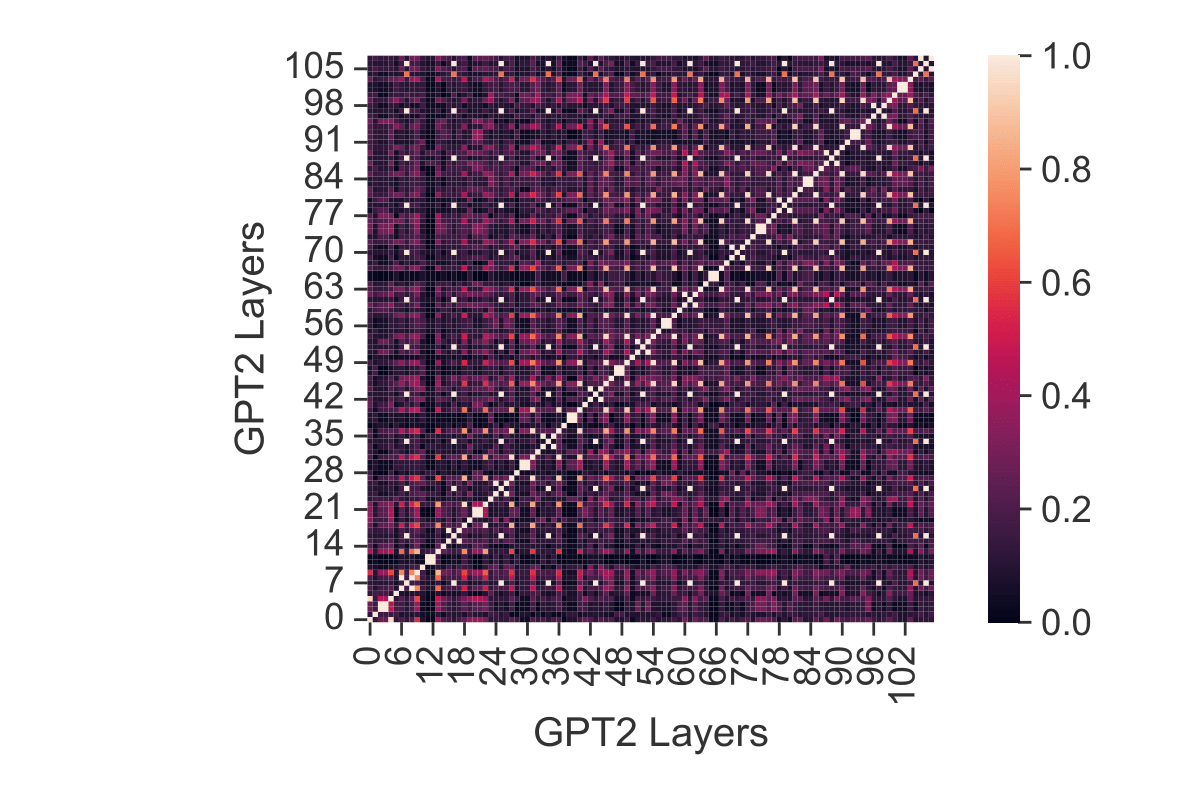}
        \subcaption{GPT2}
    \end{minipage}
    \begin{minipage}[b]{0.32\linewidth}
        \includegraphics[width=\linewidth]{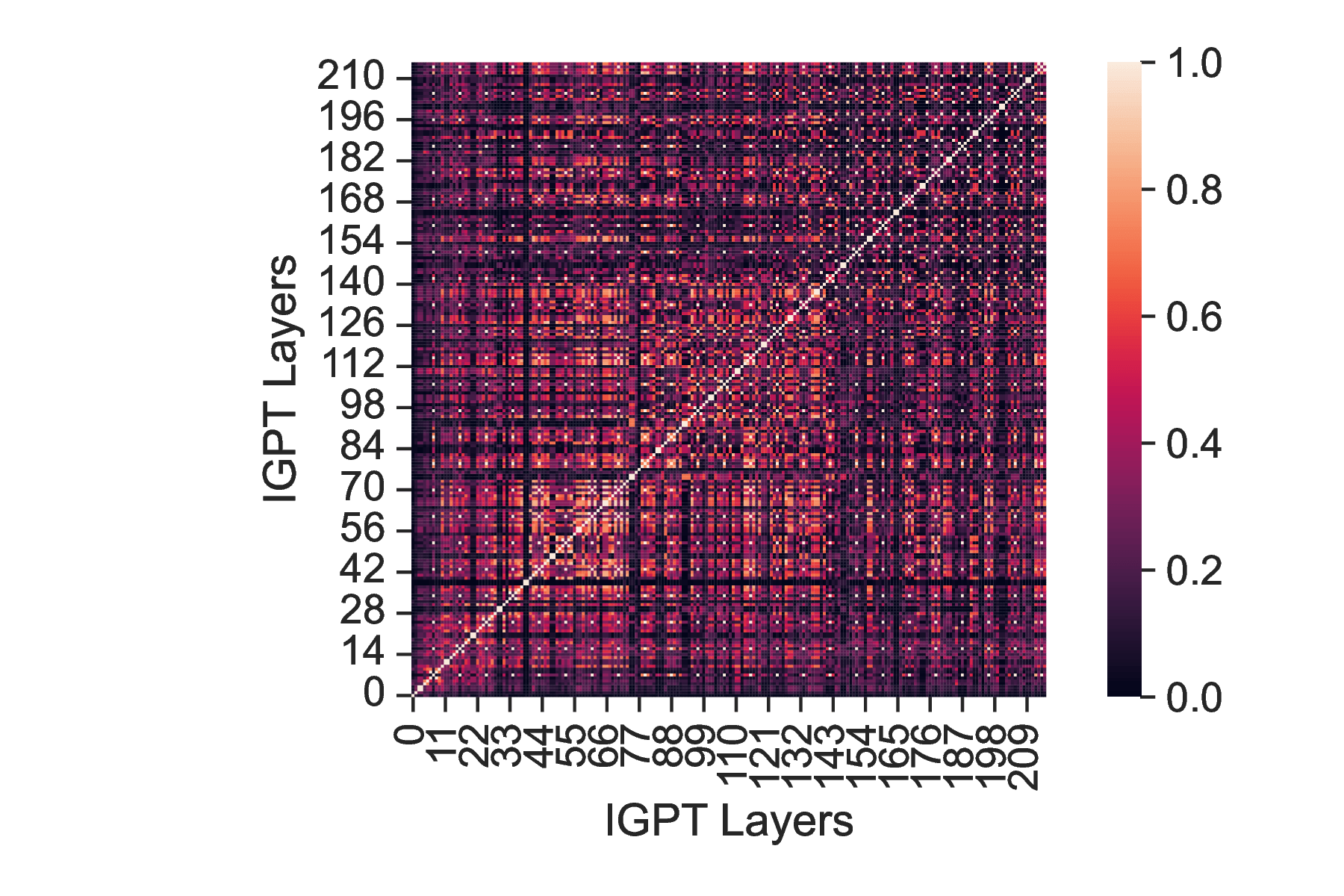}
        \subcaption{iGPT}
    \end{minipage}
    \begin{minipage}[b]{0.32\linewidth}
        \includegraphics[width=\linewidth]{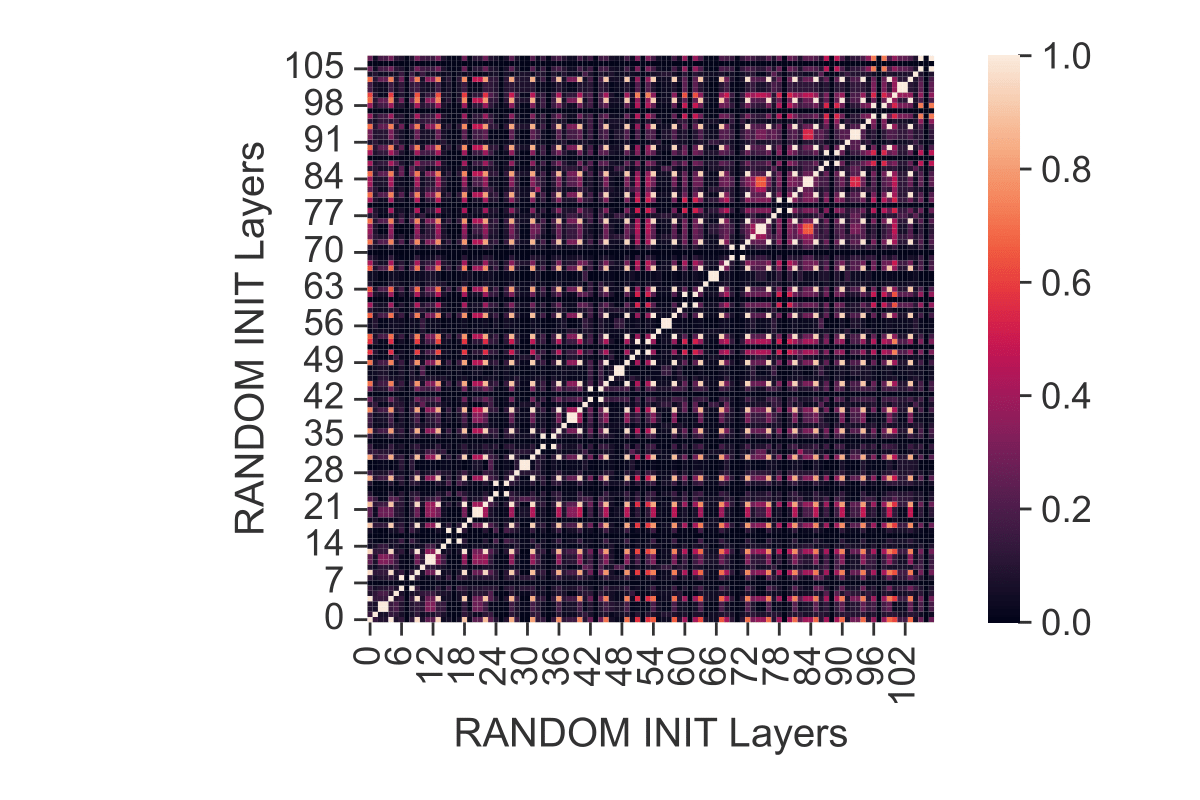}
        \subcaption{Random Initialization}
    \end{minipage}
    \caption{CKA of different layers in the same model (Walker2D \& Action).}
\end{figure}

\subsection{Mutual Information Between Hidden Representation and Data}
\label{appendix:results-for-other-conditions-mutual-information}

\begin{figure}[H]
    \centering
    \begin{minipage}[b]{0.45\linewidth}
    \includegraphics[width=\linewidth]{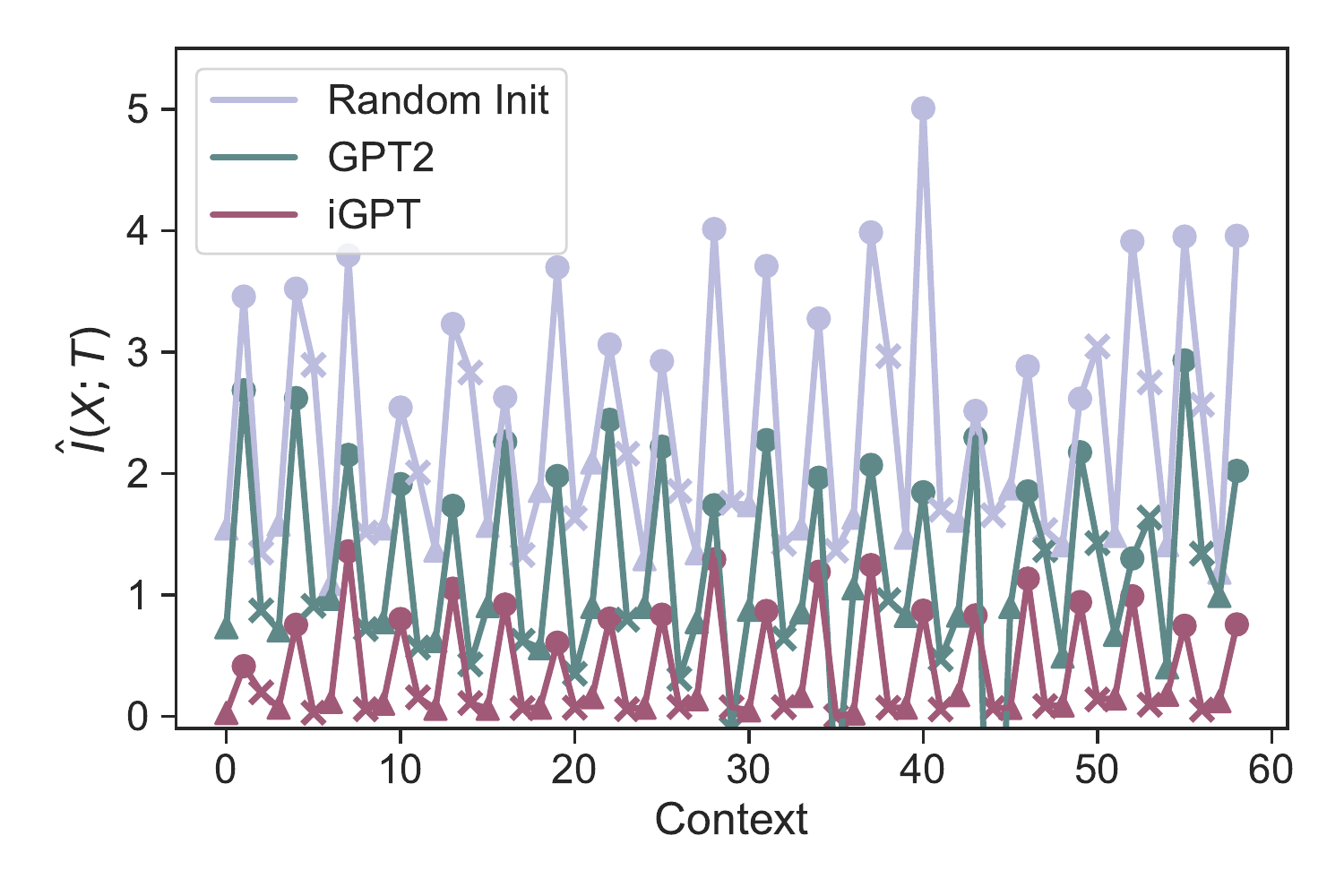}
    \subcaption{$\hat{I}(X; T)$}
    \end{minipage}
    \begin{minipage}[b]{0.45\linewidth}
    \includegraphics[width=\linewidth]{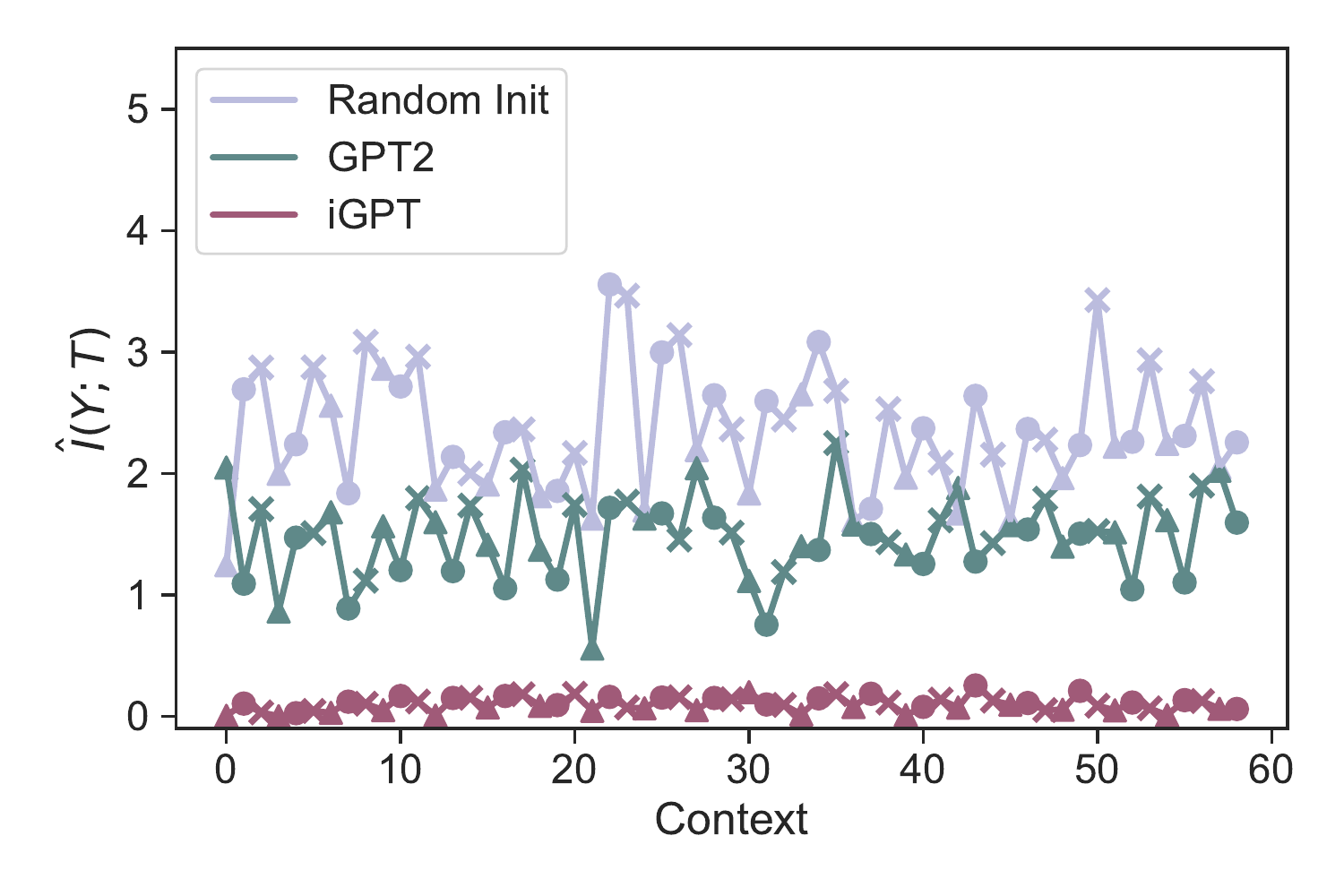}
    \subcaption{$\hat{I}(Y; T)$}
    \end{minipage}
    \caption{Estimated mutual information between data and hidden representation (Shallow).}
    \label{fig:mutual_information_context_shallow}
\end{figure}

\begin{figure}[H]
    \centering
    \begin{minipage}[b]{0.45\linewidth}
    \includegraphics[width=\linewidth]{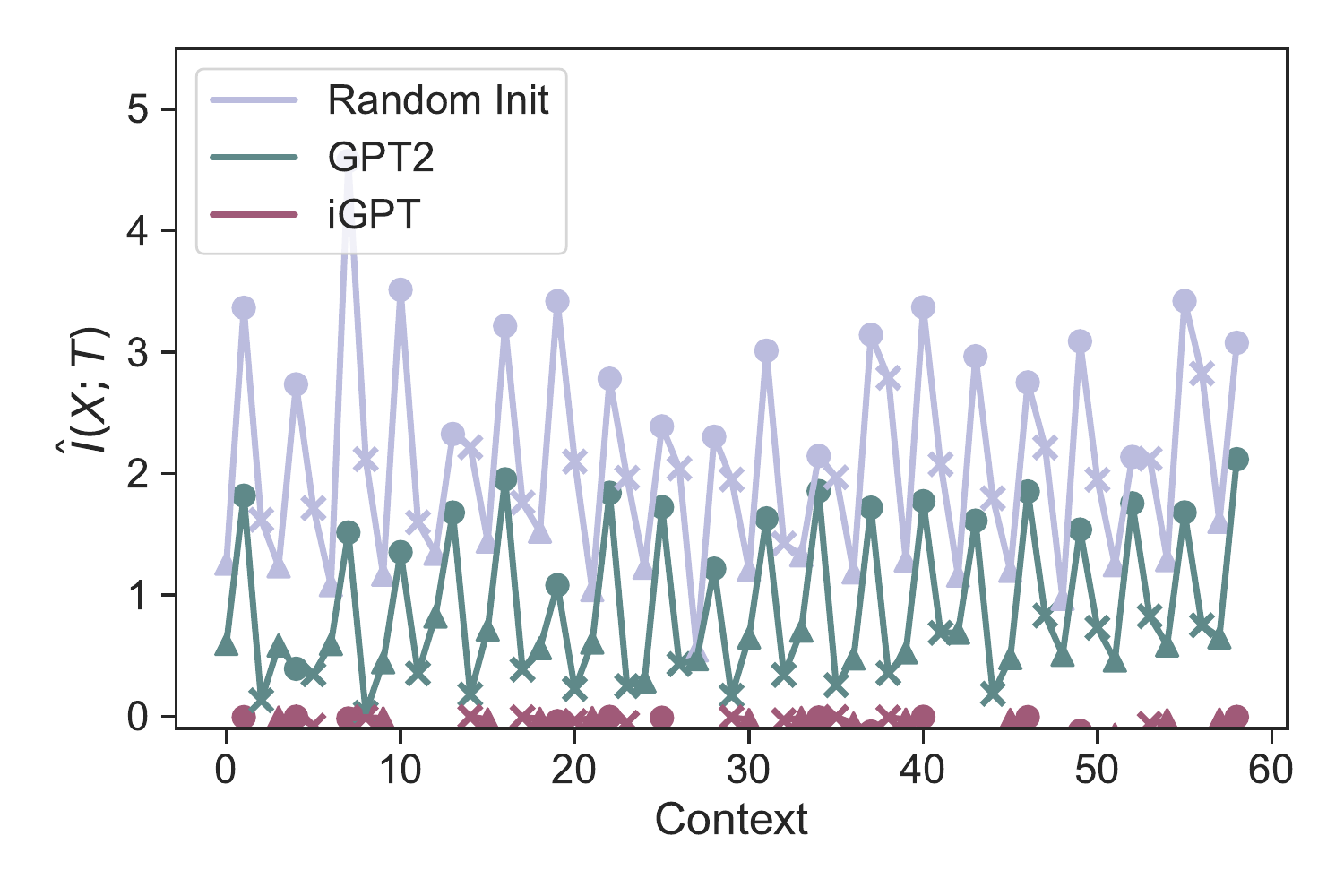}
    \subcaption{$\hat{I}(X; T)$}
    \end{minipage}
    \begin{minipage}[b]{0.45\linewidth}
    \includegraphics[width=\linewidth]{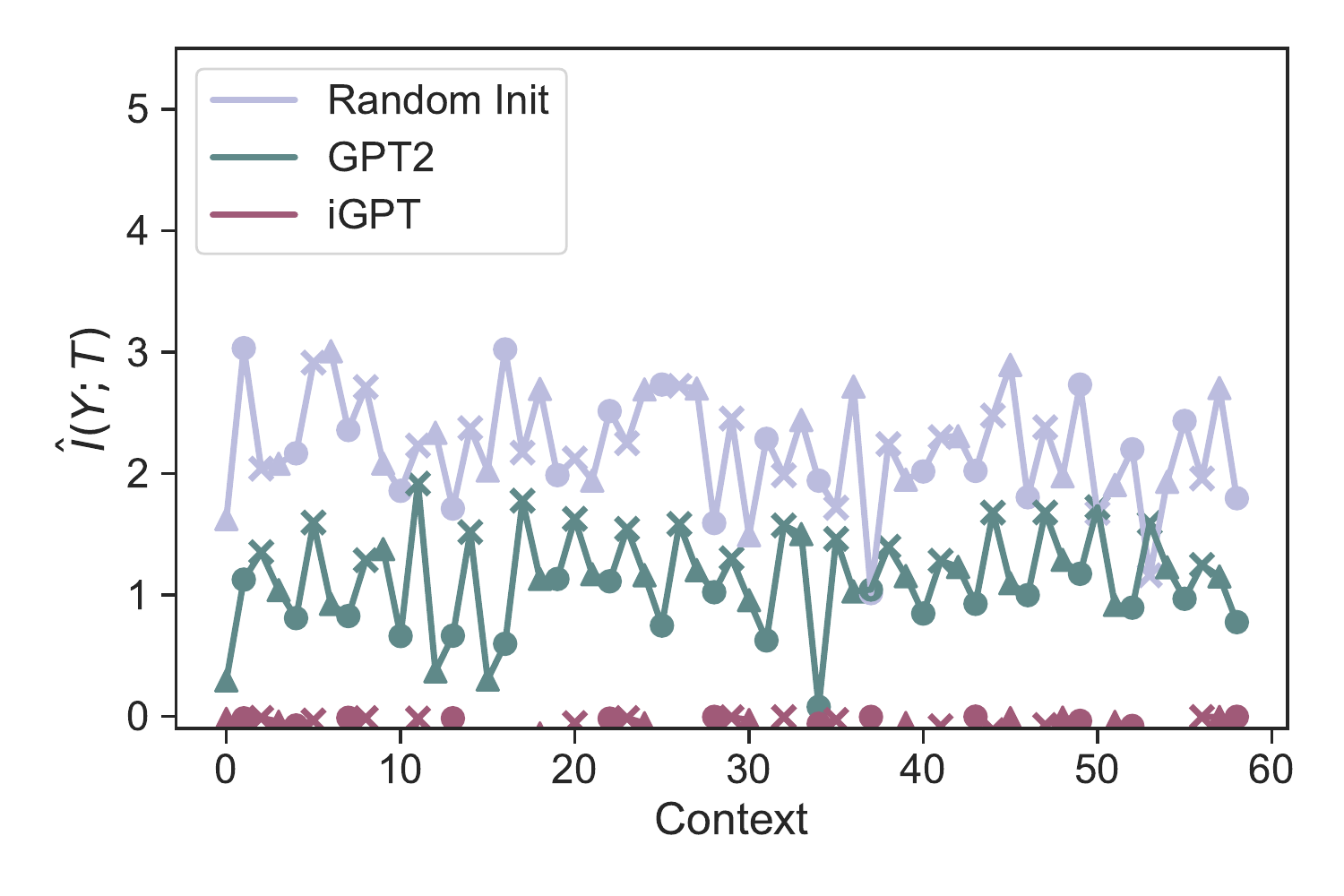}
    \subcaption{$\hat{I}(Y; T)$}
    \end{minipage}
    \caption{Estimated mutual information between data and hidden representation (Deep).}
    \label{fig:mutual_information_context_deep}
\end{figure}

\begin{figure}[H]
    \centering
    \begin{minipage}[b]{0.45\linewidth}
    \includegraphics[width=\linewidth]{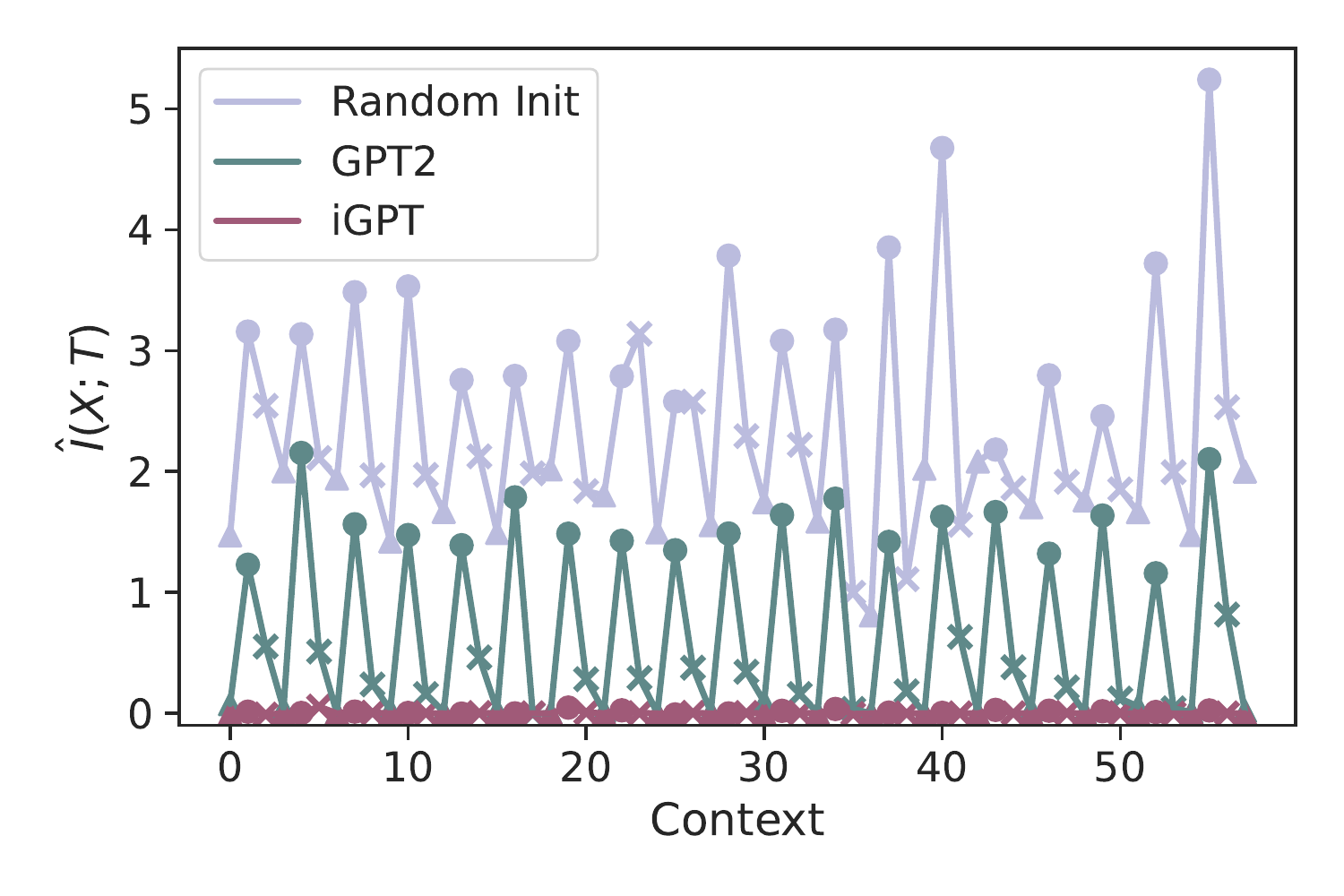}
    \subcaption{$\hat{I}(X; T)$}
    \end{minipage}
    \begin{minipage}[b]{0.45\linewidth}
    \includegraphics[width=\linewidth]{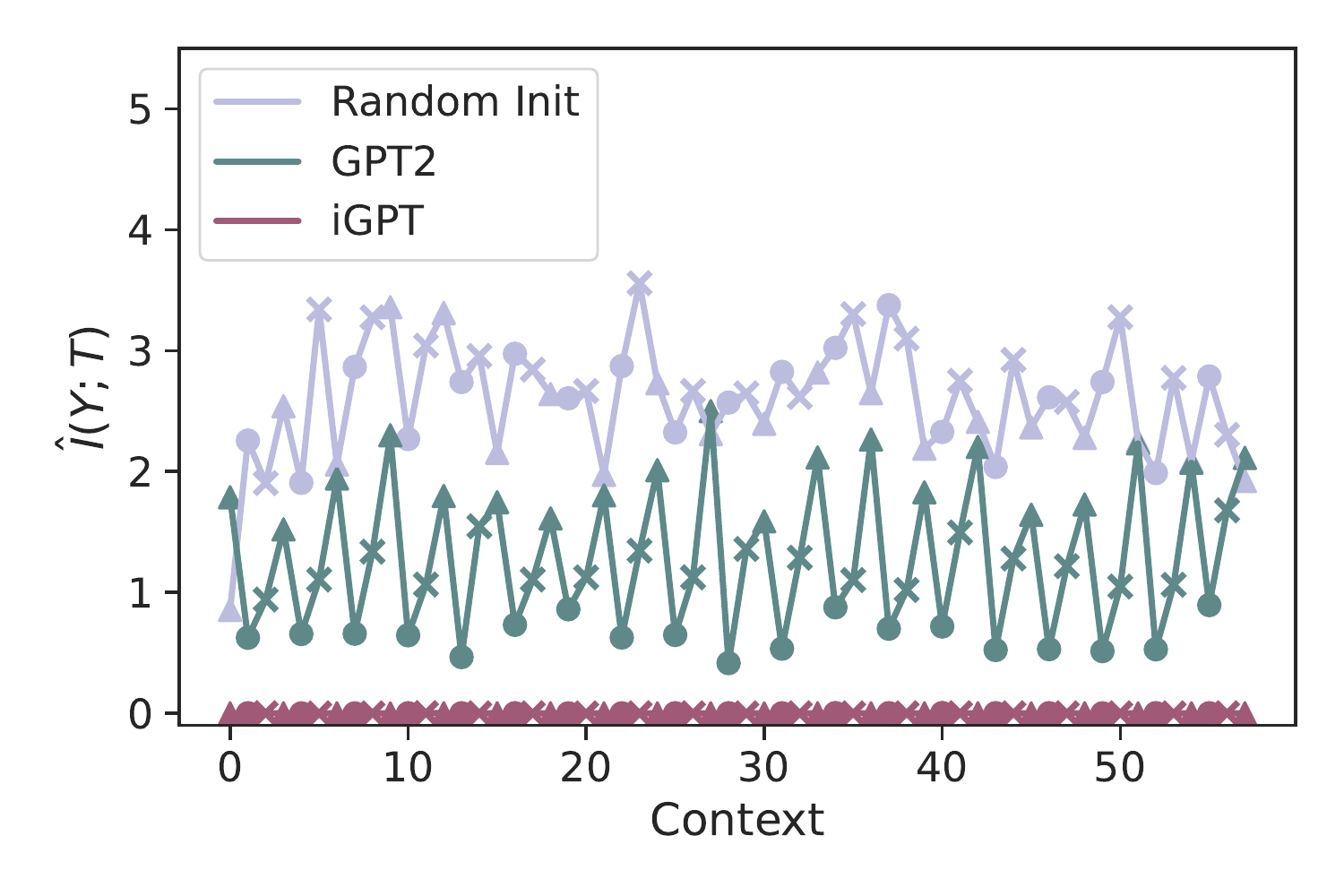}
    \subcaption{$\hat{I}(Y; T)$}
    \end{minipage}
    \caption{Estimated mutual information between data and hidden representation (Middle, Seed = 42).}
    \label{fig:mutual_information_context_middle_42}
\end{figure}

\subsection{Parameter Similarity}
\label{appendix:results-for-other-conditions-parameter-similarity}

\subsubsection{Parameter Similarity (Other Environment)}
\begin{figure}[H]
    \centering
        \includegraphics[width=\linewidth]{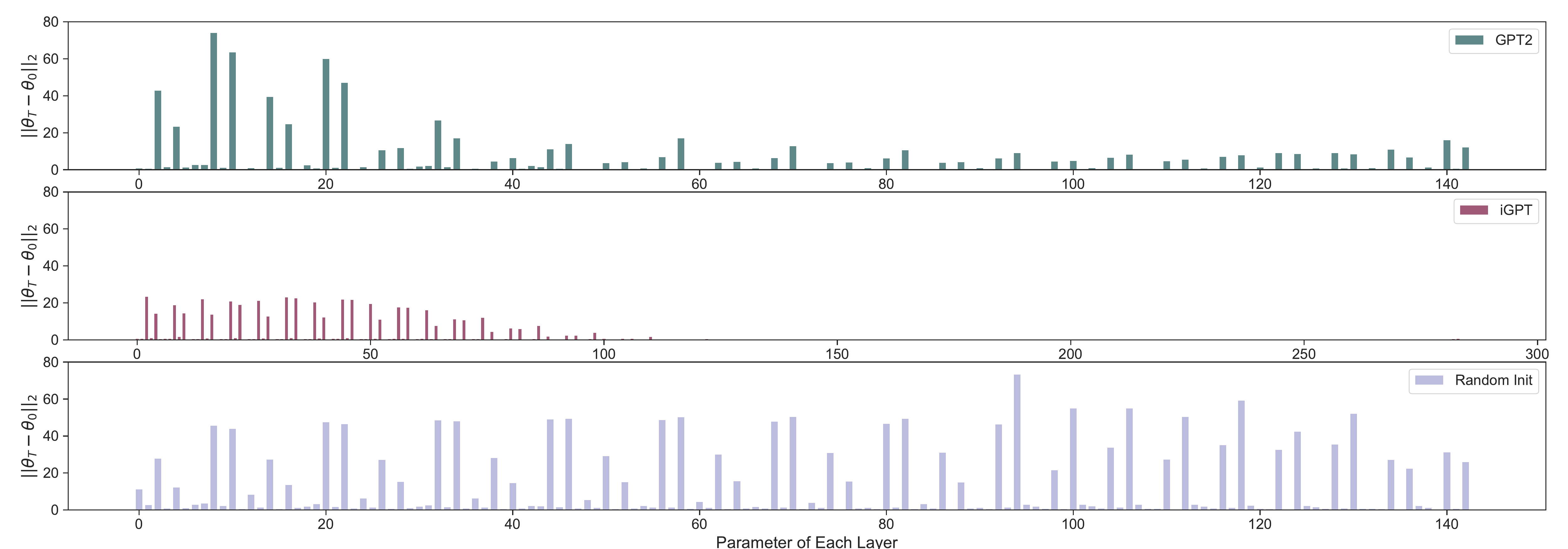}
    \caption{L2 distance of each parameter between pre post-fine-tuning (HalfCheetah).}
\end{figure}

\begin{figure}[H]
    \centering
        \includegraphics[width=\linewidth]{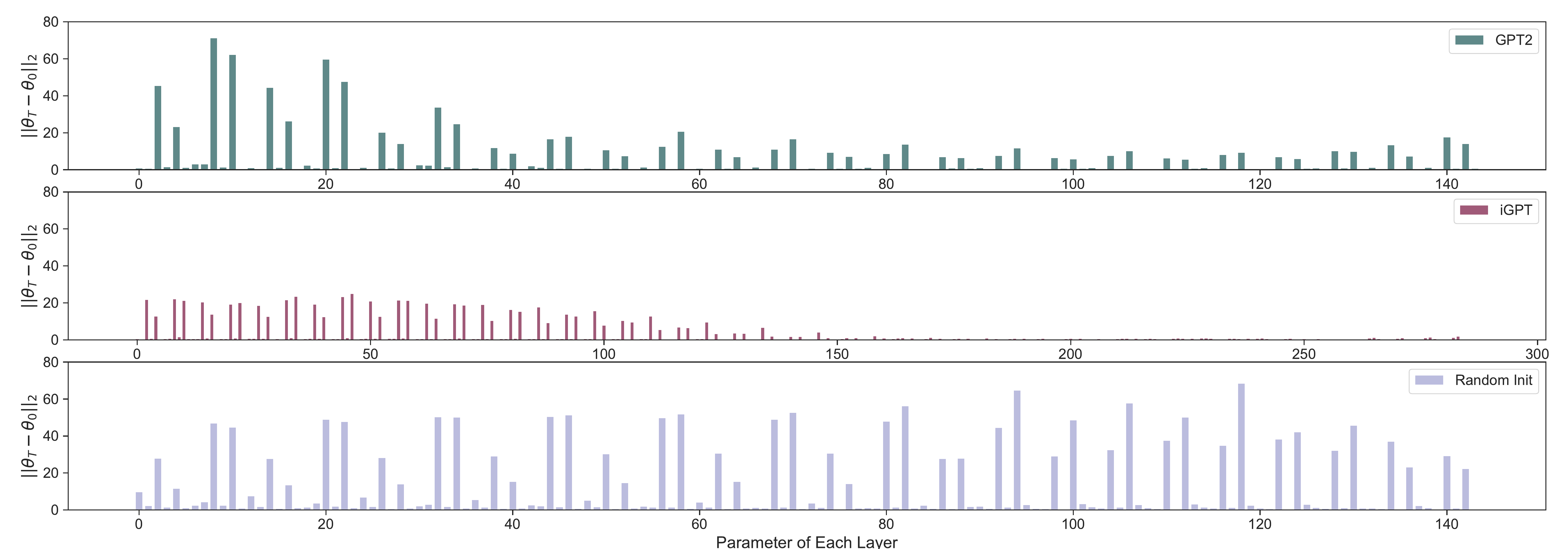}
    \caption{L2 distance of each parameter between pre post-fine-tuning (Walker2D).}
\end{figure}

\begin{figure}[H]
    \centering
        \includegraphics[width=\linewidth]{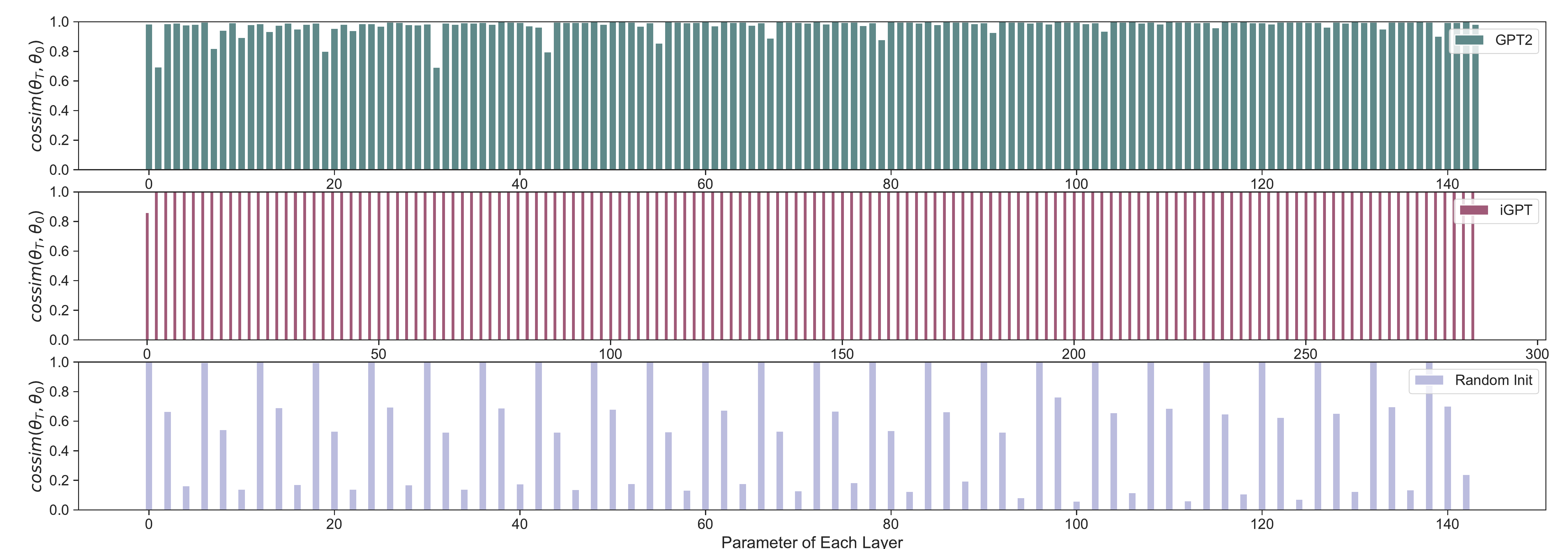}
    \caption{Cosine similarity of each parameter between pre post-fine-tuning (HalfCheetah).}
\end{figure}

\begin{figure}[H]
    \centering
        \includegraphics[width=\linewidth]{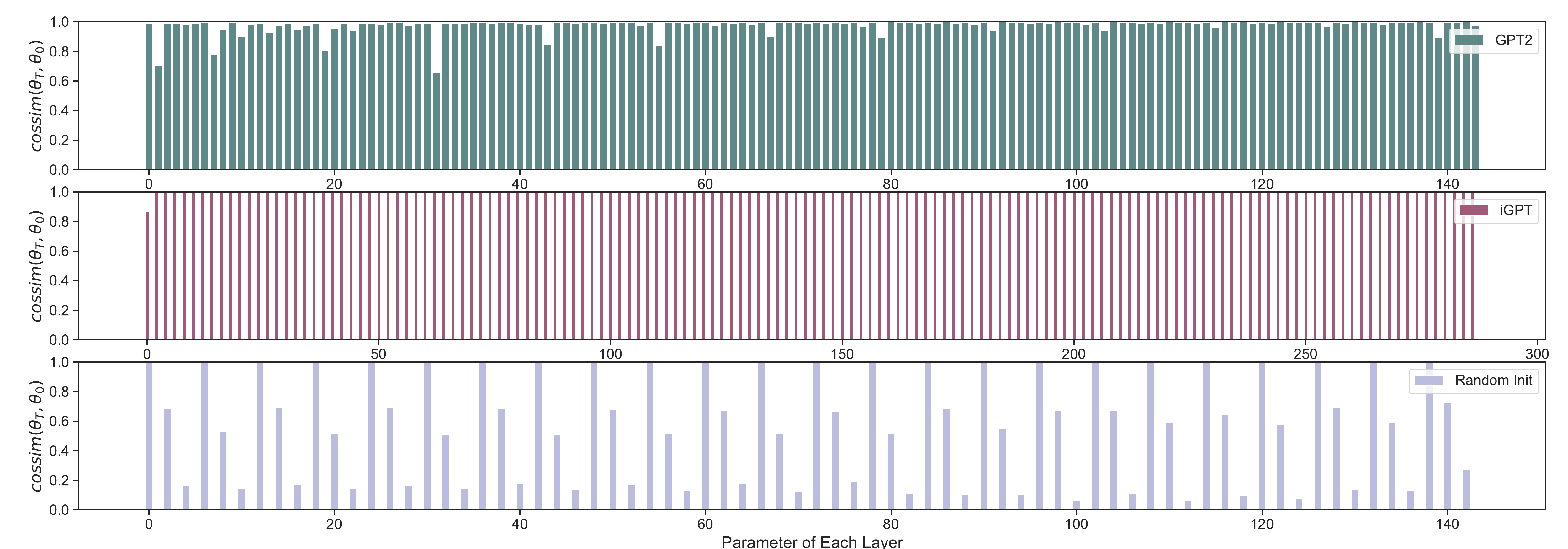}
    \caption{Cosine similarity of each parameter between pre post-fine-tuning (Walker2D).}
\end{figure}

\subsubsection{Parameter Similarity (Seed = 42)}

\begin{figure}[H]
    \centering
        \includegraphics[width=\linewidth]{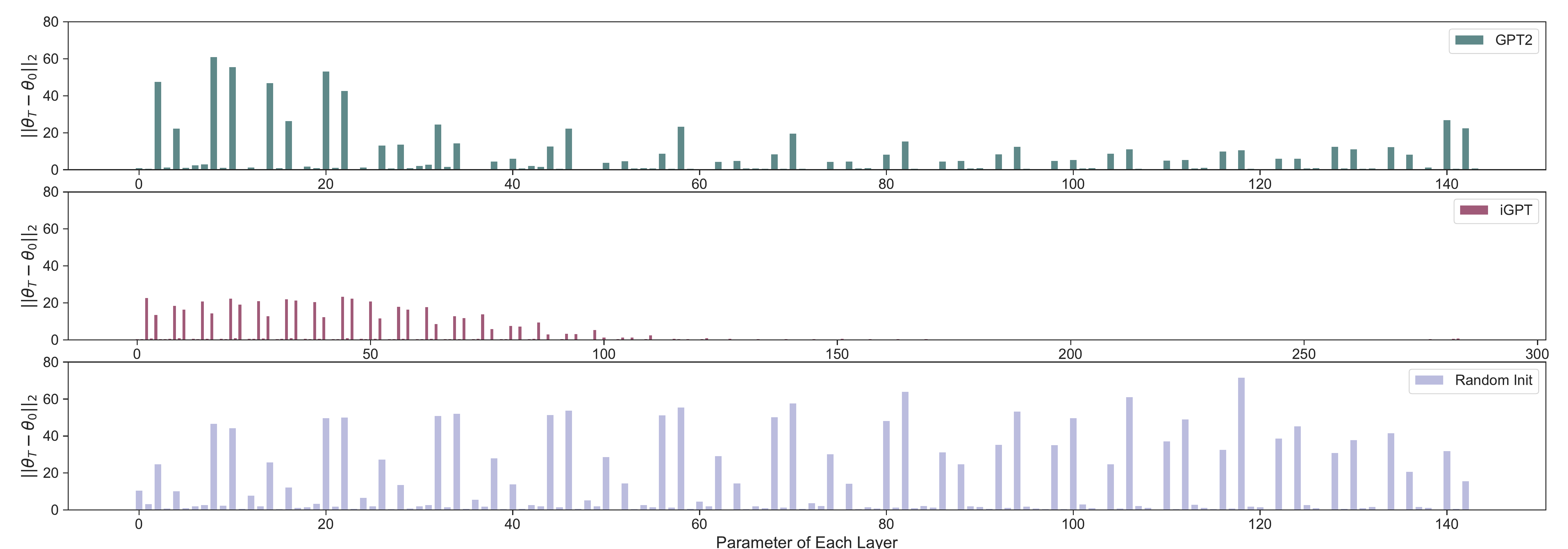}
    \caption{L2 distance of each parameter between pre post-fine-tuning (Hopper, Seed = 42).}
\end{figure}

\begin{figure}[H]
    \centering
        \includegraphics[width=\linewidth]{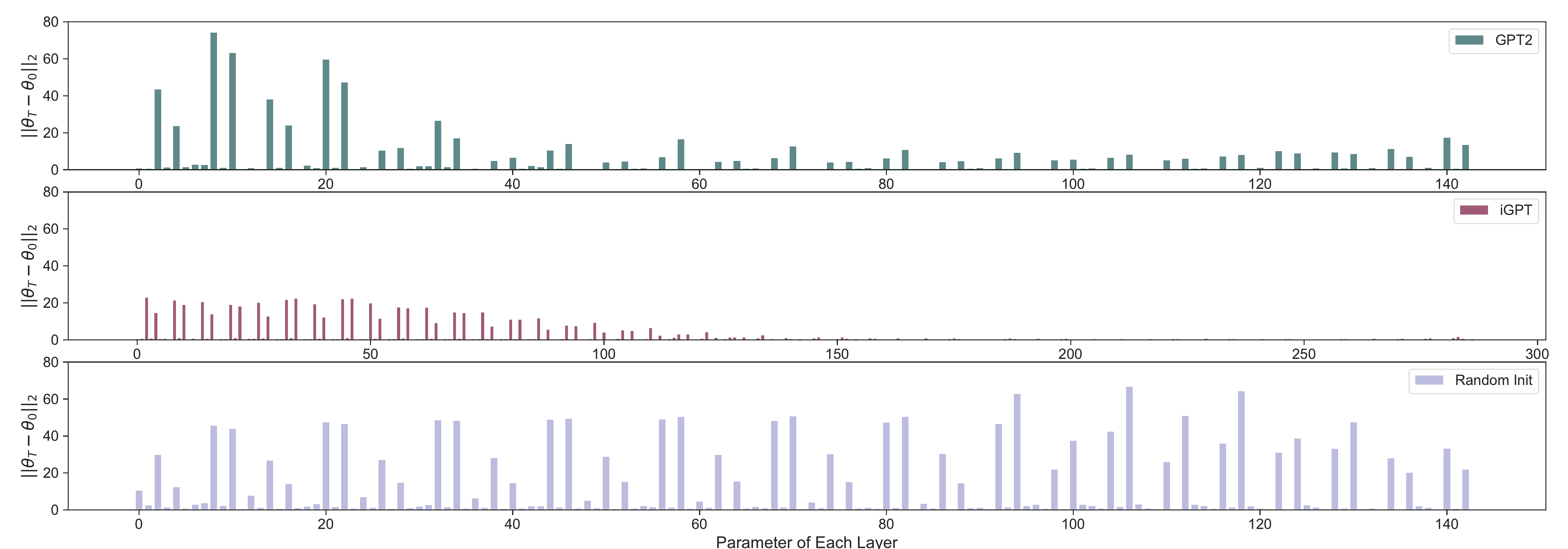}
    \caption{L2 distance of each parameter between pre post-fine-tuning (HalfCheetah, Seed = 42).}
\end{figure}

\begin{figure}[H]
    \centering
        \includegraphics[width=\linewidth]{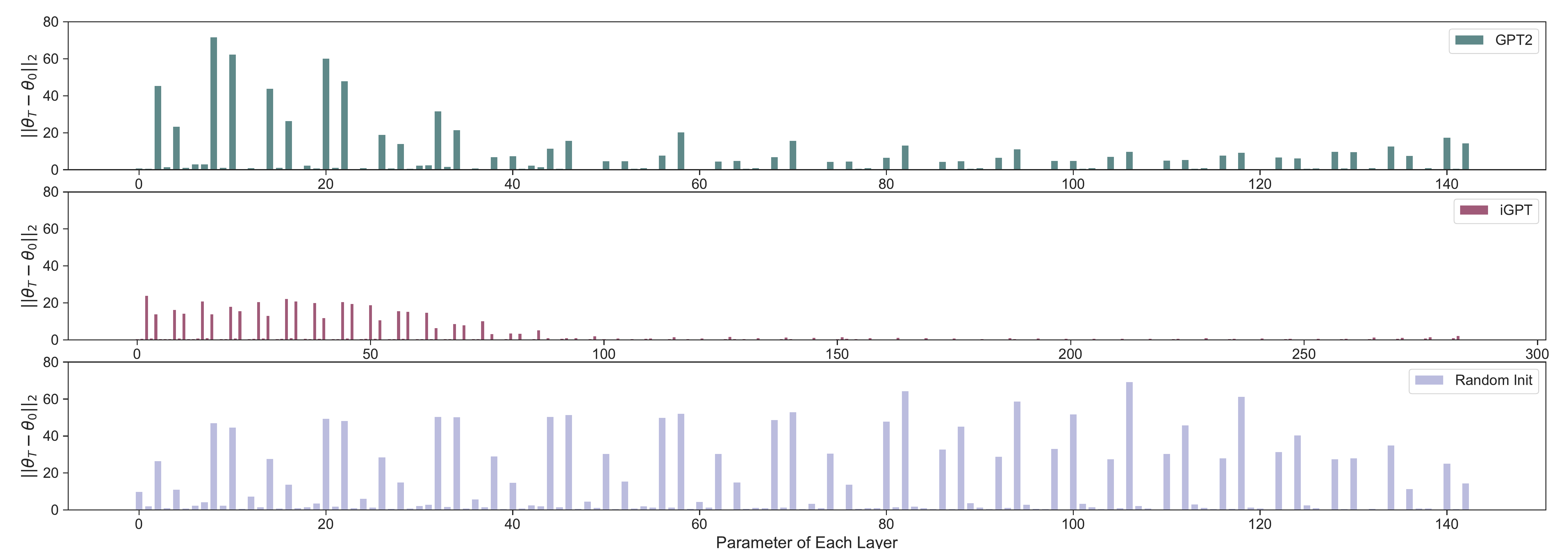}
    \caption{L2 distance of each parameter between pre post-fine-tuning (Walker2D, Seed = 42).}
\end{figure}

\begin{figure}[H]
    \centering
        \includegraphics[width=\linewidth]{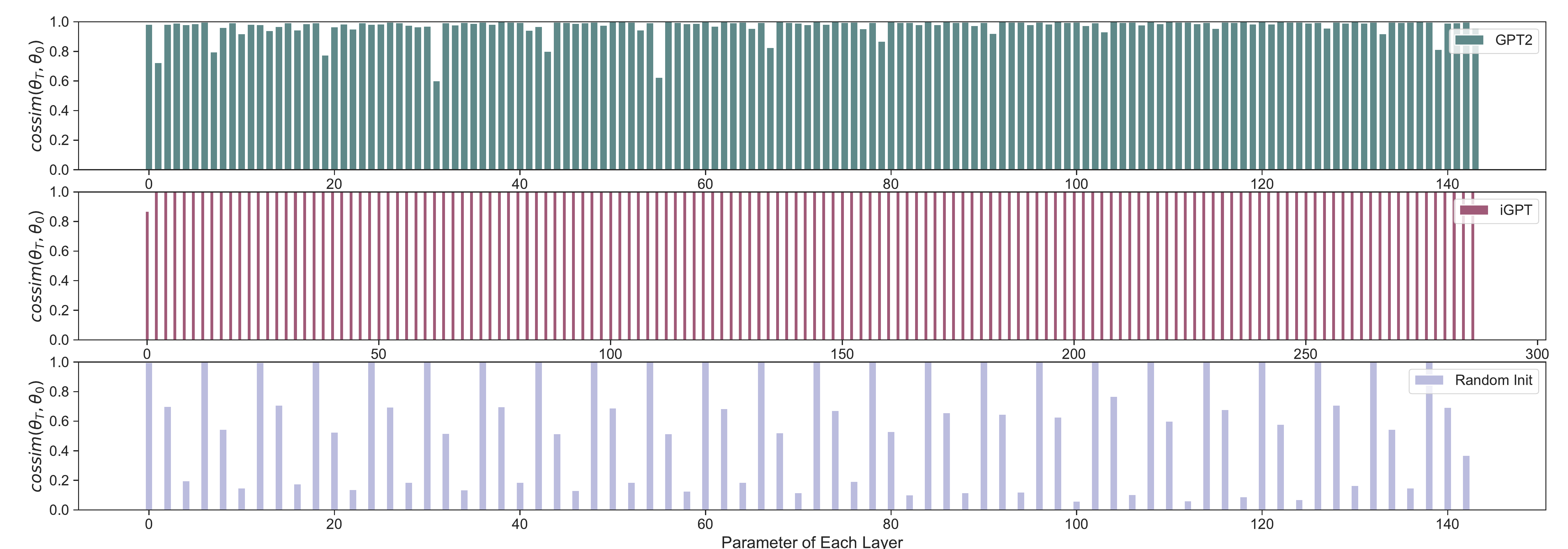}
    \caption{Cosine similarity of each parameter between pre post-fine-tuning (Hopper, Seed = 42).}
\end{figure}

\begin{figure}[H]
    \centering
        \includegraphics[width=\linewidth]{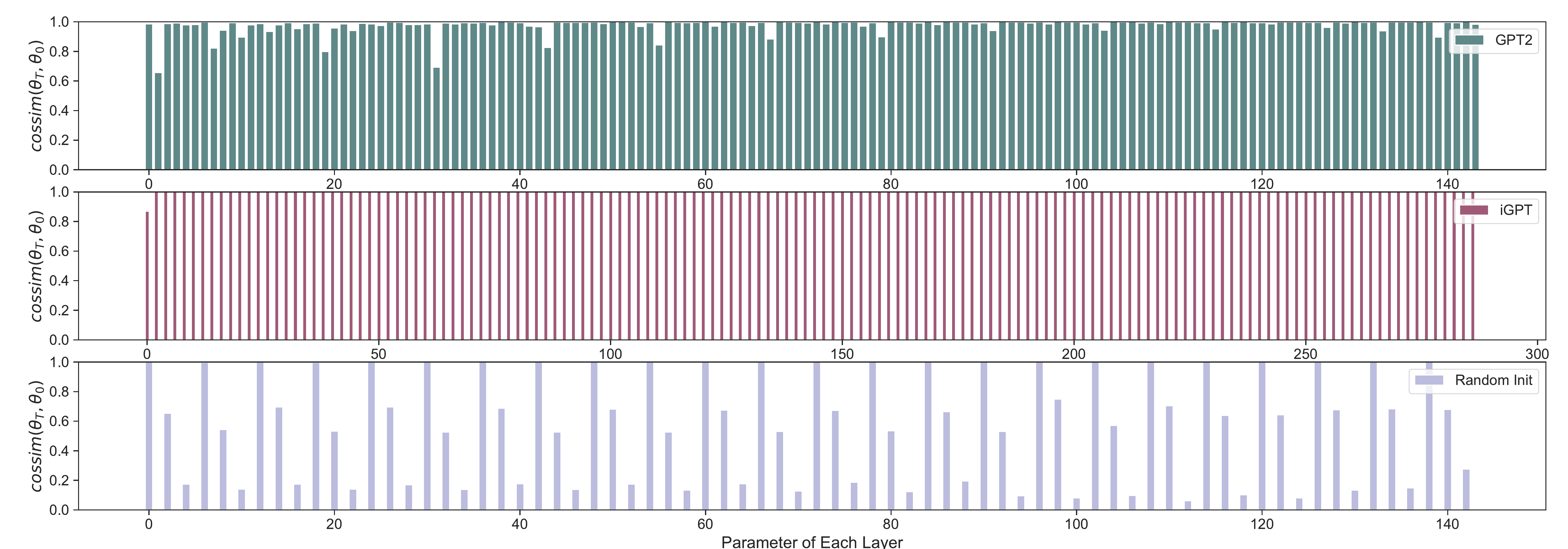}
    \caption{Cosine similarity of each parameter between pre post-fine-tuning (HalfCheetah, Seed = 42).}
\end{figure}

\begin{figure}[H]
    \centering
        \includegraphics[width=\linewidth]{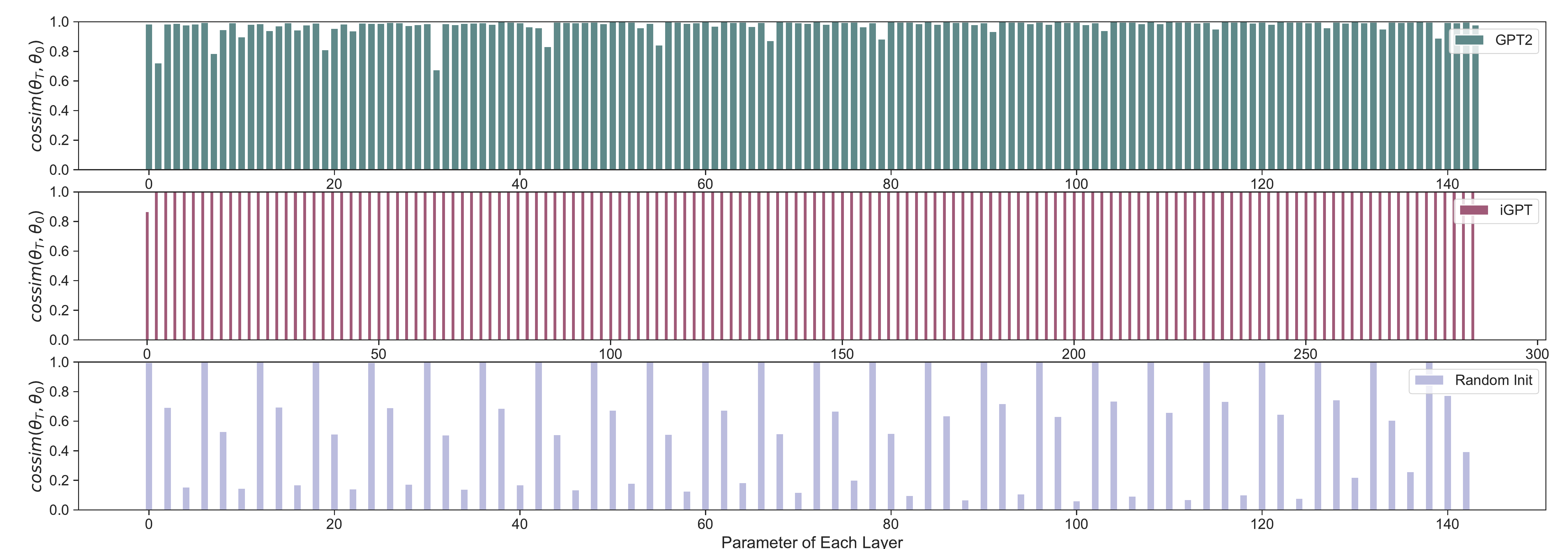}
    \caption{Cosine similarity of each parameter between pre post-fine-tuning (Walker2D, Seed = 42).}
\end{figure}

\subsection{Gradient Analysis}
\label{appendix:results-for-other-conditions-gradient-analysis}

\subsubsection{Gradient Analysis (Other Environments)}
\begin{figure}[H]
    \centering
    \begin{minipage}[b]{0.48\linewidth}
        \includegraphics[width=\linewidth]{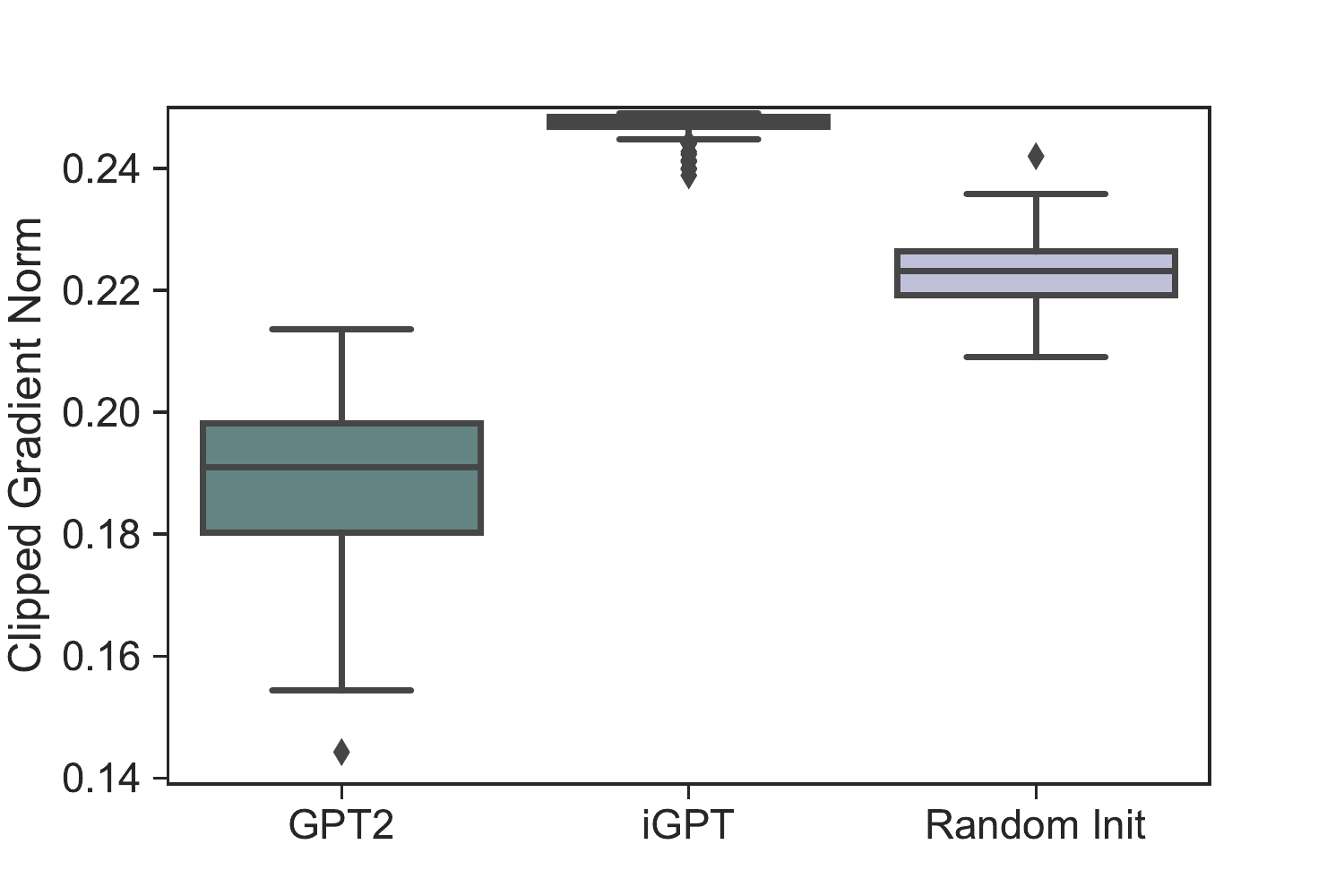}
        \subcaption{HalfCheetah}
    \end{minipage}
    \begin{minipage}[b]{0.48\linewidth}
        \includegraphics[width=\linewidth]{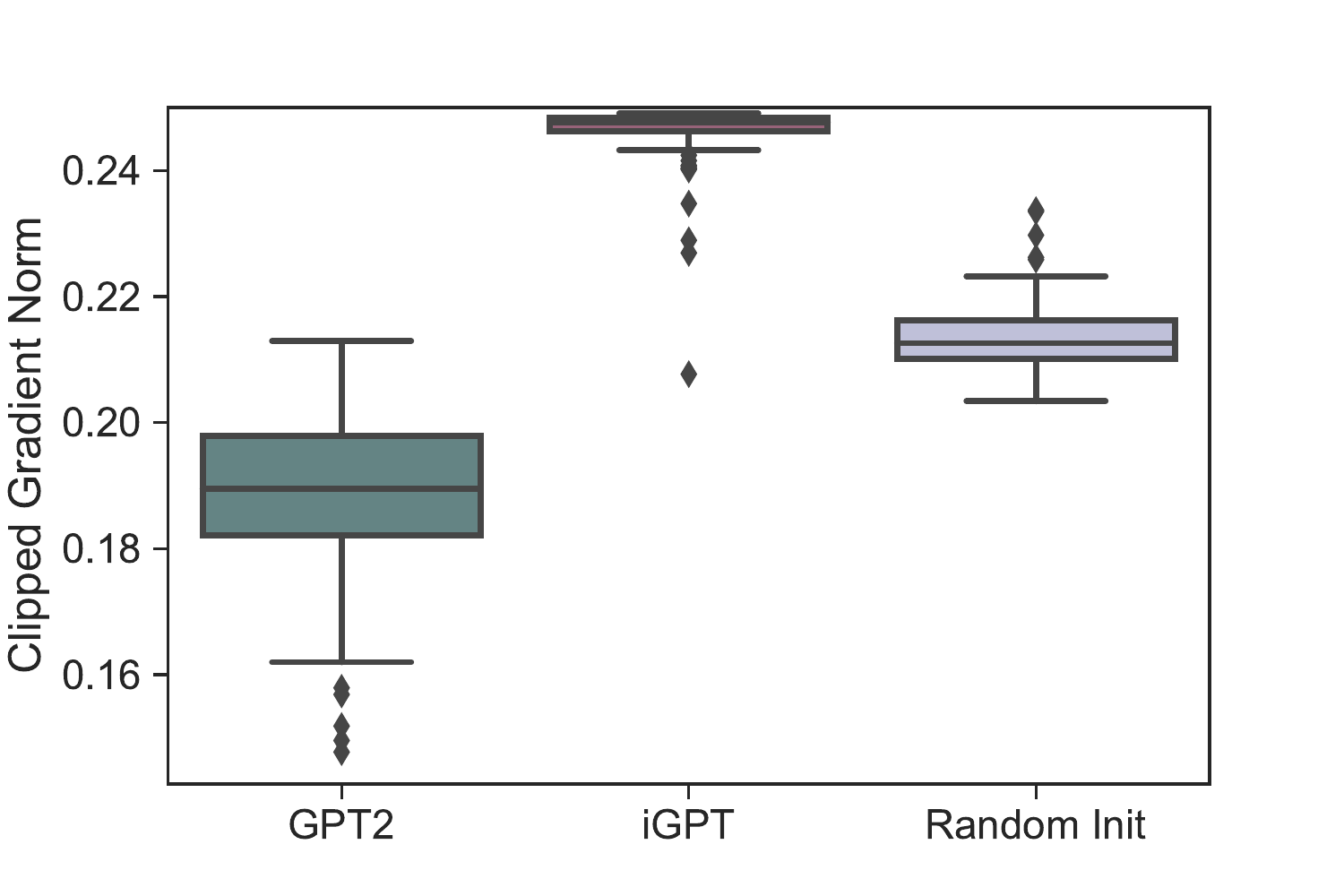}
        \subcaption{Walker2D}
    \end{minipage}
    \caption{Gradient norm.}
\end{figure}

\begin{figure}[H]
    \centering
    \begin{minipage}[b]{0.48\linewidth}
        \includegraphics[width=\linewidth]{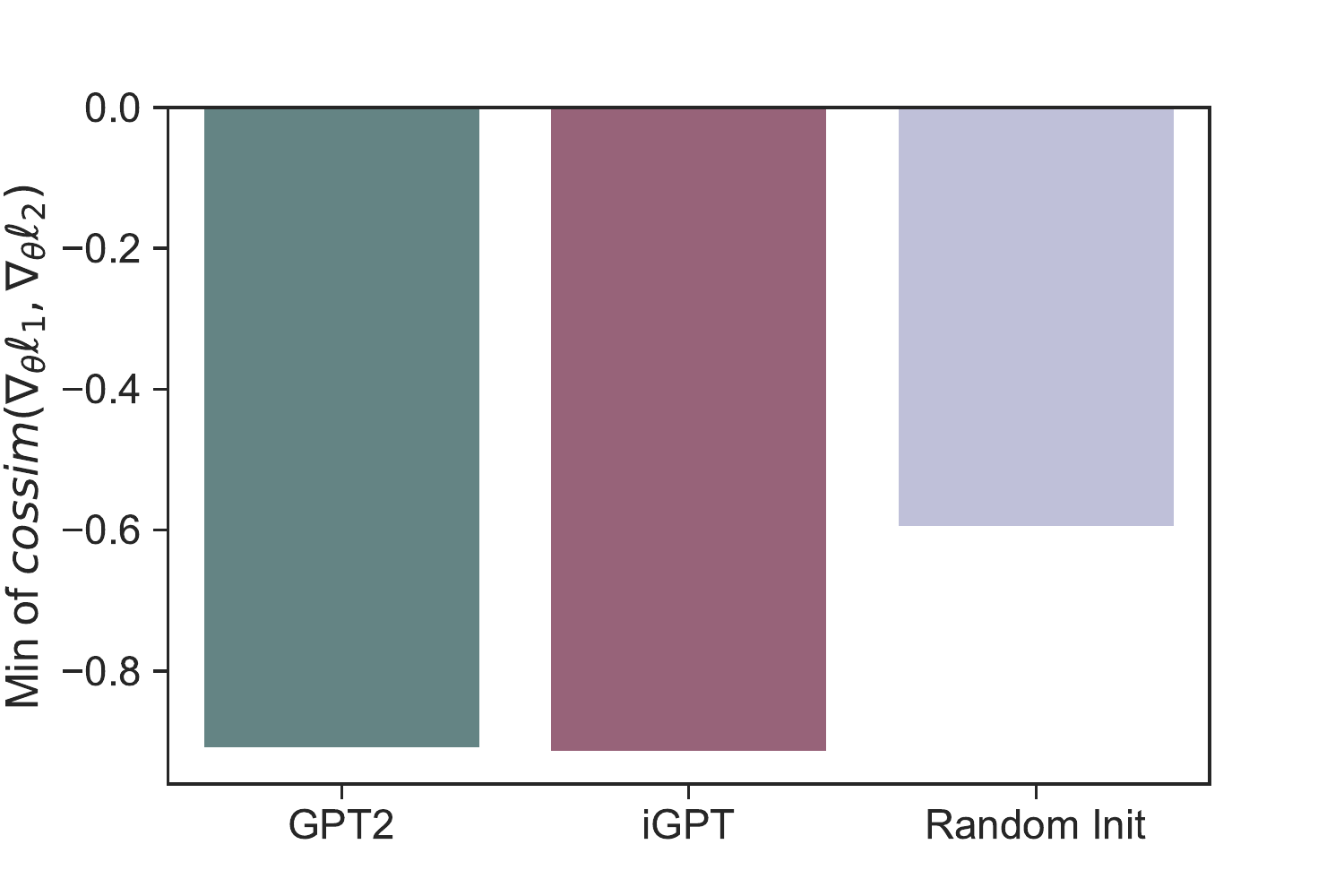}
        \subcaption{HalfCheetah}
    \end{minipage}
    \begin{minipage}[b]{0.48\linewidth}
        \includegraphics[width=\linewidth]{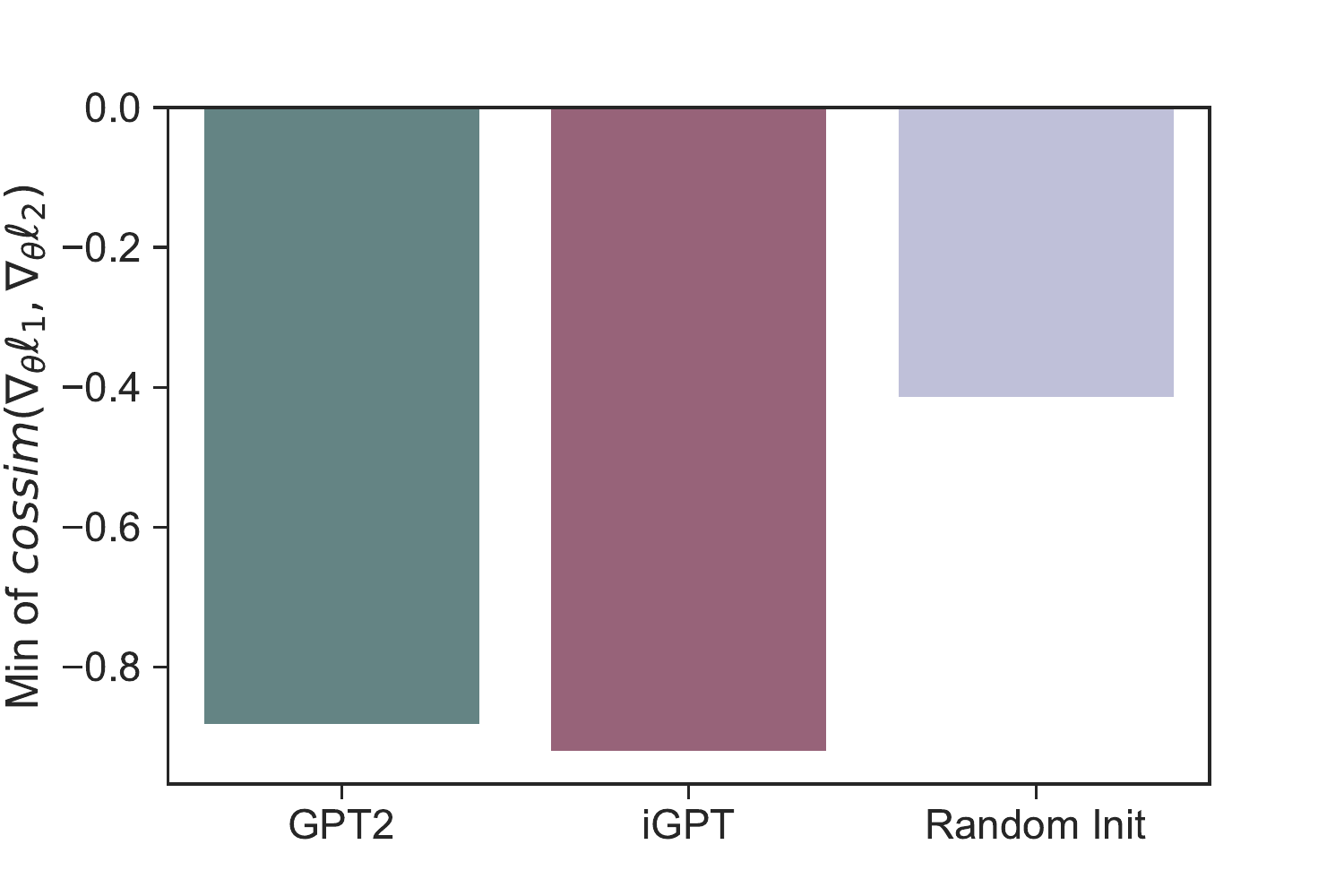}
        \subcaption{Walker2D}
    \end{minipage}
    \caption{Minimum gradient cosine similarity.}
\end{figure}

\begin{figure}[H]
    \centering
    \begin{minipage}[b]{0.48\linewidth}
        \includegraphics[width=\linewidth]{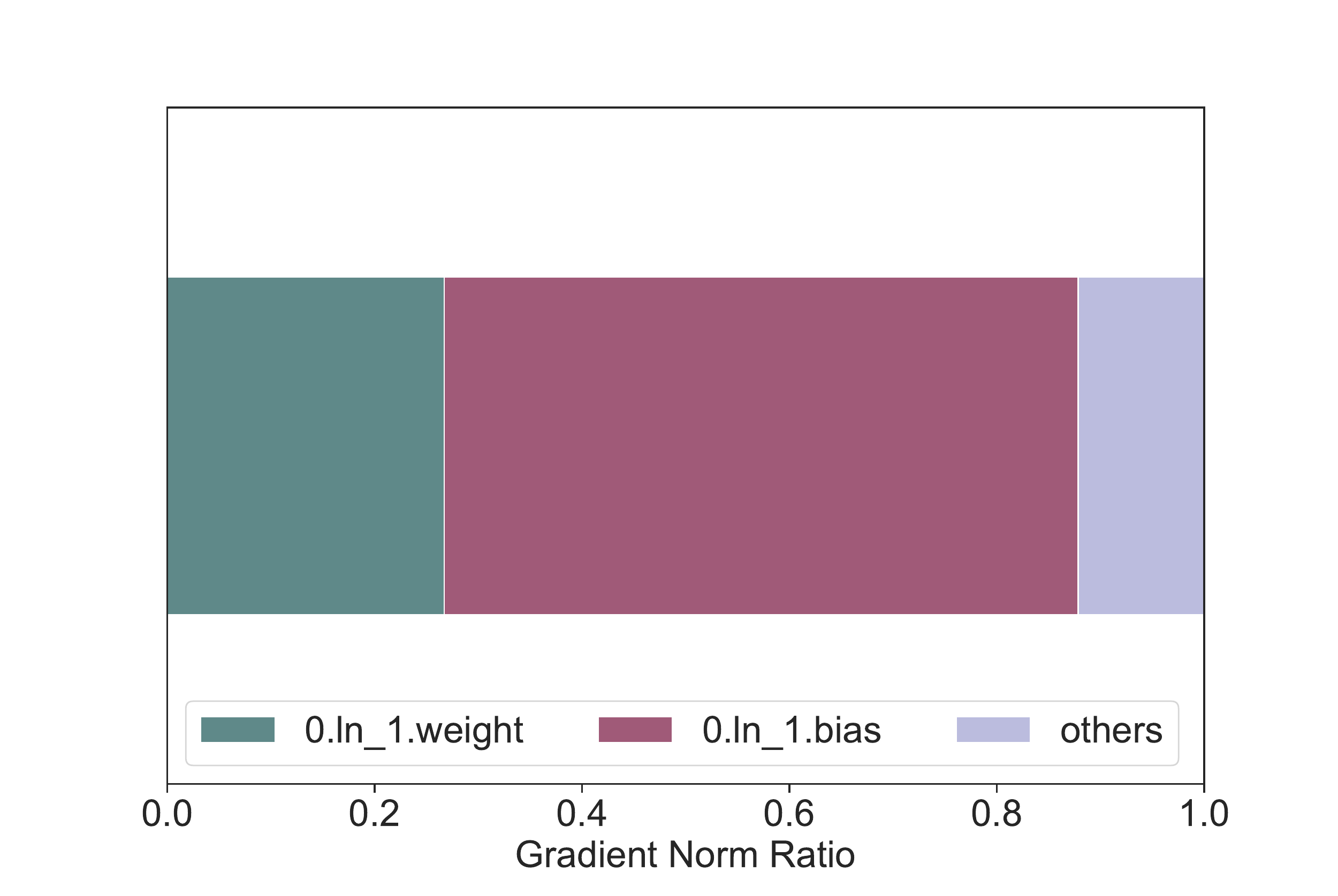}
        \subcaption{HalfCheetah}
    \end{minipage}
    \begin{minipage}[b]{0.48\linewidth}
        \includegraphics[width=\linewidth]{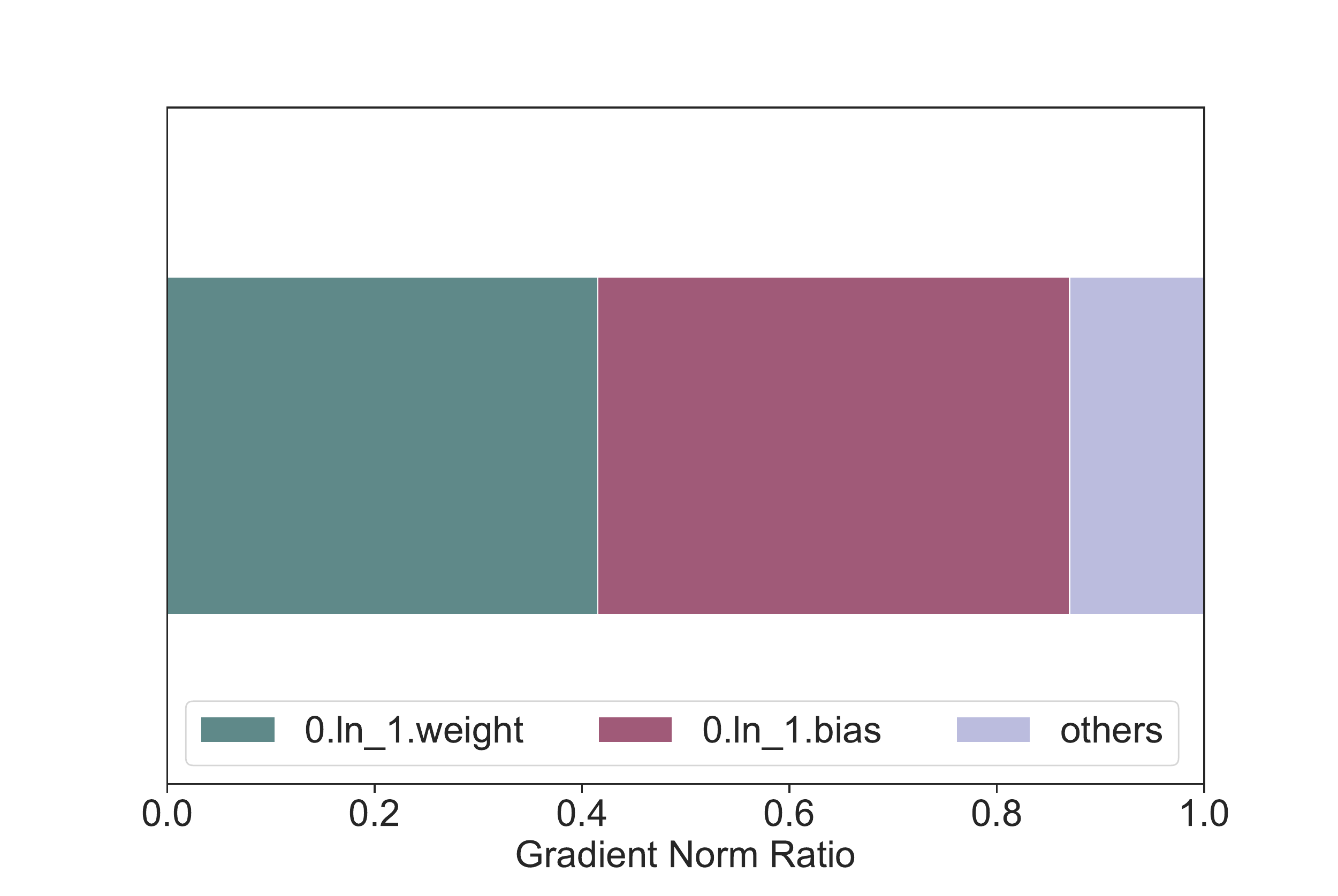}
        \subcaption{Walker2D}
    \end{minipage}
    \caption{iGPT's gradient norm ratio.}
\end{figure}

\begin{figure}[H]
    \centering
        \includegraphics[width=\linewidth]{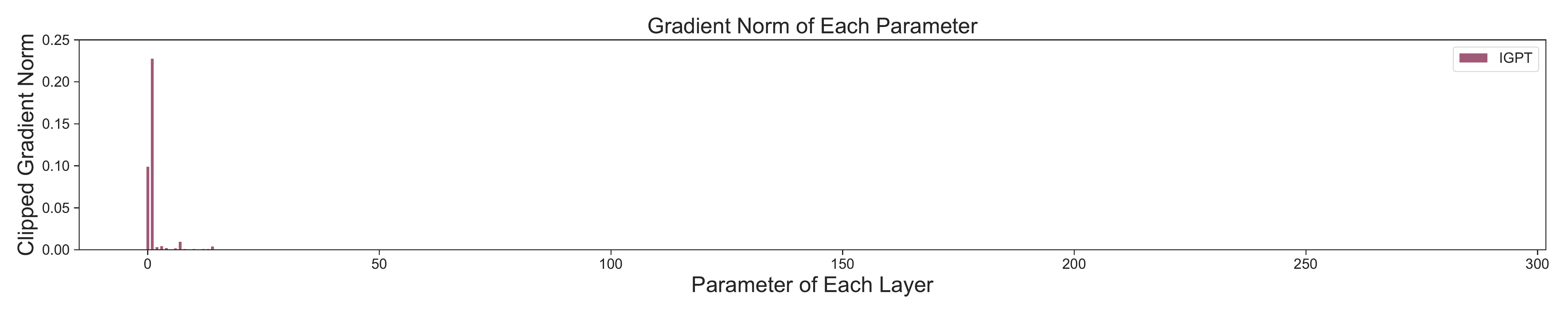}
    \caption{Gradient norm of iGPT's each parameter at epoch 1. (HalfCheetah)}
\end{figure}

\begin{figure}[H]
    \centering
        \includegraphics[width=\linewidth]{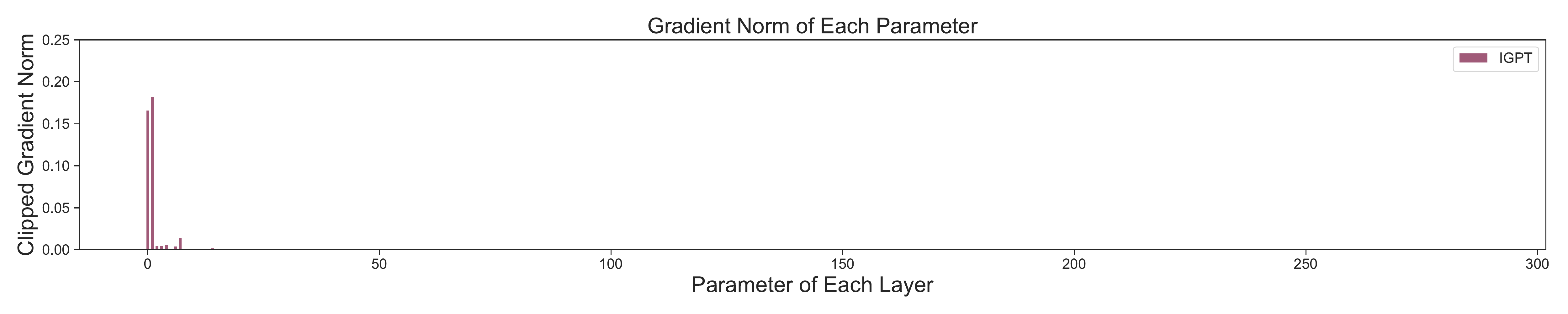}
    \caption{Gradient norm of iGPT's each parameter at epoch 1. (Walker2D)}
\end{figure}

\subsubsection{Gradient Analysis (Seed = 42)}
\begin{figure}[H]
    \centering
    \begin{minipage}[b]{0.32\linewidth}
        \includegraphics[width=\linewidth]{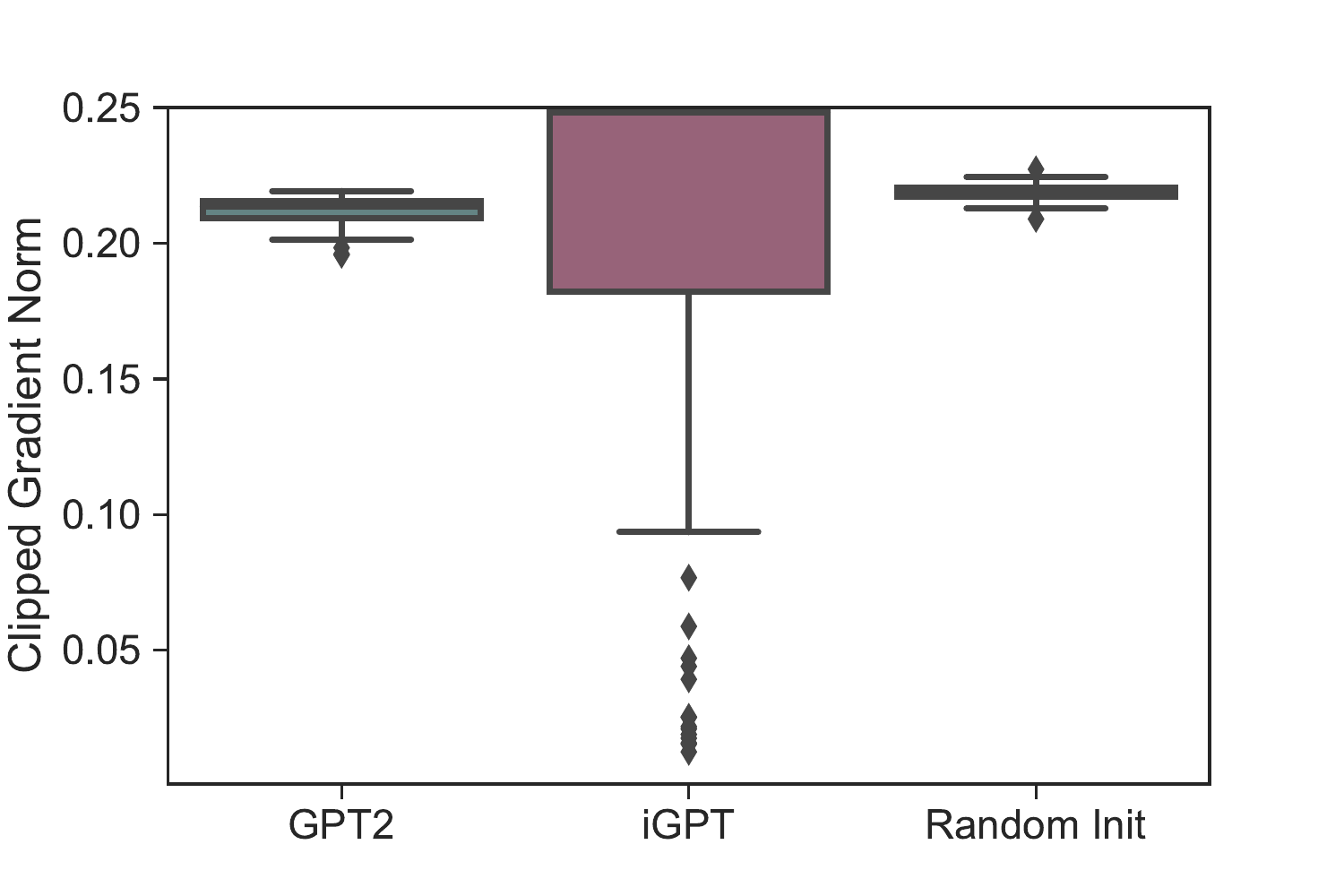}
        \subcaption{Hopper}
    \end{minipage}
    \begin{minipage}[b]{0.32\linewidth}
        \includegraphics[width=\linewidth]{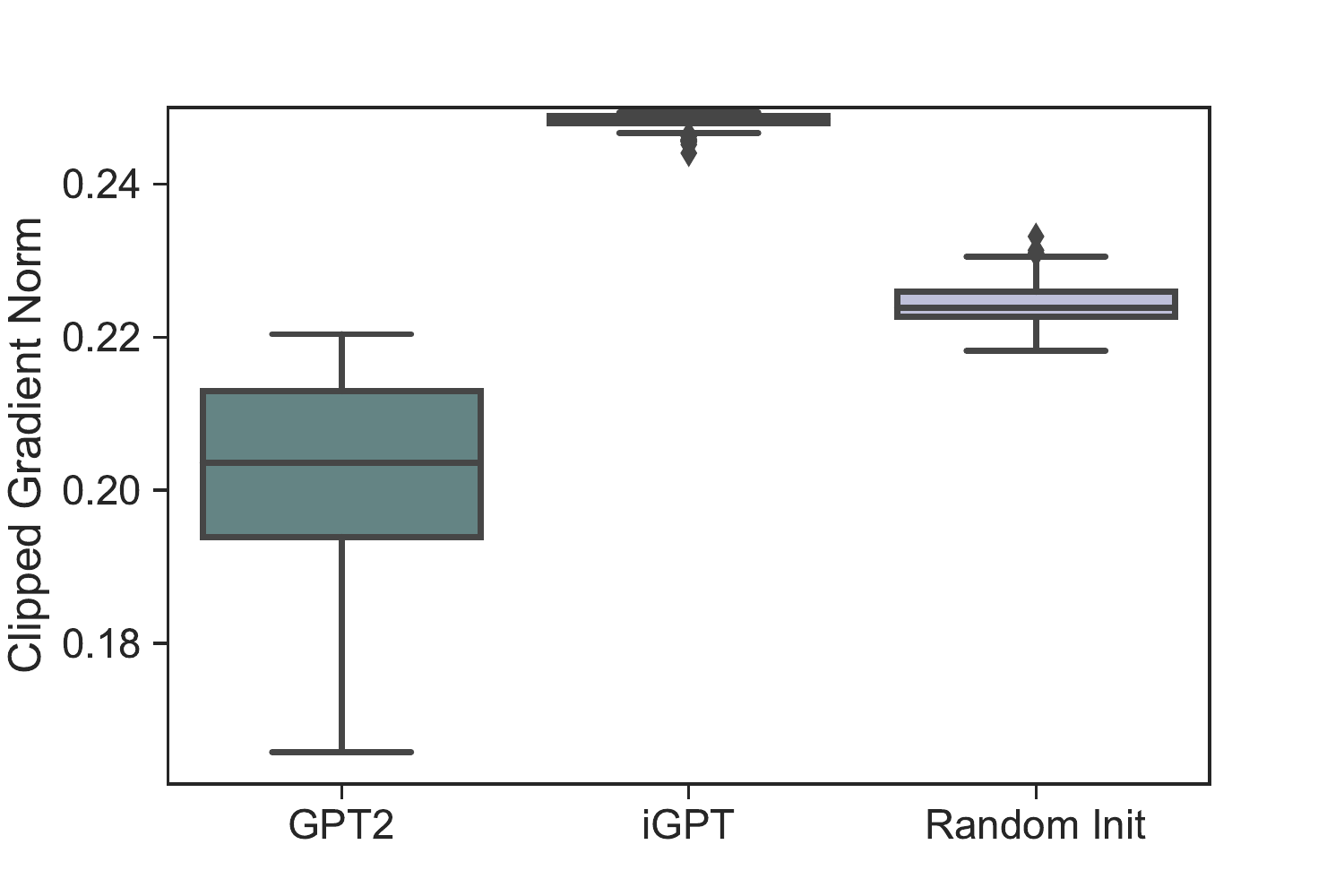}
        \subcaption{HalfCheetah}
    \end{minipage}
    \begin{minipage}[b]{0.32\linewidth}
        \includegraphics[width=\linewidth]{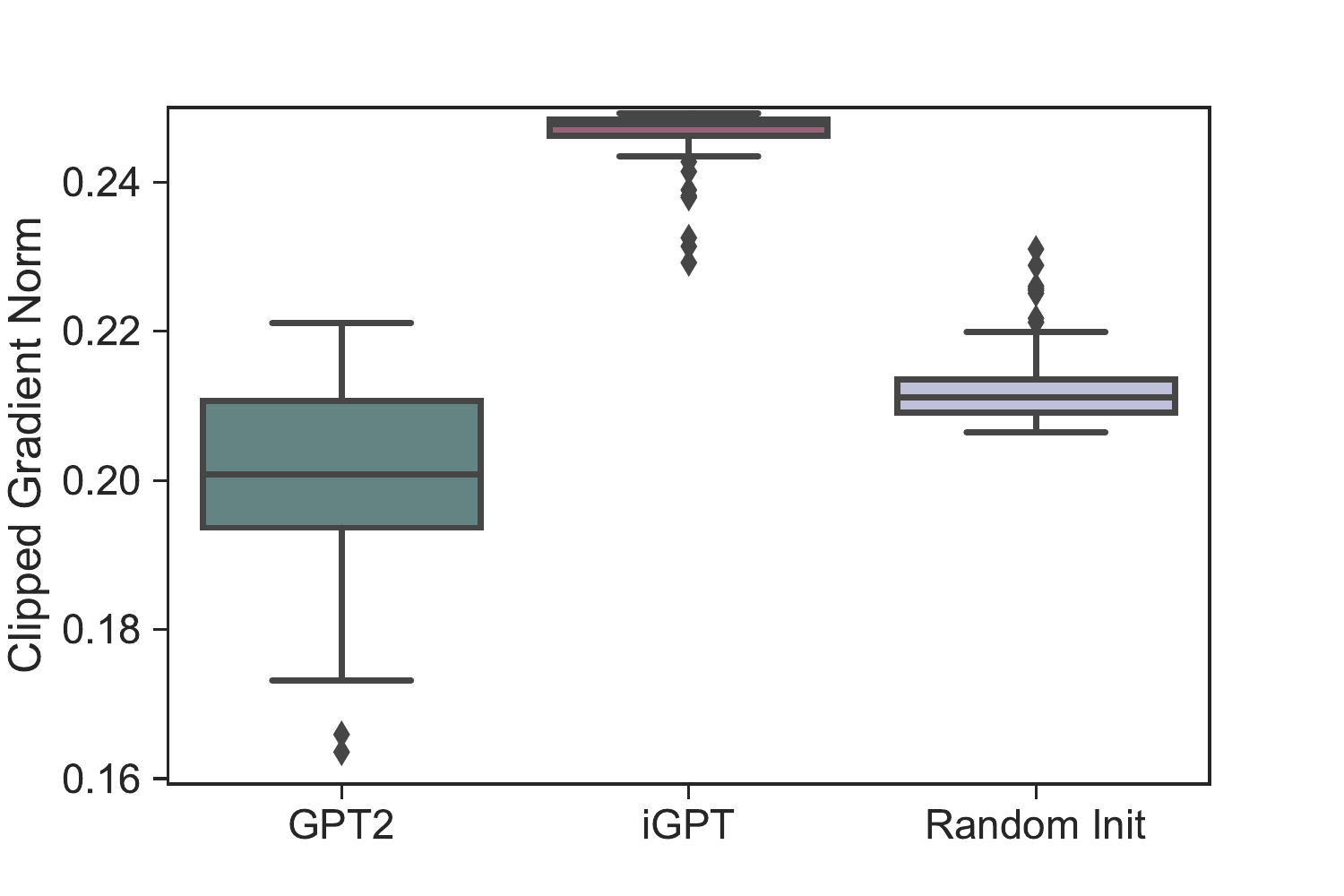}
        \subcaption{Walker2D}
    \end{minipage}
    \caption{Gradient norm (Seed = 42).}
\end{figure}

\begin{figure}[H]
    \centering
    \begin{minipage}[b]{0.32\linewidth}
        \includegraphics[width=\linewidth]{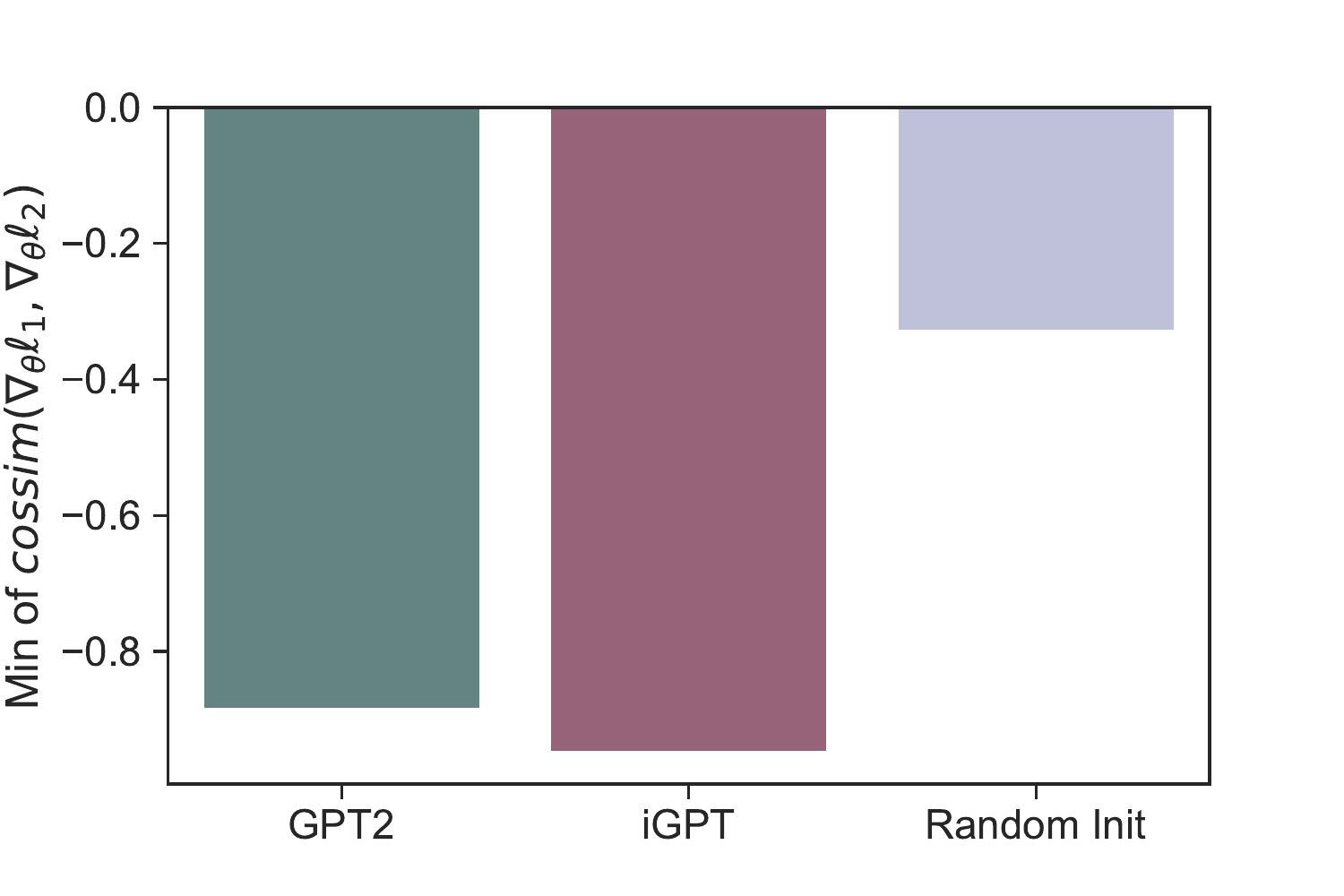}
        \subcaption{Hopper (Seed = 42)}
    \end{minipage}
    \begin{minipage}[b]{0.32\linewidth}
        \includegraphics[width=\linewidth]{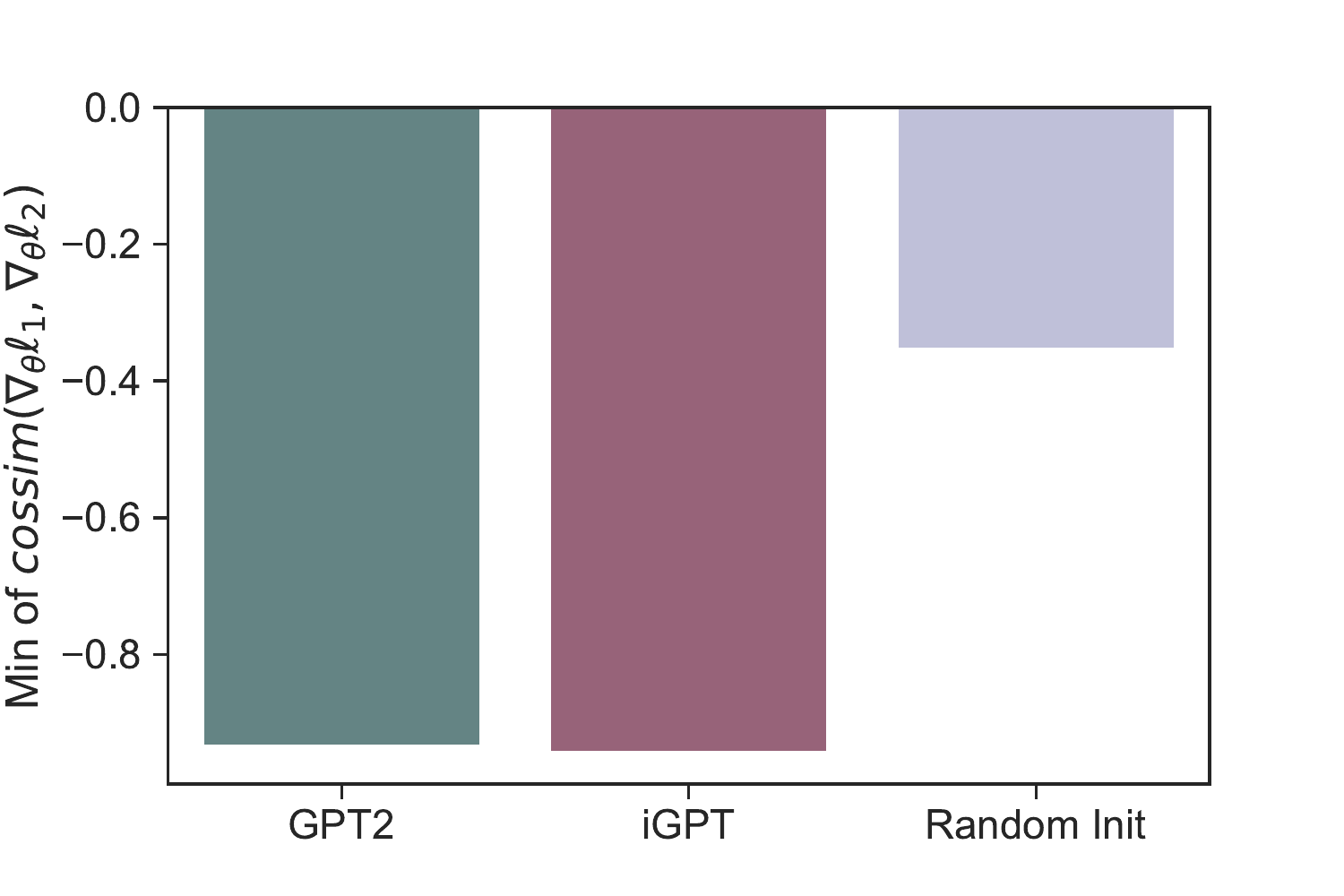}
        \subcaption{HalfCheetah (Seed = 42)}
    \end{minipage}
    \begin{minipage}[b]{0.32\linewidth}
        \includegraphics[width=\linewidth]{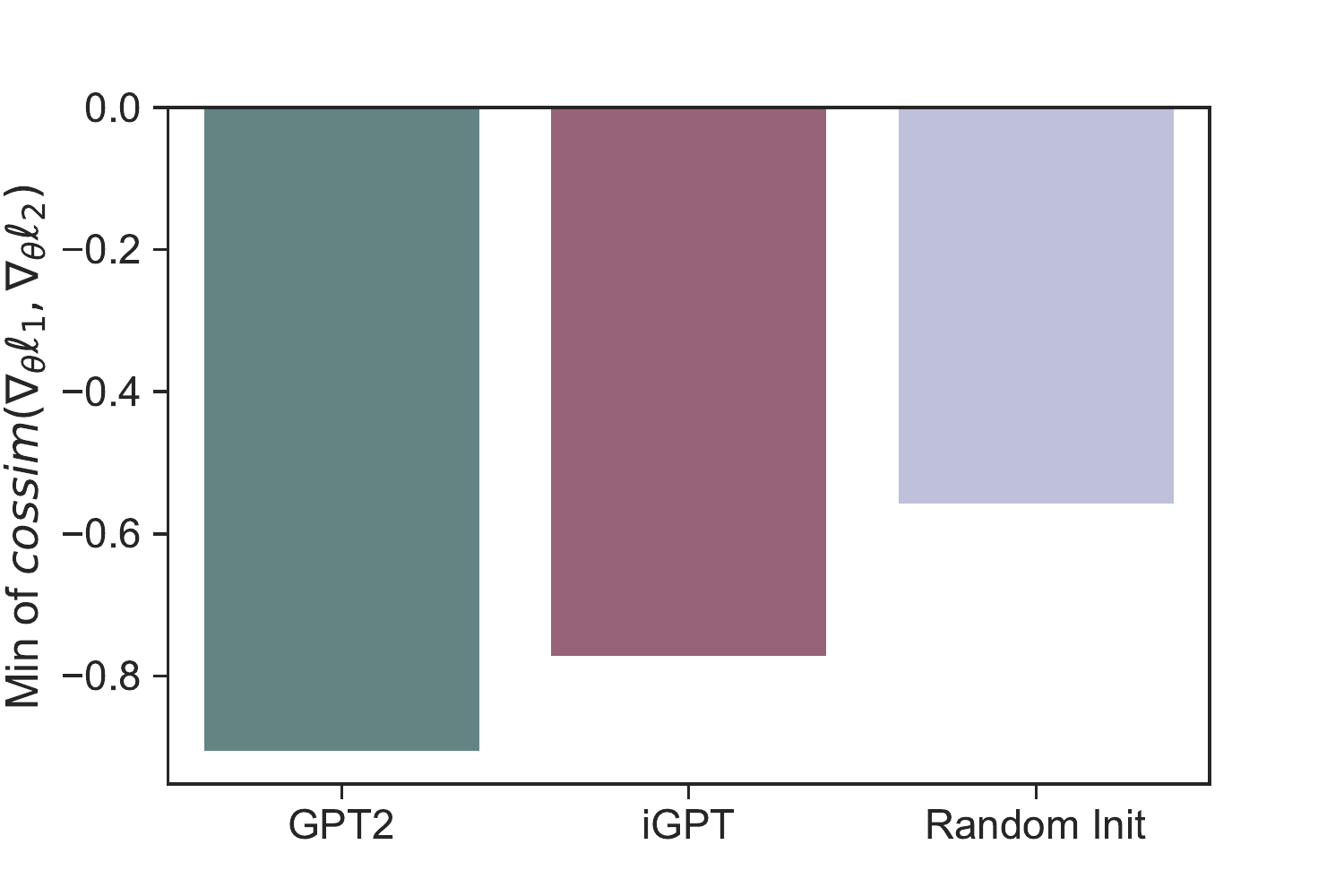}
        \subcaption{Walker2D (Seed = 42)}
    \end{minipage}
    \caption{Minimum gradient cosine similarity (seed = 42).}
\end{figure}

\begin{figure}[H]
    \centering
    \begin{minipage}[b]{0.32\linewidth}
        \includegraphics[width=\linewidth]{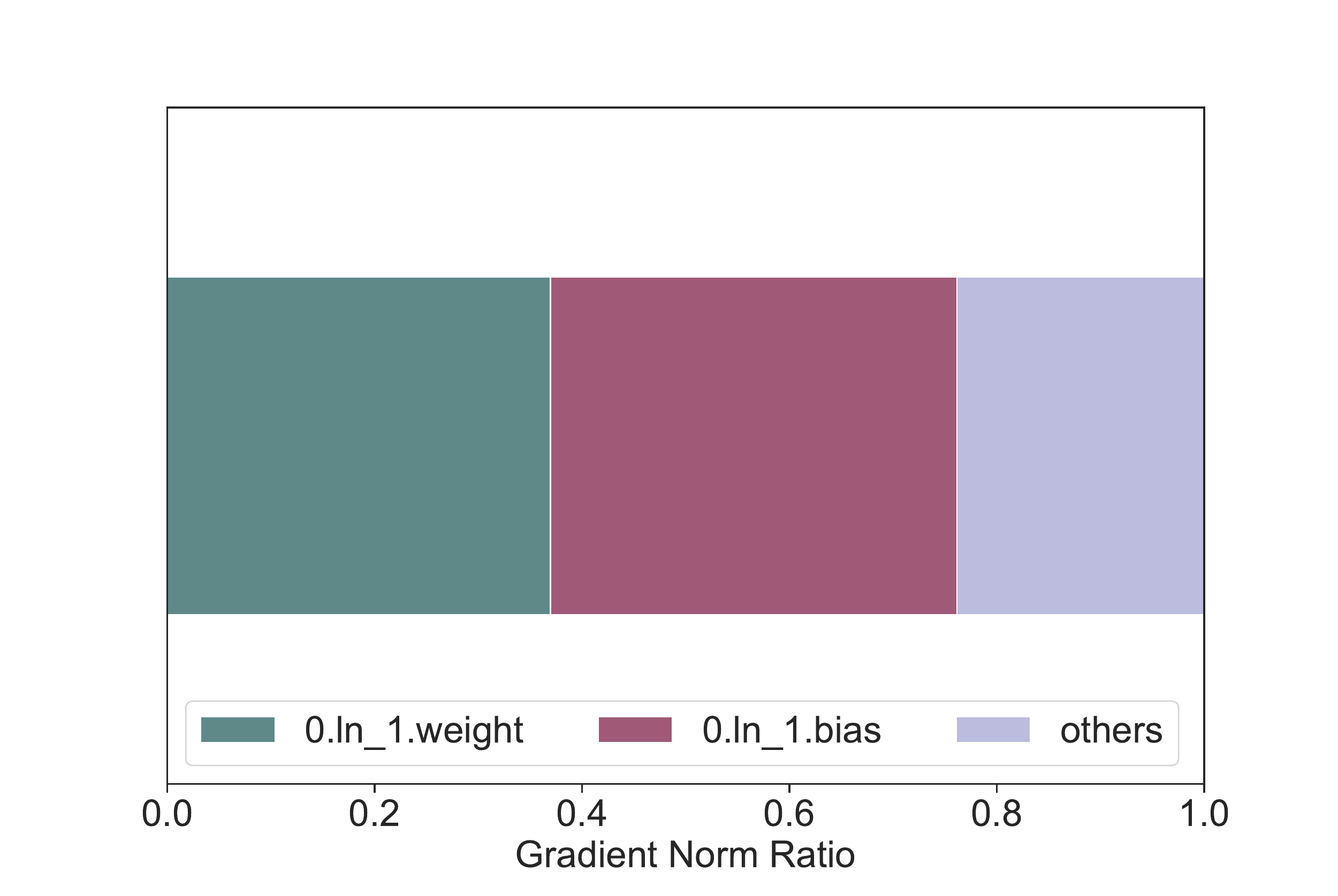}
        \subcaption{Hopper}
    \end{minipage}
    \begin{minipage}[b]{0.32\linewidth}
        \includegraphics[width=\linewidth]{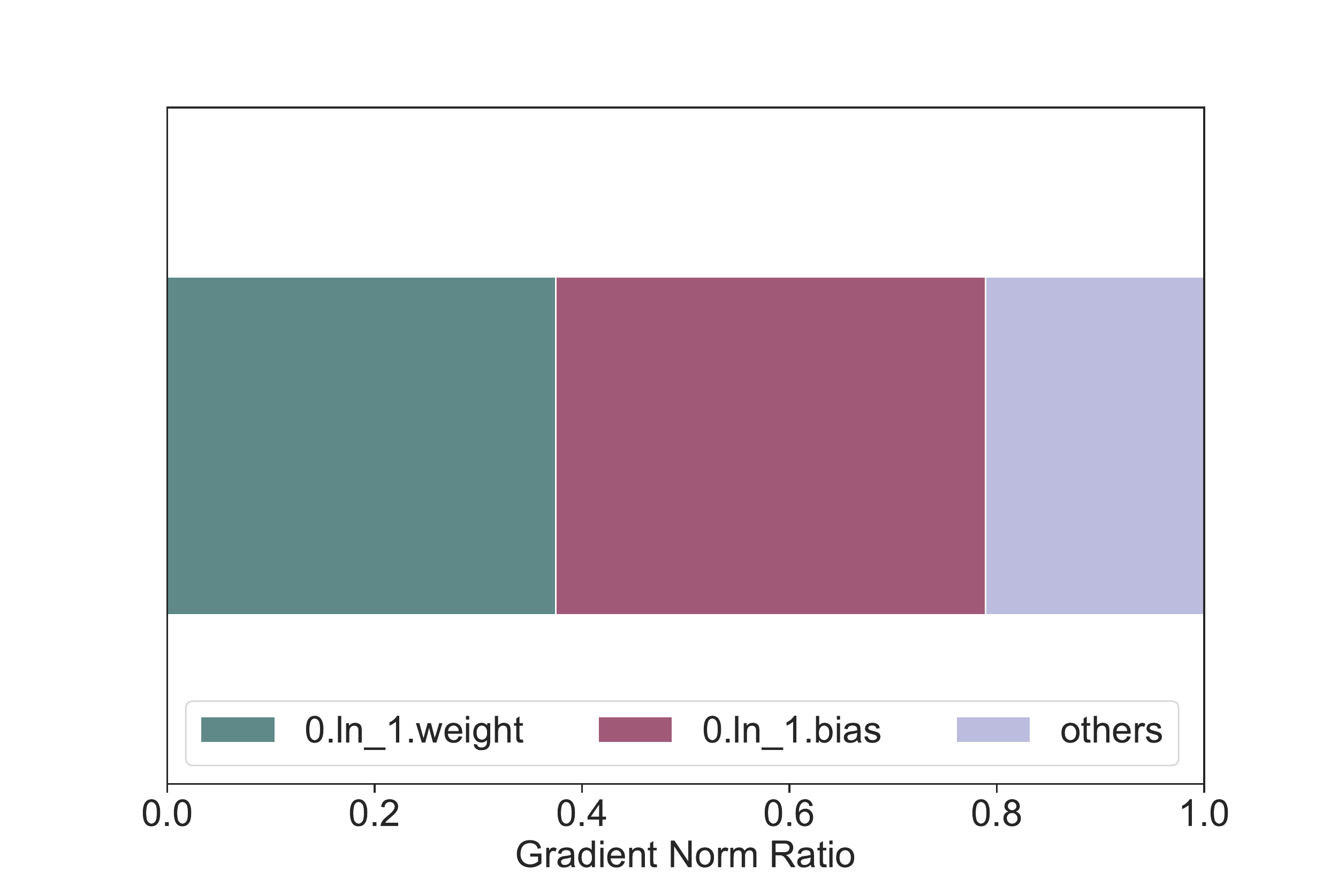}
        \subcaption{HalfCheetah}
    \end{minipage}
    \begin{minipage}[b]{0.32\linewidth}
        \includegraphics[width=\linewidth]{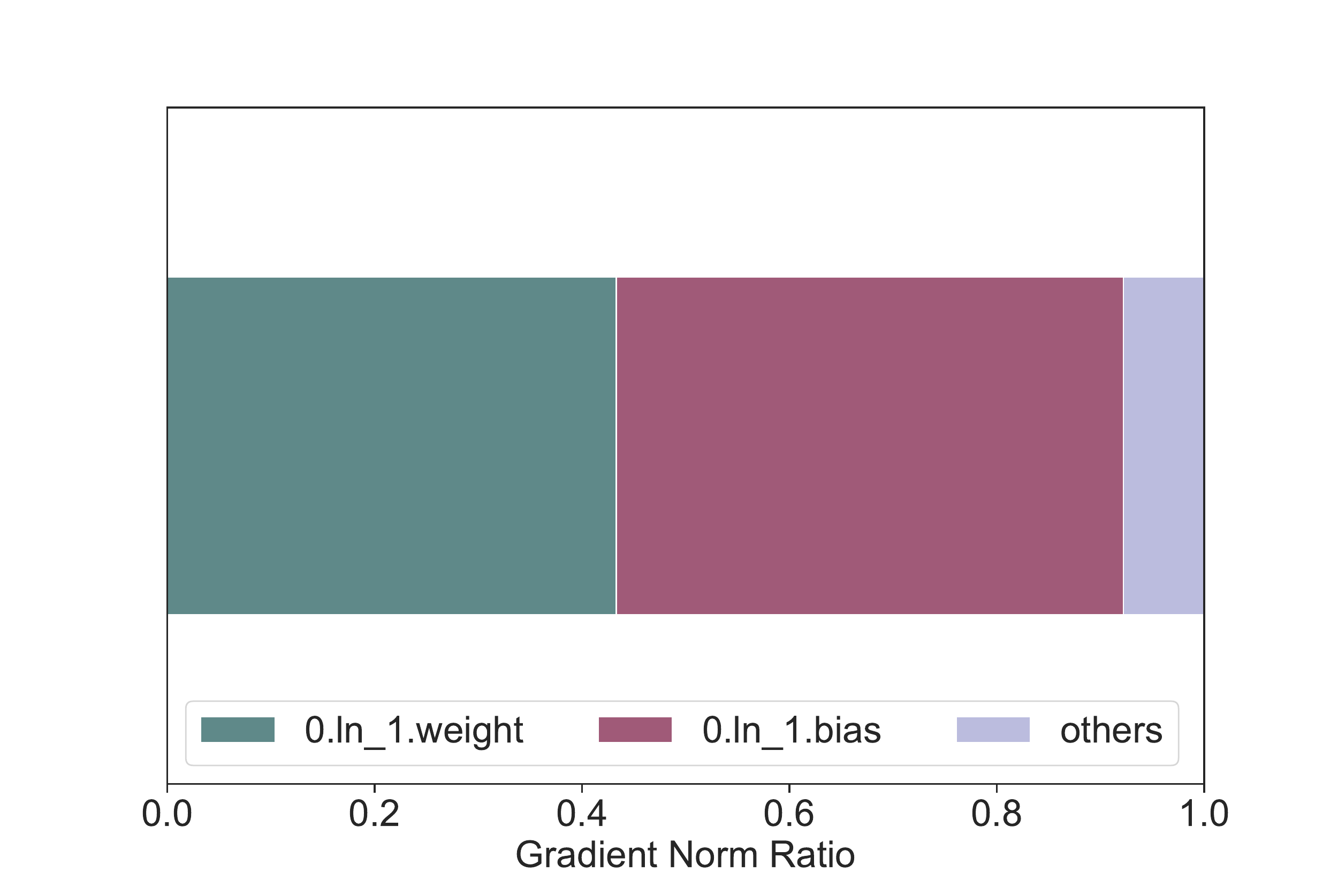}
        \subcaption{Walker2D}
    \end{minipage}
    \caption{iGPT's gradient norm ratio. (Seed = 42)}
\end{figure}

\begin{figure}[H]
    \centering
        \includegraphics[width=\linewidth]{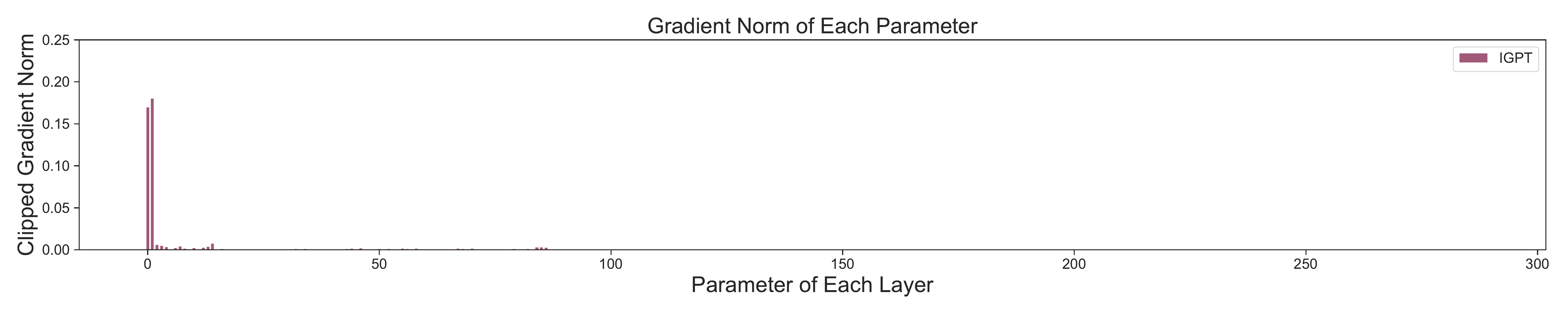}
    \caption{Gradient norm of iGPT's each parameter at epoch 1. (Hopper, Seed = 42)}
\end{figure}

\begin{figure}[H]
    \centering
        \includegraphics[width=\linewidth]{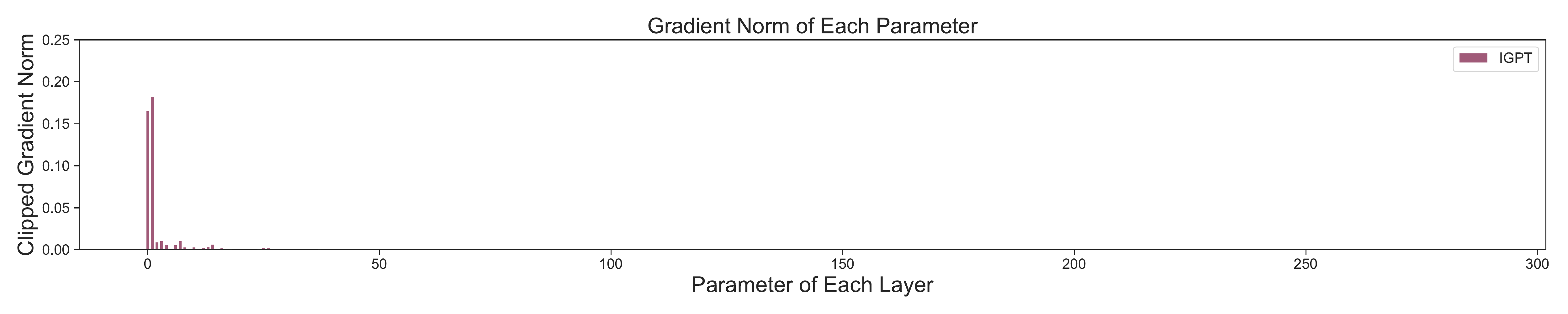}
    \caption{Gradient norm of iGPT's each parameter at epoch 1. (HalfCheetah, Seed = 42)}
\end{figure}

\begin{figure}[H]
    \centering
        \includegraphics[width=\linewidth]{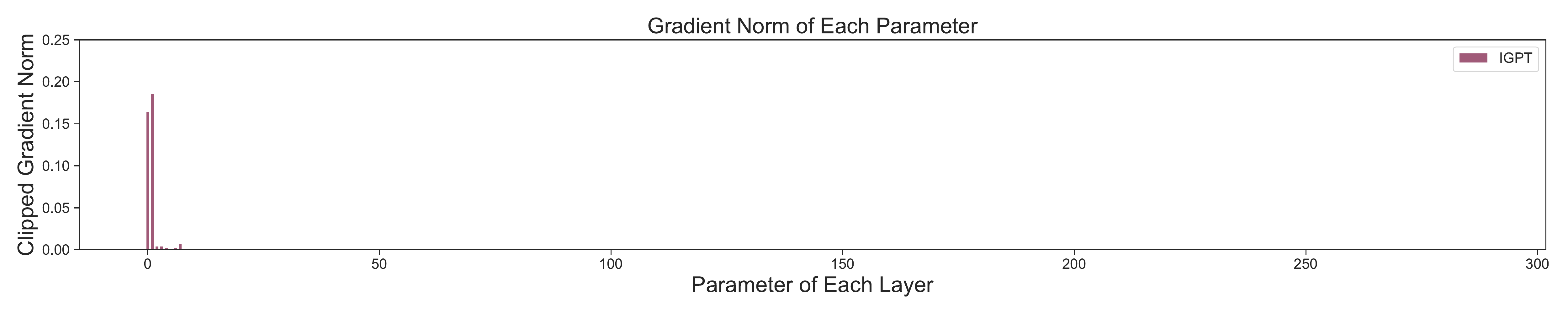}
    \caption{Gradient norm of iGPT's each parameter at epoch 1. (Walker2D, Seed = 42)}
\end{figure}

\subsection{Fine-Tuning with No Context Information}
\label{apendx:results-for-other-conditions-fine-tuning-with-no-context-information}

\begin{figure}[H]
    \centering
    \begin{minipage}[b]{0.32\linewidth}
        \includegraphics[width=\linewidth]{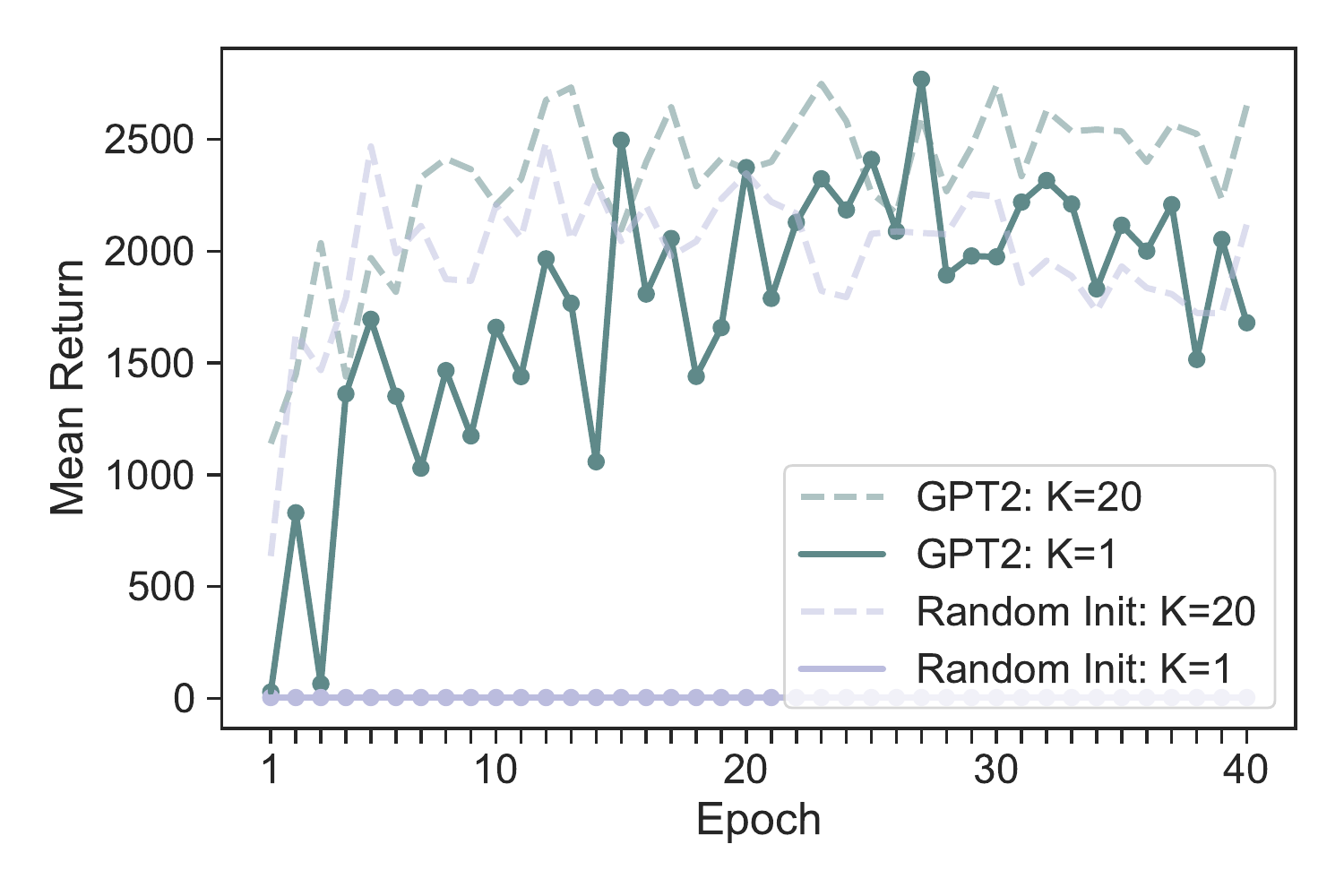}
        \subcaption{Hopper}
    \end{minipage}
    \begin{minipage}[b]{0.32\linewidth}
        \includegraphics[width=\linewidth]{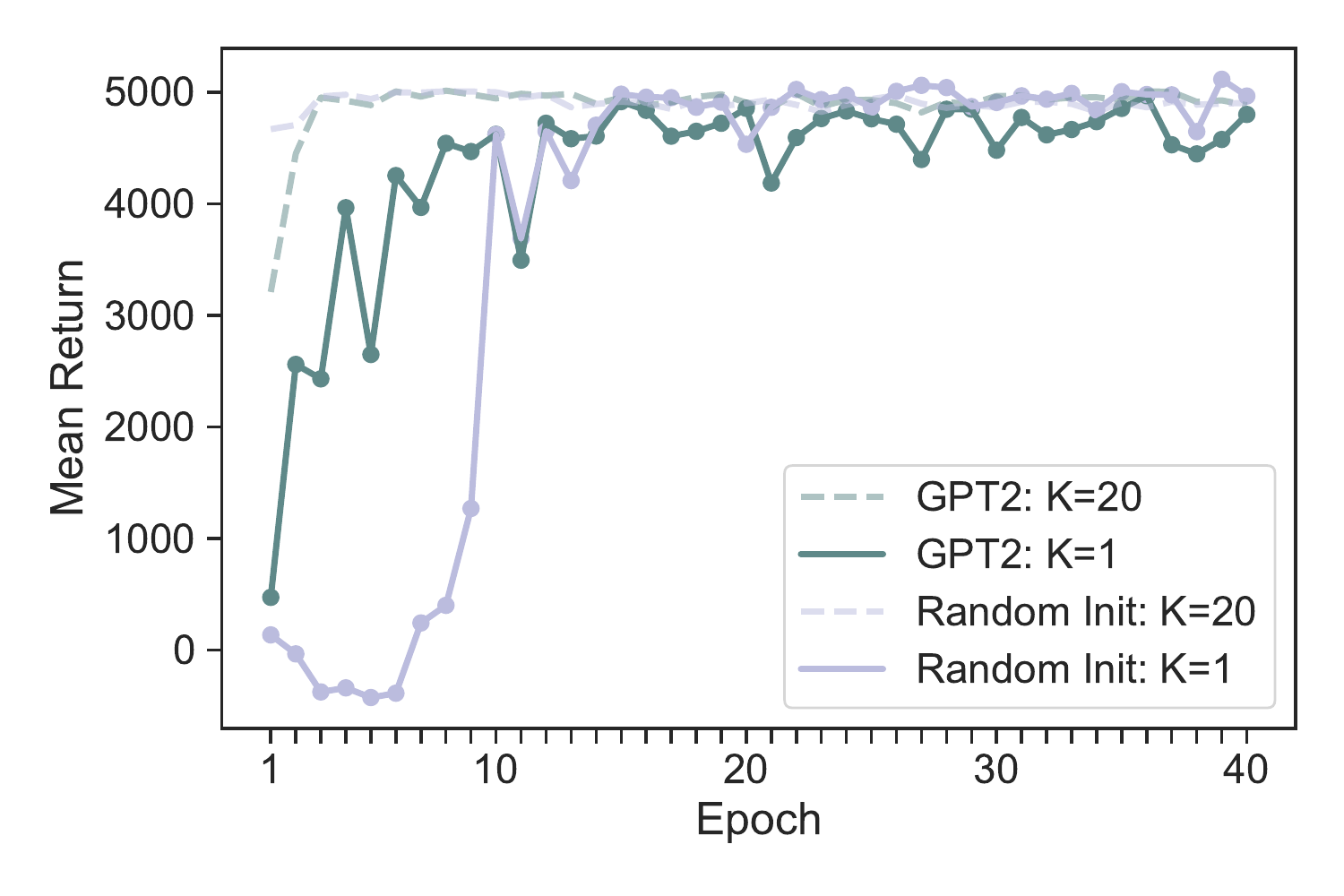}
        \subcaption{HalfCheetah}
    \end{minipage}
    \begin{minipage}[b]{0.32\linewidth}
        \includegraphics[width=\linewidth]{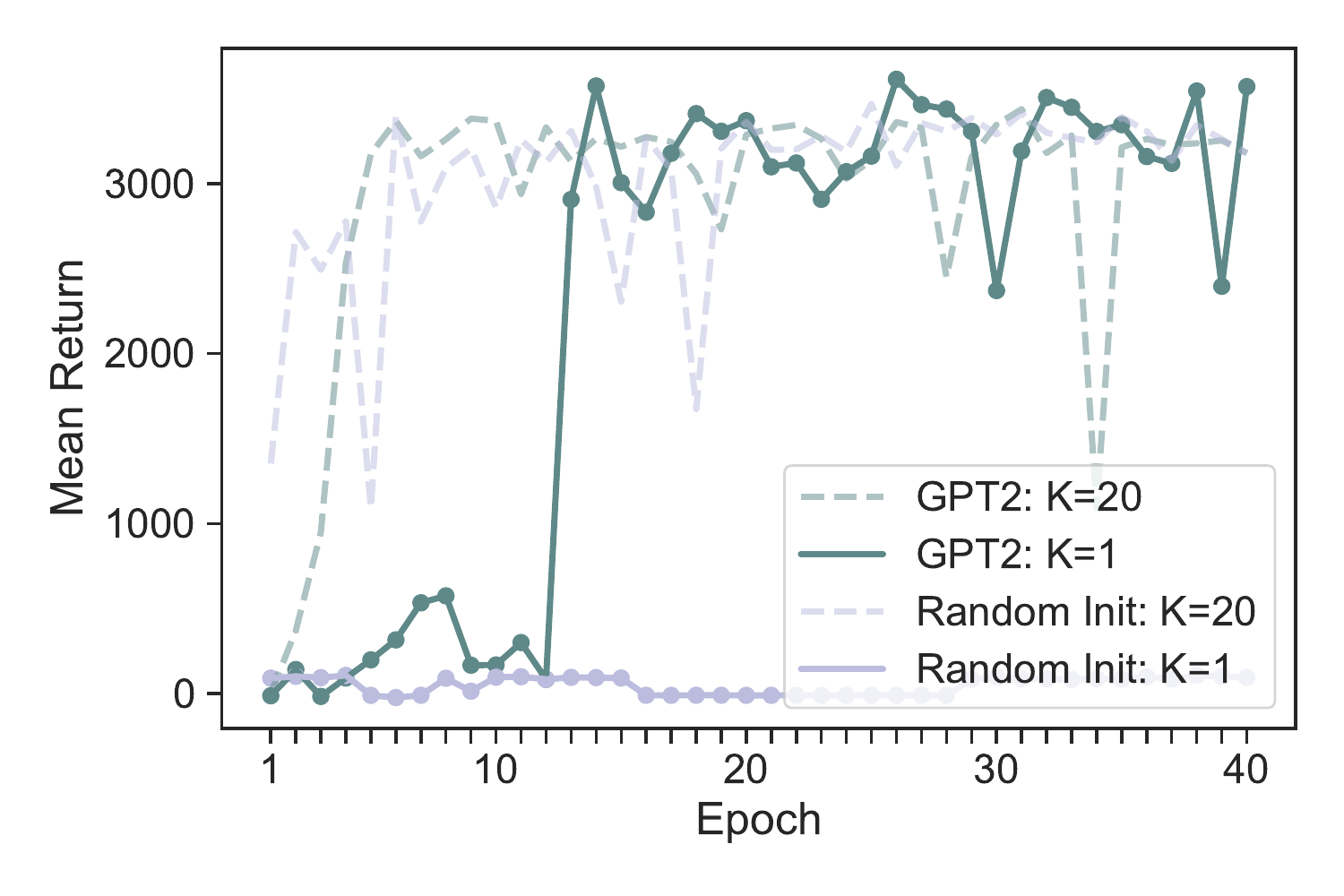}
        \subcaption{Walker2D}
    \end{minipage}
    \caption{Mean return throughout fine-tuning when access to the context information is prohibited (seed = 42).}
    \label{fig:return-mean-42}
\end{figure}

\begin{table}[H]
  \caption{Normalized return of $K = 1$ (seed = 666 \& 42).}
  \label{table:k=1-666-42}
  \centering
  \scalebox{0.8}{
  \begin{tabular}{llll}
    \toprule
    \cmidrule(r){1-4}
    Dataset     & Environment & GPT2 & Random Init \\
    \midrule
    \multirow{3}{*}{Medium} & Hopper & {$81.3 \pm 2.5$} & {$-0.3 \pm 0.1$} \\
    & HalfCheetah     & {$47.9 \pm 0.1$} & {$49.2 \pm 0.1$} \\
    & Walker 2D     & {$71.2 \pm 2.2$} & {$35.6 \pm 33.5$} \\
    \bottomrule
  \end{tabular}
  }
\end{table}

\subsection{More In-Depth Analysis of Context Dependence}
\label{appendix:results-for-other-conditions-internal-analysis-to-see-the-dependence-on-context}

\subsubsection{Replacement by the Pre-Trained Block}
\label{appendix:results-for-other-conditions-replacement}

\begin{figure}[H]
    \centering
    \begin{minipage}[b]{0.48\linewidth}
        \includegraphics[width=\linewidth]{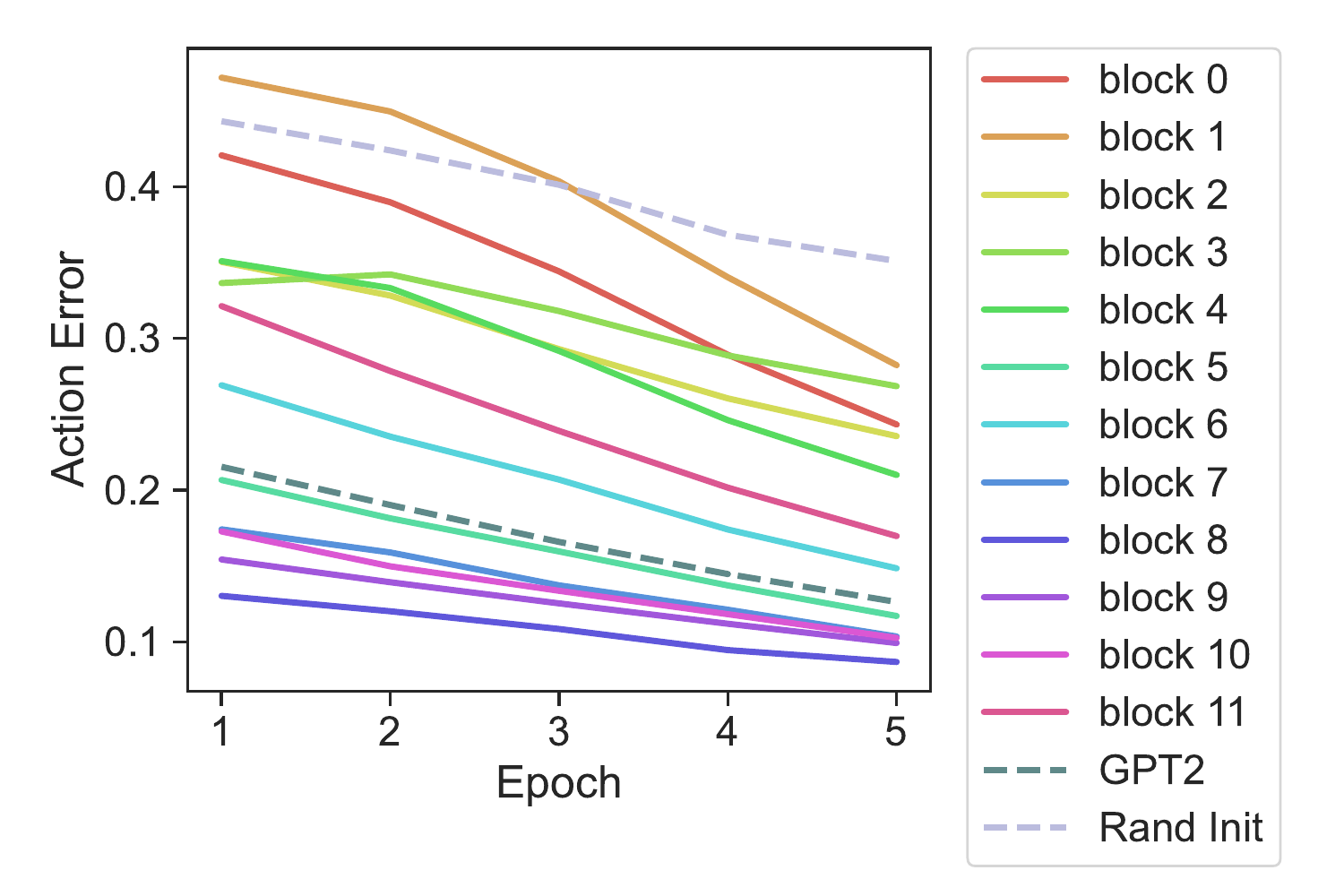}
        \subcaption{Action error}
    \end{minipage}
    \begin{minipage}[b]{0.48\linewidth}
        \includegraphics[width=\linewidth]{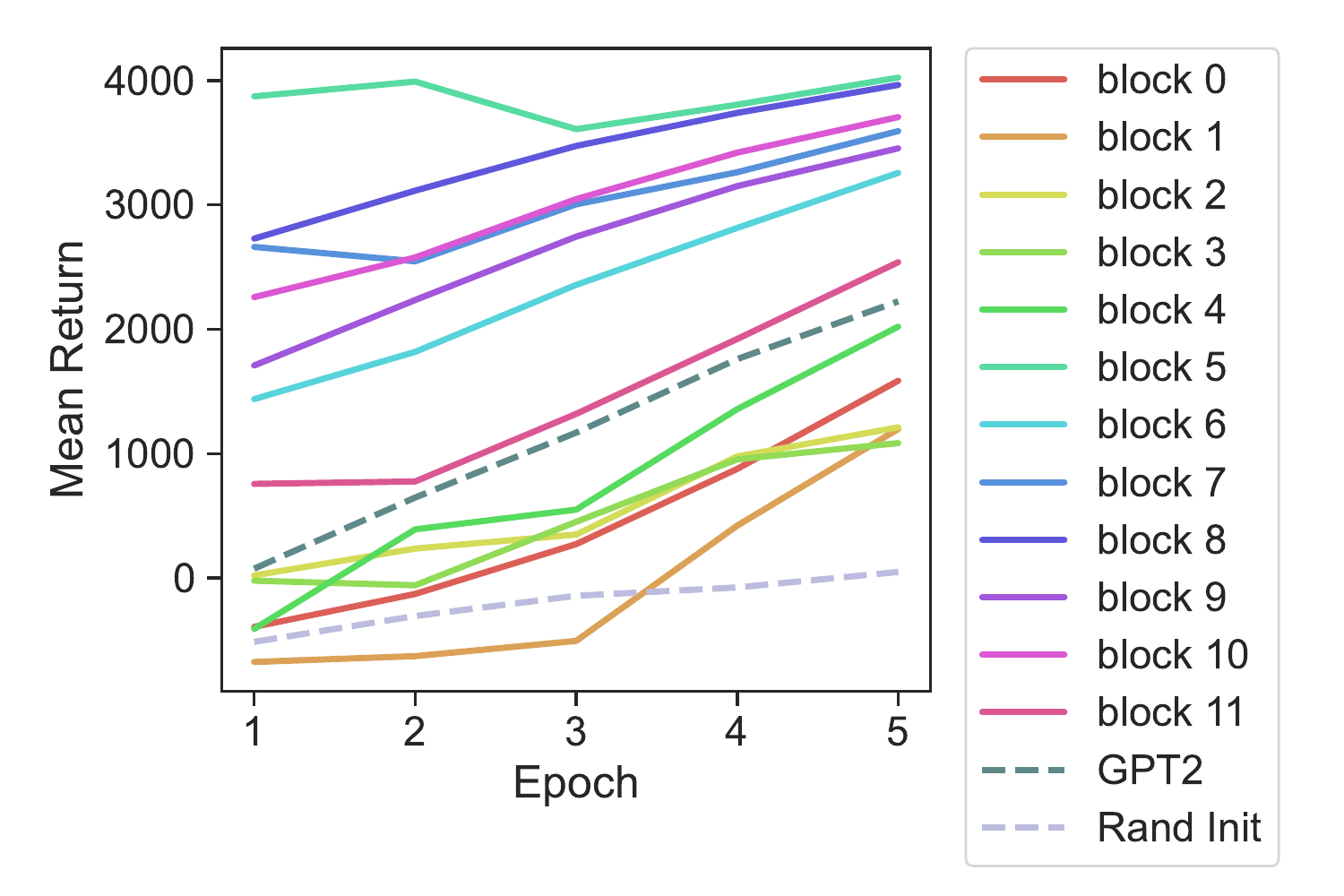}
        \subcaption{Mean return}
    \end{minipage}
    \caption{Learning curve when only a block is pre-trained (HalfCheetah, Seed = 666).}
    \label{fig:learning_curve_halfcheetah}
\end{figure}

\begin{figure}[H]
    \centering
    \begin{minipage}[b]{0.48\linewidth}
        \includegraphics[width=\linewidth]{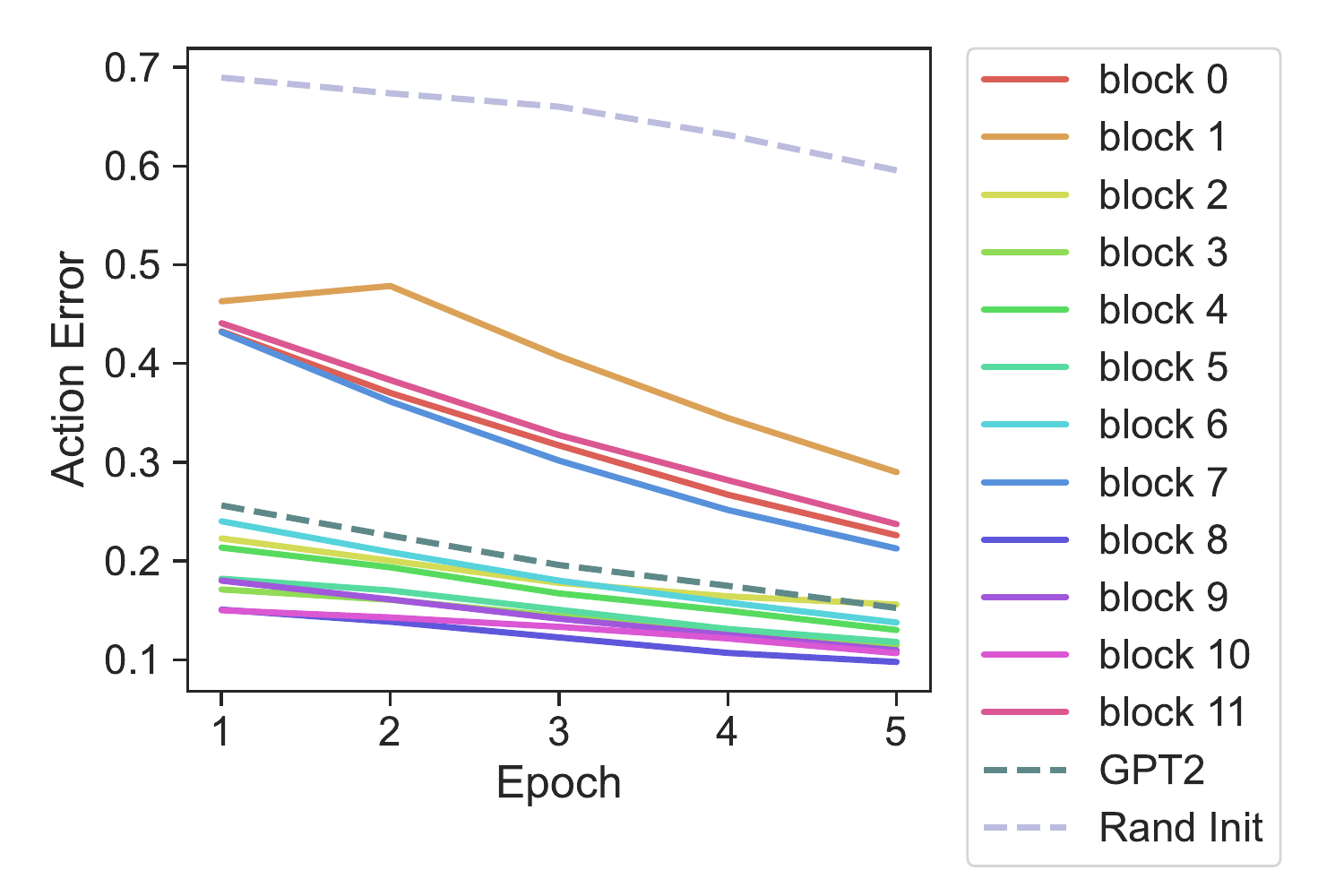}
        \subcaption{Action error}
    \end{minipage}
    \begin{minipage}[b]{0.48\linewidth}
        \includegraphics[width=\linewidth]{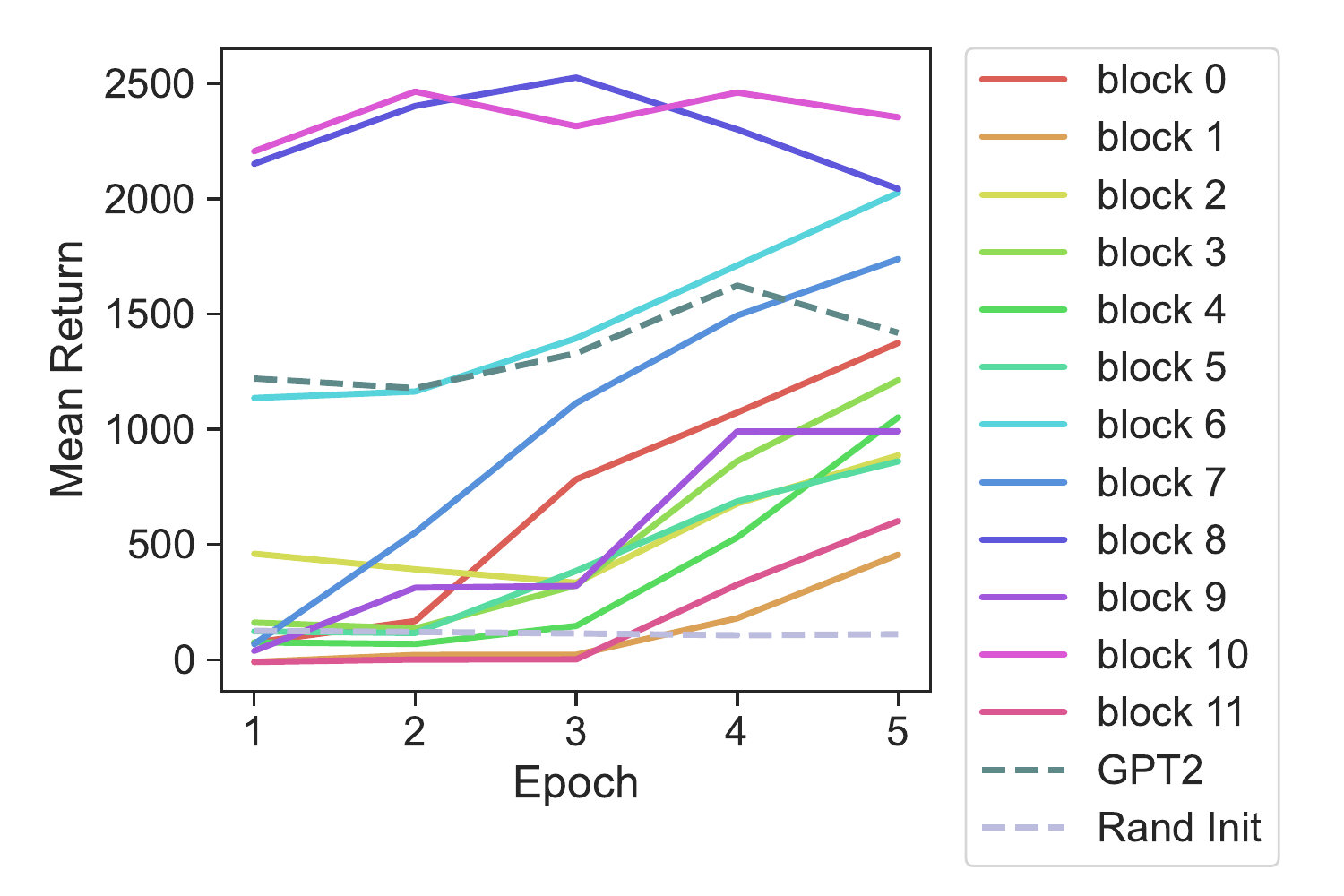}
        \subcaption{Mean return}
    \end{minipage}
    \caption{Learning curve when only a block is pre-trained (Walker2D, Seed = 666).}
    \label{fig:learning_curve_walker2d}
\end{figure}

\begin{figure}[H]
    \centering
    \begin{minipage}[b]{0.48\linewidth}
        \includegraphics[width=\linewidth]{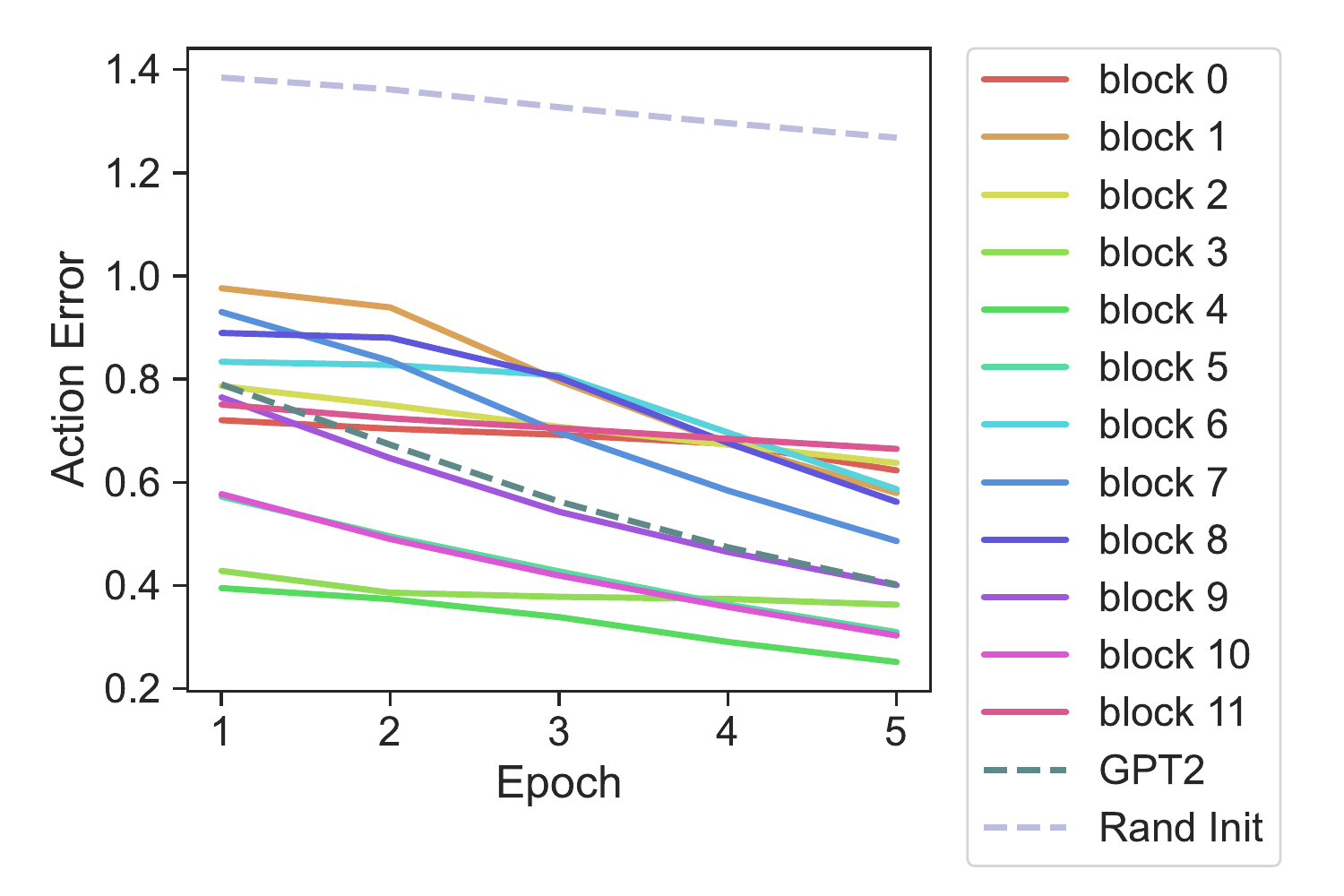}
        \subcaption{Action error}
    \end{minipage}
    \begin{minipage}[b]{0.48\linewidth}
        \includegraphics[width=\linewidth]{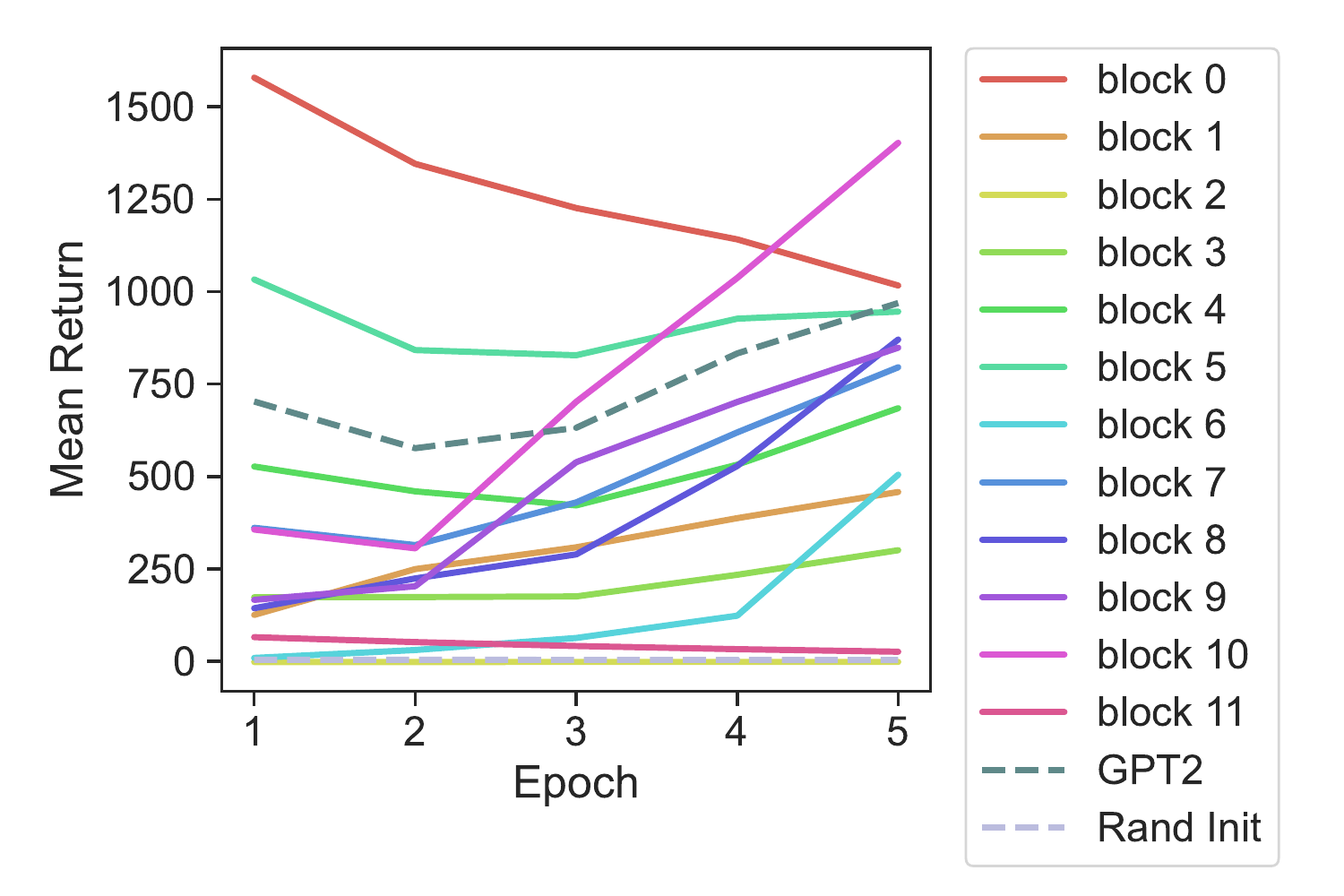}
        \subcaption{Mean return}
    \end{minipage}
    \caption{Learning curve when only a block is pre-trained (Hopper, Seed = 42).}
    \label{fig:learning_curve_hopper_42}
\end{figure}

\begin{figure}[H]
    \centering
    \begin{minipage}[b]{0.48\linewidth}
        \includegraphics[width=\linewidth]{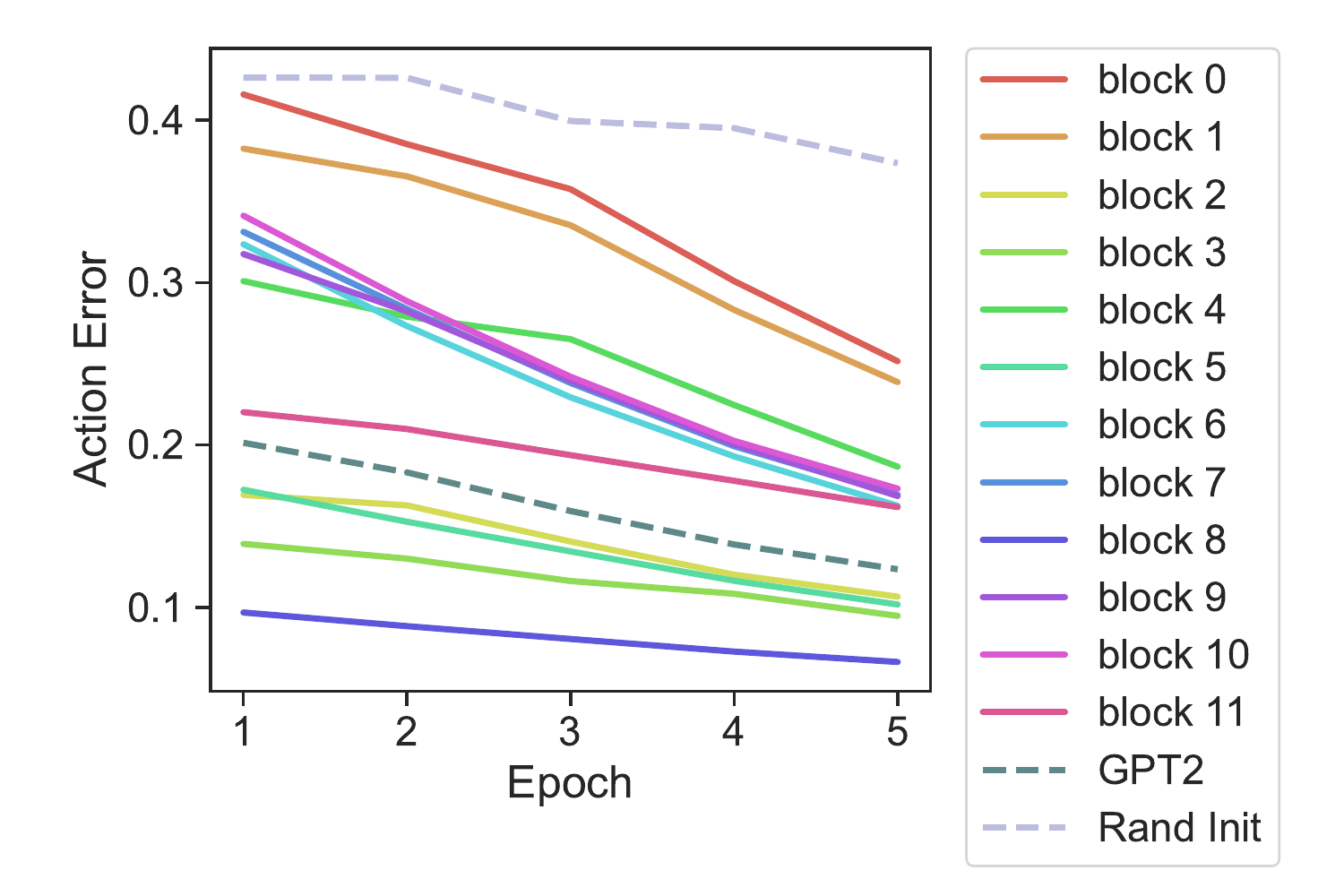}
        \subcaption{Action error}
    \end{minipage}
    \begin{minipage}[b]{0.48\linewidth}
        \includegraphics[width=\linewidth]{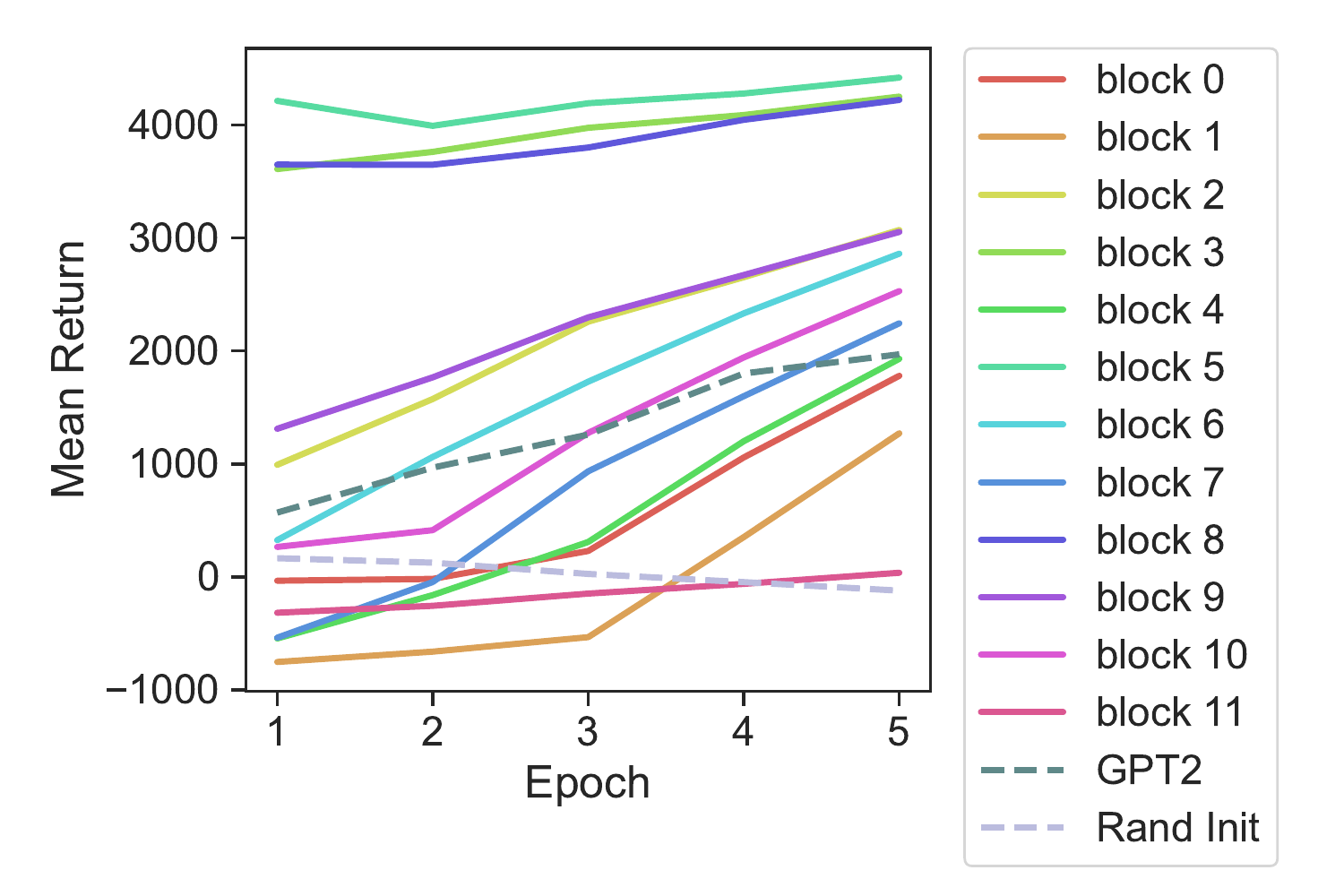}
        \subcaption{Mean return}
    \end{minipage}
    \caption{Learning curve when only a block is pre-trained (HalfCheetah, Seed = 42).}
    \label{fig:learning_curve_halfcheetah_42}
\end{figure}

\begin{figure}[H]
    \centering
    \begin{minipage}[b]{0.48\linewidth}
        \includegraphics[width=\linewidth]{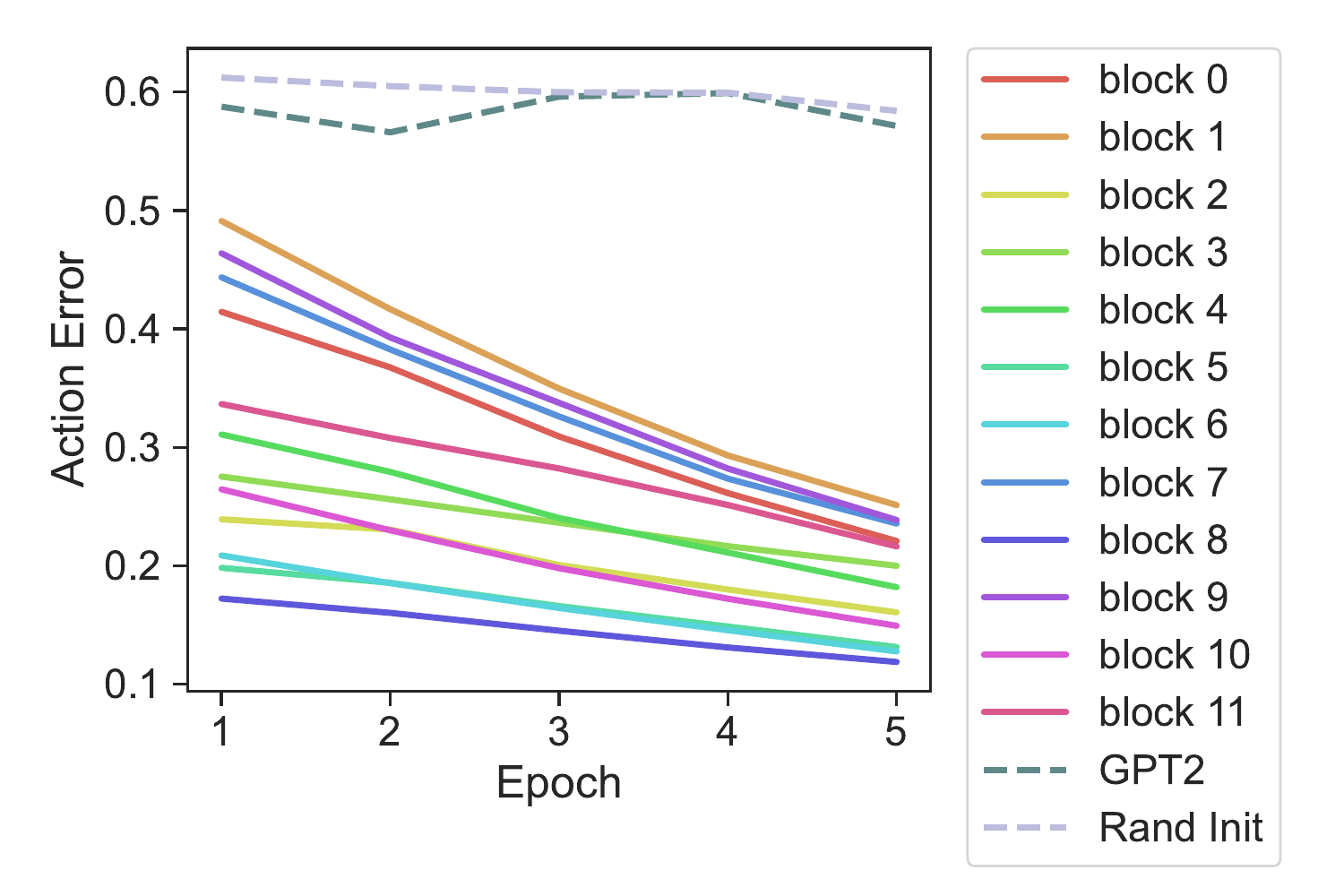}
        \subcaption{Action error}
    \end{minipage}
    \begin{minipage}[b]{0.48\linewidth}
        \includegraphics[width=\linewidth]{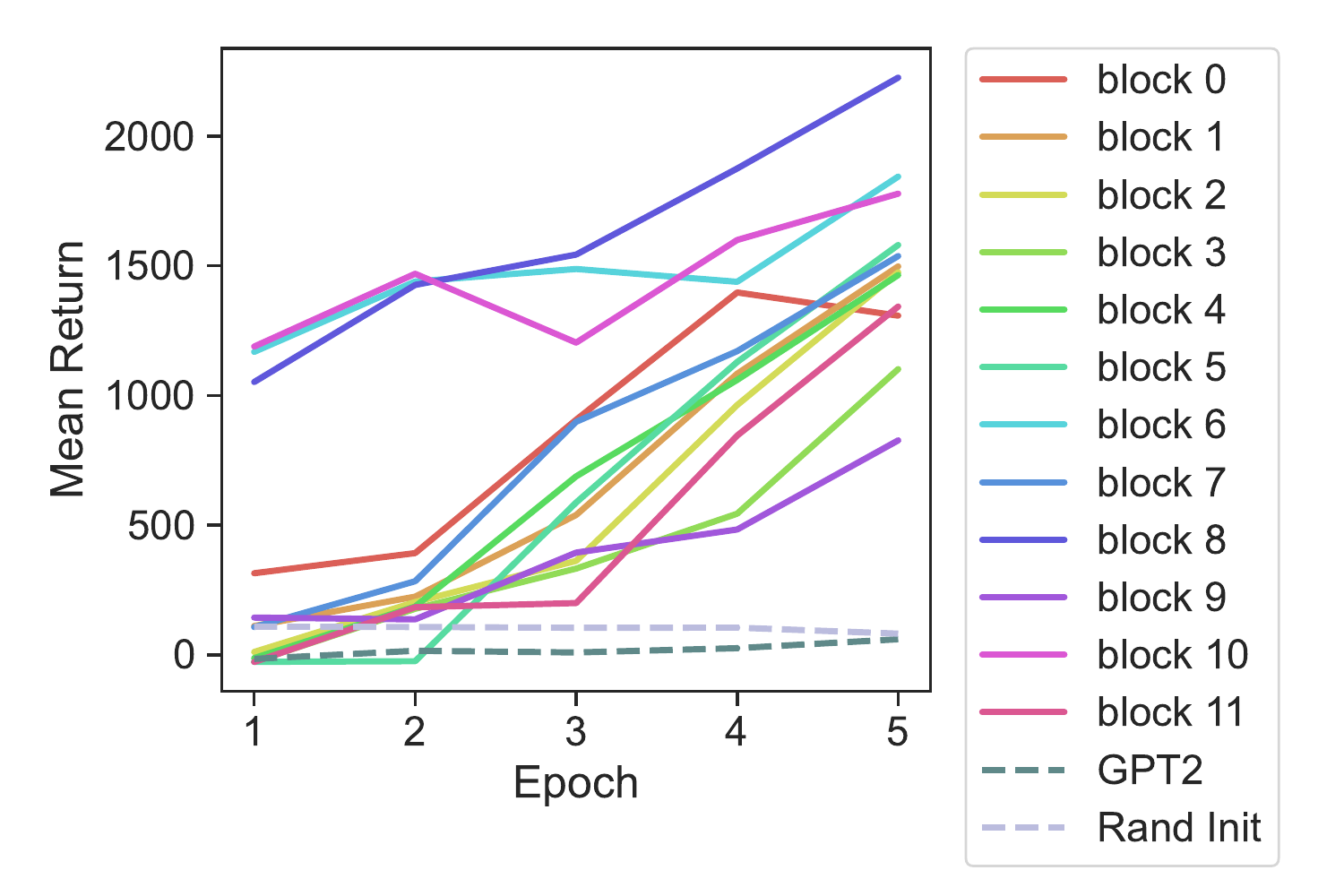}
        \subcaption{Mean return}
    \end{minipage}
    \caption{Learning curve when only a block is pre-trained (Walker2D, Seed = 42).}
    \label{fig:learning_curve_walker2d_42}
\end{figure}

\subsubsection{Attention Distance Analysis}
\label{appendix:results-for-other-conditions-attention-distance}
\begin{figure}[H]
    \centering
    \begin{minipage}[b]{0.32\linewidth}
        \includegraphics[width=\linewidth]{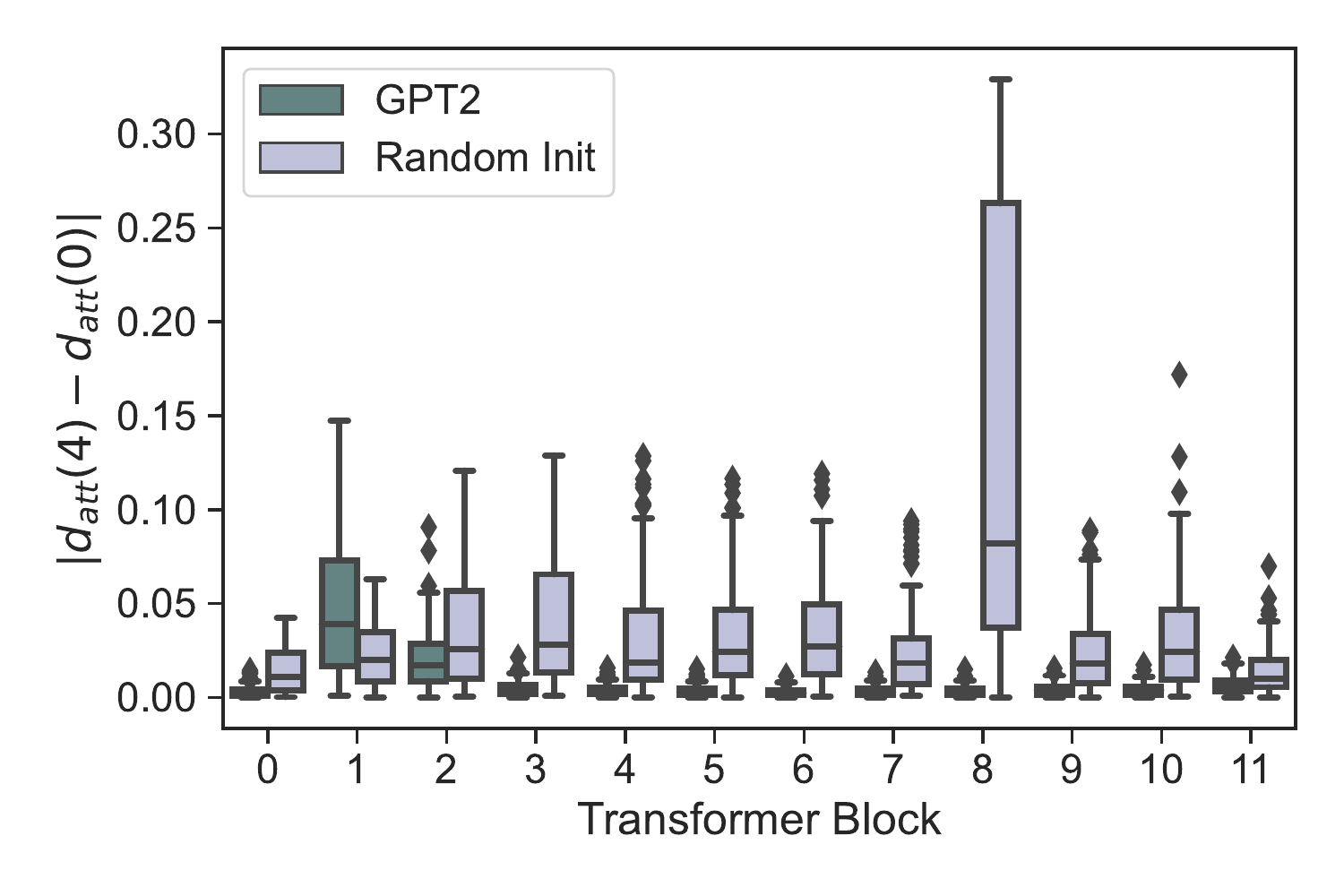}
        \subcaption{Hopper}
    \end{minipage}
    \begin{minipage}[b]{0.32\linewidth}
        \includegraphics[width=\linewidth]{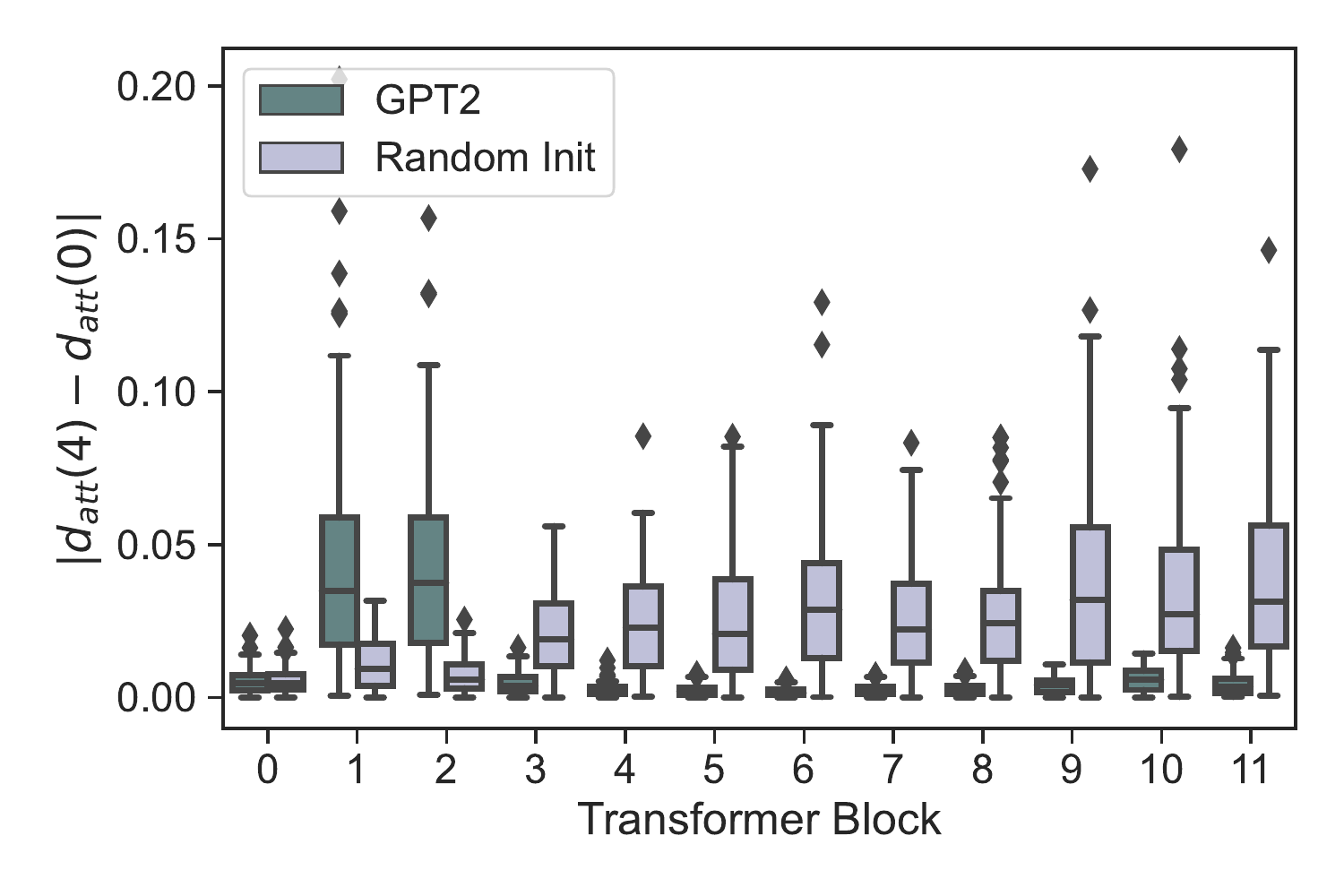}
        \subcaption{HalfCheetah}
    \end{minipage}
    \begin{minipage}[b]{0.32\linewidth}
        \includegraphics[width=\linewidth]{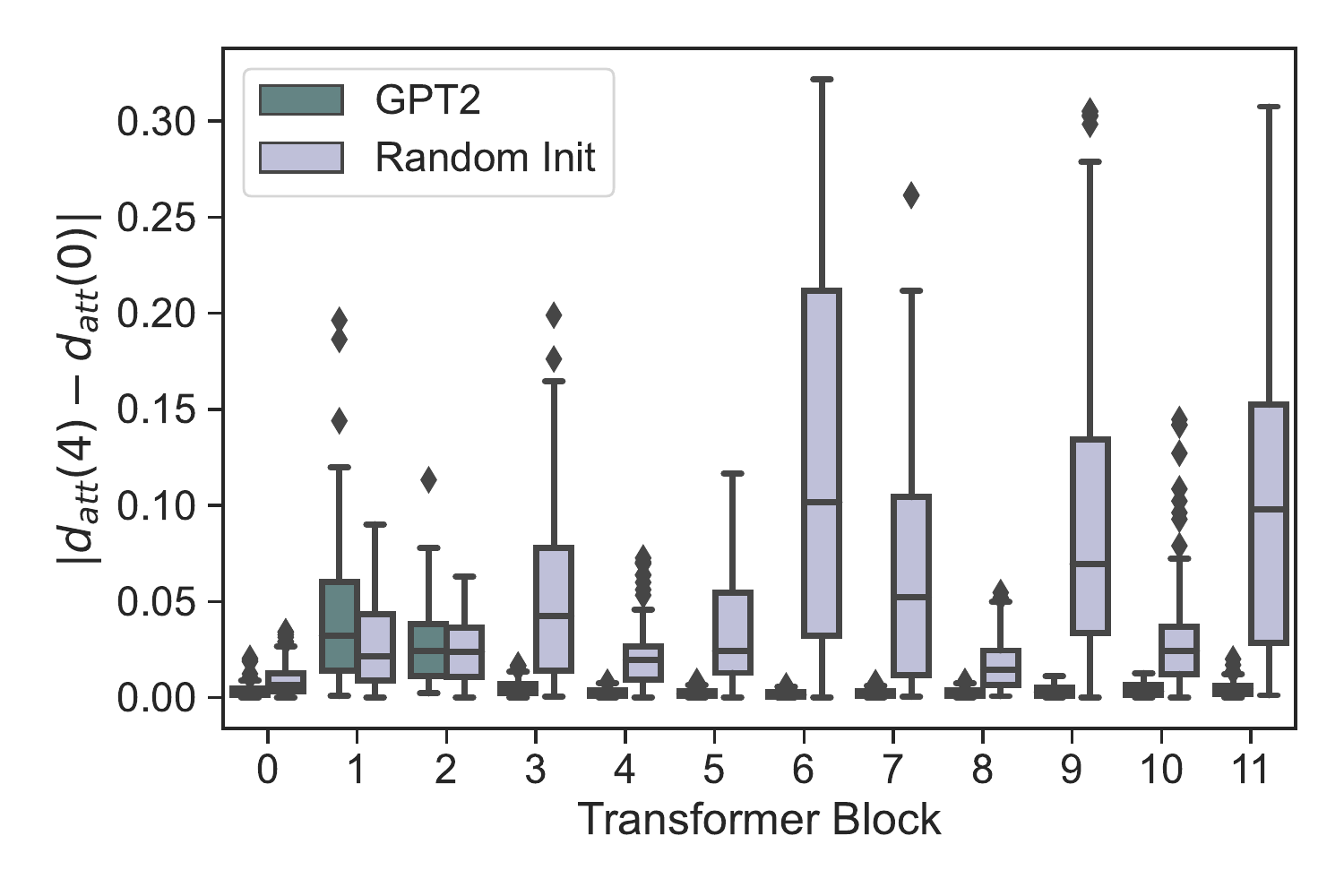}
        \subcaption{Walker2D}
    \end{minipage}
    \caption{Attention distance gap between epoch 0 and epoch 1.}
\end{figure}

\begin{figure}[H]
    \centering
    \begin{minipage}[b]{0.32\linewidth}
        \includegraphics[width=\linewidth]{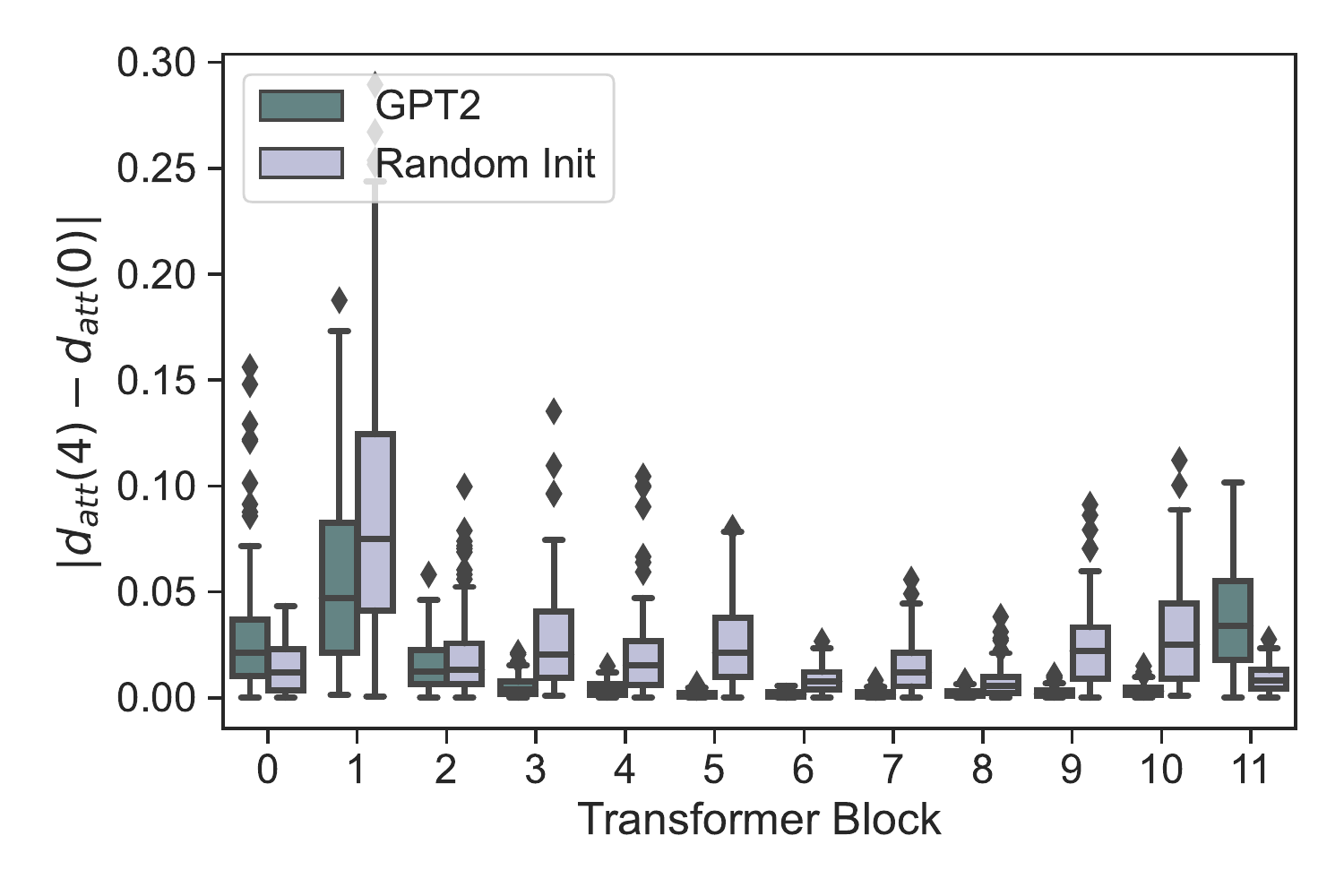}
        \subcaption{Hopper}
    \end{minipage}
    \begin{minipage}[b]{0.32\linewidth}
        \includegraphics[width=\linewidth]{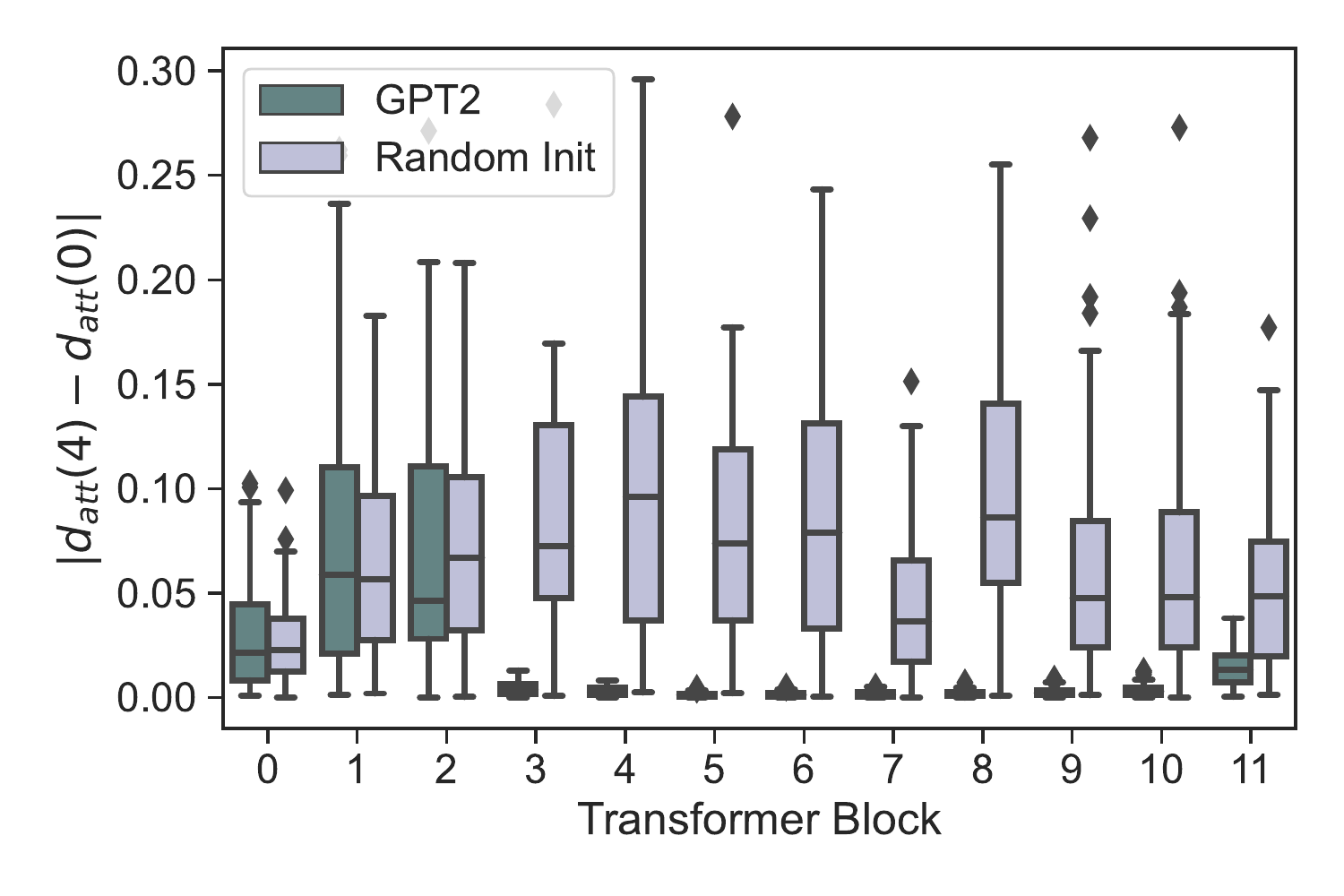}
        \subcaption{HalfCheetah}
    \end{minipage}
    \begin{minipage}[b]{0.32\linewidth}
        \includegraphics[width=\linewidth]{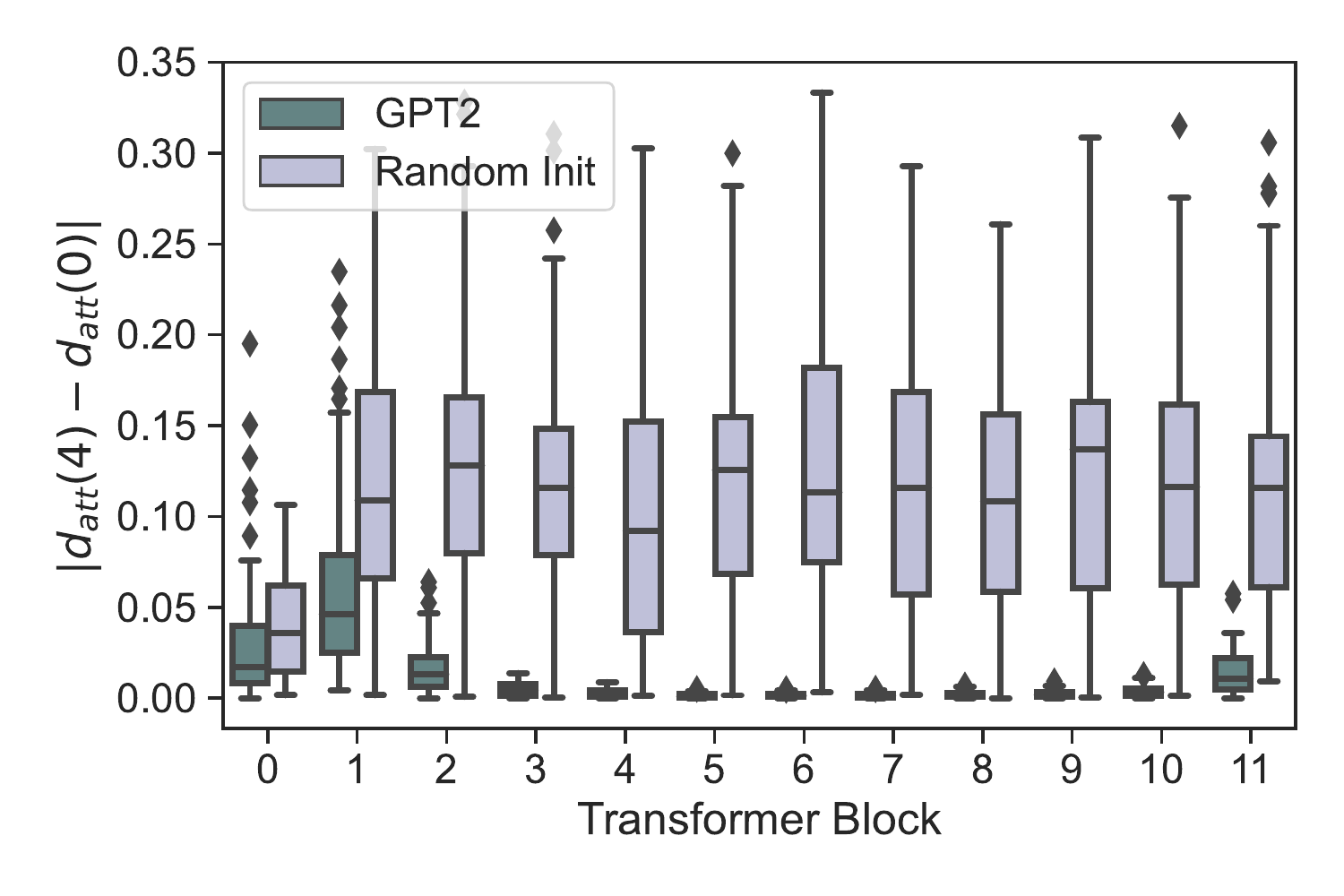}
        \subcaption{Walker2D}
    \end{minipage}
    \caption{Attention distance gap between epoch 0 and epoch 10.}
\end{figure}

\begin{figure}[H]
    \centering
    \begin{minipage}[b]{0.32\linewidth}
        \includegraphics[width=\linewidth]{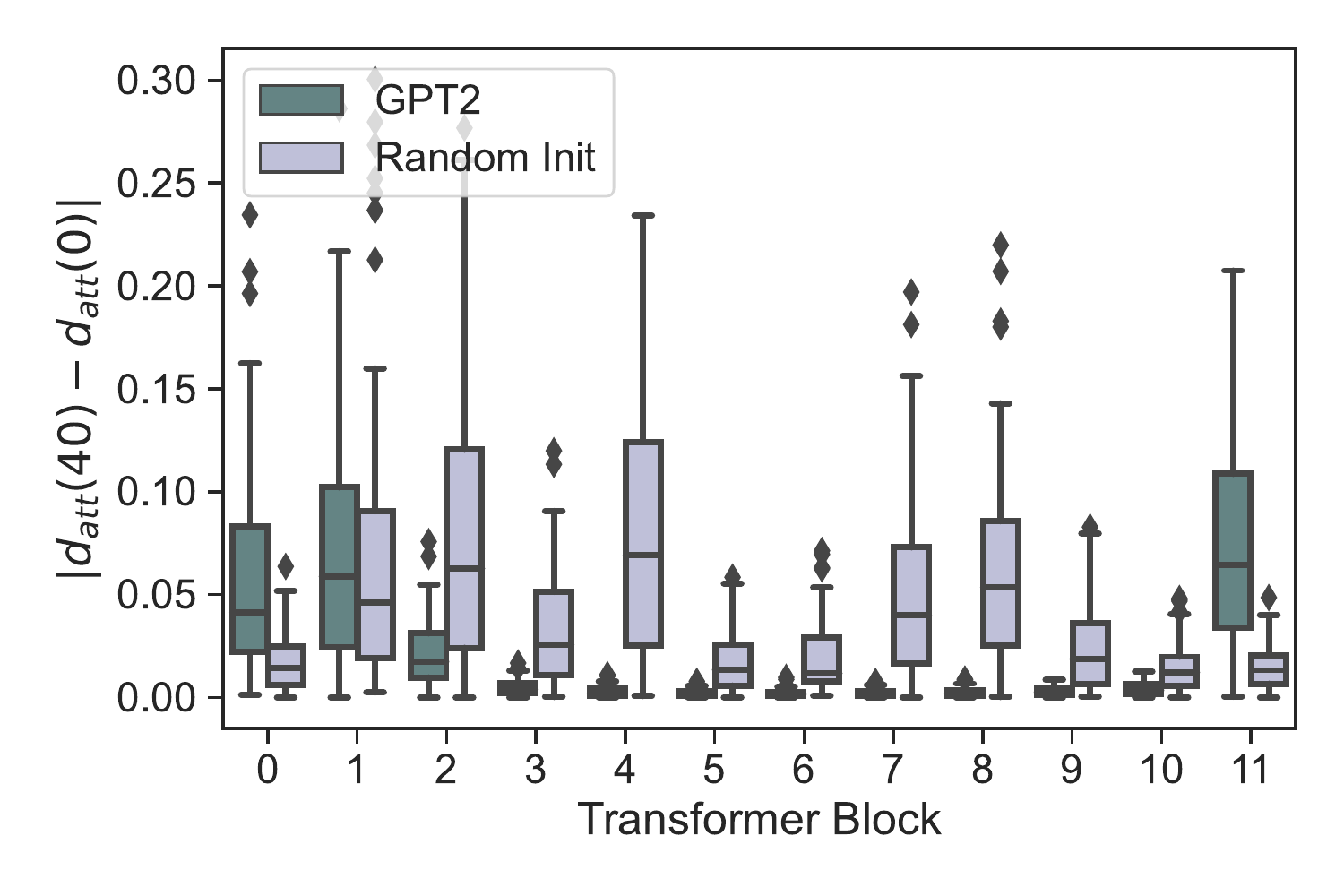}
        \subcaption{Hopper}
    \end{minipage}
    \begin{minipage}[b]{0.32\linewidth}
        \includegraphics[width=\linewidth]{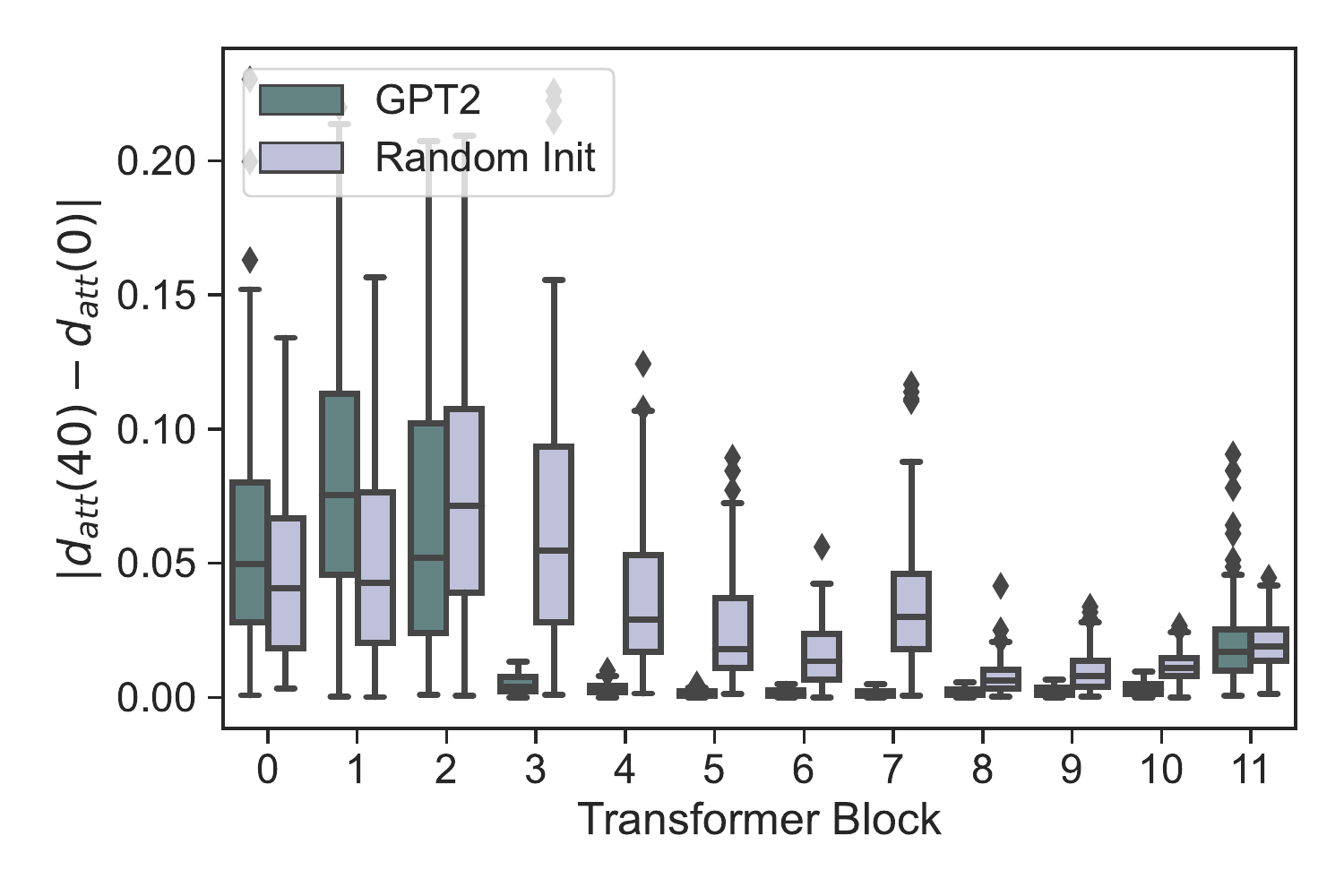}
        \subcaption{HalfCheetah}
    \end{minipage}
    \begin{minipage}[b]{0.32\linewidth}
        \includegraphics[width=\linewidth]{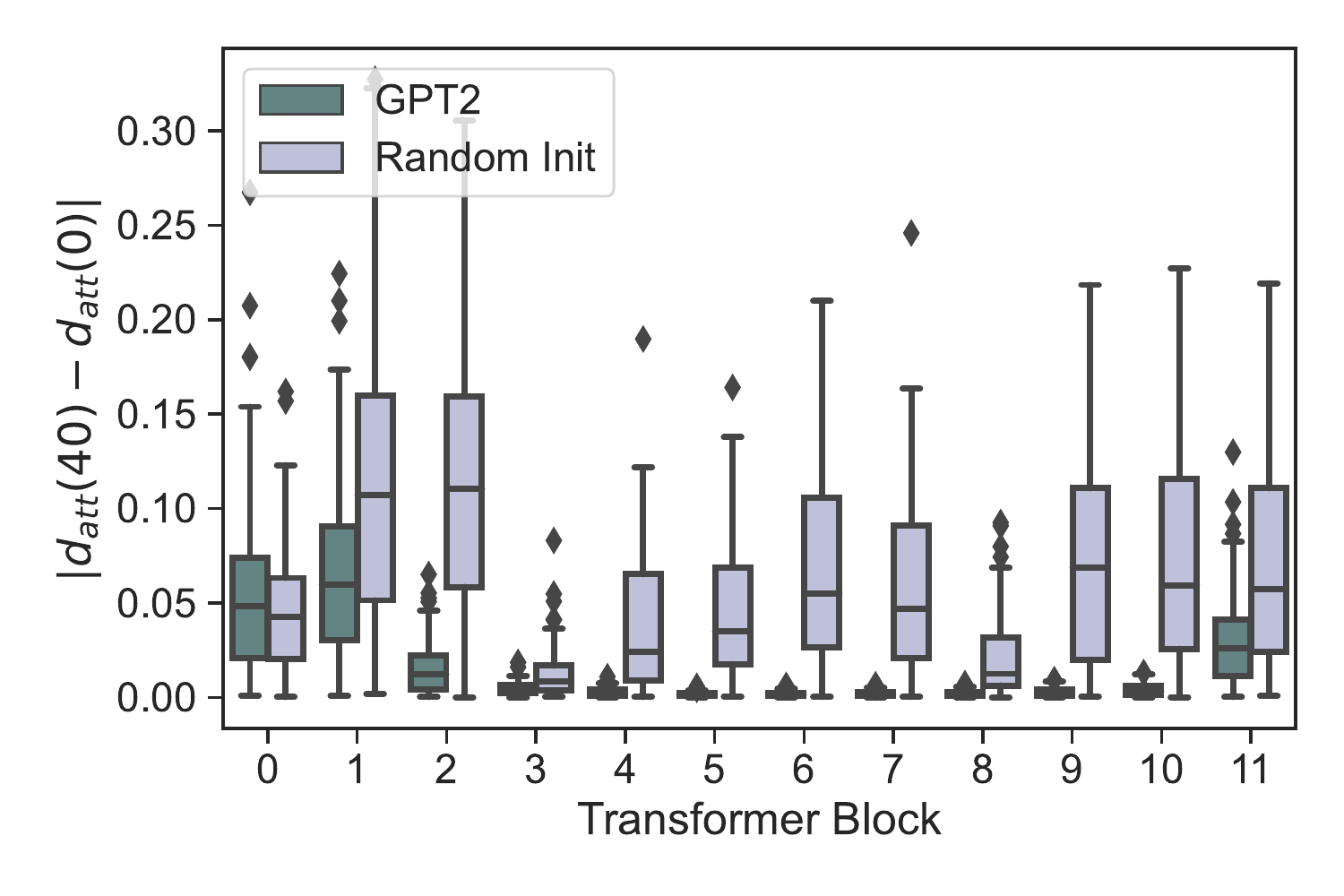}
        \subcaption{Walker2D}
    \end{minipage}
    \caption{Attention distance gap between epoch 0 and epoch 40.}
\end{figure}

\begin{figure}[H]
    \centering
    \begin{minipage}[b]{0.32\linewidth}
        \includegraphics[width=\linewidth]{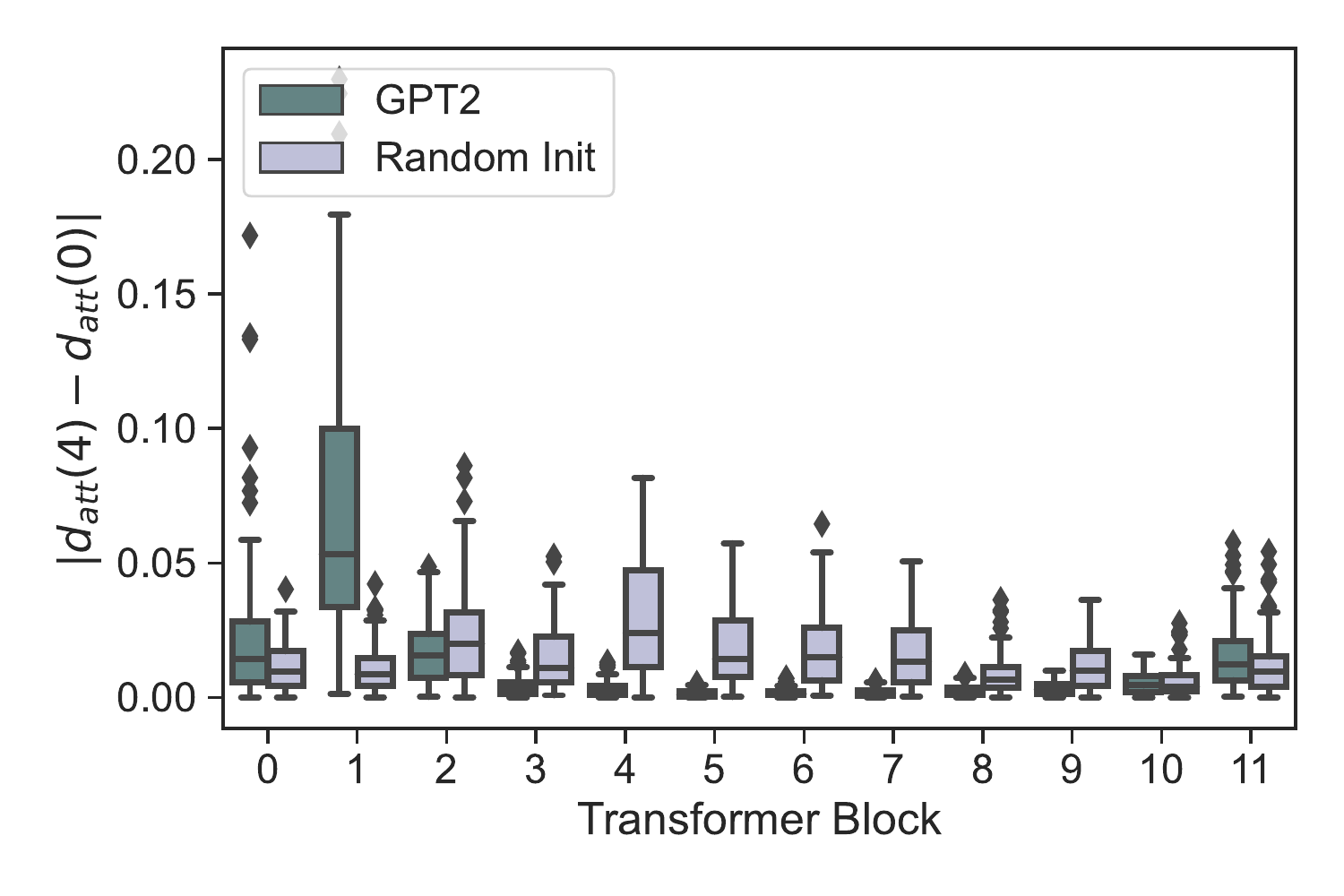}
        \subcaption{Hopper}
    \end{minipage}
    \begin{minipage}[b]{0.32\linewidth}
        \includegraphics[width=\linewidth]{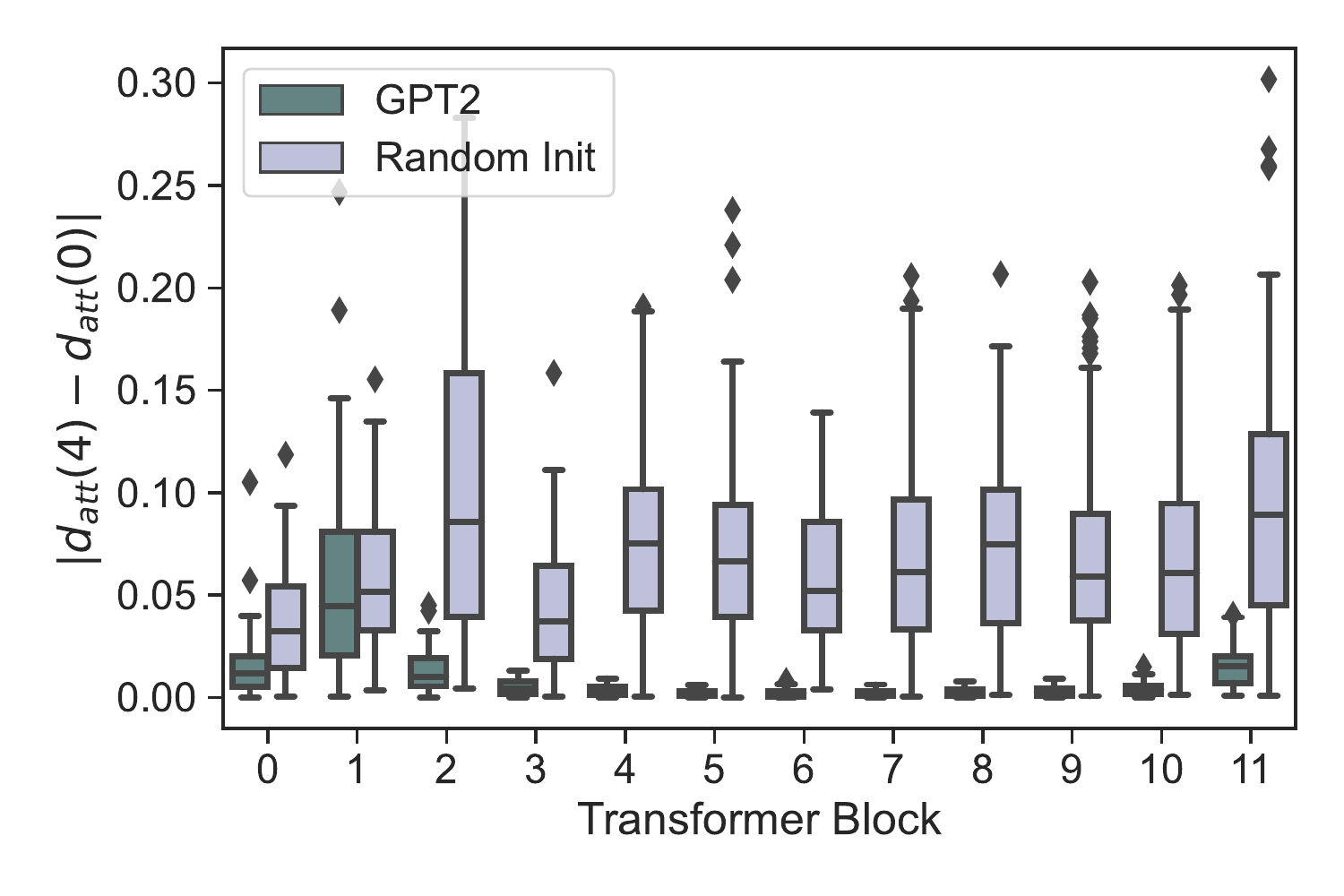}
        \subcaption{HalfCheetah}
    \end{minipage}
    \begin{minipage}[b]{0.32\linewidth}
        \includegraphics[width=\linewidth]{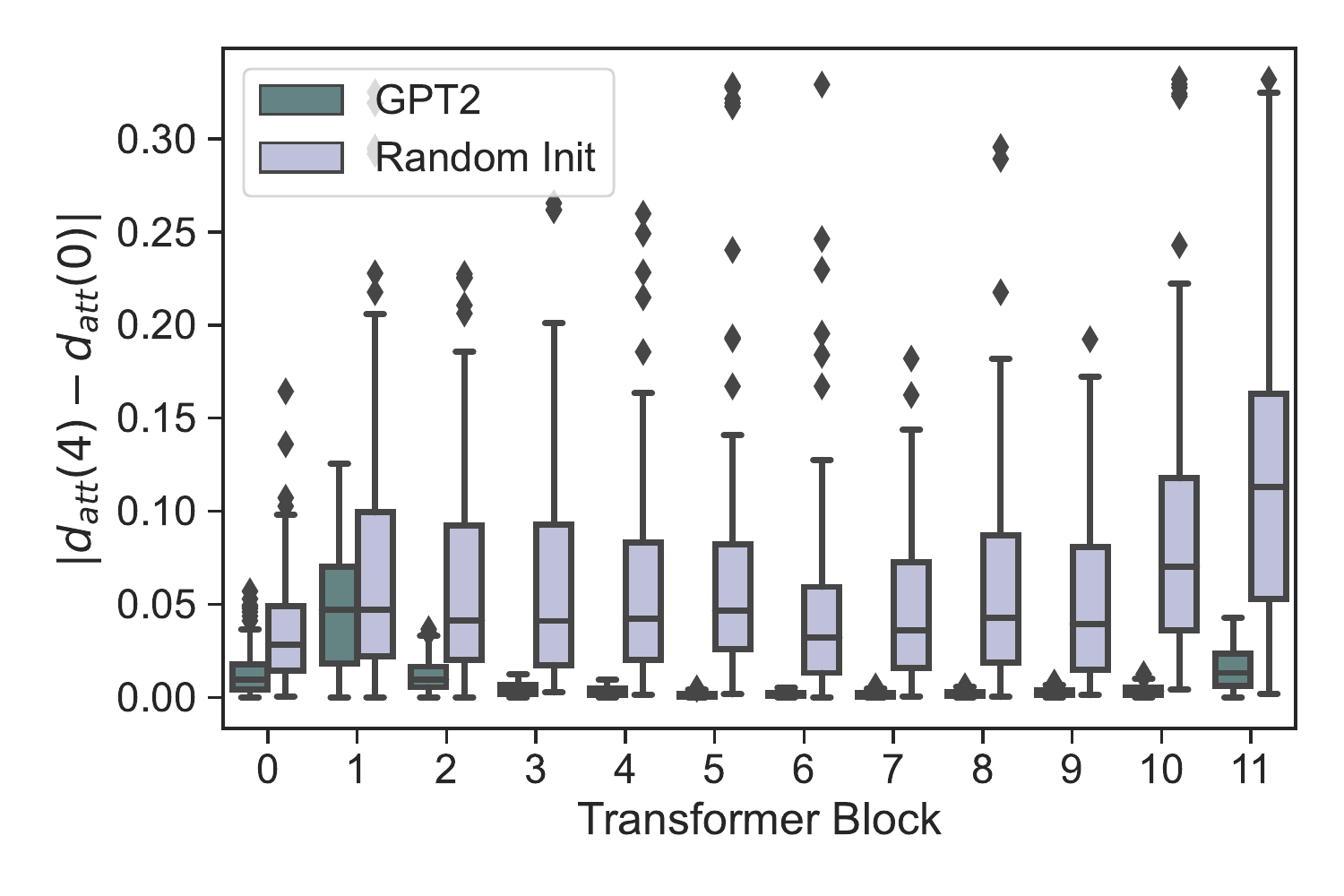}
        \subcaption{Walker2D}
    \end{minipage}
    \caption{Attention distance gap between epoch 0 and epoch 4 (seed = 42).}
\end{figure}

\section{Details for Computation}
\label{appendix:details-for-computation}
The computational environment for our experiments is as follows:
\begin{itemize}
    \item GPU: NVIDIA GeForce GTX TITAN X $\times$ 2GPU.
    \item CPU: Intel Xeon E5-2620 v3 2.4 GHz, 6 Cores $\times$ 2CPU.
    \item software: Ubuntu 20.04.4, GCC 9.4.0, CUDA 11.6.
\end{itemize}
The computational time of fine-tuning for 40 epochs is around from 24 to 72 hours for context length $K = 20$ and around from 7 to 48 hours for context length $K = 1$ per run.

\section{Licences of Assets Used for Our Experiments}
\label{appendix:licence}
\subsection{Dataset}
We use the offline RL dataset of Mujoco in D4RL \citep{fu2020d4rl}. The license of the dataset in D4RL is a  Creative Commons Attribution 4.0 License (CC BY). Thus, we can use this dataset with no consent as long as we follow the term of use. 
\subsection{Code}
The code we used does not require special consent from authors as long as we follow the terms of use. Their licenses are as follows:
\begin{itemize}
\item \href{https://github.com/machelreid/can-wikipedia-help-offline-rl}{https://github.com/machelreid/can-wikipedia-help-offline-rl}: MIT Licence.
\item \href{https://github.com/rail-berkeley/d4rl}{https://github.com/rail-berkeley/d4rl}: Apache License 2.0.
\item \href{https://github.com/google-research/google-research/tree/master/representation_similarity}{https://github.com/google-research/google-research/tree/master/representation\_similarity}: Apache License 2.0.
\item \href{https://github.com/gtegner/mine-pytorch}{https://github.com/gtegner/mine-pytorch}: MIT License.
\end{itemize}

\section{Note on the Role of Each Sub-Sections in Section \ref{section:results-and-analysis}}

Sections \ref{section:activation-similarity} and \ref{section:mutual-information-between-hidden-representation-and-input-and-label} show what may \textit{not} be a cause of the good performance. In other words, these sections eliminate some seemingly possible causes of good performance of language-pre-trained models, respectively. These analyses further highlight the importance of attention distance for performance. Without the analysis of Section \ref{section:activation-similarity}, we cannot exclude the possibility that good performance comes from re-using pre-trained representation \textit{as well}, which is common in the uni-modal case as we explained in Section \ref{section:activation-similarity}. Similarly, without the analysis of Section \ref{section:mutual-information-between-hidden-representation-and-input-and-label}, the performance can benefit from fitting better into training data \textit{as well}. In that case, it would be less clear whether utilizing some prior knowledge is solely important. Sections \ref{section:gradient-analysis}, \ref{section:dependency-on-context-informaiton}, \ref{section:replacement}, and \ref{section:attention-distance} discuss what may be a cause of the good performance, as summarized at the end of Section \ref{section:results-and-analysis}. Section \ref{section:parameter-similarity} provides complementary findings for these sections. For example, the existence of some unchanged parameters implies the possibility that some information is re-used, while the relatively larger change in only shallower layers of pre-trained models partially supports that these unchanged parameters do not seem to contradict with changed representation observed in Section \ref{section:activation-similarity}, as explained in Appendix \ref{appendix:note-on-relationship-between-parameter-change-and-representation-change}.

\section{Note on Relationship Between Parameter Change and Representation Change}
\label{appendix:note-on-relationship-between-parameter-change-and-representation-change}

In Section \ref{section:activation-similarity} we say that \textit{representation} in some layers of the pre-trained model changes, while in Section \ref{section:parameter-similarity} we say that \textit{parameters} of them in some layers do not change that much, or vice versa for the randomly initialized model. These observations of Sections \ref{section:activation-similarity} and \ref{section:parameter-similarity} are not necessarily contradictory because the change of parameters in layer $\ell$ does not correspond one-to-one with that in representation in the layer.

First, the representation of layer $\ell$ is affected by all parameters and input data prior to layer $\ell$. So, even if the parameters of layer $\ell$ have not changed, if the parameters before layer $\ell$ have changed, the representation of layer $\ell$ may change. Second, since not all of the parameters of a neural network necessarily contribute to the output, the output of a layer may not change even if some parameters of that layer change. For example, if the input value to ReLU is negative, the output will remain 0 no matter how much the parameters involved change. So, for example, it is possible that even if the parameters of the $\ell$-layer change, the representation of the $\ell$-layer remains the same. Another example is when the values of two weights $w_1^\ell$ and $w_2^\ell$ in layer $\ell$ do not change the output of the vanilla feed-forward network because of its symmetry. 

Another possible cause of this observation unique to the current analysis is that we use CKA to measure representational similarity. CKA is designed to be invariant to some transformations on the representation matrix \citep{kornblith2019similarity}. Thus, different representation matrix is regarded as the same under these transformations even when parameters are changed.

\section{Note on Why Better Action Prediction Does Not Always Result in Better Return}
\label{appendix:note-on-why-better-action-prediction-not-always-result-in-better-return}

In Section \ref{section:mutual-information-between-hidden-representation-and-input-and-label}, we explained that randomly initialized models seem to predict action better, while in Table \ref{table:sanity-check}, they perform worse than GPT2. We will add a possible explanation of why better action prediction not necessarily comes to a better return.

The most likely cause of this observation is the current problem setup. As mentioned in Appendix \ref{appendix:training-detail}, the \textit{medium} dataset is early stopped and thus does not necessarily converge to an optimal policy. This means that if the model accurately learns trajectory and predicts the next action, it does not always mean that it is the best action. Hence, we can see that a low action error does not necessarily mean a large mean return. However, We do not believe that the use of medium will hurt the validity of the analysis using this data since this dataset is not that pathological.

Another cause might be related to mutual information. The current analysis of mutual information shows that the representation of the hidden layer of the randomly initialized model holds more information as well of the input as well as the labels. Since the neural net representation is a vector representation if a single vector contains both types of information, these types of information might have been mixed. This may make it difficult to properly utilize only the label information, making it harder to accurately predict the action. Another possibility is that the pre-trained model and the randomly initialized model differ in terms of which input token type (return-to-go, state, or action) information they encode. For example, in Fig. \ref{fig:mutual_information_context} (b), the language pre-trained model encodes information uniformly from the same token type: the hidden representation with high mutual information is the representation corresponding to the input from the triangles, i.e., the part corresponding to the return-to-go. On the other hand, in the randomly initialized models, there are variations (triangle, circle, and cross). It was noted in a previous study that attention weight was strong between the same token types \citep{reid2022can}. Therefore, perhaps the pre-trained model might encode less total information, but the amount of information \textit{effectively} utilized is not that different. Finally, as pointed out in the explanation about the limitation, the limitations of mutual information as a metric might be a cause.

\end{document}